\documentclass[10pt]{article} % For LaTeX2e
\usepackage[preprint]{tmlr}
\usepackage{amsmath,amsfonts,bm}

\def\eqref#1{equation~\ref{#1}}
\def\1{\bm{1}}

\DeclareMathAlphabet{\mathsfit}{\encodingdefault}{\sfdefault}{m}{sl}
\SetMathAlphabet{\mathsfit}{bold}{\encodingdefault}{\sfdefault}{bx}{n}

\DeclareMathOperator*{\argmax}{arg\,max}

\usepackage{hyperref}
\usepackage{url}

\usepackage{graphicx}
\usepackage{booktabs}
\usepackage{amsfonts}
\usepackage{amsmath}
\usepackage{multirow}
\usepackage{algorithmic}
\usepackage{bm}
\usepackage[lined,boxed,commentsnumbered,ruled]{algorithm2e}
\usepackage{amssymb}
\usepackage{subcaption}
\usepackage[table]{xcolor}
\definecolor{lightblue}{RGB}{221,235,247}

\title{Federated Learning of Binary Neural Networks: Enabling Low-Cost Inference}

\author{\name Nitin Priyadarshini Shankar\thanks{Equal contribution.} \email ee20d425@smail.iitm.ac.in \\
      \addr Department of Electrical Engineering,\\
      Indian Institute of Technology Madras
      \AND
      \name Soham Lahiri$^*$ \email sohaml.cse.ug@jadavpuruniversity.in  \\
      \addr Department of Computer Science and Engineering, \\
      Jadavpur University
      \AND
      \name Sheetal Kalyani \email skalyani@ee.iitm.ac.in\\
      \addr Department of Electrical Engineering,\\
      Indian Institute of Technology Madras
      \AND
      \name Saurav Prakash \email saurav@ee.iitm.ac.in\\
      \addr Department of Electrical Engineering,\\
      Indian Institute of Technology Madras}

\def\month{03}  % Insert correct month for camera-ready version
\def\year{2026} % Insert correct year for camera-ready version
\def\openreview{\url{https://openreview.net/forum?id=XXXX}} % Insert correct link to OpenReview for camera-ready version

\newcommand{\sgn}{\operatorname{sign}}

\newcommand{\tr}{\operatorname{tr}}
\newcommand{\revAdd}[1]{{{#1}}}
\newcommand{\revAddd}[1]{{{#1}}}

\begin{document}

\maketitle

\begin{abstract}
Federated Learning (FL) enables distributed clients to jointly train a global model without sharing their raw data. However, deploying the resulting deep neural networks on resource-constrained edge devices remains challenging due to their inference-time computational and memory requirements. Binary Neural Networks (BNNs) offer substantial efficiency gains by representing weights and activations using a single bit, but achieving competitive accuracy under such aggressive quantization remains difficult. Rotation-aware binarization reduces quantization errors by aligning real-valued weights with their binary representations. Despite its effectiveness in centralized learning, applying rotation-aware binarization in FL is non-trivial due to discrepancies between local and global models, particularly under heterogeneous client data distributions. To address this challenge, we propose FedBNN, a federated rotation-aware binarization framework that leverages global model information to guide local binarization while incorporating locally learned rotation information into server-side construction of the global binary model. Experiments across multiple benchmark datasets, model architectures, and degrees of data heterogeneity demonstrate that the proposed integrated methodology achieves accuracy competitive with full-precision federated models while generally outperforming existing federated binary learning approaches. Overall, FedBNN enables accurate binary inference in heterogeneous federated settings, with savings of up to $58\times$ in FLOPs and $32\times$ in memory over full-precision models, offering a practical balance between predictive performance and deployment efficiency on resource-constrained edge devices.
\end{abstract}

%\keywords{Federated Learning, Binary Neural Networks} % % % 

\section{Introduction}
\label{sec:intro}

% Federated Learning (FL) enables clients to train a shared model while keeping their data decentralized. In standard FL, clients update a global model locally using private data, and a central server aggregates their updates over multiple communication rounds. Although this framework supports collaborative training without sharing raw data, it does not remove the inference-time computational and memory demands of the resulting deep neural networks. For vision applications on resource-constrained edge devices, model memory, arithmetic operations, and energy consumption can therefore remain deployment bottlenecks. Designing models that are efficient at inference is essential for practical federated vision systems.

\revAddd{Federated Learning (FL) enables clients to jointly train a global model without sharing their raw data \citep{mcmahan2017communication, kairouz2021advances}. In standard FL, clients update a global model locally using private data, and a central server aggregates their updates over multiple communication rounds. Although this framework supports collaborative training without sharing raw data, the resulting deep neural networks (DNNs) can still impose substantial computational and memory requirements during inference \citep{sze2017efficient,geng2024algorithm}. These requirements can limit their deployment on resource-constrained edge devices, where processing power, memory, and energy are often limited \citep{wang2025empowering}. Consequently, developing inference-efficient models is important for practical federated learning systems.
}

% Binary Neural Networks (BNNs)~\citep{hubara2016binarized} constrain weights and/or activations to binary values, typically $\{-1,+1\}$, replacing expensive floating-point operations with efficient bitwise operations and reducing model memory and inference cost. However, aggressive binarization can introduce substantial quantization error and degrade inference accuracy. Rotated Binary Neural Networks (RBNNs)~\citep{lin2020rotated} address this limitation by learning a rotation that reduces the angular mismatch between real-valued weights and their binary representations. This improves the fidelity of the binary approximation while retaining binary inference, but the original rotation-aware approach is designed for centralized training, where a single model is aligned with its own binary representation.

\revAddd{Binary Neural Networks (BNNs) \citep{hubara2016binarized} represent weights and activations using a single bit, typically corresponding to values in $\{-1, +1\}$, enabling efficient bitwise operations in place of floating-point arithmetic and substantially reducing inference-time computation and memory requirements. 
However, such aggressive quantization can introduce significant errors and degrade predictive accuracy. Rotated Binary Neural Networks (RBNNs) \citep{lin2020rotated} mitigate this limitation by learning rotations that reduce the angular mismatch between real-valued weights and their binary representations. By improving the fidelity of binary approximations, rotation-aware binarization helps preserve accuracy while retaining the computational and memory advantages of BNNs. However, the original rotation-aware approach was developed for centralized learning, where binarization is optimized with respect to a single model.}

% {\color{red} We can discuss FedBAT and FedMUD separately in Related Work.}

% Applying this approach directly in FL is insufficient: under heterogeneous client data, local training can cause client models to drift apart, so their locally learned rotations and binary representations need not be consistent. Averaging such updates at the server can also undo alignments established locally. Existing federated binary approaches~\citep{yang2021communication,li2024fedbat,yue2022federated,li2025panaceas} address aspects such as communication-efficient updates or quantized aggregation, but do not jointly coordinate rotation-aware binarization across clients and the server to support accurate, lightweight inference.
\revAddd{Applying rotation-aware binarization in FL is non-trivial: under heterogeneous client data, local training can cause client models to drift apart, so their locally learned rotations and binary representations need not be consistent across clients. Moreover, aggregating locally learned models at the server can also undo alignments between real-valued weights and their binary representations achieved during local training. Existing federated binary learning approaches \citep{yang2021communication,yue2022federated} enable the training and aggregation of quantized models, but do not jointly coordinate rotation-aware binarization across clients and the server to support accurate, lightweight inference.}

% In this work, we address the challenge of reducing the runtime computational complexity of FL models on resource-constrained edge devices while maintaining high performance. To this end, we propose FedBNN, which integrates three complementary mechanisms: (i) learnable fusion of local and global weights before rotation, (ii) adjustable rotated weights that account for both rotation and alignment with the global model, and (iii) rotation-aligned server-side aggregation. Together, these mechanisms make binarization explicitly aware of the federated setting, coordinating local optimization with global aggregation while retaining the parameter count of the corresponding real-valued model. To assess whether FedBNN achieves both accurate federated learning and efficient inference, we evaluate it across datasets, model architectures, and data heterogeneity settings, analyze its inference-time computational and memory costs, and compare its post-training binarization performance with that of the baselines.

\revAdd{In this work, we focus on federated learning and address the challenge of training accurate binary neural networks while enabling efficient inference on resource-constrained edge devices. To this end, we propose FedBNN, a federated rotation-aware binarization framework that coordinates local binary optimization with global model aggregation to mitigate quantization errors and inconsistencies arising from heterogeneous client distributions. Unlike existing federated binary learning approaches, FedBNN explicitly incorporates rotation-aware optimization into both client training and server-side model construction. We evaluate FedBNN across multiple benchmark datasets, model architectures, and degrees of data heterogeneity, analyzing its inference-time computational and memory costs and comparing its predictive performance against existing federated learning methods.}

Our contributions are summarized as follows:

\begin{itemize}

\item \revAdd{We propose FedBNN, a federated rotation-aware binarization framework featuring three complementary mechanisms: learnable client-server weight fusion before rotation, adaptive adjustment of rotated weights using global model information, and server-side aggregation in rotation-aligned spaces. Specifically, bi-rotation is applied to fused client-server weights at each client, followed by adaptive adjustment along rotation-aligned and server-aligned directions. The server further leverages locally learned rotations to construct a globally consistent binary model. All mixing coefficients are learned independently for each layer rather than manually tuned as hyperparameters.}

\item We evaluate FedBNN against multiple FL baselines across five benchmark datasets (FMNIST, SVHN, CIFAR10, TinyImageNet, and FEMNIST), multiple model architectures, and three data heterogeneity settings. Across $18$ experimental configurations, FedBNN achieves accuracy within $10$ percentage points of full-precision FedAvg and outperforms it in four configurations. \revAdd{Notably, when full-precision baselines are binarized post-training, FedBNN achieves at least 16.97 percentage points higher accuracy than the strongest competing baseline across all evaluated configurations.} Additionally, FedBNN outperforms existing federated binary learning methods in most evaluated settings, demonstrating its effectiveness in learning accurate binary models directly in the federated setting.

\item We analyze the inference-time computational and memory requirements of the binary models learned by FedBNN. Our analysis indicates substantial theoretical savings in arithmetic complexity and parameter storage compared with full-precision models, while the experimental results demonstrate that these savings can be achieved with competitive classification accuracy.
\end{itemize}

\section{Related Works}

\revAddd{A substantial body of research in federated learning focuses on reducing communication overhead during distributed training. FedMUD \citep{li2025panaceas} improves communication efficiency through model update decomposition, block-wise Kronecker decomposition, and aggregation-aware decomposition. \citet{hu2024practical} propose a hybrid gradient compression framework that exploits redundancies in uplink and downlink communication, while FedSMU \citep{lufedsmu} reduces communication costs by transmitting sign-based model updates. Other approaches employ binarization or sparsification to improve efficiency. FedBAT \citep{li2024fedbat} introduces a learnable binarization operator for compressing model updates during local training, whereas SpaFL \citep{kim2024spafl} uses trainable thresholds to promote structured sparsity and reduce communication and computational costs. While these approaches address communication efficiency and, in some cases, model computation, they do not investigate rotation-aware learning for binary models to enable efficient inference. \revAdd{ We demonstrate that our proposed rotation-based binarization method significantly outperforms such methods. }}

\revAddd{Beyond communication efficiency, several studies have investigated the federated training of quantized neural networks to enable resource-efficient inference. \citet{yang2021communication} (FL-BNN) propose a federated BNN framework in which clients transmit binary model parameters to the server, along with a maximum-likelihood-based parameter-updating scheme to preserve model accuracy. FedVote \citep{yue2022federated} enables clients to transmit binary or ternary weights and employs weighted plurality voting for server-side aggregation, reducing communication costs while supporting efficient inference with quantized models. While these approaches demonstrate the feasibility of learning quantized models in FL, they do not explicitly incorporate rotation-aware binarization to reduce the angular mismatch between real-valued weights and their binary representations during local training, nor do they leverage locally learned rotations for server-side construction of the global binary model.}

\revAddd{Rotation-aware binarization has been explored in centralized learning to improve the accuracy of binary neural networks. Towards this end, \citet{lin2020rotated} propose Rotated Binary Neural Networks (RBNNs), which learn orthogonal rotations to reduce the angular mismatch between real-valued weights and their binary representations. Their approach employs alternating optimization of bi-rotation matrices and an adjustable rotated-weight representation to improve the fidelity of binarization. However, the rotations are learned with respect to a single model in a centralized setting and do not account for discrepancies between local and global models encountered in FL. Building upon this approach, FedBNN incorporates global model information into client-side rotation-aware binarization while leveraging locally learned rotations for server-side construction of the global binary model. This integrated approach improves the accuracy of federated binary models while retaining the computational and memory advantages of binary inference.}

\section{Preliminaries}

\subsection{Federated Learning} 
%(FL) is a distributed machine learning paradigm that enables multiple clients to train a shared model while keeping their data decentralized collaboratively. \revAddd{Compared with centralized training, FL retains raw data on clients, but it must cope with limited communication and statistical heterogeneity across participants.} The authors of 
\citet{mcmahan2017communication} introduced the Federated Averaging (FedAvg) algorithm, which combines local stochastic gradient descent on individual clients with periodic model averaging on a central server. FedAvg targets the empirical risk minimization problem
\begin{equation}
\min_{\mathbf{w} \in \mathbb{R}^d} \mathcal{L}(\mathbf{w}) \quad \text{where} \quad \mathcal{L}(\mathbf{w}) = \frac{1}{N_s} \sum_{i=1}^{N_s} \mathcal{L}_i(\mathbf{w}).
\end{equation}
Here, \revAddd{$\mathbf{w} \in \mathbb{R}^d$ denotes the model parameters,} $\mathcal{L}_i$ is the loss on sample $(x_i, y_i)$, and $N_s$ is the total number of samples. In the federated setting with $N_c$ clients, the same objective is written as a weighted sum of client objectives:
\begin{equation}
\label{eq:fed_objective}
\mathcal{L}(\mathbf{w}) = \sum_{k=1}^{N_c} \frac{N_{sk}}{N_s}\, l^k(\mathbf{w}) \quad \text{where} \quad l^k(\mathbf{w}) = \frac{1}{N_{sk}} \sum_{i \in \mathcal{P}^k} \mathcal{L}_i(\mathbf{w}).
\end{equation}
Here, $\mathcal{P}^k$ represents the data partition on client $k$, $N_{sk} = |\mathcal{P}^k|$ \revAdd{(cardinality of the dataset partition assigned to client \(k\))}\revAddd{, and $N_s = \sum_{k=1}^{N_c} N_{sk}$}. \revAddd{In each round $t$, FedAvg selects a random subset $S_t$ of $N_c$ clients, where $|S_t|=N_{cr}$, and sends them the current global model $\mathbf{w}_t$. Each selected client starts from $\mathbf{w} \leftarrow \mathbf{w}_t$ and runs $N_e$ local epochs of SGD with batch size $N_b$ and learning rate $\eta$, applying $\mathbf{w} \leftarrow \mathbf{w} - \eta \nabla l^k(\mathbf{w}; b)$ for each mini-batch $b \subset \mathcal{P}^k$, where $l^k(\mathbf{w}; b)$ is the average of $\mathcal{L}_i$ over $b$. The resulting model $\mathbf{w}^k_{t+1}$ is returned to the server, which averages the returned models, weighting each client by its share of the round's training data:}
\revAddd{\begin{equation}
\label{eq:fedavgaggregation}
\mathbf{w}_{t+1} \leftarrow \sum_{k \in S_t} p^k\, \mathbf{w}^k_{t+1}\ ,
\qquad
p^k = \frac{N_{sk}}{\sum_{j \in S_t} N_{sj}}.
\end{equation}}%
\revAddd{With full participation ($S_t$ containing all $N_c$ clients), $p^k = N_{sk}/N_s$ in Eq. (\ref{eq:fedavgaggregation}).}

\subsection{Binarized Neural Network} 

If \( g_\psi(\cdot) \) is a CNN with \( L \) layers, its parameters are given by \( \psi = \{\mathbf{W}_1, \dots, \mathbf{W}_L\} \), where \( \mathbf{W}_l \in \mathbb{R}^{c_o \times c_i \times f \times f} \) represents the weight matrix of the \( l^\text{th} \) layer. Here, $ c_i$ and $ c_o$ represent the input and output channels, respectively, and $f$ denotes the filter size. In a Binary Neural Network (BNN), both weights ($\mathbf{W}_l$) and activations ($\mathbf{a}_l$) are binarized using the sign function:

\begin{equation}
\begin{aligned}
\mathbf{W}_l^b = \sgn(\mathbf{W}_l), \quad \mathbf{a}_l^b = \sgn(\mathbf{a}_l),
\end{aligned}
\end{equation}

and the convolution is approximated using bit-wise operations:

\begin{equation}
\label{eq:binary_convolution}
\mathbf{W}_l \ast \mathbf{a}_l \approx \mathbf{W}_l^b \circledast \mathbf{a}_l^b,
\end{equation}

where \( \circledast \) denotes bit-wise convolution (\revAddd{using} XNOR and bit count) \cite{hubara2016binarized} . Although the forward pass uses binarized values, real-valued weights and gradients are retained for backpropagation. Due to the non-differentiability of the sign function, whose derivative is zero almost everywhere, training binarized neural networks poses significant challenges, particularly in backpropagation, where meaningful gradients are required. One common approach is using a straight-through estimator (STE): if \( b = \sgn(r) \), then:

\begin{equation}
\nabla_r = \nabla_b \cdot \mathbf{1}_{|r| \leq 1},
\end{equation}

where \( \nabla_r \text{ \revAdd{is the estimate of} } \frac{\partial C}{\partial r} \), \( \nabla_b = \frac{\partial C}{\partial b} \), and \( C \) is the cost function. To ensure stable updates, real weights are clipped to the range \([-1, 1]\). We adopt the approach of \cite{hubara2016binarized} to implement binarized convolution layers, replacing floating-point operations with efficient XNOR and bit-count operations. While this drastically reduces computation and memory usage, it often comes at the cost of reduced accuracy. \revAdd{One key limitation of BNNs is the quantization error introduced when binarizing the weight vector \( \mathbf{w}_l \in \mathbb{R}^{n_l} \), obtained by flattening the weight tensor \( \mathbf{W}_l \) of the \(l\)-th convolutional layer}, where \( n_l = c_o \cdot c_i \cdot f^2 \). \revAdd{This error manifests as both norm bias, arising from the mismatch between the magnitudes of \(\mathbf{w}_l\) and \(\mathbf{w}_l^b\), and angular bias, arising from the directional mismatch between the two vectors.} While the norm bias can be mitigated by introducing an appropriate scaling factor, the angular bias, characterized by the angle \( \phi \) between \( \mathbf{w}_l \) and \( \mathbf{w}_l^b \), cannot be eliminated through scaling, potentially degrading network performance.

% One key limitation of BNNs is the quantization error caused by binarizing the weight vector \( \mathbf{w}_l \in \mathbb{R}^{n_l} \), which is the flattened form of \( \mathbf{W}_l \), where \( n_l = c_o \cdot c_i \cdot k^2 \). This error arises due to the angular bias ($\phi$) between \( \mathbf{w}_l \) and its binarized version \( \mathbf{w}_l^b \), potentially degrading network performance.  

% {Why $1$-bit rather than low-bit quantization? Quantizing weights and activations to $b>1$ bits offers a smoother accuracy--efficiency trade-off than binarization. However, for any $b>1$, the inner products that dominate convolution are still multiply--accumulate operations: the operands are cheaper, and memory falls by $32/b$. Yet, a multiplier is still required in the datapath. At $b=1$, the product of two operands in $\{+1,-1\}$ is itself in $\{+1,-1\}$, so a dot product of length $n$ collapses to an XNOR followed by a population count over $n$ bits, and the multiplier disappears entirely. This is the property we rely on, and it is what permits the runtime savings quantified in Section~\ref{sec:runtime}. The cost is that quantization error, and hence the angular bias discussed above, is largest precisely at $b=1$, which motivates methods that reduce this error \emph{during} training rather than correcting for it afterward. We therefore target the extreme-edge regime, where multiplier area and energy dominate \cite{rastegari2016xnor} the deployment budget and $32\times$ memory compression is required, and accept the accuracy cost that comes with it.}

\revAddd{\paragraph{Why $1$-bit rather than low-bit quantization?}
While quantizing weights and activations to $b>1$ bits provides a smoother accuracy--efficiency trade-off than binarization, the underlying convolutional computation remains a multiply--accumulate (MAC) operation. 
%Although lower-bit operands reduce memory and arithmetic costs, a multiplier is still required in the computational datapath for $ b> 1$. 
In contrast, when $b=1$, both operands belong to $\{+1,-1\}$, and their product also lies in $\{+1,-1\}$. Consequently, a length-$n$ inner product can be implemented using bitwise XNOR operations followed by a population count, eliminating multipliers. This property is central to our approach and enables the runtime savings reported in Section~\ref{sec:runtime}. The trade-off, however, is substantially larger quantization error at $1$ bit, including the angular bias discussed above, which a simple scaling factor cannot correct. 
%This motivates methods that explicitly reduce quantization-induced error during training rather than compensating for it after quantization. We therefore focus on the extreme $1$-bit regime, where eliminating multipliers and achieving up to $32\times$ memory compression can provide substantial deployment benefits, while addressing the accompanying accuracy degradation.

% \textcolor{blue}{The following sections detail how FedBNN addresses quantization-induced errors through integrated rotation-aware optimization at both the client and server levels.}

}

\begin{figure}[t]
    \centering
    \includegraphics[scale=0.9]{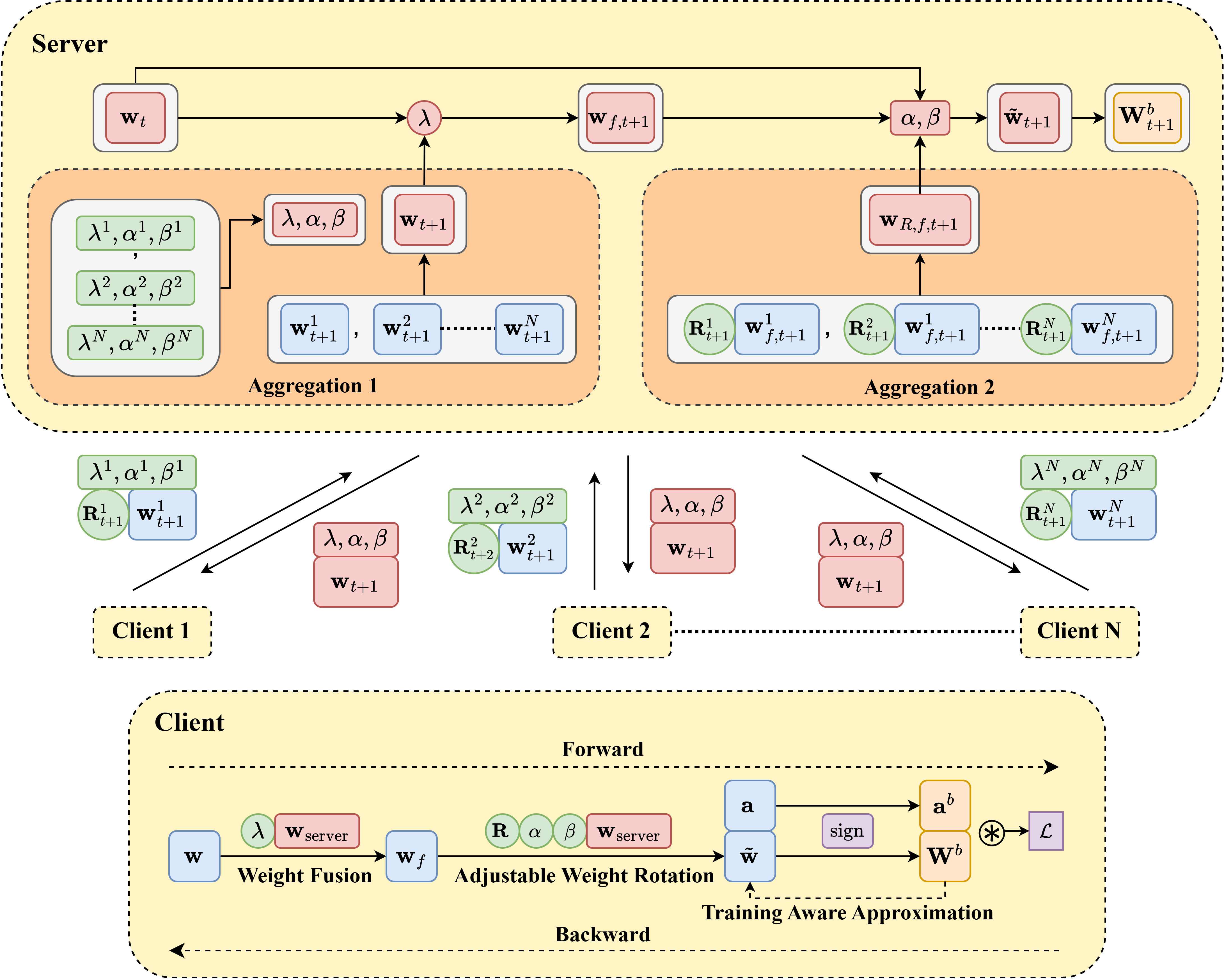}
    \caption{FedBNN overall architecture.} 
    \label{fig:architecture}
\end{figure}

\section{Proposed Method -  FedBNN}
\revAdd{In this section, we present FedBNN, our proposed framework that explicitly mitigates quantization-induced errors during training through integrated rotation-aware optimization at both the client and server levels, rather than compensating for them after quantization. An overview of the FedBNN architecture is provided in Figure \ref{fig:architecture}, and the corresponding algorithm is presented in Algorithm \ref{alg:fedbnn}.}

 % \textcolor{blue}{The substantial memory and computational benefits of binary neural networks make the extreme \(1\)-bit regime particularly attractive for resource-constrained inference, offering multiplier-free operations and up to \(32\times\) memory compression. However, the resulting quantization errors can significantly degrade model accuracy, motivating training strategies that directly mitigate these errors rather than relying on post-quantization compensation.}
\subsection{Proposed Rotation-Aware Client Training}

% \revAddd{To address the angular bias, the authors of \cite{lin2020rotated} proposed applying a rotation matrix \( \mathbf{R}_l \in \mathbb{R}^{n_l \times n_l} \) at the start of each training epoch to minimize the angle \( \phi_l \) between \( (\mathbf{R}_l)^T \mathbf{w}_l \) and \( \sgn((\mathbf{R}_l)^T \mathbf{w}_l) \).}
\revAdd{In centralized setting for binary neural networks}, to address the angular bias, the authors of \cite{lin2020rotated} proposed applying a layer-wise rotation matrix \( \mathbf{R}_l \in \mathbb{R}^{n_l \times n_l} \) at the start of each training epoch to minimize the angle \( \phi_l \) between \( (\mathbf{R}_l)^T \mathbf{w}_l \) and \( \sgn((\mathbf{R}_l)^T \mathbf{w}_l) \) for each layer \( l \).

\subsubsection{\revAdd{Federated-Aware Weight Fusion and Rotation Optimization}}
\leavevmode
     \revAddd{First, to adapt rotation for federated environments, we propose to apply a learnable fusion parameter ($\lambda_l$ per layer $l$) before rotation to merge the client and server weights, ensuring the quantization process aligns with a federated-aware representation rather than an isolated local state. The proposed fused weight \( \mathbf{w}_{fl} \) interpolates between the client and server weight representations as shown below:}
\begin{equation}
\label{eq:fused}
    \mathbf{w}_{fl} = \lambda_l \mathbf{w}_l + (1 - \lambda_l)\mathbf{w}_{\text{server},l}
\end{equation}
  Here, $\lambda_l = \text{sigmoid}(\omega_l) \in (0,1)$, $\omega_l$ is a trainable parameter, and $\mathbf{w}_{\text{server}}$ is the aggregated weight sent from the server (\revAddd{$\mathbf{w}_{t+1}$} in Eq. (\ref{eq:fedavgaggregation})) \revAddd{, $\mathbf{w}_{\text{server},l}$ is $\mathbf{w}_{\text{server}}$ for a specific layer $l$, $\mathbf{w}_{fl}$ is the fused weight for a specific layer $l$, and $\mathbf{w}_{f}$ is the fused weight of the whole architecture}. During the first local epoch of each client, the local weights are initialized from the global model, and the fused representation effectively reduces to $\mathbf{w}_{\text{server}}$. In subsequent local epochs, $\mathbf{w}_{fl}$ becomes a convex combination of the evolving client weights $\mathbf{w}_l$ and the server weights $\mathbf{w}_{\text{server}}$, as shown in \revAdd{fused weight block in  Algorithm \ref{alg:fedbnn}}. The rotation is then applied to this fused vector $\mathbf{w}_{fl}$, aligning the quantization process with a federated-aware representation while jointly capturing both local and global model information. This rotation is applied to the weights of each layer in every epoch of every round, as shown in the client diagram in Figure \ref{fig:architecture}. \revAdd{In the following, the mathematical details of the rotation process are briefly discussed.}
% Below, we discuss the entire rotation process.

For simplicity, we omit subscripts denoting the \revAddd{fused weight ($f$)}, layer \revAddd{($l$)}, client, or epoch in the following discussion. \revAddd{From \cite{lin2020rotated},} to minimize the angular bias ($\phi$), $\cos(\phi)$ needs to be maximized and hence, the optimization is formulated as follows :

\begin{equation}
\label{eq:cos}
     \cos{(\phi_{})} = \frac{\sgn ((\mathbf{R}_{})^T \mathbf{w}_{})^T ((\mathbf{R}_{})^T \mathbf{w}_{})}{\|\sgn ((\mathbf{R}_{})^T \mathbf{w}_{})\|_2 \|((\mathbf{R}_{})^T \mathbf{w}_{})\|_2},
\end{equation}

\begin{algorithm}[!tbp]
\SetKwInOut{Input}{Inputs}\SetKwInOut{Output}{Output}
\SetKw{KwTo}{in}
\SetKw{KwToo}{in parallel}
\SetKw{KwTooo}{\textbf{ClientUpdate}}
\caption{Federated Binary Neural Network (FedBNN) training. The $N_c$ clients are indexed by $k$; $N_b$ is the local minibatch size, $N_e$ is the number of local epochs, and $\eta$ is the learning rate. $N_{cr}$ is the number of clients selected per round.  $N_{eR}$ is the number of rotation optimization iterations.  $N_{lR}$ is the number of layers that require rotation. $\Theta$ is the set of all trainable parameters for all layers. \(\mathbf{w}_l\) denotes the real weights of layer \(l\), while \(\mathbf{w}\) denotes the real weights of the entire architecture. \revAddd{$(\omega, \theta, \gamma)$ are the trainable parameters that are part of the fused weight and the adjustable rotated weight.}} \label{alg:fedbnn}

$\text {\textbf{Server executes:} } $

initialize $\mathbf{w}_0$ \revAddd{and $(\omega, \theta, \gamma)$} \\
\revAddd{$\mathbf{w}_{t+1} \leftarrow \mathbf{w}_0$} \\
\For{$ \text{round}$  \KwTo $\text{range}(N_r)$}{

$S_{t} \leftarrow (\text{random set of $N_{cr}$ clients}) $

\revAddd{$\mathbf{w}_{t} \leftarrow  \mathbf{w}_{t+1}$}

\For{$ \text{each client } k \in S_t$ \KwToo }{
\revAddd{$\mathbf{w}^k_{t+1},\ (\omega^k, \theta^k, \gamma^k),\ \mathbf{R}_{t+1}^{k} \leftarrow $ \KwTooo$(k, \mathbf{w}_t, (\omega, \theta, \gamma))$} \\
}
\revAddd{$\mathbf{w}_{t+1},\ (\omega, \theta, \gamma) \leftarrow \sum_{k \in S_t} p^k\, (\mathbf{w}^k_{t+1},\ \omega^k, \theta^k, \gamma^k)$ \hfill Aggregated \emph{real-valued} model, Eq.~(\ref{eq:real_agg})}
\revAddd{$\mathbf{w}^k_{f,t+1} \leftarrow \lambda^k \mathbf{w}^k_{t+1} + (1 - \lambda^k)\, \mathbf{w}_{t+1}$ \ for $k \in S_t$ \hfill Client fused weights, Eq.~(\ref{eq:client_fused_srv})}
\revAddd{$\mathbf{w}_{f,t+1} \leftarrow \lambda\, \mathbf{w}_{t+1} + (1 - \lambda)\, \mathbf{w}_{t}$ \hfill Global fused weight, Eq.~(\ref{eq:global_fused})}\\
\revAddd{$\mathbf{w}_{R,f,t+1} \leftarrow \sum_{k \in S_t} p^k\, (\mathbf{R}^k_{t+1})^{\top} \mathbf{w}^k_{f,t+1}$ \hfill Aggregated \emph{rotation-aligned} weight, Eq.~(\ref{eq:rot_agg})}
\vspace{0.1cm}
\revAddd{$\tilde{\mathbf{w}}_{t+1} \leftarrow \mathbf{w}_{f,t+1} + \alpha\beta\, (\mathbf{w}_{R,f,t+1} - \mathbf{w}_{f,t+1}) + \alpha(1-\beta)\, (\mathbf{w}_{t} - \mathbf{w}_{f,t+1})$ \hfill Eq.~(\ref{eq:server_agg})}
\revAddd{$\mathbf{W}^b_{t+1} \leftarrow \sgn\big(\operatorname{Vec}^{-1}(\tilde{\mathbf{w}}_{t+1})\big)$ \hfill Model evaluation}
}
\vspace{0.2cm}
\revAddd{\KwTooo$(k, \mathbf{w}_{\text{server}}, (\omega, \theta, \gamma))$\textbf{:}}\\
$\mathcal{B} \leftarrow $ (split $\mathcal{P}^k$ into batches of size $N_b$)\\
$\mathbf{w} \leftarrow \mathbf{w}_{\text{server}}$\revAddd{; \ mixing parameters $\leftarrow (\omega, \theta, \gamma)$} \\
\revAddd{$\mathbf{R}_{l1}, \mathbf{R}_{l2} \leftarrow$ random orthogonal matrices for every $l$ \hfill Rotations are reset every round} \\

\For{$ \text{epoch}$  \KwTo $\text{range}(N_e)$}{

\For{$ \text{batch }b$ \KwTo $\text{range}(|\mathcal{B}|)$ }{

\For{$ l $  \KwTo $\text{range}(N_{lR})$ }{
\revAddd{$\mathbf{w}_{fl} \leftarrow \lambda_l \mathbf{w}_{l} + (1 - \lambda_l)\mathbf{w}_{\text{server,}l}$ \hfill Fused weight \\
$\mathbf{\overline{W}}_l \leftarrow \operatorname{Vec}^{-1}(\mathbf{w}_{fl})$} \\
}
\If{\revAddd{$b = 1$}}{
\For{$ e_R $  \KwTo $\text{range}(N_{eR})$}{
\For{$ l $  \KwTo $\text{range}(N_{lR})$ }{
$ \mathbf{W'}^{b}_l \leftarrow \sgn((\mathbf{R}_{l1})^T \mathbf{\overline{W}}_{l} \mathbf{R}_{l2})$ \hfill \revAddd{Rotation optimization}

$ \mathbf{U}_{1},\mathbf{S}_{1},\mathbf{V}_{1} \leftarrow \text{SVD}(\mathbf{W'}^{b}_l (\mathbf{R}_{l2})^T (\mathbf{\overline{W}}_{l})^T )$

$\mathbf{R}_{l1} \leftarrow \mathbf{V}_{1} (\mathbf{U}_{1})^T$

$ \mathbf{U}_{2},\mathbf{S}_{2},\mathbf{V}_{2} \leftarrow \text{SVD}((\mathbf{\overline{W}}_{l})^T  \mathbf{R}_{l1} \mathbf{W'}_{l}^b )$

$\mathbf{R}_{l2} \leftarrow \mathbf{U}_{2} (\mathbf{V}_{2})^T$
}
}
}
\For{$ l $  \KwTo $\text{range}(N_{lR})$ }{
\revAddd{$\Tilde{\mathbf{w}}_{l} \leftarrow \mathbf{w}_{fl} + \alpha_{l} \beta_{l}  (\mathbf{R}_{l}^T \mathbf{w}_{fl} - \mathbf{w}_{fl}) + \alpha_{l} (1 - \beta_{l})  (\mathbf{w}_{\text{server},l} - \mathbf{w}_{fl})$ \hfill Adjustable rotated weight}

}
\For{$ \mu \in \Theta$ }{$\mathbf{\mu} \leftarrow \mu - \eta \ \sigma_\mu(\mathcal{B}_b)$}
}
}

return $\mathbf{w}$, \revAddd{$(\omega, \theta, \gamma)$,}
$\mathbf{R}$ to server
\end{algorithm}

where $(\mathbf{R}_{})^T\mathbf{R}_{} = \mathbf{I}_{n_{}}$ is the $n_{}$-th order identity matrix. Note, $\|\sgn ((\mathbf{R}_{})^T \mathbf{w}_{})\|_2 = \sqrt{n_{}}$ and $\|((\mathbf{R}_{})^T \mathbf{w}_{})\|_2 = \|\mathbf{w}_{}\|_2$. \revAdd{Furthermore, the rotation optimization is done before  each training epoch while keeping $\mathbf{w}_{}$ a constant.} With the help of algebraic manipulations, we get
$\mathbf{W'}_{}^b {=} \sgn((\mathbf{R}_{1})^T\mathbf{\overline{W}}_{}\mathbf{R}_{2})$,   where Vec$(\mathbf{\overline{W}}_{}) {=} \mathbf{w}_{}$, {$\mathbf{\overline{W}}_{} {\in} \mathbb{R}^{n_1 \times n_2}$}. 
\begin{equation}
\begin{aligned}
   \mathbf{w}_{}^T \mathbf{R} = \mathbf{w}_{}^T (\mathbf{R}_{1} \otimes \mathbf{R}_{2}) = \text{Vec}((\mathbf{R}_{2})^T (\mathbf{\overline{W}}_{})^T \mathbf{R}_{1})    
\end{aligned}
\end{equation}

where $\otimes$ is the Kronecker product and the operation Vec($\cdot$) vectorizes an input matrix. The final optimization objective is given by 

\begin{equation}
\begin{aligned}
\argmax_{\mathbf{W'}_{}^b, \mathbf{R}_{1}, \mathbf{R}_{2}} &\tr(\mathbf{W'}_{}^b  (\mathbf{R}_{2})^T (\mathbf{\overline{W}}_{})^T \mathbf{R}_{1})\\
\text{s.t. } \mathbf{W'}_{}^b &\in  \{+1,-1\}^{n_{1}\times n_{2}} \\
&(\mathbf{R}_{1})^T\mathbf{R}_{1} = \mathbf{I}_{n_{1}} \\
&(\mathbf{R}_{2})^T\mathbf{R}_{2} = \mathbf{I}_{n_{2}}.
\end{aligned}
\end{equation}

Since the above optimization is non-convex, an alternating approach is used, in which one variable is updated while the other two are held fixed until convergence. We, therefore, have three optimization steps, as shown in the rotation optimization block in Algorithm \ref{alg:fedbnn}:
\begin{enumerate}
    \item \raggedright The first step is to learn $\mathbf{W'}_{}^b$ while fixing $\mathbf{R}_{1}$ and $\mathbf{R}_{2}$. It is solved by

    \begin{equation}
        \mathbf{W'}_{}^b = \sgn((\mathbf{R}_{1})^T \mathbf{\overline{W}}_{} \mathbf{R}_{2})
    \end{equation}

    \item \raggedright \revAdd{The next step updates $\mathbf{R}_{1}$ while keeping $\mathbf{W'}_{}^b$ and $\mathbf{R}_{2}$ constant. Performing}
    \[
    \mathrm{SVD}\left(
    \mathbf{W'}_{}^b (\mathbf{R}_{2})^T
    (\mathbf{\overline{W}}_{})^T
    \right)
    = \mathbf{U}_{1}\mathbf{S}_{1}(\mathbf{V}_{1})^T,
    \]
    it is solved by

    \begin{equation}
        \mathbf{R}_{1} = \mathbf{V}_{1}(\mathbf{U}_{1})^T.
    \end{equation}

    \item \raggedright Similar to the previous steps, the following step updates $\mathbf{R}_{2}$ while keeping $\mathbf{W'}_{}^b$ and $\mathbf{R}_{1}$ constant. Performing
    \[
    \mathrm{SVD}\left(
    (\mathbf{\overline{W}}_{})^T
    \mathbf{R}_{1}\mathbf{W'}_{}^b
    \right)
    = \mathbf{U}_{2}\mathbf{S}_{2}(\mathbf{V}_{2})^T,
    \]
    it is solved by

    \begin{equation}
        \mathbf{R}_{2} = \mathbf{U}_{2}(\mathbf{V}_{2})^T.
    \end{equation}
\end{enumerate}

{We fix the number of rotation optimization iterations to $N_{eR}=3$ throughout all experiments following \cite{lin2020rotated}.}
% We found that three alternating optimization iterations for the variables $\mathbf{W'}^{b}$, $\mathbf{R}_{1}$, and $\mathbf{R}_{2}$ to converge while incurring low computational overhead.}

\subsubsection{Adjustable rotated weight vector {for federated learning} with global memory}
The optimization steps described above are executed iteratively. The variables \( \mathbf{W'}^b \), \( \mathbf{R}_1 \), and \( \mathbf{R}_2 \) typically converge within three iterations. \revAddd{However, the optimization is highly non-linear due to binarization, and the process may still get trapped in a local optimum due to overshooting/undershooting, as discussed in \cite{lin2020rotated}}. 
%An adjustable, rotated weight vector scheme was used to further reduce angular bias after the bi-rotation step. However
\revAdd{Furthermore}, in a federated setting,  a client may need to align its weights not just with its own rotated representation but also with $\mathbf{w}_{\text{server}}$ even after rotation. To this end, we propose an adjustable rotated weight for a given layer $l$ as:
\begin{equation}
\begin{aligned}
\label{eq:fed_weightupdate}
\Tilde{\mathbf{w}}_{l} = \mathbf{w}_{fl} + \alpha_{l} \beta_{l}  (\mathbf{R}_{l}^T \mathbf{w}_{fl} - \mathbf{w}_{fl}) + \alpha_{l} (1 - \beta_{l})  (\mathbf{w}_{\text{server,}l} - \mathbf{w}_{fl})
\end{aligned}
\end{equation}

\revAddd{where $\mathbf{w}_{fl}$ is the fused weight}, $\alpha_{l} = \bigl|\sin(\theta_{l})\bigr|$, $\beta_{l} = \bigl|\sin(\gamma_{l})\bigr|$, $\theta_{l},\;\gamma_{l}\in\mathbb{R}$ are learnable parameters controlling the contributions from the rotated and server {weights}. {Our proposed parameterization using $\alpha_{l}\beta_{l}$ and $\alpha_{l}(1-\beta_{l})$ ensures that the resulting coefficient of $\mathbf{w}_{fl}$ in Eq. (\ref{eq:fed_weightupdate}) remains non-negative, thereby preserving update stability of the rotated and server-aligned directions.} The added regularization term {($\alpha_{l} (1- \beta_{l})  (\mathbf{w}_{\text{server},l} - \mathbf{w}_{fl})$)} updates adaptively and fuses local and global knowledge while correcting angular bias with respect to both. \revAddd{Further, the adjustable rotated weight $\Tilde{\mathbf{w}}_{l}$ is then binarized and used in the forward pass (Eq. (\ref{eq:binary_convolution})) as shown below: 
\begin{equation}
\label{eq:binarization}
    \mathbf{W}_l^b = \sgn\left(\operatorname{Vec}^{-1}(\Tilde{\mathbf{w}}_{l})\right)
\end{equation}
}
In summary, each client receives the global server weight $\mathbf{w}_{\text{server},l}$ at the beginning of each round. We introduce a learnable fusion parameter $\lambda_l$ to interpolate between the client and server weights, forming a federated-aware fused weight $\mathbf{w}_{fl}$. The bi-rotation is then applied to $\mathbf{w}_{fl}$ instead of $\mathbf{w}_l$, allowing angular correction in the shared representation space. Moreover, we introduce two additional learnable scalars $\alpha_l$ and $\beta_l$ to adaptively adjust the influence of the rotated direction and the global server model, respectively, during the binarization step.

% It draws inspiration from \cite{li2020federated}, where a proximal term is added to prevent training divergence caused by heterogeneous data. While the original method in \cite{lin2020rotated} is designed for centralized training, we extend this framework to federated learning. 

\subsubsection{Training aware approximation for federated learning} 

We build upon the training-aware approximation function introduced in~\cite{lin2020rotated}, which provides a smooth, epoch-dependent surrogate for the sign function, thereby facilitating stable gradient propagation during the early stages of training. In the centralized setting, the approximation parameters $\tau$ and $\kappa$ are updated locally as training progresses; in our federated setting, we instead \revAddd{propose to} maintain their progression across global rounds so that all participating clients follow a consistent approximation schedule. The resulting approximation is given by:

% We build upon the training-aware approximation function \cite{lin2020rotated}, which serves as a smooth, epoch-dependent surrogate for the sign function, thereby enabling better gradient flow during early training. Unlike centralized training, where $t$ and $k$ are updated locally at each epoch, our federated setup maintains these values across global rounds to ensure consistent client training behavior. The approximation is given by:
\begin{equation}
F(x) = 
\begin{cases}
\kappa \cdot \left( -\text{sign}(x) \cdot \dfrac{\tau^2 x^2}{2} + \sqrt{2}\tau x \right), & \text{if } |x| < \frac{\sqrt{2}}{\tau}, \\
\kappa \cdot \text{sign}(x), & \text{otherwise},
\end{cases}
\label{eq:adaptive_sign}
\end{equation}
where the coefficients \( \tau \) and \( \kappa \) evolve with training as:
\begin{equation}
\tau = 10^{(T_{\min}) + \frac{(tN_e + e)}{N_r N_e} (T_{\max} - T_{\min})\quad}
\end{equation}
\begin{equation}
\kappa = \max\left(\frac{1}{\tau}, 1\right)
\end{equation}
where \( T_{\min} = -2 \), \( T_{\max} = 1 \), \( N_r \) the total number of global training rounds, and \( t \) the current round index, \( N_e \) the total number of local training epochs, and \( e \) the current epoch of training. 
The derivative of this function with respect to \( x \) is:
\begin{equation}
F'(x) = \frac{\partial F(x)}{\partial x} = \max\left( \kappa \cdot (\sqrt{2}\tau  - |\tau^2 x|), 0 \right),
\label{eq:approx_gradient}
\end{equation}
which yields non-zero gradients during early training, enabling effective optimization of client parameters, and then progressively transitions to a sign-like function, thereby preserving binarization.

Using this surrogate, we compute gradients of the loss \( \mathcal{L} \) with respect 
to both activations \( \mathbf{a} \) and the real layer weights \( \mathbf{w}_{l} \) as follows:
\revAddd{\begin{align}
\sigma_{\mathbf{a}_{l}} &= \frac{\partial \mathcal{L}}{\partial \mathbf{a}_{l}} \ = \frac{\partial \mathcal{L}}{\partial F(\mathbf{a}_{l})} \cdot \frac{\partial F(\mathbf{a}_{l})}{\partial \mathbf{a}_{l}}, \\
\sigma_{\mathbf{w}_l} &= \frac{\partial \mathcal{L}}{\partial \mathbf{w}_l}  = \frac{\partial \mathcal{L}}{\partial F(\mathbf{\tilde{w}}_{l})} \cdot \frac{\partial F(\mathbf{\tilde{w}}_{l})}{\partial \mathbf{\tilde{w}}_{l}} \cdot \frac{\partial \mathbf{\tilde{w}}_{l}}{\partial \mathbf{w}_{fl}} \cdot \frac{\partial \mathbf{w}_{fl}}{\partial \mathbf{w}_l},
\end{align}}
where
% \begin{equation}
% \frac{\partial \tilde{\mathbf{w}}_{l}}{\partial \mathbf{w}_{l}} = \lambda_{l}\left((1 - \alpha_{l}) \cdot \mathbf{I}_n + \alpha_{l} \beta_{l} \cdot \mathbf{R}_{l}^\top \right),
% \end{equation}
\revAddd{\begin{equation}
\frac{\partial \tilde{\mathbf{w}}_{l}}{\partial \mathbf{w}_{fl}} = (1 - \alpha_{l}) \cdot \mathbf{I}_n + \alpha_{l} \beta_{l} \cdot \mathbf{R}_{l}^\top , \hspace{1cm} \frac{\partial {\mathbf{w}}_{fl}}{\partial \mathbf{w}_{l}} = \lambda_{l}
\end{equation}}

accounting for both the direct client path and the rotation-aligned correction. The gradients of the \revAddd{trainable} parameters \( \alpha_{} \), \( \beta_{} \) and \( \lambda_{} \), which respectively control the contributions from the rotation-aligned direction and the global server model, are computed as:
\revAddd{
\begin{align}
\sigma_{\alpha_{l}} &= \frac{\partial \mathcal{L}}{\partial \mathbf{\alpha}_{l}} = \frac{\partial \mathcal{L}}{\partial \tilde{\mathbf{w}}_{l}} \cdot \{\beta_{l}(\mathbf{R}_{}^\top \mathbf{w}_{fl} - \mathbf{w}_{\text{server},l}) + (\mathbf{w}_{\text{server},l} - \mathbf{w}_{fl}) \} \\[5pt]
 % &= , \\[5pt]
\sigma_{\beta_{l}}  &=  \frac{\partial \mathcal{L}}{\partial \mathbf{\beta}_{l}} =  \frac{\partial \mathcal{L}}{\partial \tilde{\mathbf{w}}_{l}} \cdot \{\alpha_{l}(\mathbf{R}_{}^\top \mathbf{w}_{fl} - \mathbf{w}_{\text{server},l})\} \\[5pt]
 % &= , \\[5pt]
\sigma_{\lambda_{l}}  &=  \frac{\partial \mathcal{L}}{\partial \mathbf{\lambda}_{l}} = \frac{\partial \mathcal{L}}{\partial \tilde{\mathbf{w}}_{l}} \cdot \{\alpha_{l} \beta_{l}(\mathbf{R}_{}^\top( \mathbf{w}_{l} - \mathbf{w}_{\text{server},l})) + (1 - \alpha_{l})(\mathbf{w}_{l} - \mathbf{w}_{\text{server},l})\} 
\end{align}}

% This training-aware formulation plays a critical role in stabilizing federated optimization by aligning binarization with geometric orientation and enabling meaningful gradient flow throughout local client training, as shown in Figure \ref{fig:architecture}.

In summary, at the beginning of every training epoch of each client, the rotation matrices, $\mathbf{R}_{l1}$ and $\mathbf{R}_{l2}$, are learned for a fixed $\mathbf{w}_{fl}$. At the training phase, with the fixed $\mathbf{R}_{l1}$ and $\mathbf{R}_{l2}$, the proposed model takes the sign of parameter $\tilde{\mathbf{w}}_{l}$ (Eq. (\ref{eq:binarization})) for the forward pass and the parameters $\mathbf{w}_l$, $\alpha_{l}$, $\beta_{l}$ and $\lambda_{l}$ are updated during back-propagation, which enables the network to learn a suitable value that further optimizes the application of the rotation in Eq. (\ref{eq:fed_weightupdate}). \\
\textbf{Choice of $\lambda_{l}$, $\alpha_{l}$, and $\beta_{l}$}: In the proposed method, $\lambda_{l}$, $\alpha_{l}$, and $\beta_{l}$ are treated as trainable parameters rather than hyperparameters, enabling the network to learn optimal values for each layer automatically. The variation of these parameters across training rounds for different layers is studied in depth in Appendix \ref{Alpha, beta, lambda}.

\subsection{Proposed Aggregation Strategy}
\label{sec32}

\revAddd{In FedBNN, the server keeps two representations of the global model: (i)~the real-valued FedAvg aggregate (the weights $\mathbf{w}_{t+1}$ together with the averaged mixing parameters), 
%which is the only model broadcast to clients
and (ii)~a binarization-aware evaluation weight $\tilde{\mathbf{w}}_{t+1}$, built from the clients' weights in their rotated spaces. \revAdd{However, as we describe next, the server broadcasts only the real-valued FedAvg aggregate, while,} the sign of $\tilde{\mathbf{w}}_{t+1}$ is the binary model that the server validates, selects, and finally reports; $\tilde{\mathbf{w}}_{t+1}$ itself is never transmitted to clients during training.}

\revAdd{After local training in round $t$, each participating client $k \in S_t$ uploads its real-valued weights $\mathbf{w}^k_{t+1}$, its mixing parameters $(\omega^k, \theta^k, \gamma^k)$, and its rotation matrices $\mathbf{R}^k_{t+1}$ (one pair $\mathbf{R}_1, \mathbf{R}_2$ per binarized layer, as obtained at the start of its last local epoch). The rotation matrices amount to about $1\%$ of the \revAdd{communicated real weights} for CNN4 and less than $0.6\%$ for ResNet10, ResNet18, and ConvNeXt-Tiny (Table~\ref{tab:rotation_overhead} in the Appendix \ref{table8}), so the additional uplink cost is small and decreases with model size. Each operation below is applied separately to each binarized layer; we drop the layer index $l$ from the subscripts. \revAdd{Furthermore, to differentiate between clients, we use the superscript $k$ to denote the $k$-th client.} Clients are weighted by their share of the round's training data, $p^k = N_{sk} / \sum_{j \in S_t} N_{sj}$, and the server proceeds in four steps (see also Figure~\ref{fig:architecture}).}

\textbf{Step 1: }\revAddd{The server averages the clients' real-valued parameters exactly as in FedAvg (Eq. (\ref{eq:fedavgaggregation}), restricted to the participating clients) and averages their mixing parameters in the same way:}
\revAddd{\begin{equation}
\label{eq:real_agg}
\mathbf{w}_{t+1} = \sum_{k \in S_t} p^k\, \mathbf{w}^{k}_{t+1},
\qquad
(\omega, \theta, \gamma) = \sum_{k \in S_t} p^k\, (\omega^k, \theta^k, \gamma^k).
\end{equation}}%
\revAddd{Together, $\mathbf{w}_{t+1}$ and $(\omega, \theta, \gamma)$ form the global model broadcast in round $t+1$, in which $\mathbf{w}_{t+1}$ serves as $\mathbf{w}_{\text{server}}$ in Eqs. (\ref{eq:fused}) and (\ref{eq:fed_weightupdate}).}

\textbf{Step 2: }\revAddd{The server rebuilds the fused weight of Eq. (\ref{eq:fused}) at two levels. For each participating client, it fuses the client's final weights with the new aggregate $\mathbf{w}_{t+1}$, rather than with the $\mathbf{w}_t$ that the client trained against, using the client's own learned coefficient; every client's contribution is thus expressed relative to the new global model:}
\revAddd{\begin{equation}
\label{eq:client_fused_srv}
\mathbf{w}_{f,t+1}^k = \lambda^k\, \mathbf{w}_{t+1}^k + (1 - \lambda^k)\, \mathbf{w}_{t+1},
\qquad \lambda^k = \operatorname{sigmoid}(\omega^k).
\end{equation}}%
\revAddd{At the global level, it interpolates between the new and the previous global model with the averaged coefficient, so that $\mathbf{w}_{t+1}$ plays the role of the client weight in Eq. (\ref{eq:fused}) and $\mathbf{w}_t$, the model broadcast at the start of round $t$, plays the role of $\mathbf{w}_{\text{server}}$:}
\revAddd{\begin{equation}
\label{eq:global_fused}
\mathbf{w}_{f,t+1} = \lambda\, \mathbf{w}_{t+1} + (1 - \lambda)\, \mathbf{w}_{t},
\qquad \lambda = \operatorname{sigmoid}(\omega).
\end{equation}}

\textbf{Step 3: }\revAdd{To get the aggregate for the rotated fused weight in Eq. (\ref{eq:fed_weightupdate}), the server first maps} each client's fused weight into that client's rotated weight and averages the results:
\revAddd{\begin{equation}
\label{eq:rot_agg}
\mathbf{w}_{R,f,t+1} = \sum_{k \in S_t} p^k\, (\mathbf{R}_{t+1}^{k})^{\top}\, \mathbf{w}_{f,t+1}^{k}
\end{equation}}%
Here, each term \revAdd{in the above sum} is the rotated fused weight $\mathbf{R}^{\top}\mathbf{w}_{f}$ of Eq. (\ref{eq:fed_weightupdate}), evaluated with client $k$'s final weights, the new anchor $\mathbf{w}_{t+1}$, and client $k$'s own rotation. \revAdd{We would like to emphasize that the local rotation matrices are used for this mapping: no averaged rotation is used. This is because an average of orthogonal matrices is in general not orthogonal.} Furthermore, each client draws fresh random orthogonal $\mathbf{R}_1, \mathbf{R}_2$ at the start of every round before optimizing them as in Algorithm~\ref{alg:fedbnn}.

\textbf{Step 4: }\revAddd{The server combines these quantities through the server-side counterparts of Eq. (\ref{eq:fed_weightupdate}), i.e.,  $\mathbf{w}_{f,t+1}$ replaces $\mathbf{w}_{f}$, the rotation-aligned aggregate $\mathbf{w}_{R,f,t+1}$ replaces $\mathbf{R}^{\top}\mathbf{w}_{f}$, and the previous global model $\mathbf{w}_t$ replaces $\mathbf{w}_{\text{server}}$:}
\revAddd{\begin{equation}
\label{eq:server_agg}
\tilde{\mathbf{w}}_{t+1} = \mathbf{w}_{f,t+1}
  + \alpha\beta\,(\mathbf{w}_{R,f,t+1} - \mathbf{w}_{f,t+1})
  + \alpha(1-\beta)\,(\mathbf{w}_{t} - \mathbf{w}_{f,t+1}),
\qquad \alpha = |\sin\theta|,\ \beta = |\sin\gamma|,
\end{equation}}%
\revAddd{where $\theta$ and $\gamma$ are the averages from Eq. (\ref{eq:real_agg}). Equivalently, $\tilde{\mathbf{w}}_{t+1} = (1-\alpha)\,\mathbf{w}_{f,t+1} + \alpha\beta\,\mathbf{w}_{R,f,t+1} + \alpha(1-\beta)\,\mathbf{w}_{t}$; since $\alpha, \beta \in [0,1]$, the evaluation weight is a convex combination of the fused, rotation-aligned, and previous global weights, and its three coefficients are the quantities tracked over training in Figures~\ref{alphabetalambdafmnist}--\ref{alphabetalambdafemnist} of the Appendix \ref{Alpha, beta, lambda}. The server binarizes it as in Eq. (\ref{eq:binarization}), $\mathbf{W}^b_{t+1} = \sgn\big(\operatorname{Vec}^{-1}(\tilde{\mathbf{w}}_{t+1})\big)$, evaluates $\mathbf{W}^b_{t+1}$ on the validation set, and retains the round with the highest validation accuracy; the binary model of that round is the one evaluated on the test set.}

\revAdd{We also emphasize that the binarization step used to obtain $\mathbf{W}^{b}_{t+1}$ does not affect client-side training or the model parameters communicated between clients and the server. The evaluation weight only determines the binary model selected for inference.} In round $t+1$, every selected client initializes its weights and mixing parameters from $\mathbf{w}_{t+1}$ and $(\omega, \theta, \gamma)$, and randomized rotation matrices, and trains as in Algorithm~\ref{alg:fedbnn}. Building $\tilde{\mathbf{w}}_{t+1}$ in this way makes model selection binarization-aware: the server scores the same kind of adjustable rotated weight that clients binarize in their forward passes (Eq. (\ref{eq:fed_weightupdate})), using each client's own rotation. \revAdd{We observe through empirical evaluations that} aggregating in the rotation-aligned space before forming $\tilde{\mathbf{w}}_{t+1}$ gives higher accuracy than averaging the clients' adjustable weights $\tilde{\mathbf{w}}^k$ directly in 15 of 18 settings, by up to $18.2$ percentage points under the strongest heterogeneity (CIFAR10, ResNet10, Non-IID~2; Table~\ref{tab:ablation2}, \revAdd{Appendix \ref{aggregatingwtilda}}). Setting $\lambda^k = \lambda = \beta = 1$ removes every server anchor and reduces Eq. (\ref{eq:server_agg}) to $\mathbf{w}_{t+1} + \alpha\big(\sum_{k \in S_t} p^k (\mathbf{R}^k_{t+1})^{\top}\mathbf{w}^k_{t+1} - \mathbf{w}_{t+1}\big)$; this variant lowers accuracy in 16 of the 18 settings of Table~\ref{tab:ablationserver} (Section~\ref{sec:ablation_server}).

\subsection{Runtime computation savings for binary models}
\label{sec:runtime}
From \cite{10505944}, we estimate the number of runtime multiplication and addition operations in a $2$D CNN for comparison. For a convolution between real-valued \( \mathbf{W}_l \in \mathbb{R}^{c_o \times c_i \times f \times f} \) and input \( \mathbf{a}_l \in \mathbb{R}^{c_i \times h_{in}^w \times h_{in}^h} \), the output is \( \mathbf{a}_{l+1} \in \mathbb{R}^{c_o \times h_{out}^w \times h_{out}^h} \). The number of multiplications is \( c_i \cdot f^2 \cdot h_{out}^w \cdot h_{out}^h \cdot c_o \), and additions are roughly of the same order. Thus, the total FLOPs for the \( l^\text{th} \) layer is approximately \( 2 \cdot c_i \cdot f^2 \cdot h_{out}^w \cdot h_{out}^h \cdot c_o \). Also, we consider every parameter to be of $32$ bits. Hence, to calculate the total memory, we multiply the total parameters by $32$. By binarizing weights and activations to \(\{+1, -1\}\), convolutions are replaced by efficient XNOR and bit-count operations. And all the weights will require only $1$ bit of storage. The work in \cite{rastegari2016xnor} reports a FLOPs reduction of \(58 \times\) and memory savings of \(32 \times\). Hence, to compare FedBNN with real models, we use this conversion factor to estimate the computational savings in FLOPs and memory.

\section{Experimental Evaluation}

\subsection{Datasets and Partitioning}

To comprehensively evaluate the effectiveness of FedBNN, we conduct experiments on several widely used federated learning benchmarks (\cite{li2024fedbat,kim2024communication,liu2024fedlpa}), namely, FMNIST, SVHN, CIFAR10, TinyImageNet, and FEMNIST. The client models are trained on the partitioned training data for all experiments. The testing data is split into two equal sets: validation and testing. The best model is selected on the server after aggregation, based on \revAdd{binarized accuracy on the }validation set. The final performance of the model is reported on the unseen test set. To thoroughly assess our approach's performance on data heterogeneity, we evaluate under both IID and non-IID data distribution scenarios, following federated learning benchmarks \cite{mcmahan2017communication,li2024fedbat}. Under IID partitioning, each client is assigned an equal share of randomly sampled data, resulting in locally similar datasets. The non-IID setting comprises two configurations: Non-IID 1 and Non-IID 2. In Non-IID 1, samples are distributed among clients according to a Dirichlet distribution \cite{hsu2019measuring}, with the Dirichlet parameter $\nu$ modulating the degree of statistical skewness, set to $0.3$ across all datasets. Non-IID 2 represents an extreme case of heterogeneity, where each client receives data from only a subset of possible labels. These partitioning strategies enable a systematic examination of model performance as data distributions on clients become increasingly disparate, closely mirroring realistic federated deployment scenarios.

\subsection{Simulation Setup}

All experiments are implemented using the PyTorch framework and executed on a server with two NVIDIA RTX PRO 6000 GPUs. We conduct federated training with $N_c = 100$ clients per experiment for all datasets. The training follows the standard federated averaging (FedAvg) protocol for model synchronization. We use stochastic gradient descent with learning rates of $0.1$ for CNN4, ResNet10, and ResNet18. \revAdd{For ConvNeXt-Tiny, we use a learning rate of $0.01$. Including ConvNeXt-Tiny allows us to evaluate the scalability of FedBNN beyond lightweight convolutional architectures using a more challenging, high-capacity model.} The learning rate is decreased by a factor of $2$ throughout training at an interval of $200$ rounds. Furthermore, $10$ clients are randomly sampled for local model updates in each round. Each federated training round consists of $15$ local epochs ($5$ for FMNIST and $10$ for SVHN), with a mini-batch size of $64$. The global training process runs for $1500$ rounds in total ($500$ for FMNIST and SVHN). 
%{To evaluate scalability beyond lightweight convolutional networks, we additionally consider ConvNeXt-Tiny, a modern high-capacity architecture that provides a substantially more challenging benchmark for binary federated learning than ResNet-based models. }

\subsection{Simulation results}
 
\revAdd{We first compare FedBNN with federated methods that also deploy binary models (Section~\ref{sec:binary_comparison}) and then with full-precision and communication-efficient federated methods (Section~\ref{sec:fp_comparison}), in both cases on CIFAR10 with ResNet10 and ConvNeXt-Tiny and on TinyImageNet with ResNet18. Appendix~\ref{sec:additional_results} gives the comparison with full-precision and communication-efficient methods on additional datasets.}
 
\subsubsection{Comparison with inference-efficient binary federated learning methods}
\label{sec:binary_comparison}
 
\begin{table*}[!ht]
\caption{Comparison with inference-efficient binary federated learning methods for $N_c = 100$. Accuracy is the model's classification accuracy on the validation set, and binarized accuracy is its test accuracy after post-training binarization (Section~\ref{sec:post-training}). FL-BNN and FedBNN perform inference directly with binary weights, so their two accuracies coincide. The best result in each column of each block is shown in \textbf{bold}.}
\centering
\label{tab:inference_comparison}
% \resizebox{\textwidth}{!}{
\begin{tabular}{|c|c|c|c|c|c|c|c|}
\hline
\multirow{2}{*}{\textbf{Method}} &
\multirow{2}{*}{\textbf{Dataset}} &
\multicolumn{3}{c|}{\textbf{Accuracy}} &
\multicolumn{3}{c|}{\textbf{Binarized Accuracy}} \\
\cline{3-5}\cline{6-8}
& \multirow{2}{*}{\textbf{(Model)}}
& \textbf{IID} & \textbf{Non -} & \textbf{Non -}
& \textbf{IID} & \textbf{Non -} & \textbf{Non -} \\
& & & \textbf{IID 1} & \textbf{IID 2}
&  & \textbf{IID 1} & \textbf{IID 2} \\
\hline
 
FL-BNN &
\multirow{2}{*}{CIFAR10}
& 78.86 & 57.36 & 56.50
& 78.86 & 57.36 & 56.50 \\
FedVote & \multirow{2}{*}{(ResNet10)}
& 68.54 & 57.40 & 45.12
& 70.16 & 59.60 & 44.36 \\
{FedBNN} &
& \textbf{89.95} & \textbf{82.84} & \textbf{73.82}
& \textbf{89.95} & \textbf{82.84} & \textbf{73.82} \\
\hline
 
FL-BNN &
\multirow{2}{*}{CIFAR10}
& 65.40 & \textbf{68.96} & \textbf{67.26}
& 65.40 & \textbf{68.96} & \textbf{67.26} \\
FedVote & \multirow{2}{*}{(ConvNeXt-Tiny)}
& 32.70 & 32.16 & 32.06
& 32.52 & 31.96 & 32.46 \\
{FedBNN} &
& \textbf{72.08} & 67.08 & 63.00
& \textbf{72.08} & 67.08 & 63.00 \\
\hline
 
FL-BNN &
\multirow{2}{*}{TinyImageNet}
& 26.96 & 26.88 & 23.92
& 26.96 & 26.88 & 23.92 \\
FedVote & \multirow{2}{*}{(ResNet18)}
& 24.56 & 18.80 & 20.96
& 25.28 & 20.06 & 21.02 \\
{FedBNN} &
& \textbf{46.54} & \textbf{43.00} & \textbf{45.74}
& \textbf{46.54} & \textbf{43.00} & \textbf{45.74} \\
\hline
\end{tabular}
\end{table*}
 
\revAdd{Table~\ref{tab:inference_comparison} compares FedBNN with FL-BNN~\citep{yang2021communication} and FedVote~\citep{yue2022federated}. FedBNN is the most accurate method in seven of the nine settings. On CIFAR10 (ResNet10), it reaches $89.95\%$, $82.84\%$, and $73.82\%$ under IID, Non-IID~1, and Non-IID~2, against $78.86\%$, $57.36\%$, and $56.50\%$ for FL-BNN and $68.54\%$, $57.40\%$, and $45.12\%$ for FedVote, so its margin over FL-BNN grows from $11.1$ percentage points under IID data to $25.5$ points under Non-IID~1. On TinyImageNet, FedBNN is $16.1$ to $21.8$ points more accurate than FL-BNN and $22.0$ to $24.8$ points more accurate than FedVote. The two exceptions are CIFAR10 (ConvNeXt-Tiny) under Non-IID~1 and Non-IID~2, where FL-BNN reaches $68.96\%$ and $67.26\%$ against $67.08\%$ and $63.00\%$ for FedBNN; under IID data, FedBNN is more accurate ($72.08\%$ versus $65.40\%$), and FedVote stays near $32\%$ in all three partitions.}
 
\revAdd{The larger margins under heterogeneous data on ResNet10 are consistent with where binarization enters each pipeline. In FL-BNN, clients train binary networks locally but upload only their binary parameters, and the server recovers real-valued parameters through maximum-likelihood estimation. FedVote resolves disagreement by voting over signs that each client produced independently of the others. In neither method is cross-client sign consistency an objective of training, so the binary structures learned at different clients need not agree, and aggregation must reconcile them after the fact. FedBNN instead anchors each client's binarization to the global model: it rotates the fused client--server weight before taking the sign, so each client binarizes relative to a shared representation rather than its own drifted weights. Client weights diverge most under heterogeneous partitions, which is precisely where averaging independently binarized models is least reliable. The advantage is not universal, however: with ConvNeXt-Tiny, FL-BNN is more accurate under both non-IID partitions.}
 
\subsubsection{Comparison with full-precision and communication-efficient methods}
\label{sec:fp_comparison}
 
\revAdd{Table~\ref{tab:performance} compares FedBNN on the same three settings with full-precision FedAvg~\citep{mcmahan2017communication} and with FedBAT~\citep{li2024fedbat} and FedMUD~\citep{li2025panaceas}, which reduce communication during training but deploy real-valued models. This comparison highlights a complementary trade-off: unlike FedBAT and FedMUD, which prioritize communication efficiency during training but retain real-valued models for inference, FedBNN does not explicitly reduce communication costs; instead, it prioritizes inference efficiency through binary model deployment, making it particularly suitable for resource-constrained edge devices where computational and memory budgets are critical.
On CIFAR10 (ResNet10), FedBNN reaches $89.95\%$ under IID data, $0.91$ percentage points below FedAvg and above FedBAT ($89.38\%$) and FedMUD ($88.74\%$). Under Non-IID~1, its $82.84\%$ is below FedAvg ($86.28\%$) and FedMUD ($84.22\%$) but $10.04$ points above FedBAT, and under Non-IID~2, FedBNN is the most accurate method, with $73.82\%$ against $70.62\%$ for FedAvg.}
 
\begin{table}[!ht]
    \caption{Comparison with full-precision FedAvg and the communication-efficient methods FedBAT and FedMUD for $N_c = 100$, on the settings of Table~\ref{tab:inference_comparison}; Table~\ref{tab:performance_appendix} gives FMNIST, SVHN, and FEMNIST. Accuracy is the test accuracy of the model selected on the validation set, evaluated in the precision in which it was trained: real-valued for FedAvg, FedBAT, and FedMUD, and binary for FedBNN. Binarized accuracy is the test accuracy after post-training binarization (Section~\ref{sec:post-training}); for FedBNN, which is binary by construction, the two are identical. FLOPs and memory are estimated for inference as described in Section~\ref{sec:runtime}. The best result in each column of each block is shown in bold.}
    \label{tab:performance}
    \centering
    % \resizebox{\columnwidth}{!}{%
    \begin{tabular}{|p{1.3cm}|p{1.8cm}|p{0.9cm}|p{0.9cm}|p{0.9cm}|p{1.55cm}|p{1.1cm}|p{0.9cm}|p{0.9cm}|p{0.9cm}|}
 
    \hline
       Method&Dataset&\multicolumn{3}{|c|}{Accuracy} &FLOPs&Memory&\multicolumn{3}{|c|}{Binarized Accuracy}\\
 
       \cline{3-5}\cline{8-10}
       &(Model)&IID & Non -IID 1 & Non -IID 2 & &(MB)&IID & Non -IID 1 & Non -IID 2   \\
 
              \hline
       FedAvg  &  &$\mathbf{90.86}$&$\mathbf{86.28}$&${70.62}$&$4.40\times10^8$&$19.6170$&$17.2$&$11.38$&$12.74$\\
       FedBAT   &CIFAR10&{$89.38$}&{$72.80$}&{$63.70$}&$4.40\times10^8$&$19.6170$&$13.62$&$10.94$&$10.26$ \\
       FedMUD  &(ResNet10)&$88.74$&$84.22$&$67.22$&$4.40\times10^8$&$19.6170$&$15.54$&$10.78$&$18.98$\\
        FedBNN   &&$89.95$&{$82.84$}&$\mathbf{73.82}$&$\mathbf{1.11\times10^7}$&$\mathbf{0.6130}$&$\mathbf{89.95}$&$\mathbf{82.84}$&$\mathbf{73.82}$ \\
        % FedBNN   & (HViT-tiny)&$ $&{$ $}&{$ $}&$\mathbf{\times10^x}$&$\mathbf{}$&$\mathbf{}$&$\mathbf{}$&$\mathbf{}$ \\
        % FedBNN (Ul-B)& &&&&&&&&&&&& \\
           \hline
 
       FedAvg  & \multirow{2}{*}{CIFAR10} & $65.22$ & $60.20$ & $61.84$ & $2.98\times10^9$ & $ 111.640$ & $18.04$ & $21.64$ & $16.56$ \\
FedBAT  & \multirow{2}{*}{(ConvNeXt} & $66.04$ & $64.22$ & $61.88$ & $2.98\times10^9$ & $111.640 $ & $14.90$ & $21.42$ & $16.54$ \\
FedMUD  & \multirow{2}{*}{-Tiny)} & $53.92$ & $54.30$ & $38.18$ & $2.98\times10^9$ & $111.640 $ & $40.46$ & $35.44$ & $29.50$ \\
FedBNN  &  & $\mathbf{72.08}$ & $\mathbf{67.08}$ & $\mathbf{63.00}$ & $\mathbf{6.07\times10^7}$ & $ \mathbf{3.4887}$ & $\mathbf{72.08}$ & $\mathbf{67.08}$ & $\mathbf{63.00}$ \\
        
       % \hline
 
       %        FedAvg  &  &${64.52}$&$\mathbf{63.42}$&$\mathbf{53.36}$&$1.11\times10^9$&$ 45.090$&$1.16 $&$0.88 $&$0.94 $\\
       % FedBAT   &CIFAR100&$42.14 $&$33.88$&$ 26.26$&$1.11\times10^9$&$ 45.090$&$1.10 $&$0.82 $&$1.10 $\\
       % FedMUD  &(ResNet18)& $\mathbf{65.14}$&$47.48$&$52.20 $&$1.11\times10^9$&$ 45.090$&$46.56 $&$22.54 $&$ 44.08$\\
       %  FedBNN   &&$60.04$&$51.14$&$42.92$&$\mathbf{2.26 \times10^7}$&$ \mathbf{1.41}$&$\mathbf{60.04} $&$\mathbf{51.14} $&$\mathbf{42.92} $\\
       %  % FedBNN (Ul-B)& &&&&&&&&&&&& \\
 
                      \hline
        
       FedAvg  &\multirow{2}{*}{Tiny-}& $\mathbf{55.00}$ & $\mathbf{52.62}$ & $\mathbf{54.54}$ & $4.44\times10^9$ & $45.090$ & $0.52$ &$ 0.52 $& $0.56$ \\
       FedBAT  &  \multirow{2}{*}{ImageNet}        & $27.30$ & $32.12$ & $20.90$ & $4.44\times10^9$ & $45.090$ & $0.60$ & $0.48 $&$ 0.80$ \\
       FedMUD  &   \multirow{2}{*}{(ResNet18)}               & $47.20 $& $44.16$ & $46.06$ & $4.44\times10^9$ & $45.090$ & $16.24$ & $12.80$ & $15.60$ \\
        FedBNN  &  &$46.54$ &$43.00$ & $45.74$ & $\mathbf{9.05\times10^7}$ & $\mathbf{1.41}$ & $\mathbf{46.54}$ & $\mathbf{43.00} $& $\mathbf{45.74}$ \\
        \hline

    \end{tabular}%
    % }
 
\end{table}
 
\revAdd{With ConvNeXt-Tiny, FedBNN is the most accurate method under all three partitions, with $72.08\%$, $67.08\%$, and $63.00\%$, that is, $6.86$, $6.88$, and $1.16$ percentage points above FedAvg, while FedMUD falls to $38.18\%$ under Non-IID~2. This suggests that binarization acts as a regularizer for this high-capacity model ($28$M parameters): under IID data, the validation loss of FedAvg starts to rise after about $100$ communication rounds, indicating overfitting, whereas that of FedBNN stays lower for the rest of training (Figure~\ref{fig:cifar10_convnext_fedavg_fedbnn}(b)). Validation curves for all datasets, architectures, and partitions are given in Appendix~\ref{convergence}, and the evolution of the learned mixing coefficients in Appendix~\ref{Alpha, beta, lambda}.}
 
\revAdd{On TinyImageNet (ResNet18), FedBNN reaches $46.54\%$, $43.00\%$, and $45.74\%$. This is $8.5$ to $9.6$ percentage points below FedAvg and at most $1.2$ points below FedMUD, but $10.9$ to $24.8$ points above FedBAT, which falls to $20.90\%$ under Non-IID~2. Across the nine settings of Table~\ref{tab:performance}, FedBNN is more accurate than FedAvg in four, than FedBAT in all nine, and than FedMUD in five; over all $18$ settings, including those in Appendix~\ref{sec:additional_results}, the counts are $4$, $16$, and $11$. FedBNN thus gives up some accuracy relative to FedAvg in most settings, at most $9.6$ percentage points, in exchange for a binary model that is far cheaper at inference (Section~\ref{sec:inference_savings}).}

\subsubsection{Post-training binarization performance comparison}
\label{sec:post-training}
 
\revAdd{A real-valued model can also be made cheap at inference by binarizing it after training. We binarize the trained models of FedAvg, FedBAT, and FedMUD by applying a sign function to the weights and activations with layer-wise norm scaling, i.e., $\mathbf{w}_b = \frac{\lVert \mathbf{w} \rVert_2\text{sign}(\mathbf{w})}{ \lVert  \text{sign}(\mathbf{w}) \rVert_2}$, so that all methods are compared at similar FLOPs and memory. The binarized-accuracy columns of Table~\ref{tab:performance} show the result: the baselines fall to between $10.26\%$ and $18.98\%$ on CIFAR10 with ResNet10 and to between $14.90\%$ and $40.46\%$ with ConvNeXt-Tiny, and on TinyImageNet, FedAvg and FedBAT fall below $1\%$ and FedMUD to between $12.80\%$ and $16.24\%$. Because FedBNN is trained as a binary network, its binarized accuracy equals its accuracy, and it exceeds the best binarized baseline by $30.1$ to $72.8$ percentage points in these nine settings. FMNIST, SVHN, and FEMNIST show the same pattern (Appendix~\ref{sec:additional_results}). } 
\revAdd{Compared with FL-BNN and FedVote, FedBNN achieves consistently higher binarized accuracy across all three settings on CIFAR10 with ResNet10 and TinyImageNet with ResNet18. On CIFAR10 with ConvNeXt-Tiny, FedBNN also achieves the highest binarized accuracy under IID data. These results highlight that FedBNN's advantage is particularly pronounced when compared with methods that do not consistently preserve accuracy across different data distributions and model architectures.}
 
\subsubsection{Computation and memory savings during inference}
\label{sec:inference_savings}
 
\revAdd{FedAvg, FedBAT, and FedMUD deploy real-valued models, so their inference cost is that of the full-precision network; compressing the updates exchanged during training does not reduce it. FedBNN's binary model replaces the multiply--accumulate operations of the binarized layers with XNOR and bit-count operations and stores each of their weights in $1$ bit (Section~\ref{sec:runtime}). On CIFAR10, FedBNN reduces the FLOPs of ResNet10 from $4.40\times10^8$ to $1.11\times10^7$ ($39.6\times$) and those of ConvNeXt-Tiny from $2.98\times10^9$ to $6.07\times10^7$ ($49.1\times$); on TinyImageNet, it reduces the FLOPs of ResNet18 from $4.44\times10^9$ to $9.05\times10^7$ ($49.1\times$). Memory shrinks by $32\times$ in each case, from $19.62$ to $0.61$~MB, from $111.64$ to $3.49$~MB, and from $45.09$ to $1.41$~MB, respectively. Appendix~\ref{sec:additional_results} reports the same quantities for FMNIST, SVHN, and FEMNIST, and Appendix~\ref{budget} compares FedBNN with real-valued networks scaled down to the same inference budget.}
 
\subsubsection{SVD complexity and calls during training}
\revAdd{For each binarized layer, every alternating iteration (Eqs.~(11)--(13)) forms an $n_1 \times n_1$ and an $n_2 \times n_2$ matrix and computing their SVDs has a cost of $\mathcal{O}(n_1^3 + n_2^3)$. With $N_{eR} = 3$ iterations at the start of every local epoch, this gives 6 SVD calls per binarized layer per local epoch. Summing over all binarized layers and local epochs in one client round gives the call counts in Table~\ref{tab:svd_time} (CNN4: 4 layers; ResNet10: 11; ResNet18: 19; ConvNeXt-Tiny: 57).}
 
\begin{table}[t]
\centering
\caption{{Per-client SVD overhead per federated round (IID, 10 clients).
Values are mean \(\pm\) std of CUDA-synchronized \texttt{torch.svd} wall time on an NVIDIA RTX PRO~6000 Blackwell Server Edition. Call counts are deterministic for a given architecture and local schedule.}}
\label{tab:svd_time}
 
% \rowcolors{1}{blue!10}{white} % alternating row colors
 
\begin{tabular}{|l|l|r|r|r|}
\hline
% \toprule
% \rowcolor{blue!25} % header color
\textbf{Dataset} & \textbf{Model} & \textbf{SVD calls} & \textbf{SVD time (s)} & \textbf{Train time (s)} \\
\hline
FMNIST & CNN4 & $120$ & \(0.495 \pm 0.014\) & $10$ \\
SVHN & CNN4 & $240$ & \(0.990 \pm 0.025\) & $51$ \\
CIFAR10 & ResNet10 & $990$ & \(7.488 \pm 0.766\) & $67$ \\
CIFAR10 & ConvNeXt-Tiny & $5130$ & \(25.608 \pm 1.444\) & $200$ \\
TinyImageNet & ResNet18 & $1710$ & \(33.504 \pm 2.324\) & $164$ \\
\hline
% \bottomrule
\end{tabular}
\end{table}
 
\revAdd{These timings measure the rotation and SVD overhead within each client's local training round, excluding communication. For CNN4, the overhead is about \(0.5\) to \(1\)\,s per round; it increases with depth and width, consistent with the $\mathcal{O}(n_1^3 + n_2^3)$ cost of each SVD and the larger number of SVD calls. This overhead is incurred only during training: the deployed model consists of binary weights alone, so no rotation or SVD is computed at inference.}

\subsubsection{Ablation Study: Importance of server alignment}
\label{sec:ablation_server}
 
\begin{table*}[!ht]
\caption{Ablation study for $N_c = 100$ comparing with and without server alignment.}
\label{tab:ablationserver}
\centering
\begin{tabular}{|l|l|c|c|c|}
\hline
Method & Dataset (Model) & IID & Non-IID 1 & Non-IID 2 \\
\hline
FedBNN & FMNIST & $88.46$ & $\mathbf{87.76}$ & $\mathbf{83.38}$ \\
FedBNN ($\beta = 1$, $\lambda = 1$) & (CNN4) & $\mathbf{88.54}$ & $87.68$ & $82.48$ \\
\hline
FedBNN & SVHN & $\mathbf{86.89}$ & $\mathbf{85.63}$ & $\mathbf{84.40}$ \\
FedBNN ($\beta = 1$, $\lambda = 1$) & (CNN4) & $86.77$ & $85.38$ & $84.13$ \\
\hline
FedBNN & CIFAR10 & $\mathbf{89.95}$ & $\mathbf{82.84}$ & $\mathbf{73.82}$ \\
FedBNN ($\beta = 1$, $\lambda = 1$) & (ResNet10) & $89.34$ & $82.82$ & $68.54$ \\
\hline
FedBNN & TinyImageNet & $\mathbf{46.54}$ & $\mathbf{43.00}$ & $\mathbf{45.74}$ \\
FedBNN ($\beta = 1$, $\lambda = 1$) & (ResNet18) & $42.86$ & $40.62$ & $43.90$ \\
\hline
FedBNN & FEMNIST & $\mathbf{79.73}$ & $\mathbf{80.31}$ & $79.13$ \\
FedBNN ($\beta = 1$, $\lambda = 1$) & (ResNet18) & $79.10$ & $79.61$ & $\mathbf{80.68}$ \\
\hline
FedBNN & CIFAR10 & $\mathbf{72.08}$ & $\mathbf{67.08}$ & $\mathbf{63.00}$ \\
FedBNN ($\beta = 1$, $\lambda = 1$) & (ConvNeXt-Tiny) & $70.78$ & $61.78$ & $58.50$ \\
\hline
\end{tabular}
\end{table*}
 
\revAdd{To evaluate the role of server-side alignment, we set $\beta = 1$ and $\lambda = 1$ both on the clients (Eqs. (\ref{eq:fused}) and (\ref{eq:fed_weightupdate})) and on the server (Eqs. (\ref{eq:client_fused_srv})--(\ref{eq:server_agg})). This removes every server anchor: clients rotate and binarize only their own weights, and the evaluation weight in Eq. (\ref{eq:server_agg}) becomes
\begin{equation}
\tilde{\mathbf{w}}_{t+1} = \mathbf{w}_{t+1} + \alpha \left( \mathbf{w}_{R,t+1} - \mathbf{w}_{t+1} \right),
\qquad
\mathbf{w}_{R,t+1} = \sum_{k \in S_t} p^k\, (\mathbf{R}^k_{t+1})^{\top} \mathbf{w}^k_{t+1}.
\end{equation}}
 
\revAdd{Table~\ref{tab:ablationserver} shows that server alignment improves accuracy in 16 of the 18 settings. The largest gains are $5.28$ percentage points on CIFAR10 (ResNet10) under Non-IID~2, $5.30$ and $4.50$ points on CIFAR10 (ConvNeXt-Tiny) under Non-IID~1 and Non-IID~2, and $1.84$ to $3.68$ points on TinyImageNet. On FMNIST, SVHN, and FEMNIST, the differences are at most $1.55$ points in either direction, and the two exceptions occur there: FMNIST under IID data ($88.46\%$ versus $88.54\%$) and FEMNIST under Non-IID~2 ($79.13\%$ versus $80.68\%$). Appendix~\ref{sec:ablation_components} disables fused weighting and server alignment separately on CIFAR10 and TinyImageNet, and Appendix~\ref{aggregatingwtilda} compares FedBNN's evaluation weight with averaging the clients' adjustable rotated weights $\tilde{\mathbf{w}}^k$ directly.}
 
% ======================== END BLOCK 1 ========================
 
\section{Conclusion}

\revAdd{We propose FedBNN, a rotation-aware binary neural network framework for federated learning that reduces inference FLOPs by up to $58\times$ and memory by $32\times$ relative to real-valued models. FedBNN addresses two key challenges of binarization in federated learning, namely binary representations that become inconsistent across heterogeneous clients and server aggregation that undoes the alignment between real-valued weights and their binary representations, through learnable client--server weight fusion before rotation, adaptive rotation-aware adjustment toward the global model, and rotation-informed server aggregation. Against binary federated methods, FedBNN is more accurate than FL-BNN and FedVote in seven of nine settings; against real-valued baselines, it stays within $10$ percentage points of full-precision FedAvg, surpasses it in four of $18$ settings, is more accurate than FedBAT and FedMUD in $16$ and $11$ settings, respectively, and exceeds the best of their post-training binarized versions by $17$ to $78$ percentage points in every setting. Future work includes a formal convergence analysis of federated training with learned rotations, deployment on hardware with native bitwise support, such as FPGAs and ASICs, to measure wall-clock gains, and extensions of the fused-weight rotation to low-bit quantization and other aggregation strategies.}

\bibliography{main}
\bibliographystyle{tmlr}

\clearpage

\appendix
\section{Appendix}

% ======================= BEGIN BLOCK 2 =======================
\subsection{Results on FMNIST, SVHN, and FEMNIST}
\label{sec:additional_results}
 
\revAdd{Table~\ref{tab:performance_appendix} completes Table~\ref{tab:performance} with the remaining three settings: FMNIST and SVHN with CNN4, and FEMNIST with ResNet18.}
 
\begin{table}[!ht]
    \caption{Comparison with FedAvg, FedBAT, and FedMUD on FMNIST, SVHN, and FEMNIST for $N_c = 100$. The columns are as in Table~\ref{tab:performance}.}
    \label{tab:performance_appendix}
    \centering
    % \resizebox{\columnwidth}{!}{%
    \begin{tabular}{|p{1.3cm}|p{1.8cm}|p{0.9cm}|p{0.9cm}|p{0.9cm}|p{1.55cm}|p{1.1cm}|p{0.9cm}|p{0.9cm}|p{0.9cm}|}
 
    \hline
       Method&Dataset&\multicolumn{3}{|c|}{Accuracy} &FLOPs&Memory&\multicolumn{3}{|c|}{Binarized Accuracy}\\
 
       \cline{3-5}\cline{8-10}
       &(Model)&IID & Non -IID 1 & Non -IID 2 & &(MB)&IID & Non -IID 1 & Non -IID 2   \\
 
              \hline
       FedAvg  &  &$\mathbf{92.24}$&$\mathbf{91.44}$&$\mathbf{89.28}$&$2.02\times10^7$&$1.5635$&$53.42$&$63.68$&$54.72$\\
       FedBAT   &FMNIST&$89.12$&$87.66$&$85.56$&$2.02\times10^7$&$1.5635$&$14.34$&$16.98$&$8.00$ \\
       FedMUD  &(CNN4)& $89.60$&$88.60$&$86.00$&$2.02\times10^7$&$1.6291$&$63.20$&$66.5$&$66.08$\\
        FedBNN   &&$ 88.46 $&$87.76$&$83.38$&$\mathbf{3.48\times10^5}$&$\mathbf{0.0489}$&$\mathbf{88.46}$&$\mathbf{87.76}$&$\mathbf{83.38}$ \\
        % FedBNN (Ul-B)& &&&&&&&&&&&& \\
 
                      \hline
       FedAvg  &  &$\mathbf{92.10}$&$\mathbf{90.60}$&$\mathbf{89.34}$&$3.00\times10^7$&$1.5965$&$28.01$&$22.56$&$16.92$\\
       FedBAT   &SVHN&$86.01$&$80.83$&$75.78$&$3.00\times10^7$&$1.5965$&$50.35$&$26.69$&$34.71$\\
       FedMUD  &(CNN4)&$86.31$&$84.38$&$83.14$&$3.00\times10^7$&$1.6127$&$69.92$&$50.87$&$51.19$\\
        FedBNN   &&$86.89$&$85.63$&$84.40$&$\mathbf{5.19\times10^5}$&$\mathbf{0.0498}$&$\mathbf{86.89}$&$\mathbf{85.63}$&$\mathbf{84.40}$ \\
        % FedBNN (Ul-B)& &&&&&&&&&&&& \\
 
                      \hline
        
       FedAvg  &         & $\mathbf{80.24}$ & $\mathbf{81.12}$ & $\mathbf{80.32}$ & $9.13\times10^8$ & $45.090$ & $2.08$ & $1.66$ &$ 1.62$ \\
       FedBAT  &  FEMNIST    & $76.44$ & $74.31$ & $78.41$ & $9.13\times10^8$ & $45.090$ & $0.38$ &$ 2.08$ &$ 2.40$ \\
       FedMUD  &   (ResNet18)          & $78.79$ & $80.11$ & $76.68$ & $9.13\times10^8$ & $45.090$ & $25.74$ & $0.76$ & $0.80$ \\
       FedBNN  &  & $79.73 $ & $80.31$ & $79.13$ & $\mathbf{1.84\times10^7}$ & $\mathbf{1.41}$ & $\mathbf{79.73}$ & $\mathbf{80.31}$ & $\mathbf{79.13}$ \\

        % FedBNN   & (HViT-tiny)&$ $&{$ $}&{$ $}&$\mathbf{\times10^x}$&$\mathbf{}$&$\mathbf{}$&$\mathbf{}$&$\mathbf{}$ \\
        % FedBNN (Ul-B)& &&&&&&&&&&&& \\

       \hline

    \end{tabular}%
    % }
 
\end{table}
 
\revAdd{On FMNIST, FedBNN reaches $88.46\%$, $87.76\%$, and $83.38\%$ under IID, Non-IID~1, and Non-IID~2, which is $3.68$ to $5.90$ percentage points below FedAvg and $0.84$ to $2.62$ points below FedMUD; it is $0.10$ points above FedBAT under Non-IID~1 and below it under the other two partitions. On SVHN, FedBNN reaches $86.89\%$, $85.63\%$, and $84.40\%$, which is $4.94$ to $5.21$ points below FedAvg but above FedBAT (by $0.88$, $4.80$, and $8.62$ points) and FedMUD (by $0.58$, $1.25$, and $1.26$ points) in every partition. On FEMNIST, FedBNN reaches $79.73\%$, $80.31\%$, and $79.13\%$, within $1.2$ points of FedAvg and above FedBAT and FedMUD in every partition.}
 
\revAdd{Post-training binarization costs the baselines between $16.4$ and $79.5$ percentage points in these settings; the most accurate binarized baseline reaches $69.92\%$ (FedMUD on SVHN under IID data), and on FEMNIST none exceeds $25.74\%$. FedBNN loses nothing and exceeds the best binarized baseline by $17.0$ to $78.2$ points. It reduces the FLOPs of CNN4 by about $58\times$, from $2.02\times10^7$ to $3.48\times10^5$ on FMNIST and from $3.00\times10^7$ to $5.19\times10^5$ on SVHN, and those of ResNet18 on FEMNIST by $49.6\times$; memory shrinks by $32\times$ in all three settings.}

\subsection{Aggregating \(\tilde{\mathbf{w}}\)} \label{aggregatingwtilda}

% \section{Alternative Server-Side Update via Client-Specific Auxiliary Models}

{
In this section, we describe an alternative server-side aggregation formulation for model evaluation. 
Instead of first aggregating rotation-aligned weights and then forming the global model, the server constructs the adjustable rotated weight vector ($\tilde{\mathbf{w}}^k$) for each client and subsequently aggregates them. After local training, each client $k$ sends its updated real-valued weights $\mathbf{w}^k$ together with the learned rotation matrices $\mathbf{R}^k$ to the server. 
Rather than aggregating the rotated weights in advance, the server computes $\tilde{\mathbf{w}}^k$ for each client directly from the local weights and their corresponding rotation-aligned representations.
}

\revAddd{\begin{equation}
\tilde{\mathbf{w}}_{t+1}^k =
\mathbf{w}^k_{f,t+1}
+ \alpha^k \beta^k
\left(
(\mathbf{R}_{t+1}^k)^\top \mathbf{w}^k_{f,t+1}
-
\mathbf{w}^k_{f,t+1}
\right)
+ \alpha^k (1-\beta^k)
\left(
\mathbf{w}_{t}
-
\mathbf{w}^k_{f,t+1}
\right)
\end{equation}
where $\mathbf{w}^k_{f,t+1}$ is given by Eq. (\ref{eq:client_fused_srv}), and $\alpha^k = |\sin\theta^k|$ and $\beta^k = |\sin\gamma^k|$ are client $k$'s own coefficients.}

{
Each $\tilde{\mathbf{w}}_{t+1}^k$ incorporates the client-specific rotation-aligned weights while preserving the influence of the client's local update. 
These auxiliary models are then aggregated using the same weighting factors as in FedAvg:
}

\revAddd{\begin{equation}
\tilde{\mathbf{w}}_{t+1} = \sum_{k \in S_t} p^k\, \tilde{\mathbf{w}}_{t+1}^k 
\end{equation}}

{
The aggregated auxiliary model $\tilde{\mathbf{w}}$ is used only for server-side evaluation and model selection and is never transmitted to clients. 
The model broadcast to clients for the next communication round remains the standard FedAvg model obtained by aggregating the real-valued weights.
}

\begin{table*}[htbp]
    \caption{Ablation study under different aggregation strategies for $N_c = 100$.}
    \label{tab:ablation2}
    \centering
    \begin{tabular}{|c|c|c|c|c|}
    \hline
    Method  & Dataset & \multicolumn{3}{|c|}{Accuracy}  \\
    \cline{3-5}
     & (Model) & IID & Non-IID 1 & Non-IID 2  \\
    \hline
    
    FedBNN &\multirow{1}{*}{FMNIST}& $\mathbf{88.46}$ & $\mathbf{87.76}$ & $\mathbf{83.38}$  \\
    % Strategy 2 & \multirow{2}{*}{(CNN4)} & $19.24$ & $15.18$ & $18.80$ & $19.24$ & $15.18$ & $18.80$ \\
    Aggregating $\tilde{\mathbf{w}}$ & \multirow{1}{*}{(CNN4)} & $87.88$ & $87.24$ & $80.40$  \\
    \hline
    
    FedBNN  &\multirow{1}{*}{SVHN}& $\mathbf{86.89}$ & $\mathbf{85.63}$ & $\mathbf{84.40}$  \\
    % Strategy 2 & \multirow{2}{*}{(CNN4)} & $19.70$ & $19.67$ & $19.65$ & $19.70$ & $19.67$ & $19.65$ \\
    Aggregating $\tilde{\mathbf{w}}$ & \multirow{1}{*}{(CNN4)} & $86.25$ & $84.74$ & $80.40$  \\
    \hline
    
    FedBNN & \multirow{1}{*}{CIFAR10} & $\mathbf{89.95}$ & $\mathbf{82.84}$ & $\mathbf{73.82}$  \\
    % Strategy 2 & \multirow{2}{*}{(ResNet10)} & $11.34$ & $12.94$ & $12.76$ & $11.34$ & $12.94$ & $12.76$ \\
    Aggregating $\tilde{\mathbf{w}}$ & \multirow{1}{*}{(ResNet10)} & $89.82$ & $82.22$ & $55.62$ \\
    \hline
    
    % Strategy 1 & \multirow{2}{*}{CIFAR100} & $\mathbf{60.04}$ & $\mathbf{51.14}$ & $\mathbf{42.92}$  \\
    % % Strategy 2 & \multirow{2}{*}{(ResNet18)} & $0.88$ & $0.90$ & $1.36$ & $0.88$ & $0.90$ & $1.36$ \\
    % Strategy 2 &  & $59.10$ & $51.04$ & $36.38$  \\
    % \hline
    
    FedBNN & TinyImageNet & $\mathbf{46.54}$ & $\mathbf{43.00}$ & $\mathbf{45.74}$  \\
    % Strategy 2 & ImageNet & $0.46$ & $0.52$ & $0.58$ & $0.46$ & $0.52$ & $0.58$ \\
    Aggregating $\tilde{\mathbf{w}}$ & (ResNet18) & $46.36$ & $42.46$ & $43.84$  \\
    \hline
    
    FedBNN & \multirow{1}{*}{FEMNIST} & $79.73$ & $80.31$ & $\mathbf{79.13}$  \\
    % Strategy 2 & \multirow{2}{*}{(ResNet18)} & $4.98$ & $5.00$ & $5.26$ & $4.98$ & $5.00$ & $5.26$ \\
    Aggregating $\tilde{\mathbf{w}}$ & \multirow{1}{*}{(ResNet18)} & $\mathbf{80.56}$ & $\mathbf{80.76}$ & $76.41$  \\
    \hline

    FedBNN & \multirow{1}{*}{CIFAR10} & $72.08$ & $\mathbf{67.08}$ & $\mathbf{63.00}$  \\
    % Strategy 2 & \multirow{2}{*}{(Convnext-Tiny)} & $12.56$ & $13.08$ &  & $12.56$ & $13.08$ & \\
    Aggregating $\tilde{\mathbf{w}}$ & \multirow{1}{*}{(ConvNeXt-Tiny)} & $\mathbf{72.62}$ & $66.60$ & $62.70$ \\
    \hline
    \end{tabular}
\end{table*}

% \subsection{Comparison with Alternative Aggregation Strategy}

\revAddd{Table \ref{tab:ablation2} summarizes performance across multiple datasets, models, and data-heterogeneity settings. Overall, aggregating the rotation-aligned weights before forming  $\tilde{\mathbf{w}}^k$ yields better performance across most settings. For FMNIST with CNN4, the proposed method achieves higher accuracy across all data distributions, obtaining 88.46\%, 87.76\%, and 83.38\% in the IID, Non-IID 1, and Non-IID 2 settings, respectively, compared to 87.88\%, 87.24\%, and 80.40\% for the alternative aggregation. Similarly, on SVHN with CNN4, FedBNN improves accuracy across all settings, reaching 86.89\%, 85.63\%, and 84.40\%, outperforming the alternative approach, which achieves 86.25\%, 84.74\%, and 80.40\%.}

\revAddd{The benefits are particularly pronounced on more challenging datasets. For CIFAR10 with ResNet10 under the most heterogeneous Non-IID 2 setting, the proposed aggregation achieves 73.82\% accuracy, compared to only 55.62\% with $\tilde{\mathbf{w}}^k$, indicating significantly improved robustness to client heterogeneity. On TinyImageNet with ResNet18, the proposed method also provides consistent gains, achieving 46.54\%, 43.00\%, and 45.74\% across the three distributions, compared to 46.36\%, 42.46\%, and 43.84\% for the alternative strategy. In the FEMNIST dataset with ResNet18, the proposed aggregation achieves superior performance under the Non-IID 2 setting. Finally, for CIFAR10 with ConvNeXt-Tiny, the differences are small under all settings, but the proposed aggregation outperforms under heterogeneous data distributions. These results indicate that performing aggregation in the rotation-aligned space before constructing the auxiliary update better preserves the binarization structure across clients. In contrast, aggregating the auxiliary models $\tilde{\mathbf{w}}^k$ introduces averaging after computing $\tilde{\mathbf{w}}^k$, potentially diluting the benefits of rotation. }

\revAddd{We observe a slight drop for FedBNN relative to aggregating $\tilde{\mathbf{w}}^k$ in three settings: FEMNIST under IID and Non-IID~1, and CIFAR10 (ConvNeXt-Tiny) under IID, in each case by at most $0.83$ percentage points. Both strategies average real-valued vectors and binarize only afterward; they differ in whether each client's own coefficients $\alpha^k, \beta^k$ are applied before averaging (aggregating $\tilde{\mathbf{w}}^k$) or the averaged coefficients $\alpha, \beta$ are applied once to the averaged quantities (FedBNN, Eq. (\ref{eq:server_agg})). When client data are similar, the clients' weights and coefficients are similar as well, so the two orders yield nearly the same model, and small differences in either direction are expected. As heterogeneity increases, the client-specific corrections diverge: under Non-IID~2, FedBNN is better in all six cases, most clearly on CIFAR10 (ResNet10), with $73.82\%$ versus $55.62\%$.}

% \subsection{Ablation Study: Importance of server alignment, additional results}

% For the ConvNeXt-Tiny architecture considered in Section \ref{sec:ablation_server}, Table \ref{tab:ablationserver_contd} further highlights the importance of server alignment. The full FedBNN framework achieves $72.08\%$, $67.08\%$, and $63.00\%$ accuracy under IID, Non-IID 1, and Non-IID 2 settings, respectively. When server alignment is removed by setting $\beta = 1$ and $\lambda = 1$, the performance drops to $70.78\%$, $61.78\%$, and $58.50\%$. The larger degradation under Non-IID distributions indicates that server alignment is particularly important for maintaining consistency across heterogeneous clients.

% \begin{table*}[htbp]
% \caption{Ablation study for $N_c = 100$ comparing with and without server aggregation.}
% \label{tab:ablationserver_contd}
%     \centering
%     \begin{tabular}{|l|l|c|c|c|}
%     \hline
%     Method  & Dataset & \multicolumn{3}{|c|}{Accuracy}  \\
%     \cline{3-5}
%      & (Model) & IID & Non-IID 1 & Non-IID 2  \\
%     \hline
% FedBNN & CIFAR10 & $\mathbf{72.08}$ & $\mathbf{67.08}$ & $\mathbf{63.00}$ \\
% FedBNN ($\beta = 1$, $\lambda = 1$) & (ConvNeXt-Tiny) & $70.78$ & $61.78$ & $58.50$ \\
% \hline
% \end{tabular}
% \end{table*}

\subsection{Comparison with Real-Valued Networks of Similar Complexity} \label{budget}

Table~\ref{tab:less_complex_real} compares FedBNN with real-valued ResNet models whose architecture is reduced such that both the training and inference complexity approximately match the FLOPs and memory footprint of FedBNN at inference. This differs from the experiment presented in Section \ref{sec:post-training} of the main paper, where full-sized real-valued models are trained, and only the inference complexity is reduced through post-training binarization.

\begin{table*}[htbp]
    \caption{Performance comparison for $N_c = 100$. The FLOPs and memory values are calculated during runtime.}
    \label{tab:less_complex_real}
    \centering
    % \resizebox{\columnwidth}{!}{%
    \begin{tabular}{|p{1.25cm}|p{2.2cm}|p{4.5cm}|p{0.85cm}|p{0.85cm}|p{0.85cm}|p{1.5cm}|p{1.2cm}|}

    \hline
       Method&Dataset &Model&\multicolumn{3}{|c|}{Accuracy} &FLOPs& Memory \\

       \cline{4-6}
       &&&IID & Non-IID 1 & Non-IID 2 && (MB)  \\

              \hline
        FedBNN        &  &ResNet10&${89.95}$ & ${82.84}$ & $\mathbf{73.82}$&$1.11\times10^7$&$ 0.61$\\
        % FedBNN        &  &ResNet10 (less filters)&${67.04}$&${56.72}$&${52.54}$&$7.36\times10^5$&$ 0.015$\\
        FedAvg  &  \multirow{2}{*}{CIFAR10}&ResNet10&$\mathbf{90.86}$&$\mathbf{86.28}$&${70.62}$&$4.40\times10^8$&$19.62$\\
        FedAvg        &  &ResNet10 (FLOPs matched)&${84.74}$&${81.06}$&${66.16}$&$1.12\times10^7$&${0.49}$\\
        FedAvg        &  &ResNet10 (Memory matched) &${86.60}$&${81.50}$&${68.78}$&$1.35 \times10^7$&${0.59}$\\
        \hline 
        FedBNN        &  &ResNet18& ${46.54}$ & ${43.00}$ & ${45.74}$ &$9.05\times10^7$&$1.41 $\\
        FedAvg  &  \multirow{2}{*}{TinyImageNet}&ResNet18& $\mathbf{55.00}$ & $\mathbf{52.62}$ & $\mathbf{54.54}$ &$4.44\times10^9$&$45.09$ \\
        FedAvg        &   &ResNet18 (FLOPs matched)&${41.06}$&${37.72}$&${35.66}$&$8.99 \times10^7$&${0.95}$\\
        FedAvg        &  &ResNet18 (Memory matched) &${43.16}$&${40.54}$&${39.24}$&$1.33 \times10^8$&${1.40}$\\

       \hline
    \end{tabular}%
    % }

\end{table*}

When real-valued networks are scaled down (by reducing the number of output filters) to match the computational and memory constraints of FedBNN, their performance drops significantly. On CIFAR10, FedBNN with ResNet10 achieves $73.82\%$ accuracy in the most challenging Non-IID~2 setting, outperforming the FLOPs-matched and memory-matched real-valued models by $7.66\%$ and $5.04\%$, respectively. Similar trends are observed for IID and Non-IID~1 settings, where FedBNN maintains competitive or superior accuracy while operating at comparable computational and memory budgets.

A similar pattern is observed on TinyImageNet with ResNet18. Under Non-IID~2, FedBNN achieves $45.74\%$ accuracy, which is substantially higher than the FLOPs-matched and memory-matched real-valued models that achieve $35.66\%$ and $39.24\%$, respectively. Even under IID and moderately non-IID settings, FedBNN consistently performs better when the model complexity is constrained.

These results highlight that when the entire model capacity must be reduced to satisfy strict computational and memory limits, training directly with binarized networks is significantly more effective than shrinking real-valued architectures to the same complexity. Combined with the results in Section~\ref{sec:post-training} that study post-training binarization, this analysis demonstrates that FedBNN provides strong performance across both practical deployment scenarios.

\subsection{Isolating fused weighting and server alignment}
\label{sec:ablation_components}

\revAddd{Section~\ref{sec:ablation_server} disables fused weighting and server alignment together. To separate the effects here, we disable each one on its own: $\lambda = 1$ removes the fused weight (Eq.~(\ref{eq:fused}) on the clients and Eqs.~(\ref{eq:client_fused_srv})--(\ref{eq:global_fused}) on the server), and $\beta = 1$ removes the server-alignment term $\alpha(1-\beta)(\cdot)$ of Eqs.~(\ref{eq:fed_weightupdate}) and~(\ref{eq:server_agg}). Table~\ref{tab:ablation_components} reports both variants next to the full method and the joint ablation of Table~\ref{tab:ablationserver}.}

\begin{table*}[htbp]
\caption{\revAddd{Accuracy (\%) with fused weighting ($\lambda = 1$) or server alignment ($\beta = 1$) disabled separately, for $N_c = 100$. The last row of each block repeats the joint ablation of Table~\ref{tab:ablationserver}. The best result in each column of each block is shown in bold.}}
\label{tab:ablation_components}
\centering
\begin{tabular}{|l|l|c|c|c|}
\hline
Method & Dataset (Model) & IID & Non-IID 1 & Non-IID 2 \\
\hline
FedBNN & & $\mathbf{89.95}$ & $82.84$ & $\mathbf{73.82}$ \\
FedBNN ($\lambda = 1$) & CIFAR10 & $89.02$ & $\mathbf{84.12}$ & $70.12$ \\
FedBNN ($\beta = 1$) & (ResNet10) & $88.44$ & $83.36$ & $70.34$ \\
FedBNN ($\beta = 1$, $\lambda = 1$) & & $89.34$ & $82.82$ & $68.54$ \\
\hline
FedBNN & & $\mathbf{46.54}$ & $43.00$ & $\mathbf{45.74}$ \\
FedBNN ($\lambda = 1$) & TinyImageNet & $45.24$ & $41.70$ & $43.36$ \\
FedBNN ($\beta = 1$) & (ResNet18) & $44.90$ & $\mathbf{43.24}$ & $45.44$ \\
FedBNN ($\beta = 1$, $\lambda = 1$) & & $42.86$ & $40.62$ & $43.90$ \\
\hline
\end{tabular}
\end{table*}

\revAddd{The full method is more accurate than either single-component variant in 9 of the 12 comparisons. The largest gap is on CIFAR10 under Non-IID~2, where disabling either component costs $3.5\%$ to $3.7\%$ ($73.82\%$ versus $70.12\%$ with $\lambda = 1$ and $70.34\%$ with $\beta = 1$). On TinyImageNet, disabling fused weighting costs $1.3\%$ to $2.4\%$ in every setting, whereas disabling server alignment $(\beta = 1)$ matters mainly under IID data (1.6 \%). The exceptions are CIFAR10 under Non-IID~1, where both single-component variants are slightly more accurate ($84.12\%$ and $83.36\%$ versus $82.84\%$), and TinyImageNet under Non-IID~1 with $\beta = 1$ ($43.24\%$ versus $43.00\%$). Disabling both components together gives the lowest accuracy in four of the six settings, for example, $68.54\%$ on CIFAR10 under Non-IID~2.}

\subsection{Communication overhead for transmitting R1 and R2 from client to the server} \label{table8}

\begin{table*}[htbp]
\centering
\caption{\revAddd{Parameter overhead of the rotation matrices $\mathbf{R}_1$ and $\mathbf{R}_2$ in FedBNN, relative to the weights $\mathbf{w}$ of the binarized layers. The overhead is about $1\%$ for CNN4 and at most $0.56\%$ for the larger architectures, and it decreases with model size.}}
\label{tab:rotation_overhead}
\begin{tabular}{|l|l|l|p{2cm}|p{2cm}|p{2cm}|}
\hline
Method & Dataset & Model & No of parameters for $\mathbf{w}$  &  No of parameters for $\mathbf{R}_1, \mathbf{R}_2$ & Overhead Percentage\\
\hline
\multirow{6}{*}{FedBNN}&FMNIST &CNN4&$387,360$&$4,070$&$1.05\%$ \\
&SVHN           &CNN4&$387,936$&$4,142$&$1.07\%$ \\
&CIFAR10        &ResNet10&$4,890,624$&$27,552 $&
$0.56\%$ \\
&FEMNIST        &ResNet18&$11,157,504$&$62,272$&$0.56\%$ \\
&TinyImageNet   &ResNet18&$11,157,504$&$62,272$&$0.56\%$ \\
&CIFAR10        &ConvNeXt-Tiny&$27,751,392$&$78,084
$&$0.28\%$ \\
\hline
\end{tabular}
\end{table*}

Table~\ref{tab:rotation_overhead} reports the parameter overhead introduced by the rotation matrices $\mathbf{R}_1$ and $\mathbf{R}_2$ used in FedBNN across different datasets and model architectures. The results show that the additional parameters required for the rotation matrices are negligible compared to the total number of weight parameters $\mathbf{w}$. For smaller models, such as CNN4 on FMNIST and SVHN, the overhead is approximately $1\%$, whereas for larger architectures, such as ResNet10 and ResNet18, the overhead drops to around $0.56\%$. For even larger models such as ConvNeXt-Tiny, the overhead further decreases to only $0.28\%$. This trend occurs because the number of rotation parameters grows much more slowly than the total number of model weights. Overall, the results demonstrate that FedBNN introduces minimal parameter overhead (compared to FedAvg) while enabling rotation-based alignment, making the method scalable to larger architectures without significantly increasing model size or communication cost.

\subsection{Convergence of FedBNN vs FedAvg} \label{convergence}
\begin{figure*}[htbp]
\centering

\begin{subfigure}{1.0\textwidth}
\centering
\includegraphics[width=\linewidth]{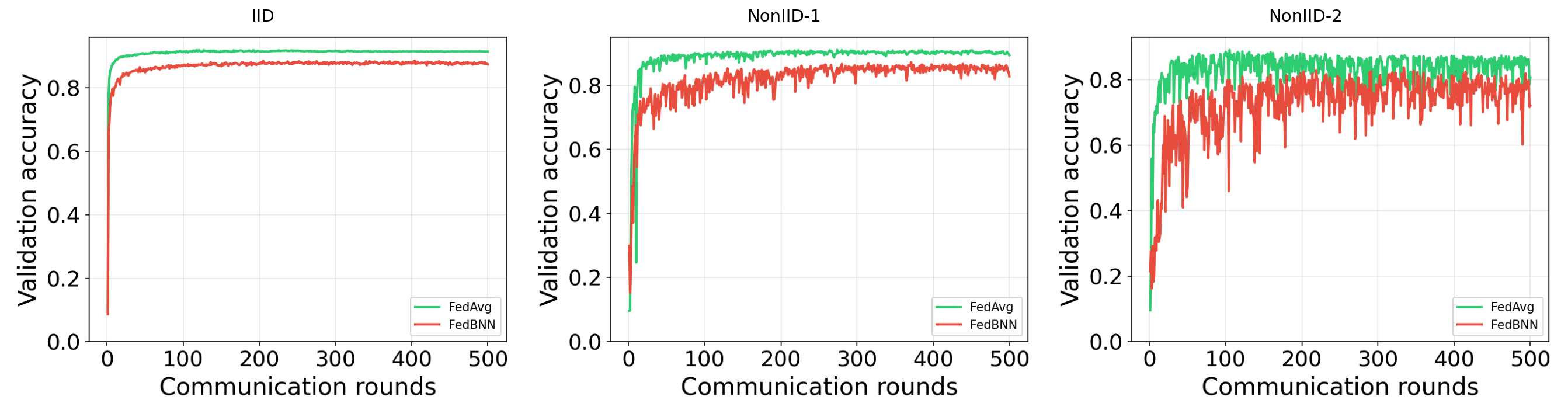}
\caption{FMNIST CNN4 — Validation Accuracy}
\end{subfigure}

\vspace{0.3em}

\begin{subfigure}{1.0\textwidth}
\centering
\includegraphics[width=\linewidth]{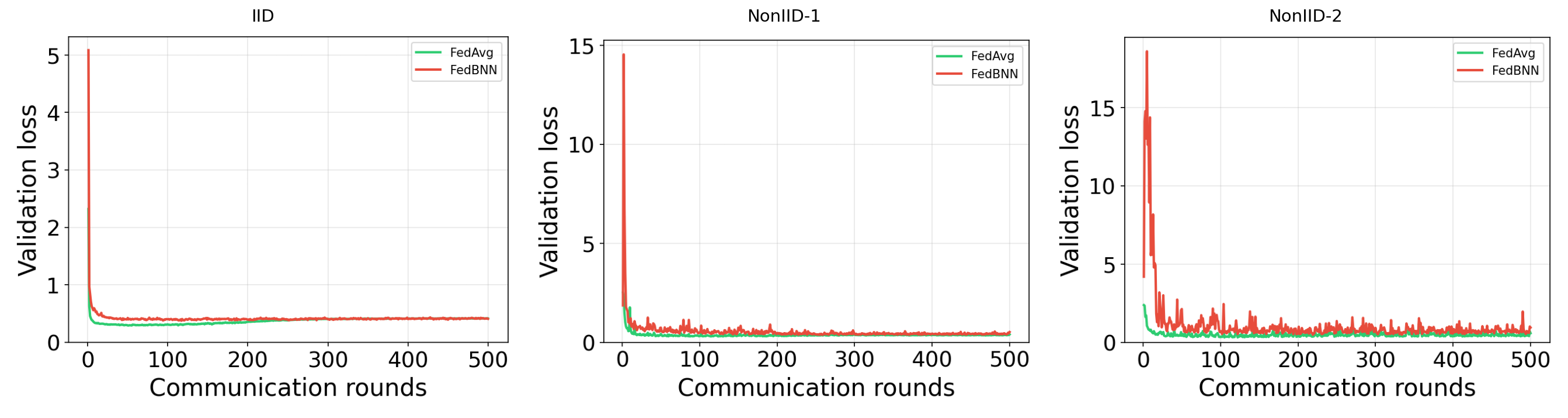}
\caption{FMNIST CNN4 — Validation Loss}
\end{subfigure}

\caption{Performance comparison between FedAvg and FedBNN on FMNIST using the CNN4 architecture under IID, Non-IID1 (Dirichlet $\nu=0.3$), and Non-IID2 (label-skew) data partitions.}
\label{fig:fmnist_fedavg_fedbnn}
\end{figure*}

\begin{figure*}[htbp]
\centering

\begin{subfigure}{0.9\textwidth}
\centering
\includegraphics[width=\linewidth]{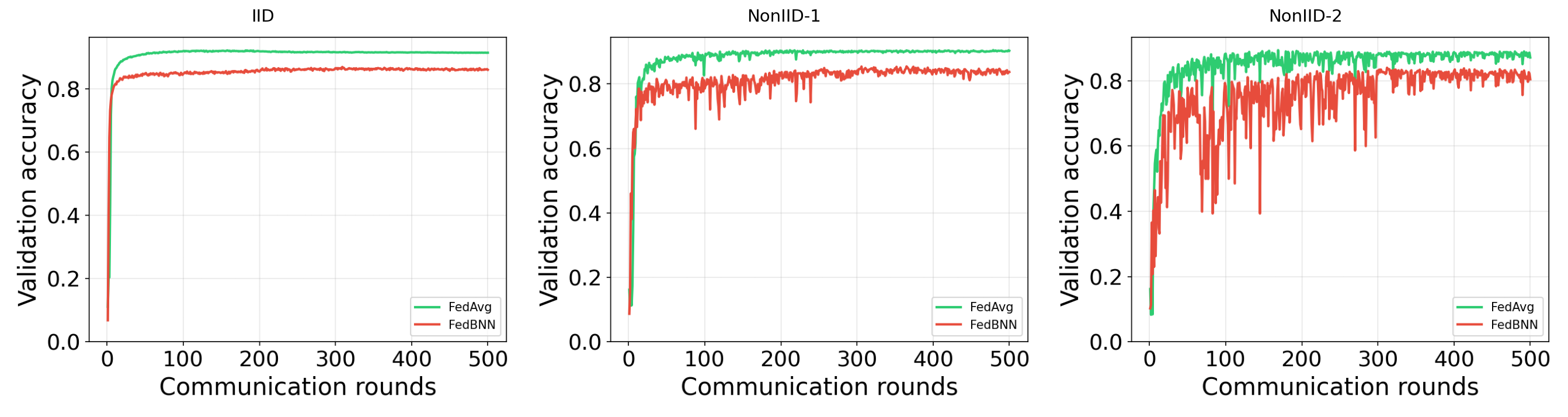}
\caption{SVHN CNN4 — Validation Accuracy}
\end{subfigure}

\vspace{0.3em}

\begin{subfigure}{0.9\textwidth}
\centering
\includegraphics[width=\linewidth]{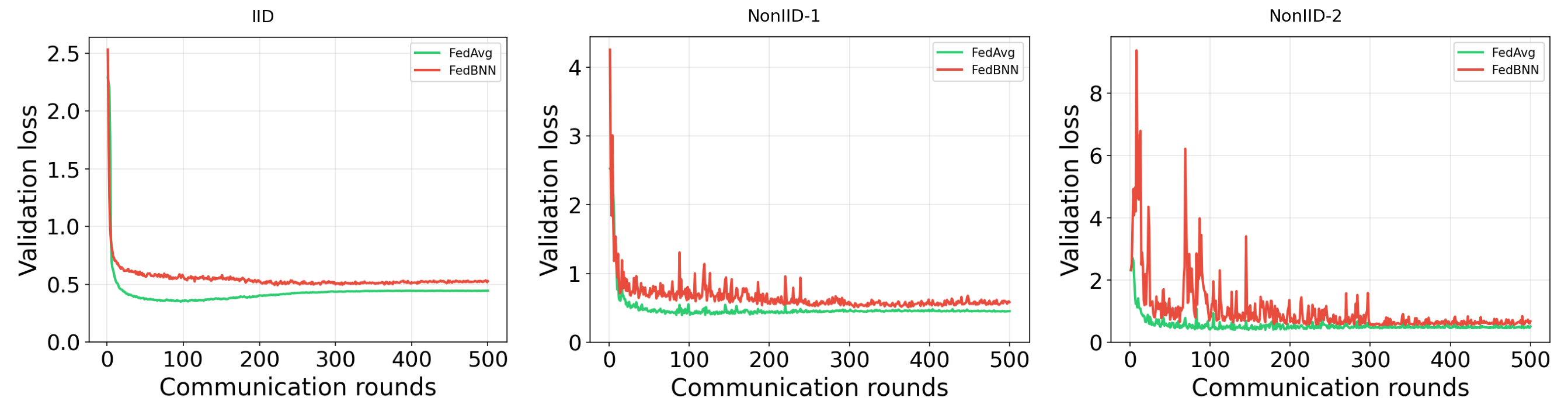}
\caption{SVHN CNN4 — Validation Loss}
\end{subfigure}

\caption{Performance comparison between FedAvg and FedBNN on SVHN using the CNN4 architecture under IID, Non-IID1 (Dirichlet $\nu=0.3$), and Non-IID2 (label-skew) data partitions.}
\label{fig:svhn_fedavg_fedbnn}
\end{figure*}

\begin{figure*}[htbp]
\centering

\begin{subfigure}{0.9\textwidth}
\centering
\includegraphics[width=\linewidth]{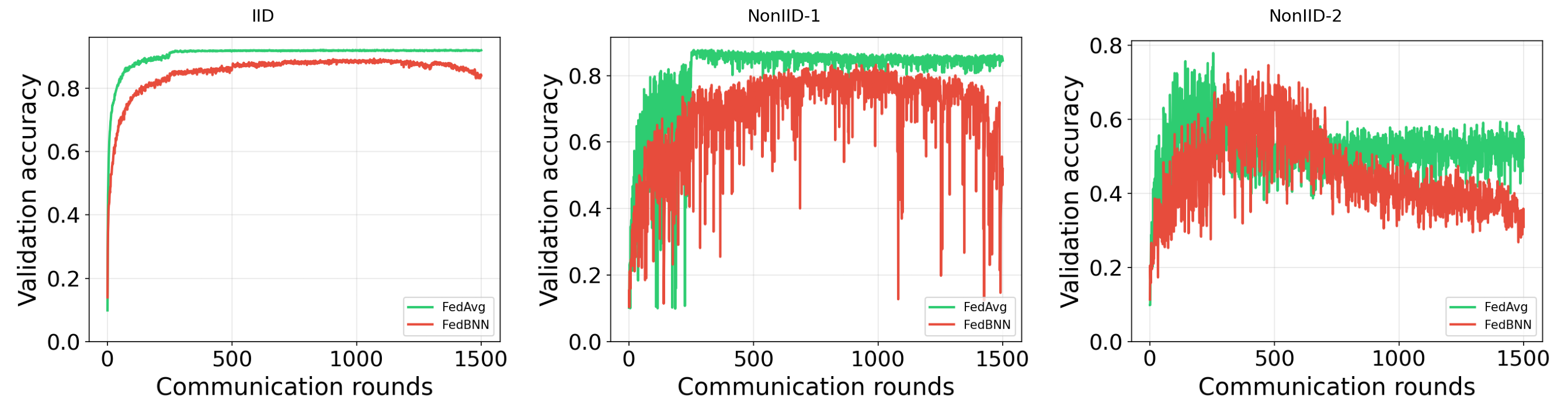}
\caption{CIFAR10 ResNet10 — Validation Accuracy}
\end{subfigure}

\vspace{0.3em}

\begin{subfigure}{0.9\textwidth}
\centering
\includegraphics[width=\linewidth]{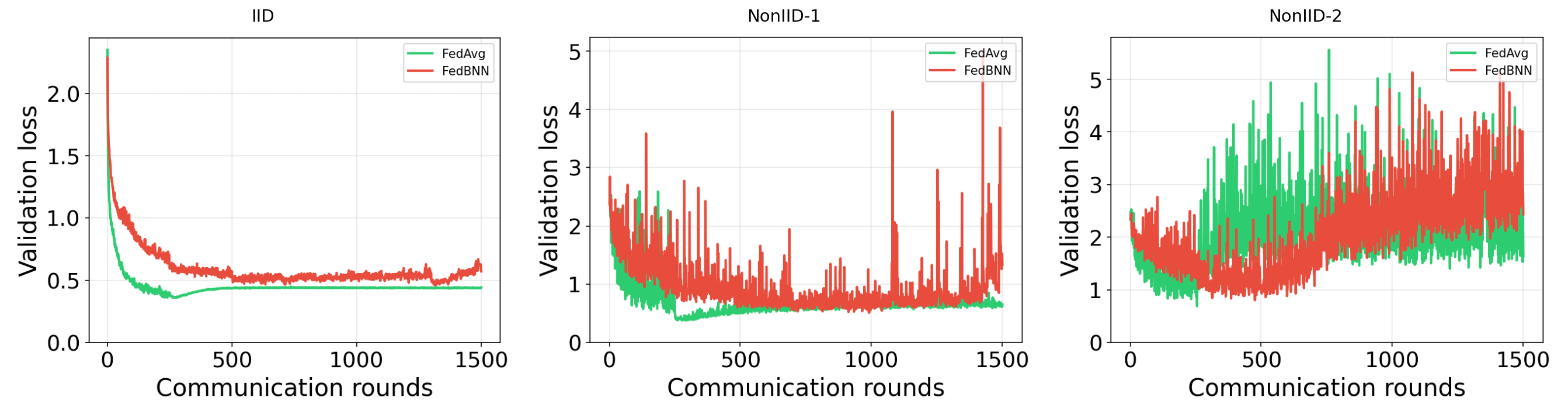}
\caption{CIFAR10 ResNet10 — Validation Loss}
\end{subfigure}

\caption{Performance comparison between FedAvg and FedBNN on CIFAR10 using the ResNet10 architecture under IID, Non-IID1 (Dirichlet $\nu=0.3$), and Non-IID2 (label-skew) data partitions.}
\label{fig:cifar10_resnet_fedavg_fedbnn}
\end{figure*}

\begin{figure*}[htbp]
\centering

\begin{subfigure}{1.0\textwidth}
\centering
\includegraphics[width=\linewidth]{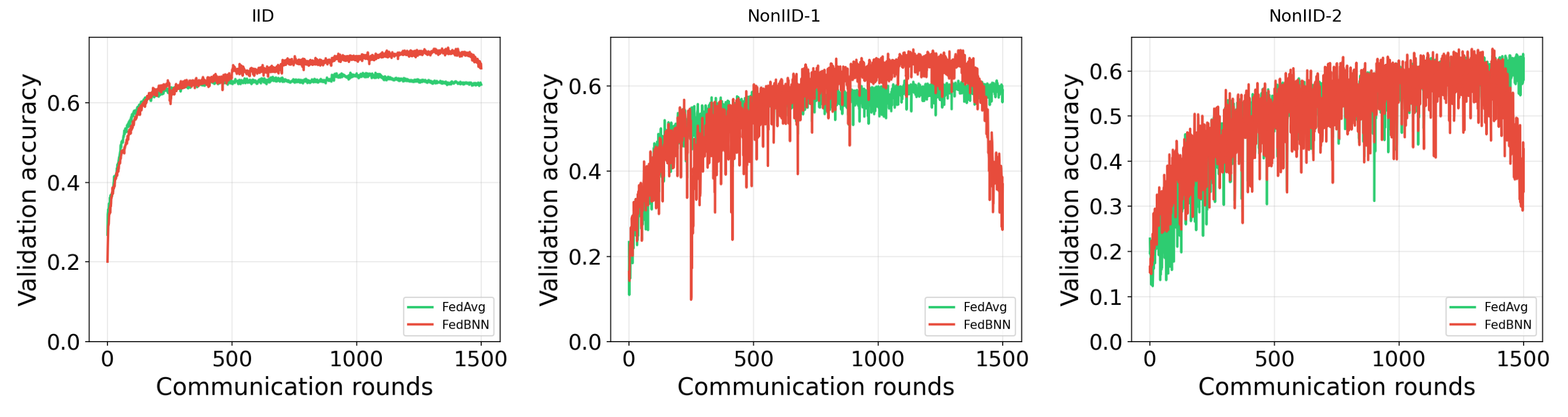}
\caption{CIFAR10 ConvNeXt-Tiny — Validation Accuracy}
\end{subfigure}

\vspace{0.3em}

\begin{subfigure}{0.9\textwidth}
\centering
\includegraphics[width=\linewidth]{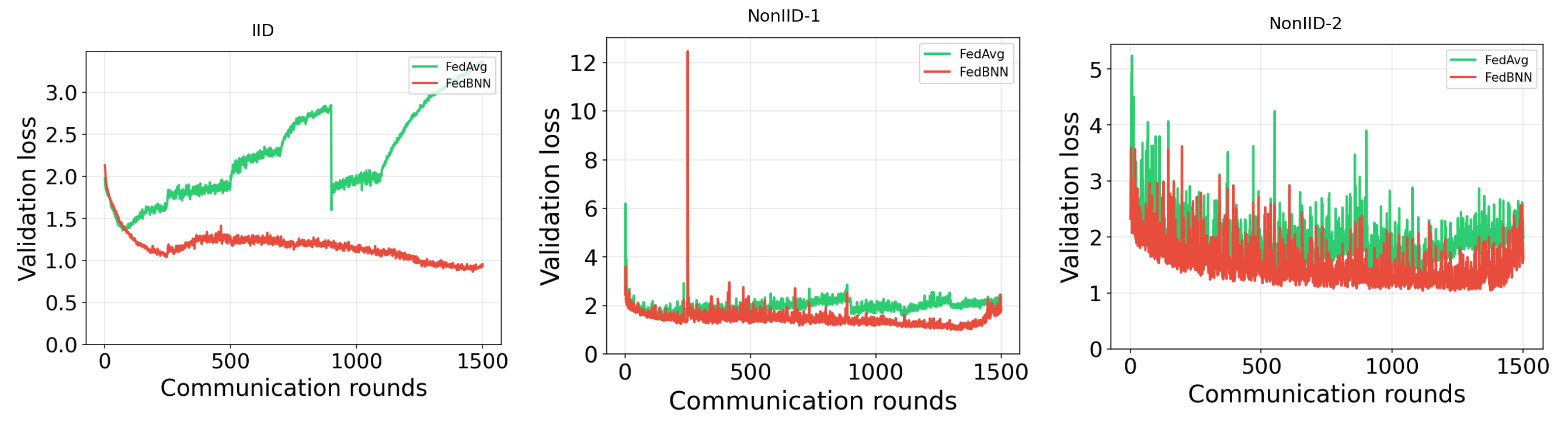}
\caption{CIFAR10 ConvNeXt-Tiny — Validation Loss}
\end{subfigure}

\caption{Performance comparison between FedAvg and FedBNN on CIFAR10 using the ConvNeXt-Tiny architecture under IID, Non-IID1 (Dirichlet $\nu=0.3$), and Non-IID2 (label-skew) data partitions. {The adaptive coefficients also exhibit an annealing behavior, with $1-\alpha \rightarrow 1$ and $\alpha\beta \rightarrow 0$ during optimization, indicating that the rotation correction gradually vanishes as the model converges.}}
\label{fig:cifar10_convnext_fedavg_fedbnn}
\end{figure*}

\begin{figure*}[htbp]
\centering

\begin{subfigure}{0.9\textwidth}
\centering
\includegraphics[width=\linewidth]{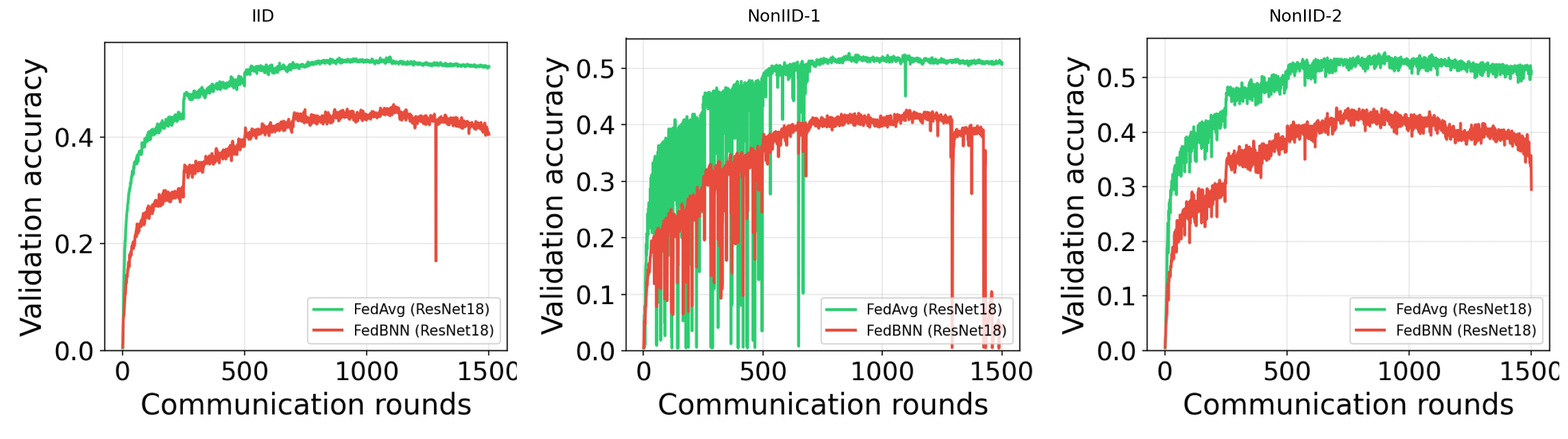}
\caption{TinyImageNet ResNet18 — Validation Accuracy}
\end{subfigure}

\vspace{0.3em}

\begin{subfigure}{0.9\textwidth}
\centering
\includegraphics[width=\linewidth]{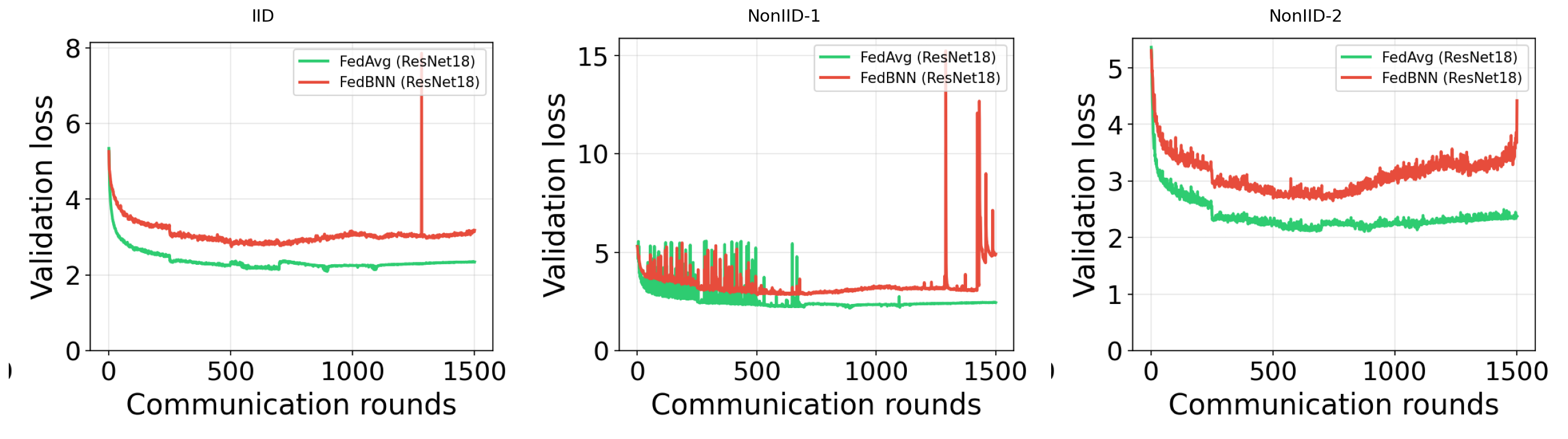}
\caption{TinyImageNet ResNet18 — Validation Loss}
\end{subfigure}

\caption{Performance comparison between FedAvg and FedBNN on TinyImageNet using the ResNet18 architecture under IID, Non-IID1 (Dirichlet $\nu=0.3$), and Non-IID2 (label-skew) data partitions.}
\label{fig:tinyimagenet_fedavg_fedbnn}
\end{figure*}

\begin{figure*}[htbp]
\centering

\begin{subfigure}{0.9\textwidth}
\centering
\includegraphics[width=\linewidth]{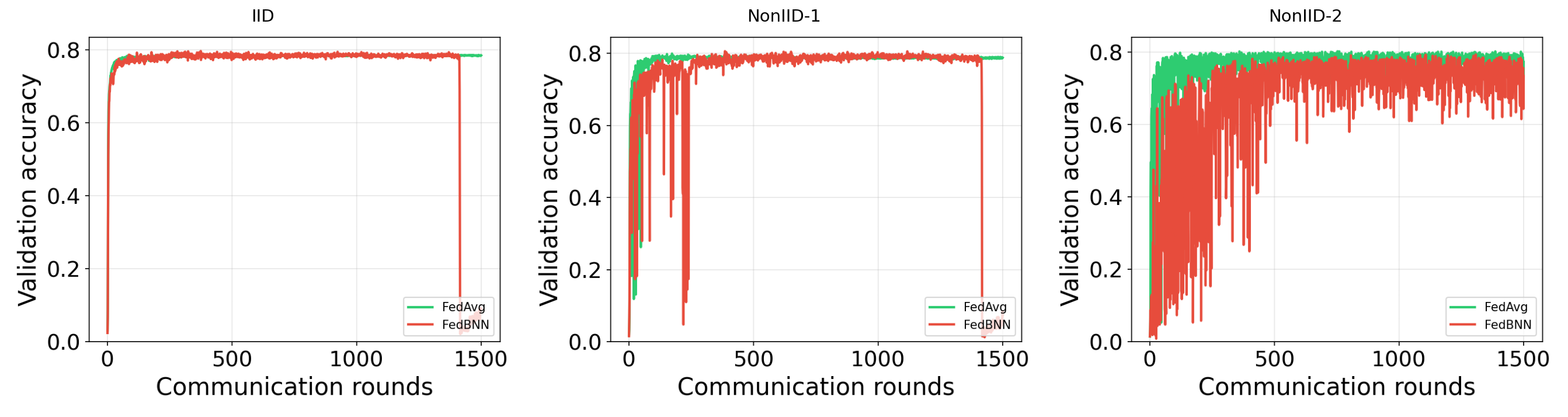}
\caption{FEMNIST ResNet18 — Validation Accuracy}
\end{subfigure}
\vspace{0.3em}

\begin{subfigure}{0.9\textwidth}
\centering
\includegraphics[width=\linewidth]{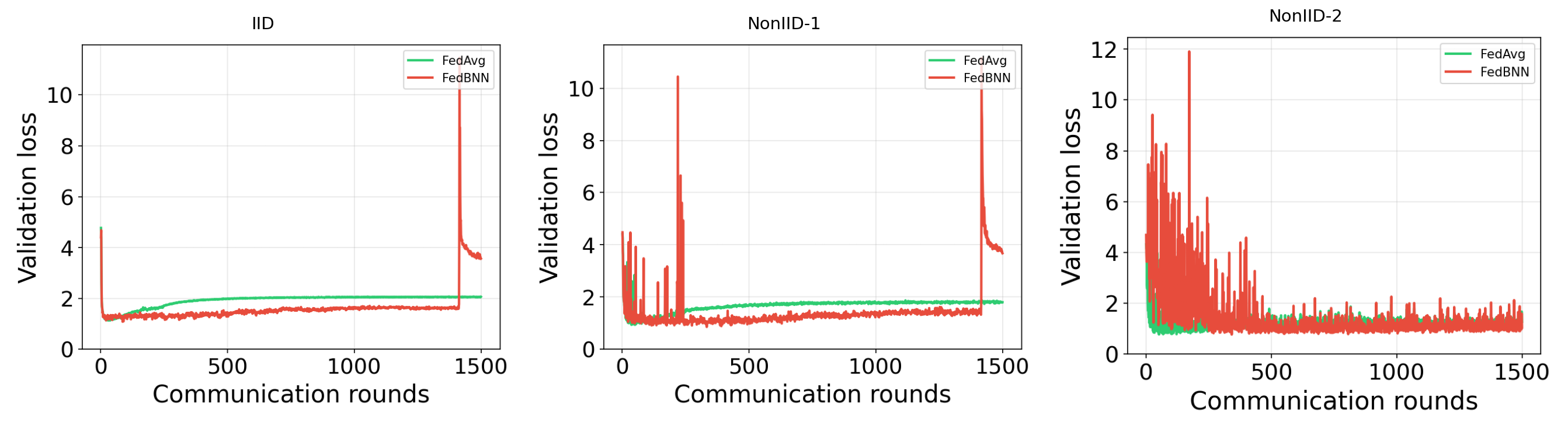}
\caption{FEMNIST ResNet18 — Validation Loss}
\end{subfigure}

\caption{Performance comparison between FedAvg and FedBNN on FEMNIST using the ResNet18 architecture under IID, Non-IID1 (Dirichlet $\nu=0.3$), and Non-IID2 (label-skew) data partitions.}
\label{fig:femnist_fedavg_fedbnn}
\end{figure*}

{
We compare the validation accuracy and validation loss of FedBNN against FedAvg across multiple datasets (FMNIST, SVHN, CIFAR10, FEMNIST, TinyImageNet), architectures (CNN-4, ResNet-10, ConvNeXt-Tiny, ResNet-18), and data-heterogeneity settings (IID, Non-IID-1, and Non-IID-2). The validation accuracy and validation loss curves (see Figures~\ref{fig:fmnist_fedavg_fedbnn}, \ref{fig:svhn_fedavg_fedbnn}, \ref{fig:cifar10_resnet_fedavg_fedbnn}, \ref{fig:cifar10_convnext_fedavg_fedbnn}, \ref{fig:tinyimagenet_fedavg_fedbnn}, \ref{fig:femnist_fedavg_fedbnn}) show that FedBNN achieves competitive performance relative to FedAvg under both IID and Non-IID settings. FedBNN even outperforms FedAvg when considering the ConvNeXt-Tiny architecture. \revAddd{With ConvNeXt-Tiny, the IID validation loss of FedAvg rises after about 100 rounds while that of FedBNN stays lower (Figure~\ref{fig:cifar10_convnext_fedavg_fedbnn}), which is consistent with binarization acting as a regularizer for this high-capacity model.} Overall, these results indicate that FedBNN is a promising approach for federated learning with binary neural networks, offering a favorable accuracy–efficiency trade-off without sacrificing (and at times improving) generalization compared to full-precision FedAvg.}

% \vspace{-1cm}

\newpage
\subsection{Alpha, beta, lambda} \label{Alpha, beta, lambda}
\begin{figure*}[htbp]
\centering
\vspace{1.8cm}
% Row 1: (1-alpha)
\begin{subfigure}{0.32\textwidth}
\includegraphics[width=\linewidth]{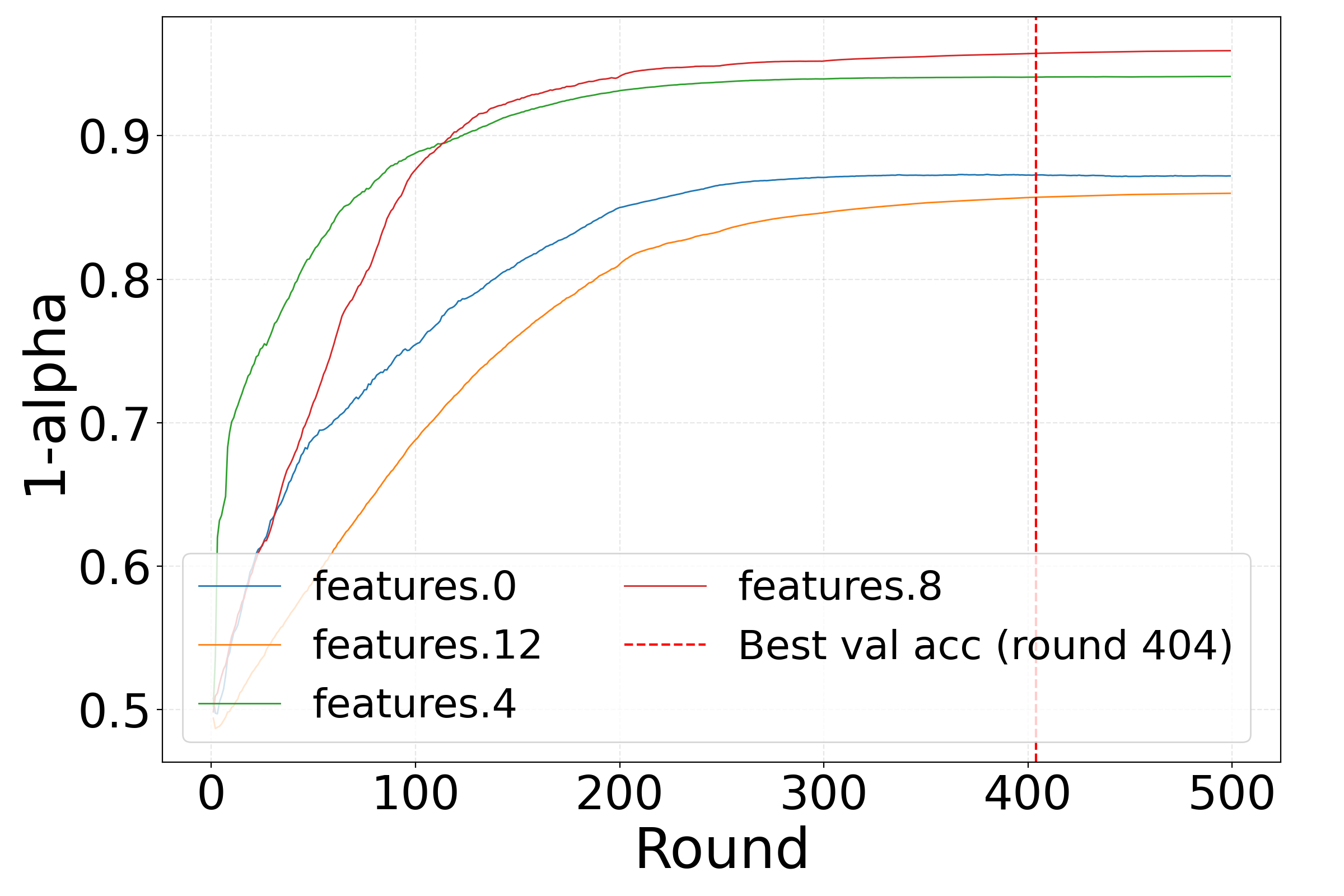}
\caption{IID: $(1-\alpha)$}
\end{subfigure}
\begin{subfigure}{0.32\textwidth}
\includegraphics[width=\linewidth]{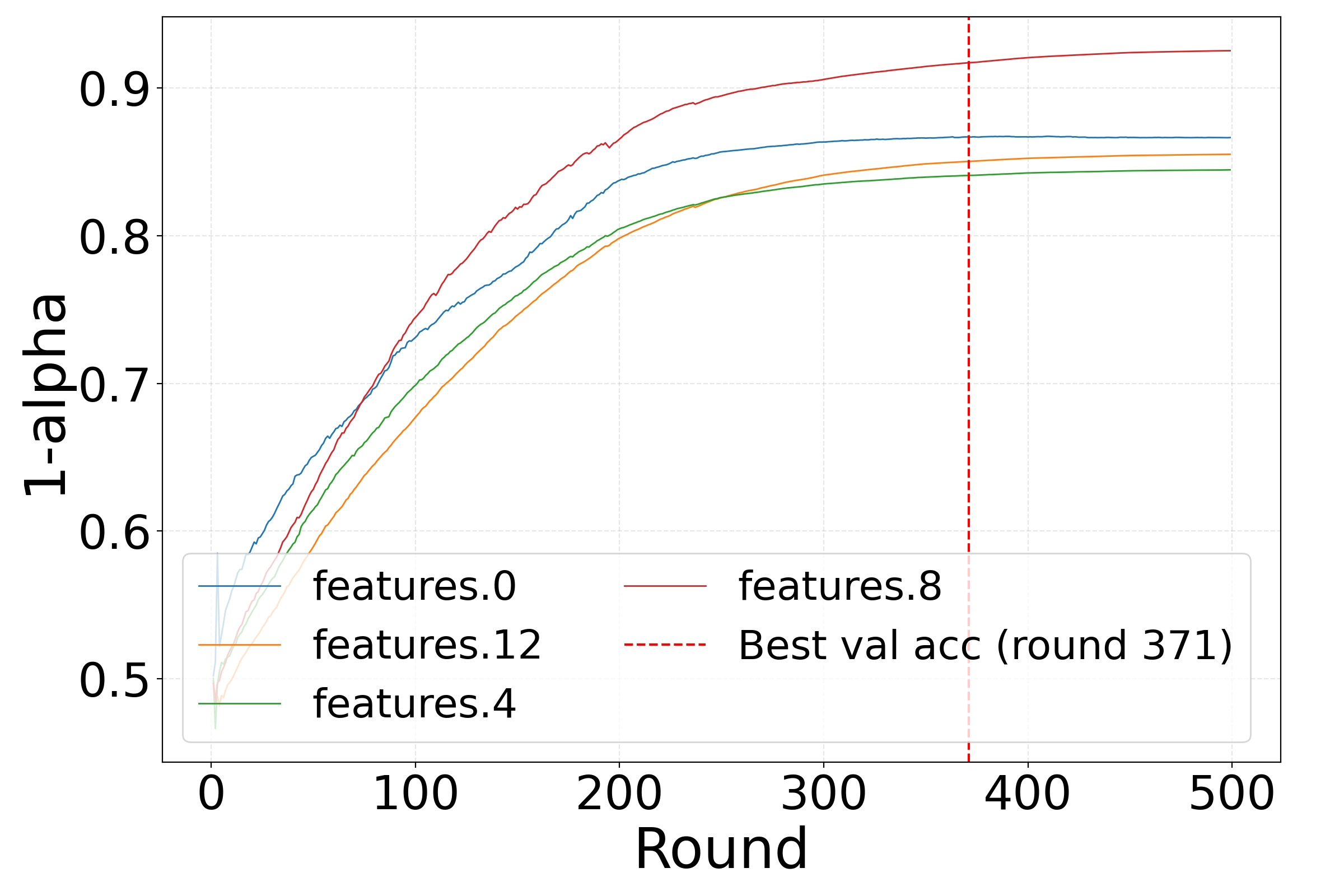}
\caption{Non-IID 1: $(1-\alpha)$}
\end{subfigure}
\begin{subfigure}{0.32\textwidth}
\includegraphics[width=\linewidth]{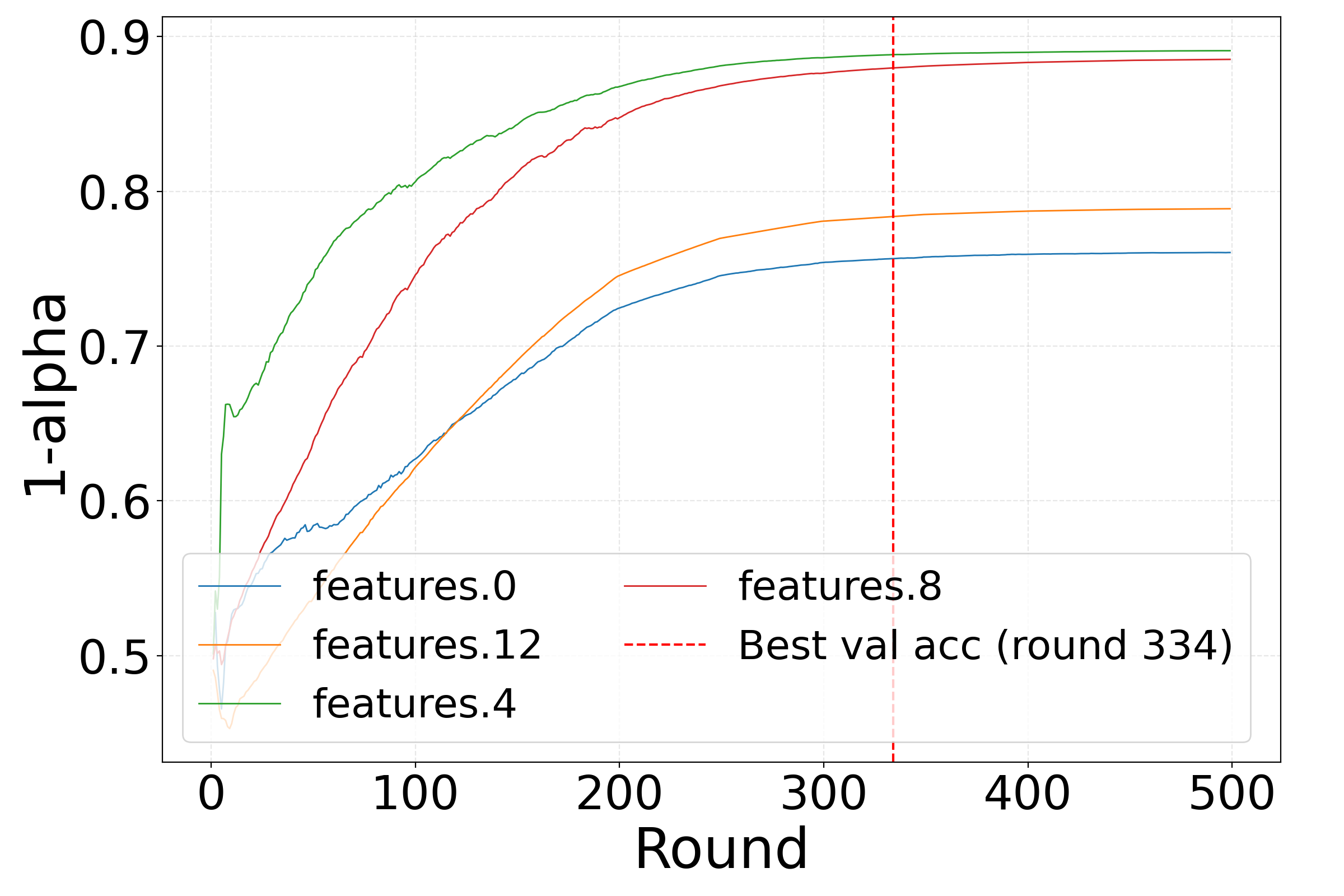}
\caption{Non-IID 2: $(1-\alpha)$}
\end{subfigure}

\vspace{0.0em}

% Row 2: alpha beta
\begin{subfigure}{0.32\textwidth}
\includegraphics[width=\linewidth]{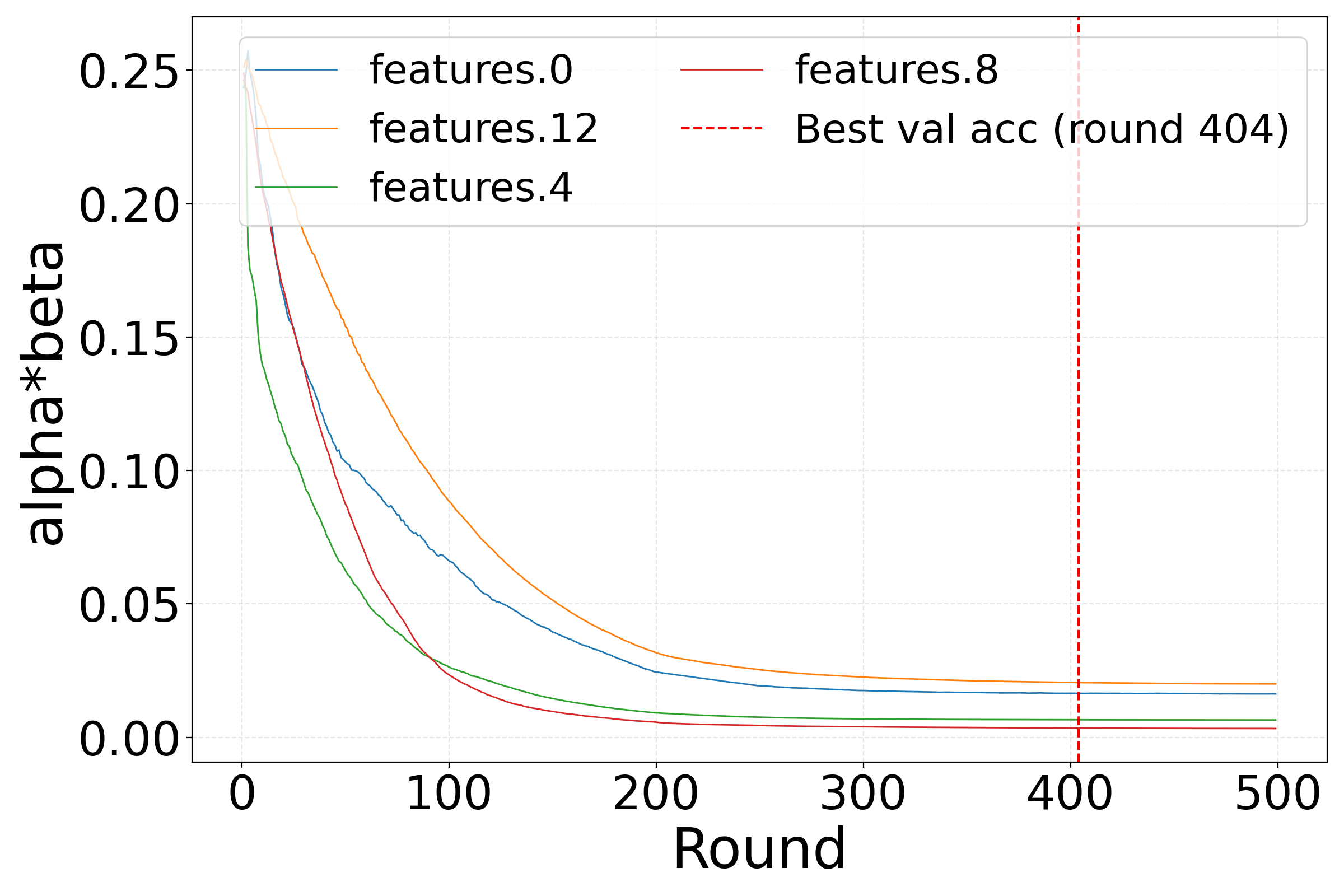}
\caption{IID: $\alpha\beta$}
\end{subfigure}
\begin{subfigure}{0.32\textwidth}
\includegraphics[width=\linewidth]{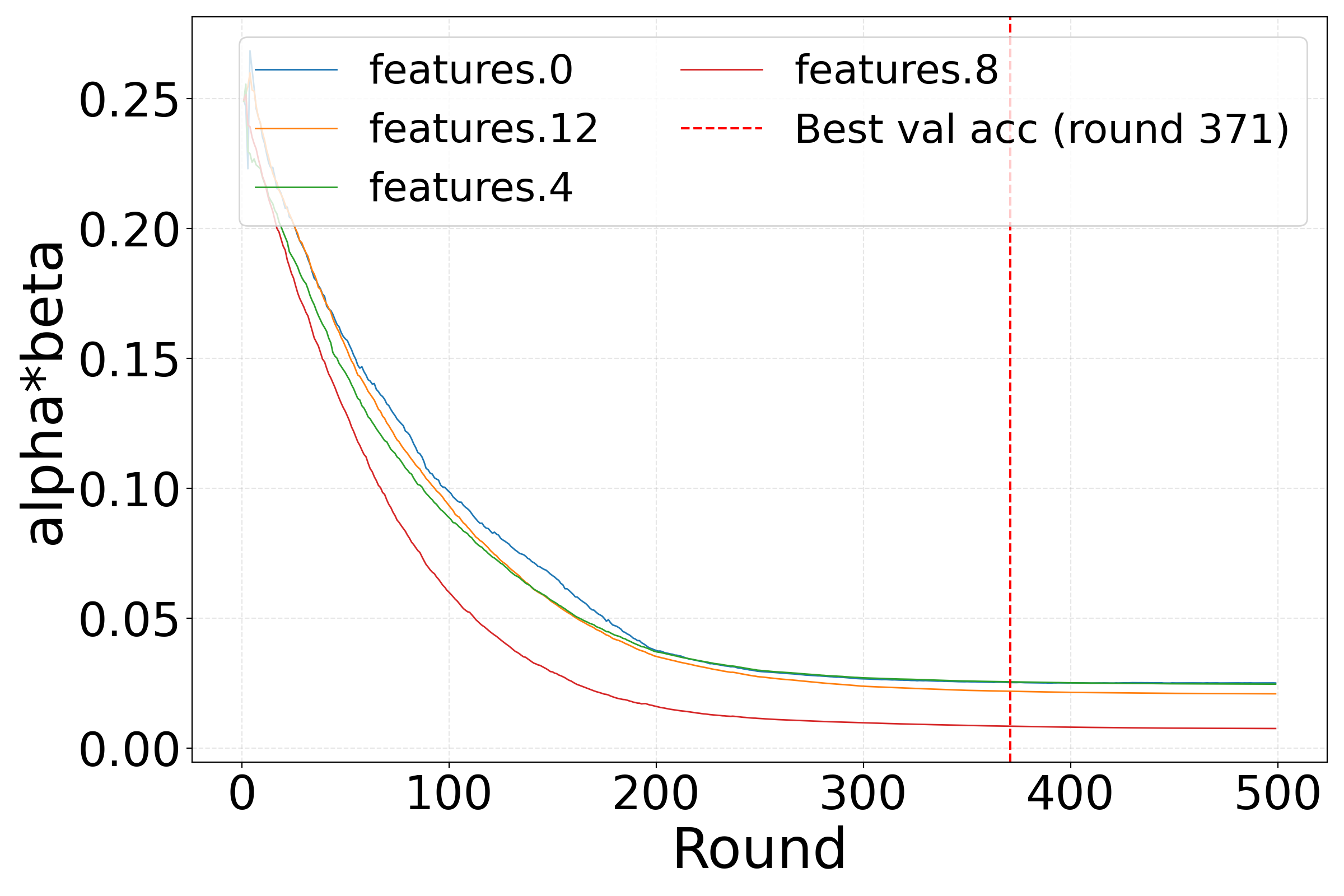}
\caption{Non-IID 1: $\alpha\beta$}
\end{subfigure}
\begin{subfigure}{0.32\textwidth}
\includegraphics[width=\linewidth]{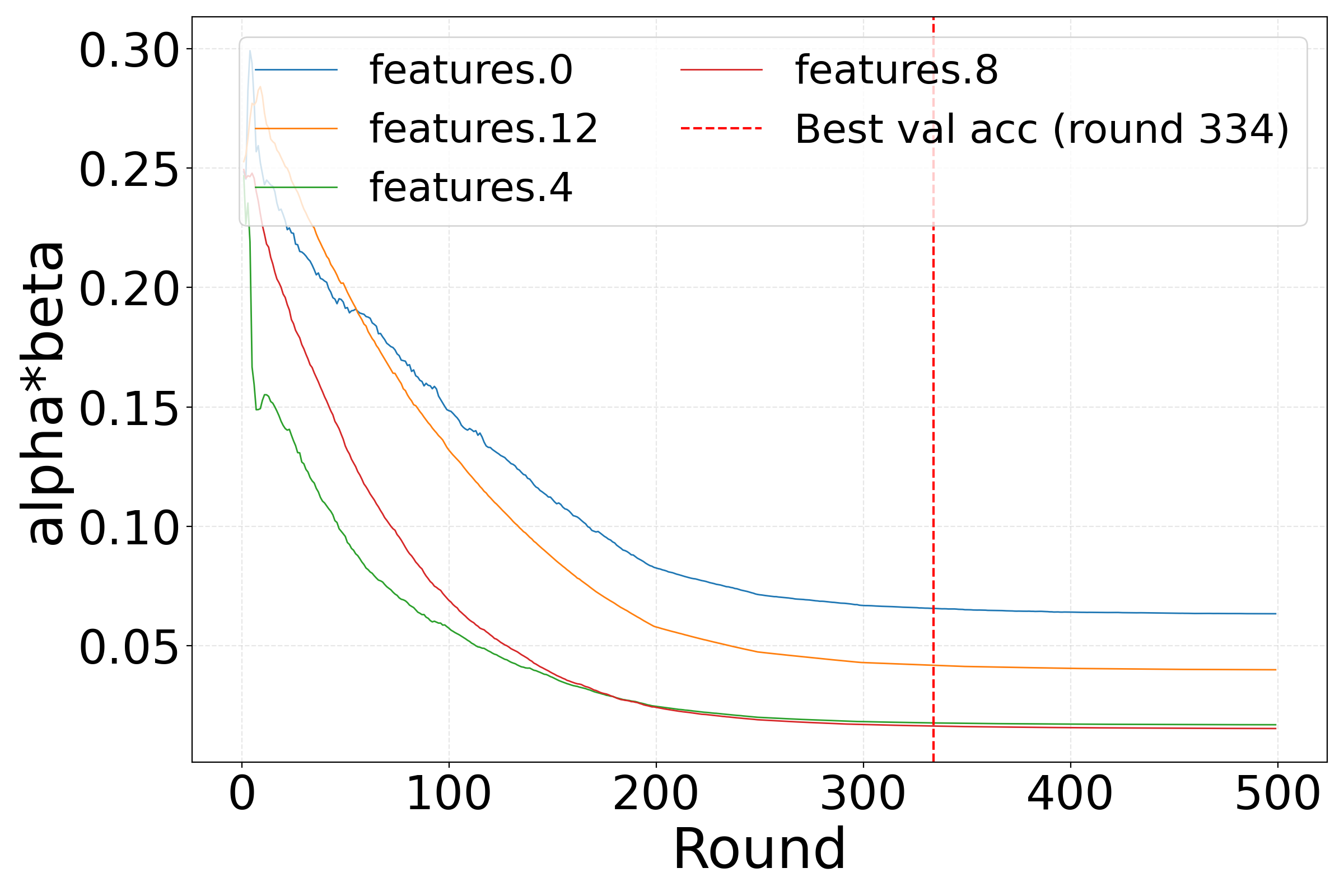}
\caption{Non-IID 2: $\alpha\beta$}
\end{subfigure}

\vspace{0.0em}

% Row 3: alpha(1-beta)
\begin{subfigure}{0.32\textwidth}
\includegraphics[width=\linewidth]{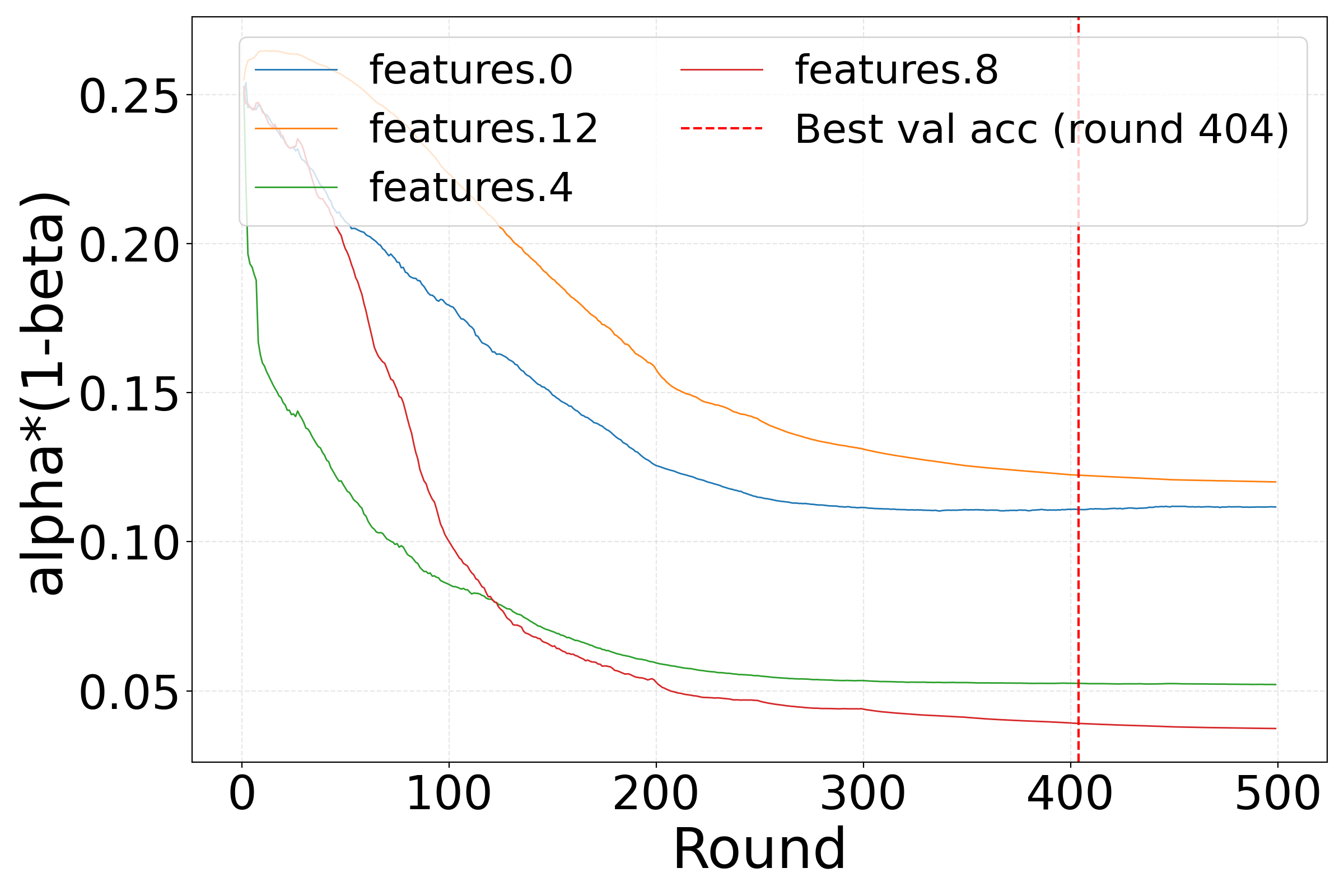}
\caption{IID: $\alpha(1-\beta)$}
\end{subfigure}
\begin{subfigure}{0.32\textwidth}
\includegraphics[width=\linewidth]{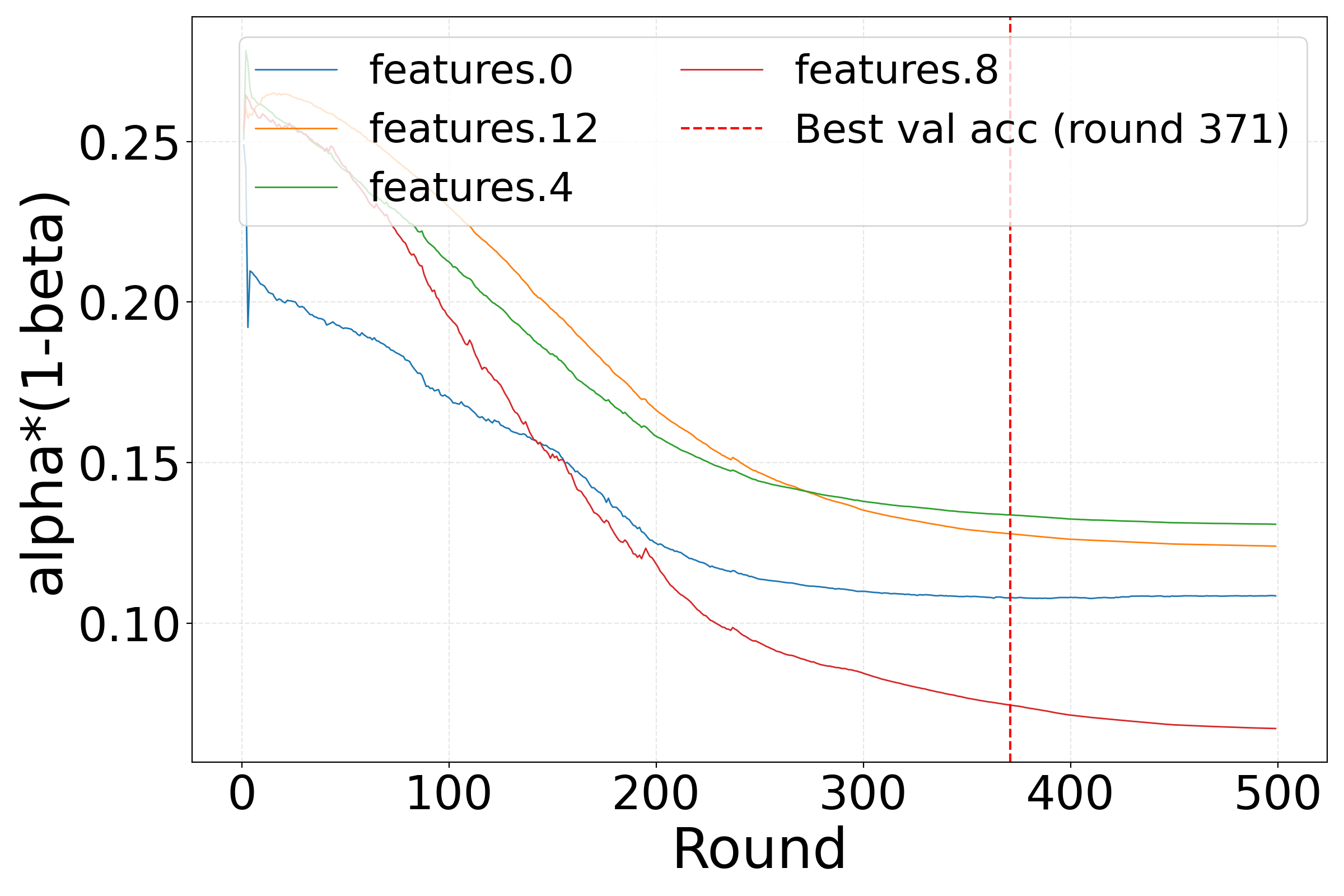}
\caption{Non-IID 1: $\alpha(1-\beta)$}
\end{subfigure}
\begin{subfigure}{0.32\textwidth}
\includegraphics[width=\linewidth]{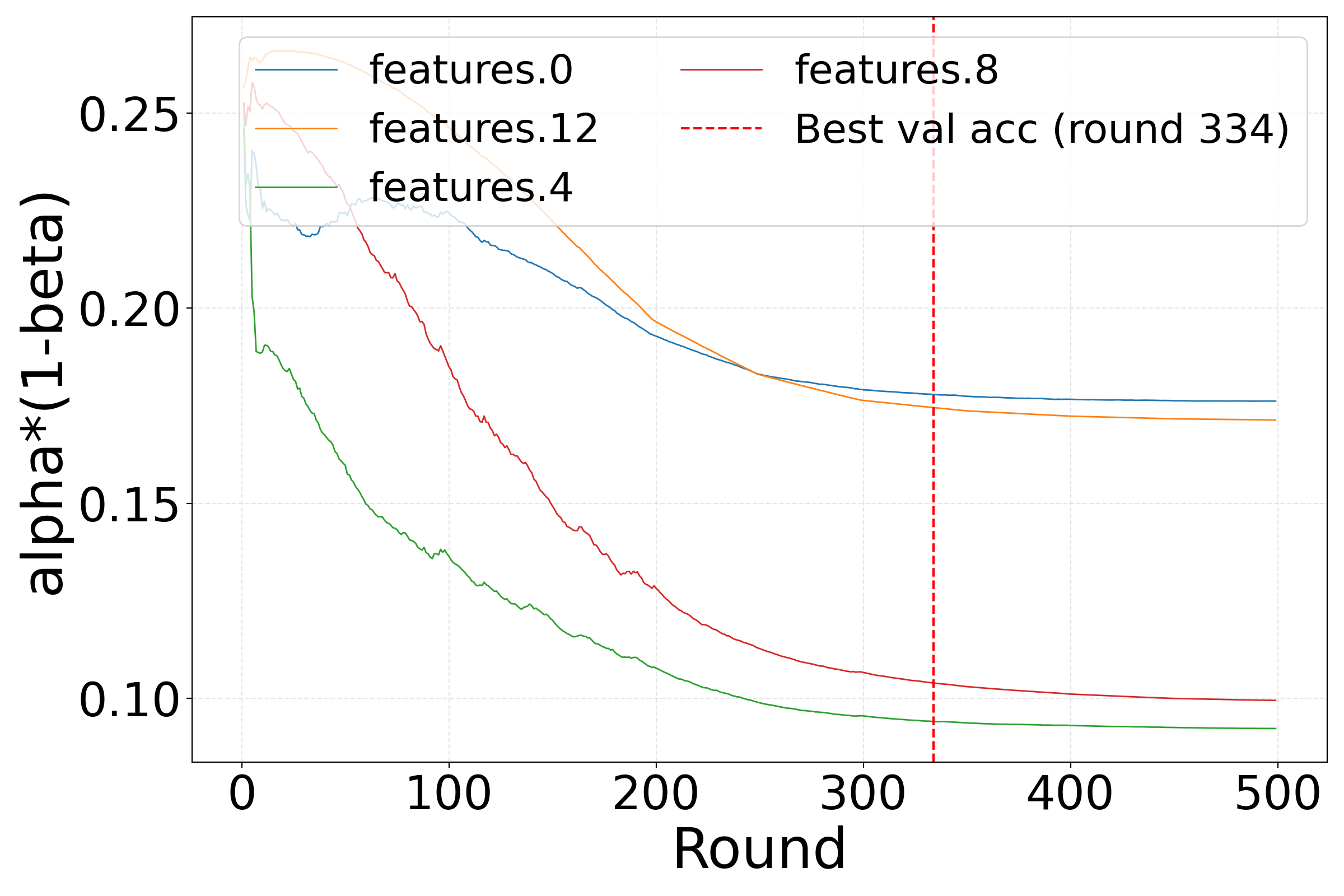}
\caption{Non-IID 2: $\alpha(1-\beta)$}
\end{subfigure}

\vspace{0.0em}

% Row 4: lambda
\begin{subfigure}{0.32\textwidth}
\includegraphics[width=\linewidth]{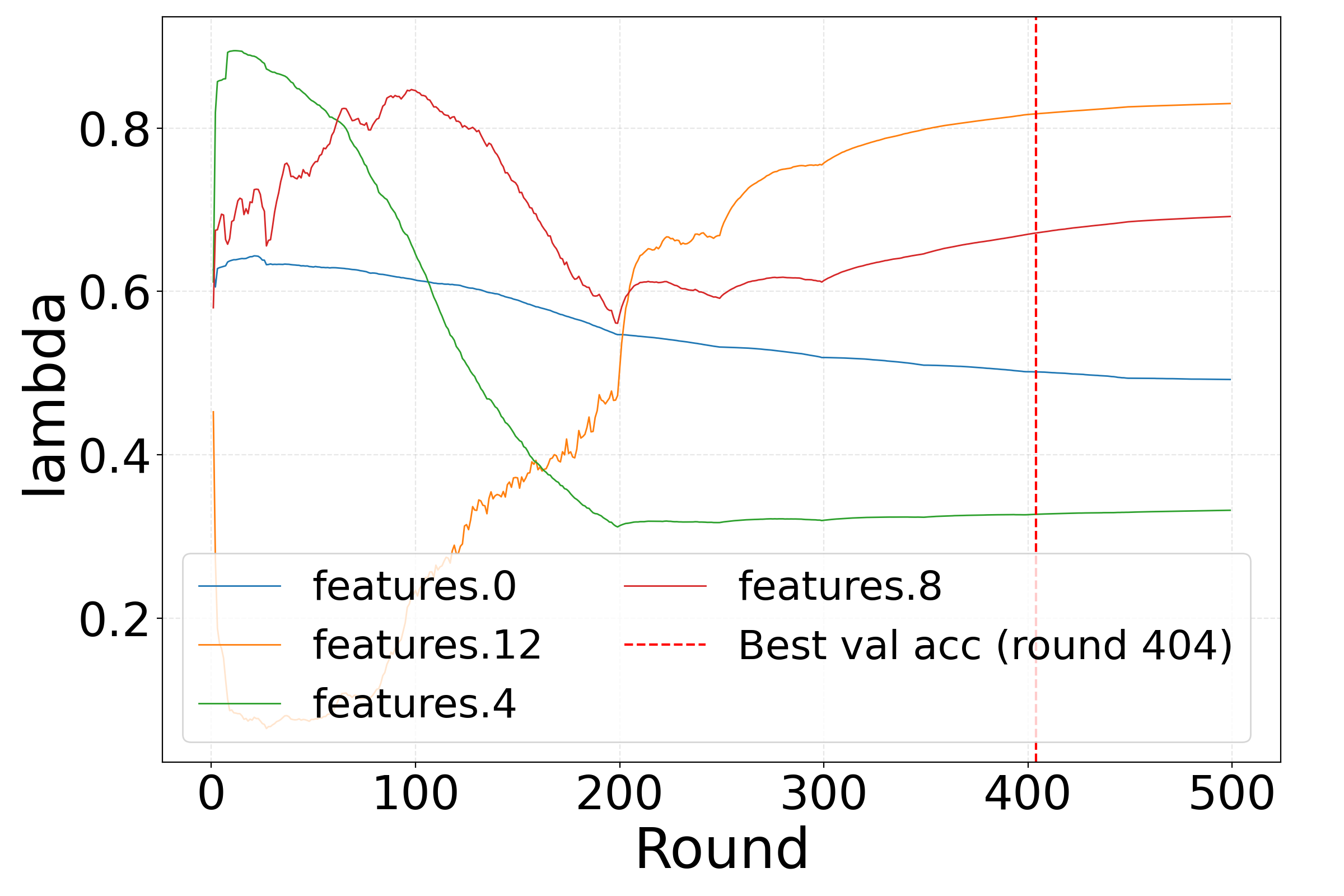}
\caption{IID: $\lambda$}
\end{subfigure}
\begin{subfigure}{0.32\textwidth}
\includegraphics[width=\linewidth]{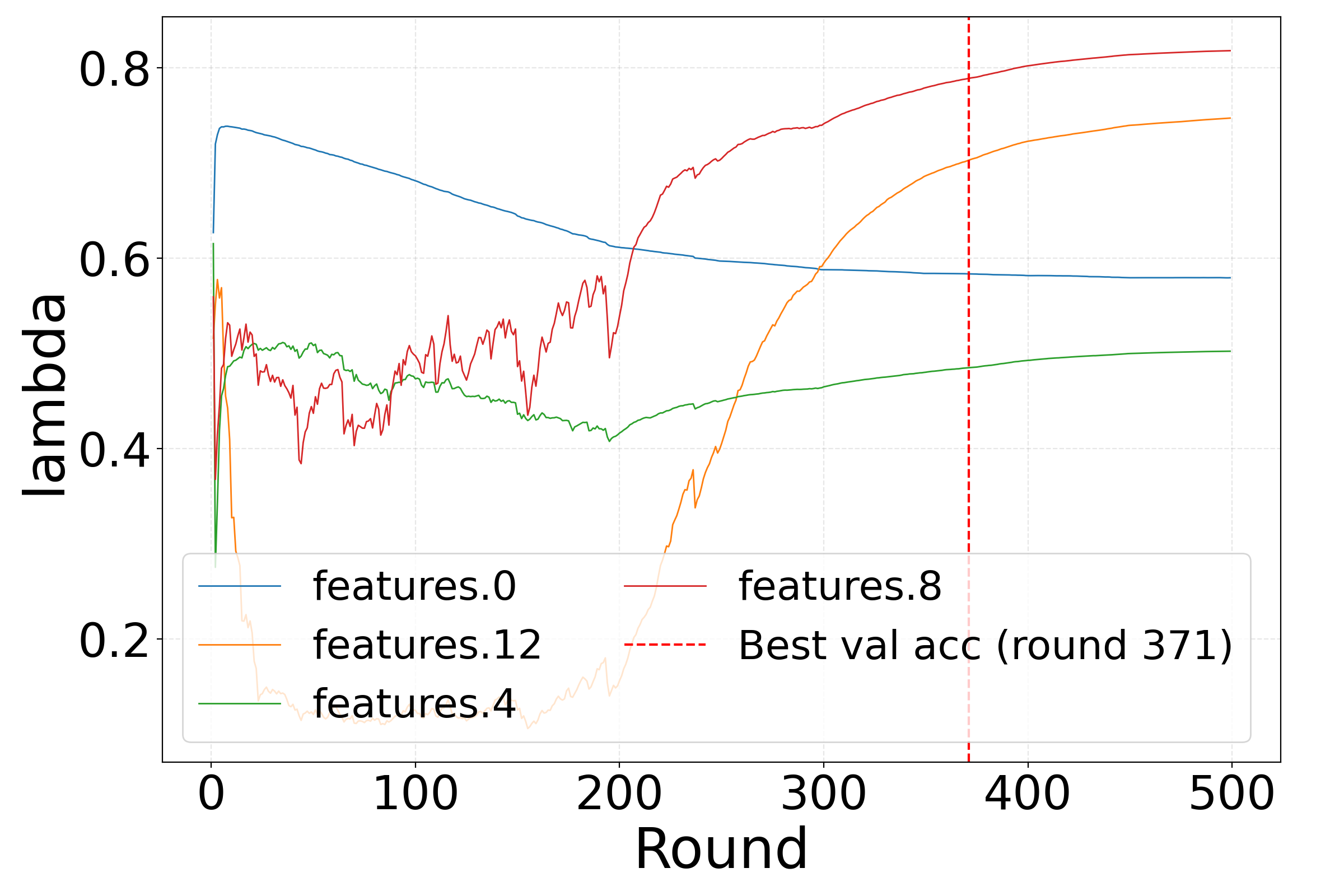}
\caption{Non-IID 1: $\lambda$}
\end{subfigure}
\begin{subfigure}{0.32\textwidth}
\includegraphics[width=\linewidth]{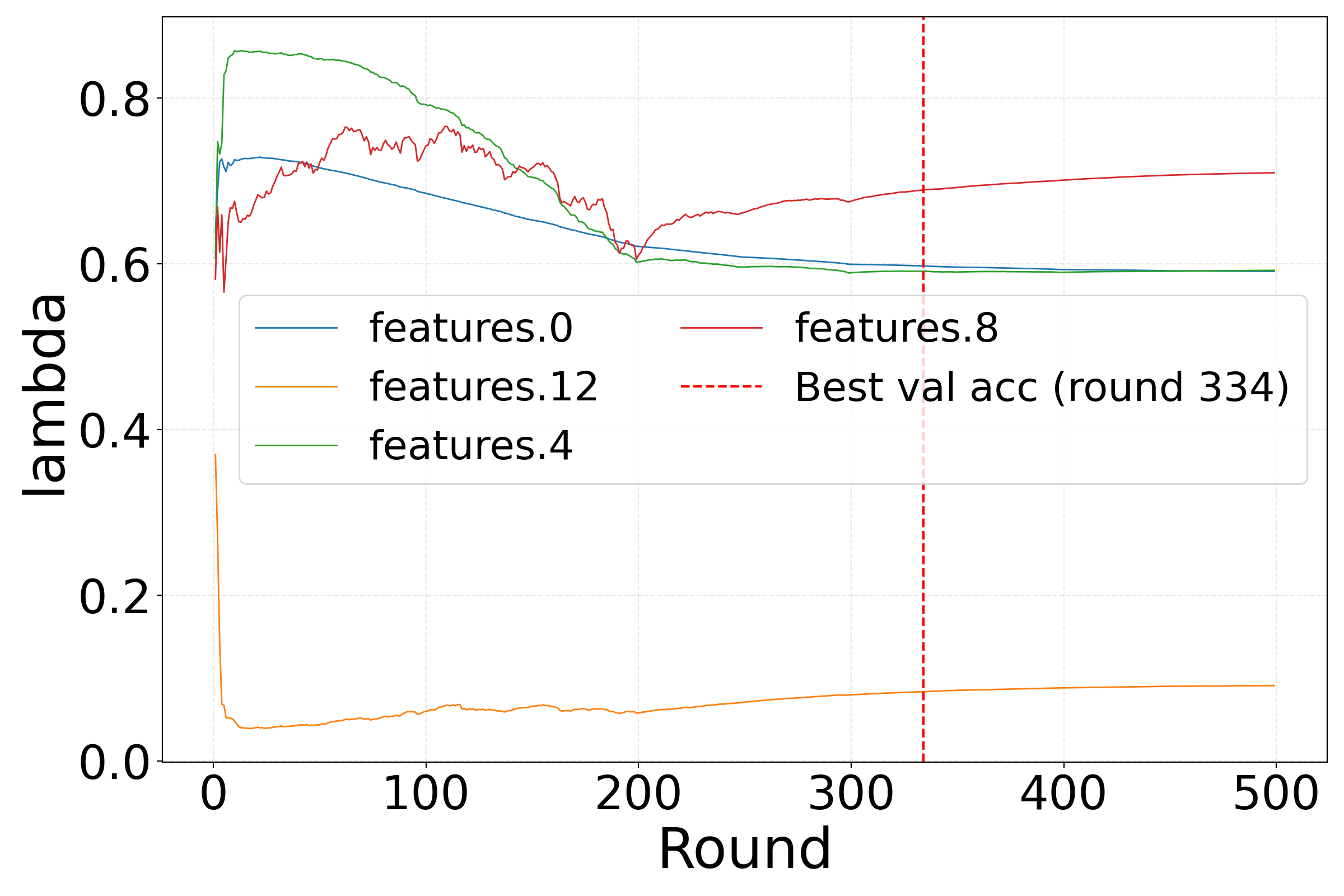}
\caption{Non-IID 2: $\lambda$}
\end{subfigure}

\caption{Mixing parameter variation on FMNIST under IID, Non-IID1, and Non-IID2.}
\label{alphabetalambdafmnist}
\end{figure*}

\begin{figure*}[htbp]
\centering

% Row 1: (1-alpha)
\begin{subfigure}{0.32\textwidth}
\includegraphics[width=\linewidth]{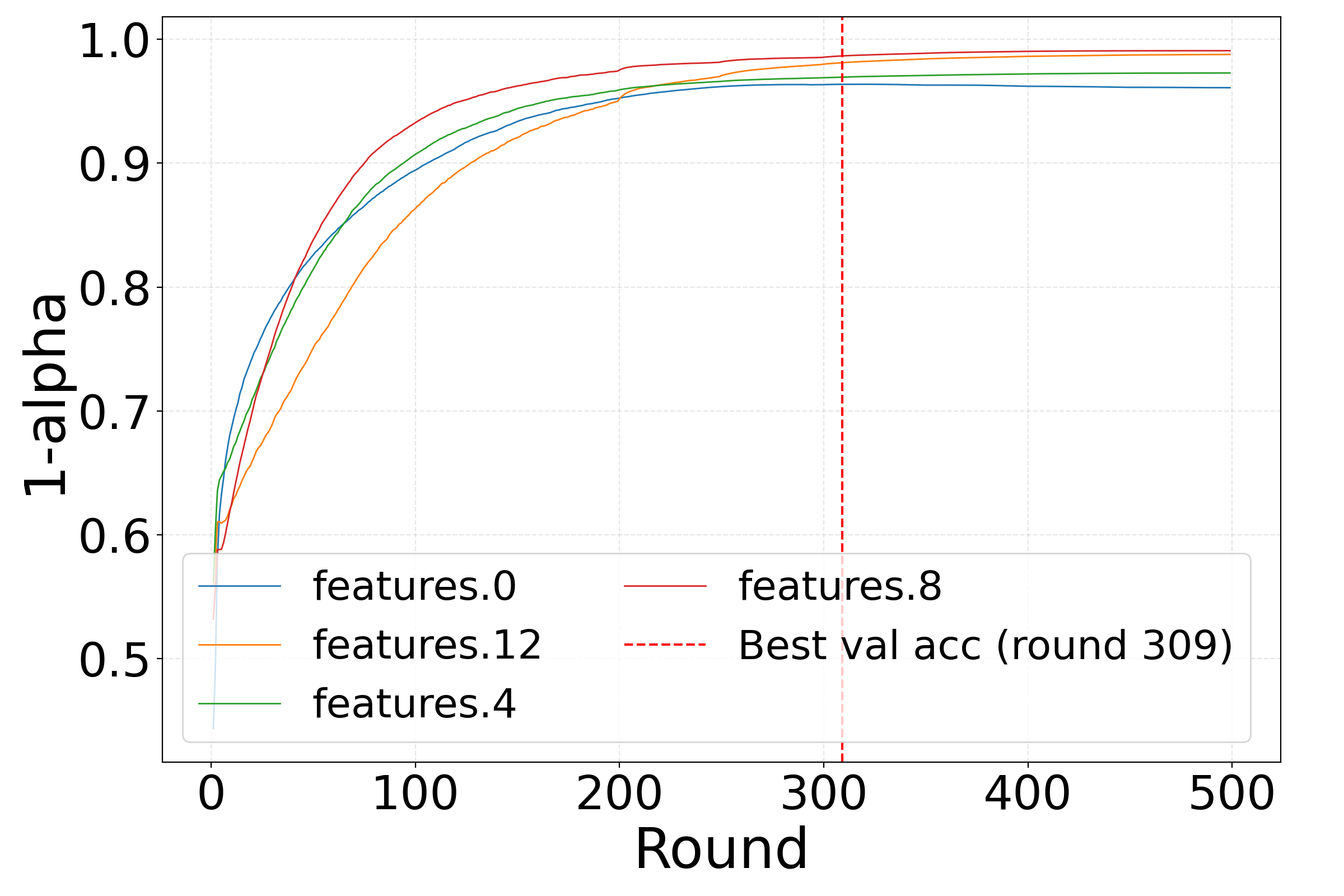}
\caption{IID: $(1-\alpha)$}
\end{subfigure}
\begin{subfigure}{0.32\textwidth}
\includegraphics[width=\linewidth]{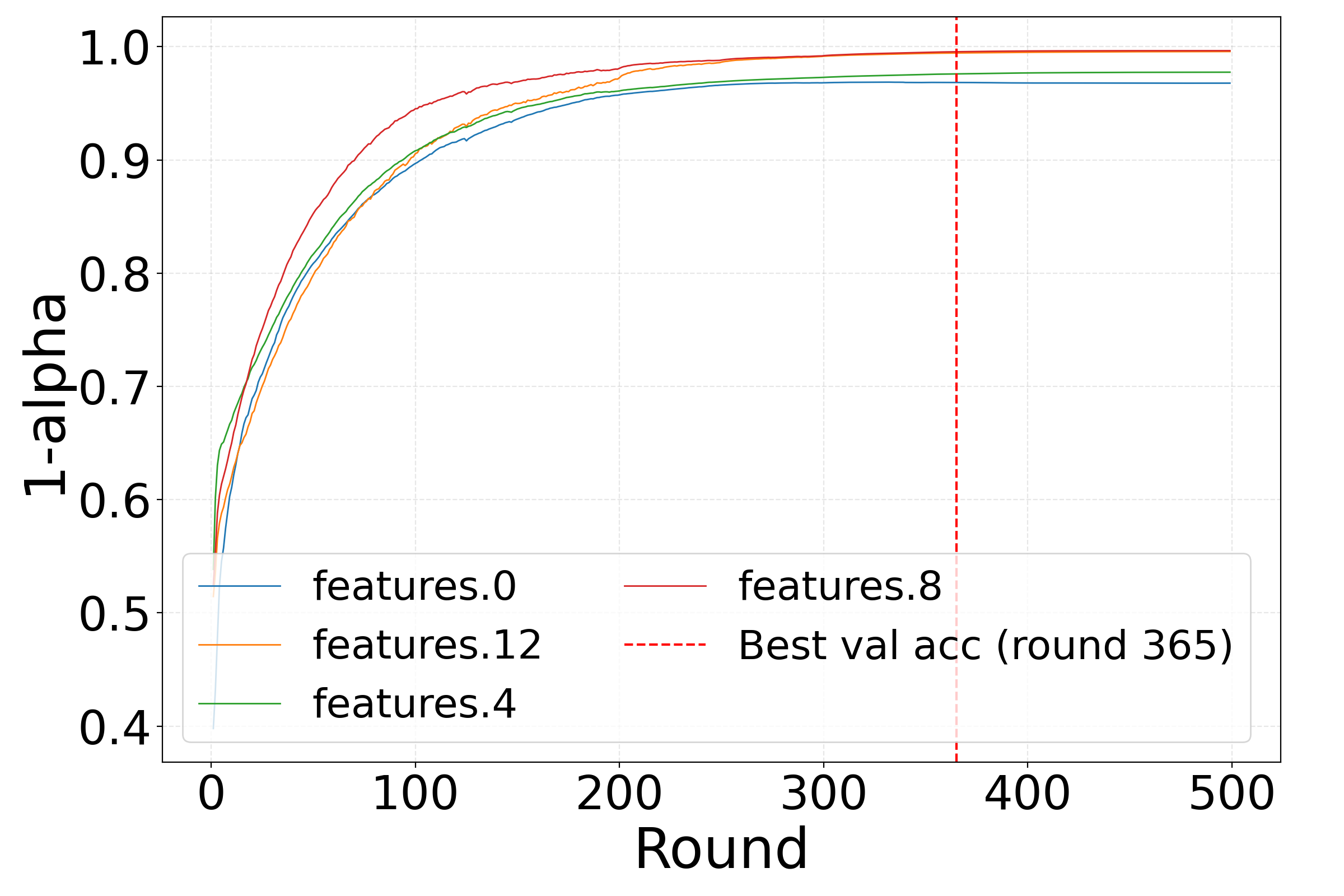}
\caption{Non-IID1: $(1-\alpha)$}
\end{subfigure}
\begin{subfigure}{0.32\textwidth}
\includegraphics[width=\linewidth]{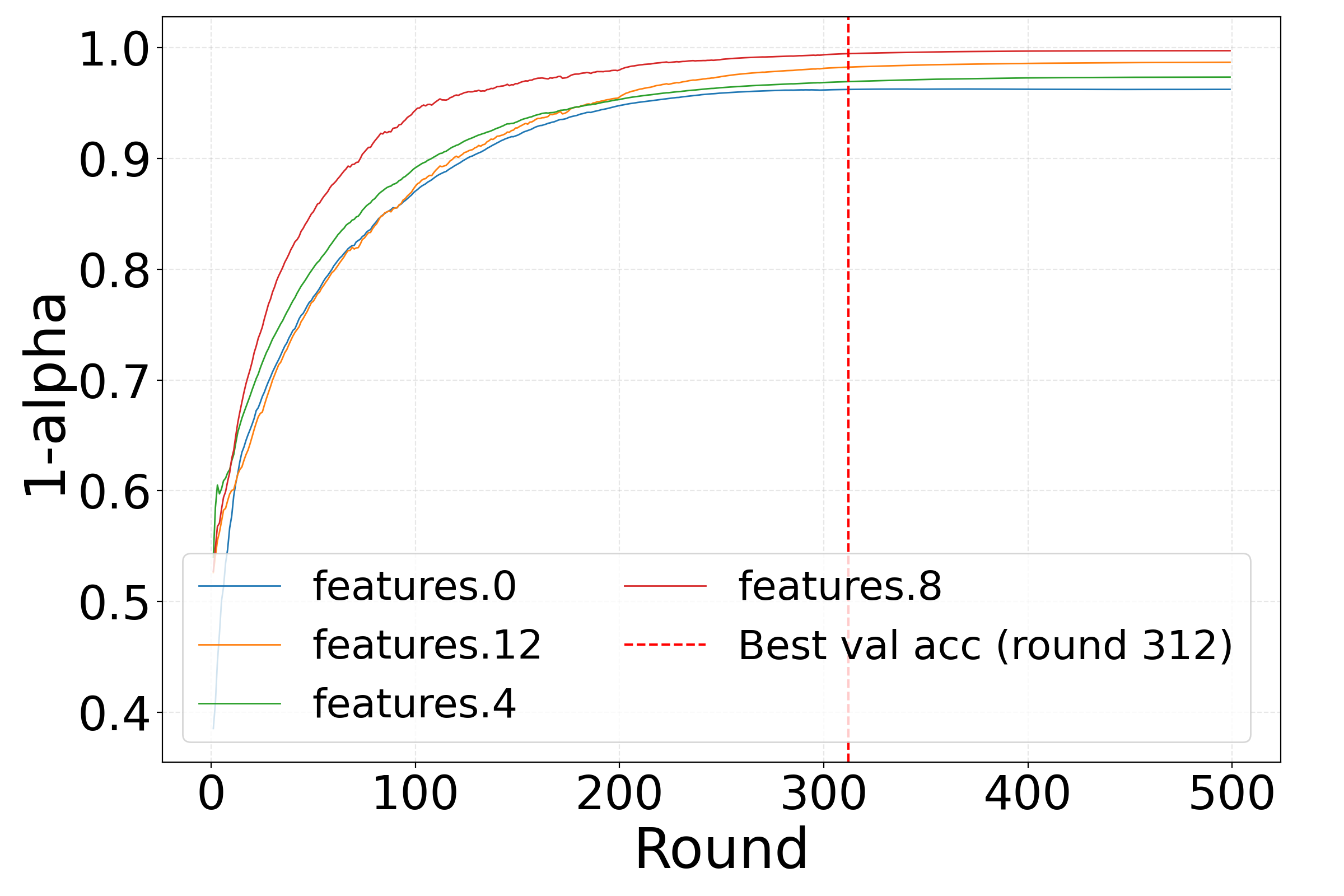}
\caption{Non-IID2: $(1-\alpha)$}
\end{subfigure}

\vspace{0.0em}

% Row 2: alpha beta
\begin{subfigure}{0.32\textwidth}
\includegraphics[width=\linewidth]{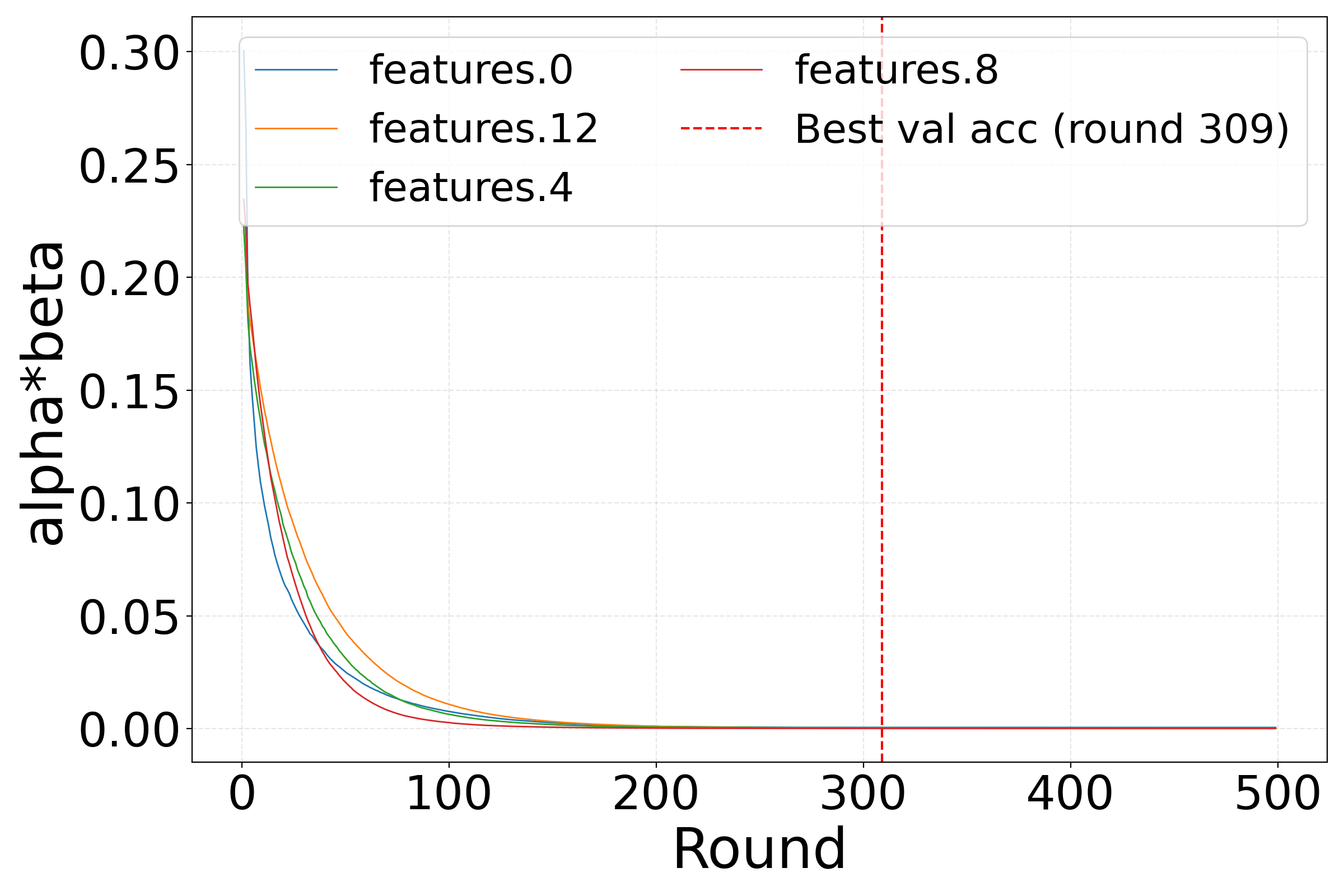}
\caption{IID: $\alpha\beta$}
\end{subfigure}
\begin{subfigure}{0.32\textwidth}
\includegraphics[width=\linewidth]{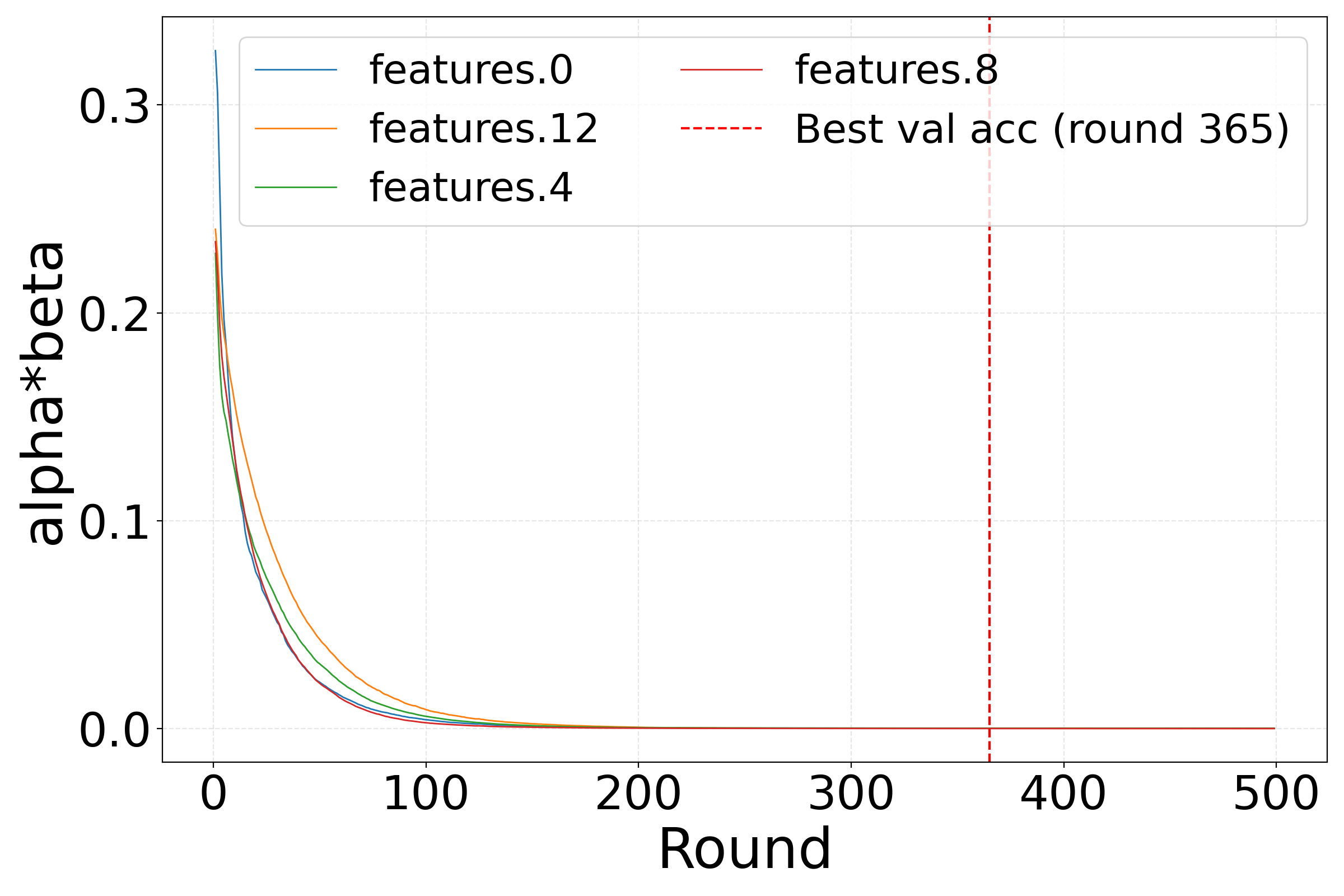}
\caption{Non-IID1: $\alpha\beta$}
\end{subfigure}
\begin{subfigure}{0.32\textwidth}
\includegraphics[width=\linewidth]{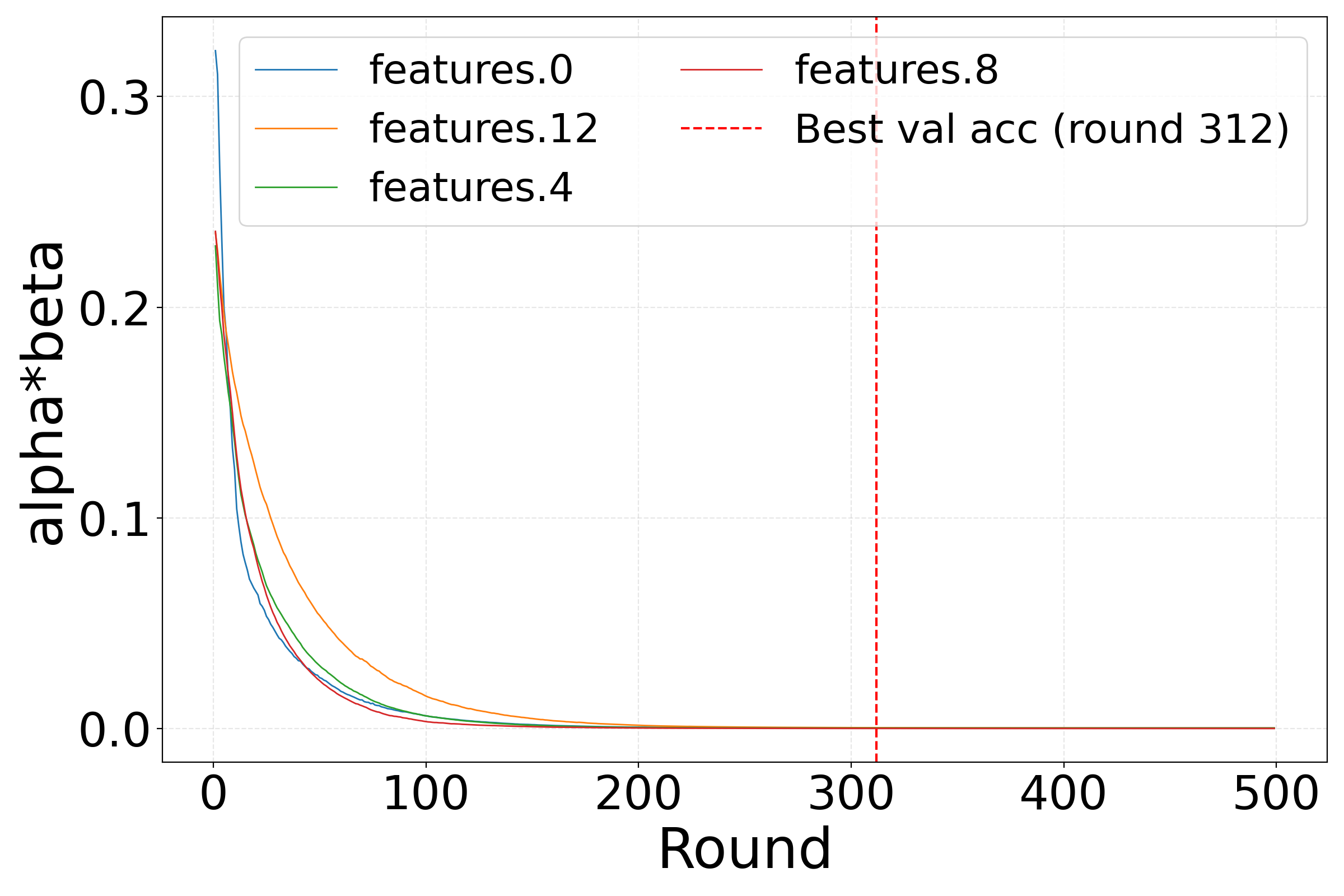}
\caption{Non-IID2: $\alpha\beta$}
\end{subfigure}

\vspace{0.0em}

% Row 3: alpha(1-beta)
\begin{subfigure}{0.32\textwidth}
\includegraphics[width=\linewidth]{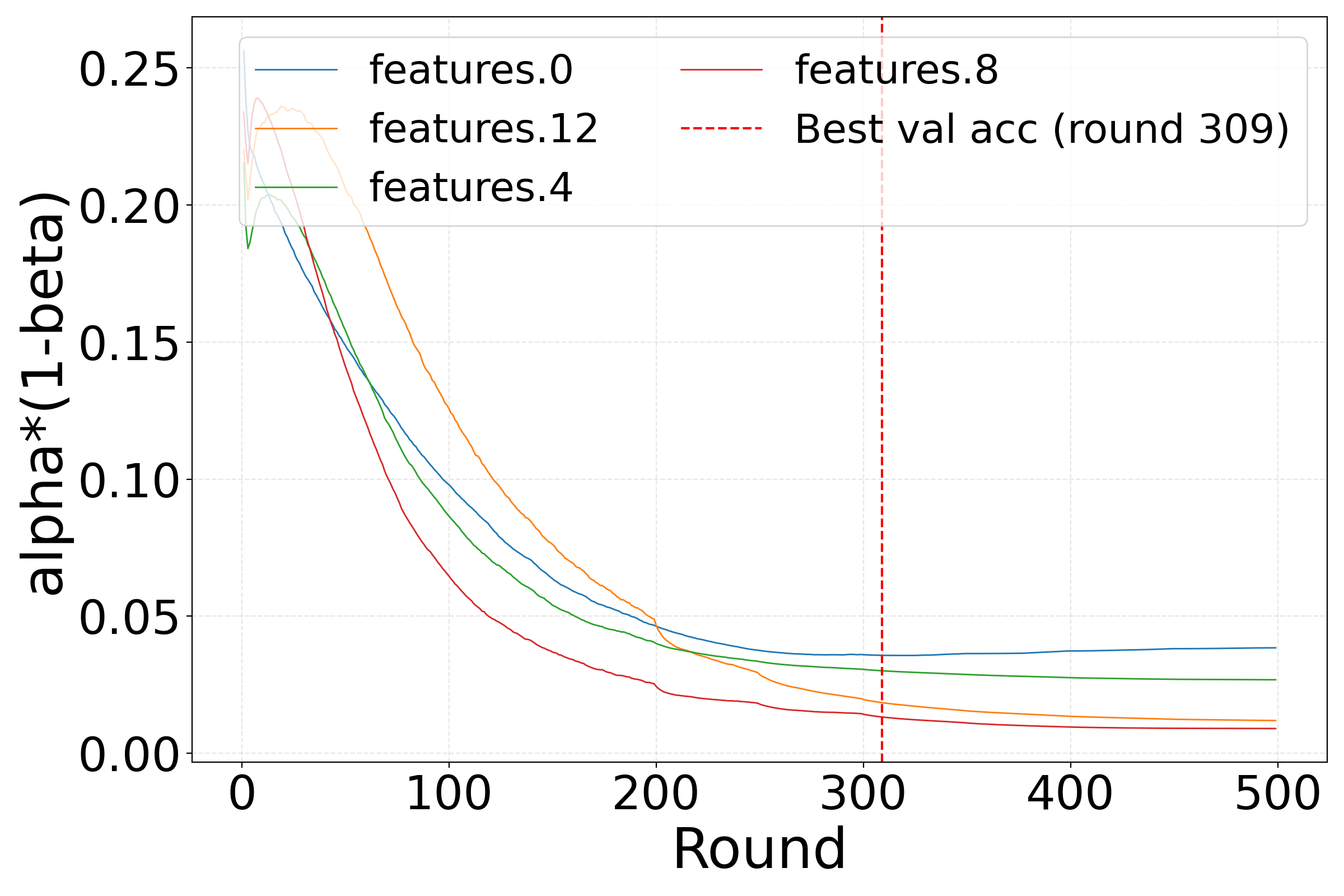}
\caption{IID: $\alpha(1-\beta)$}
\end{subfigure}
\begin{subfigure}{0.32\textwidth}
\includegraphics[width=\linewidth]{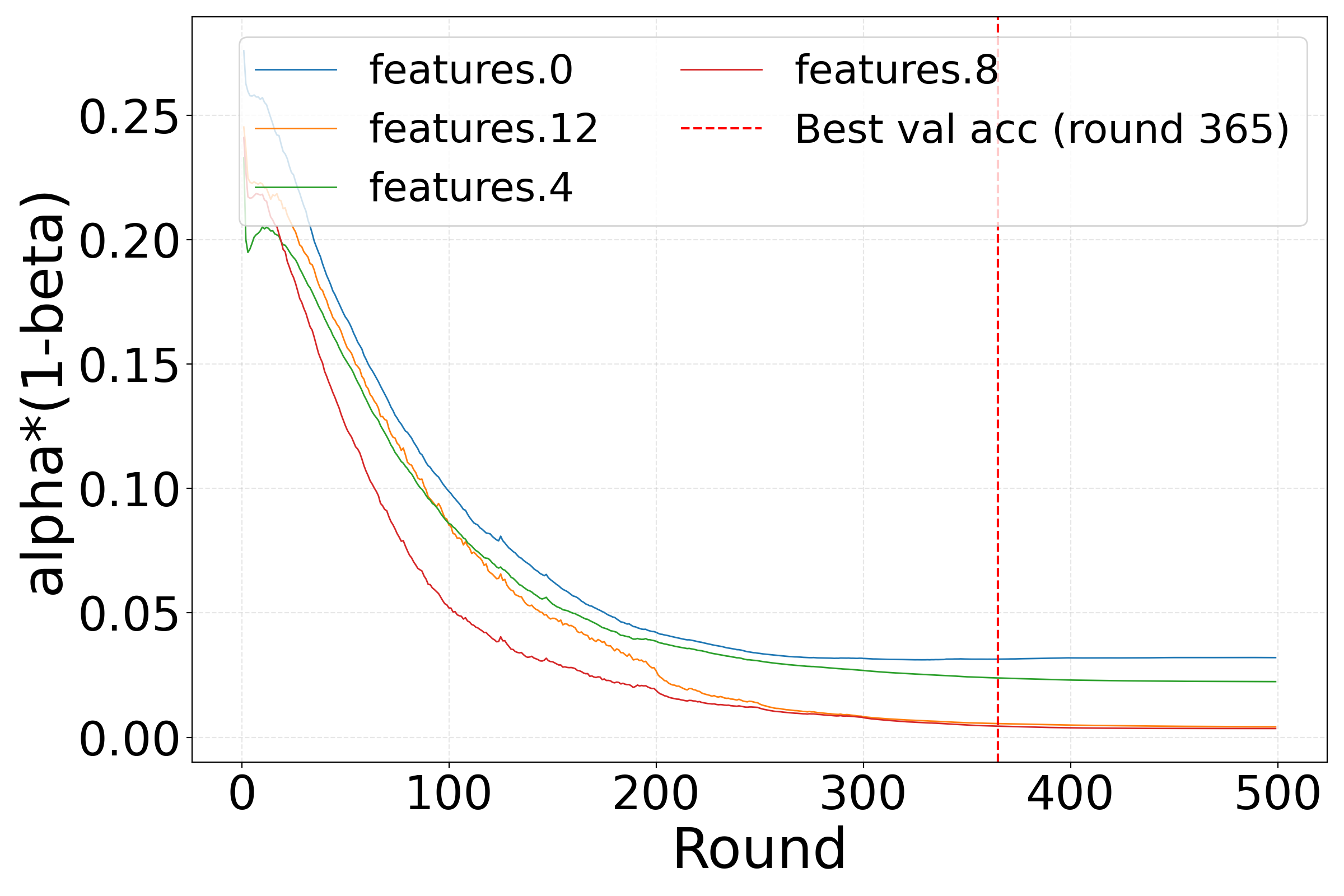}
\caption{Non-IID1: $\alpha(1-\beta)$}
\end{subfigure}
\begin{subfigure}{0.32\textwidth}
\includegraphics[width=\linewidth]{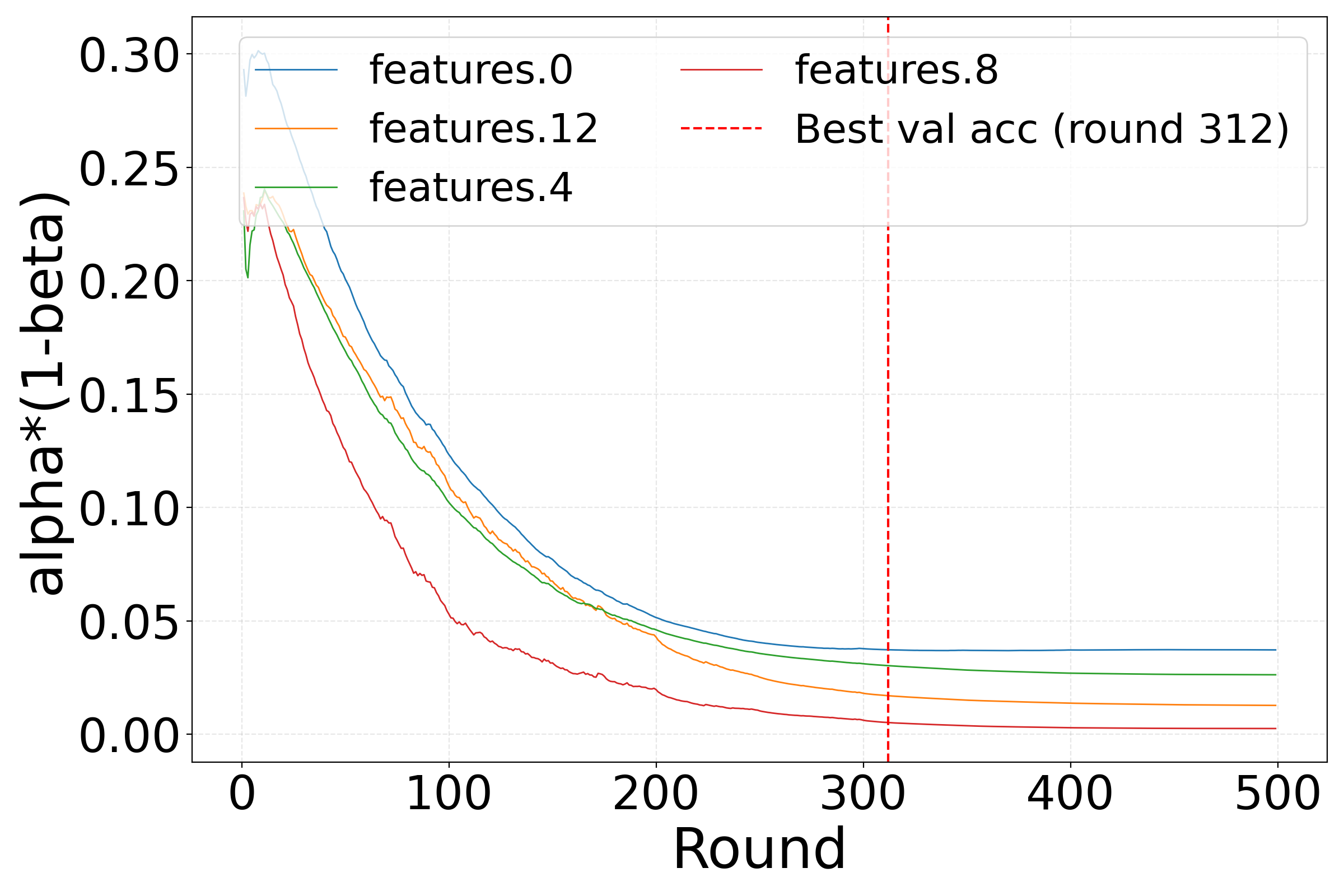}
\caption{Non-IID2: $\alpha(1-\beta)$}
\end{subfigure}

\vspace{0.0em}

% Row 4: lambda
\begin{subfigure}{0.32\textwidth}
\includegraphics[width=\linewidth]{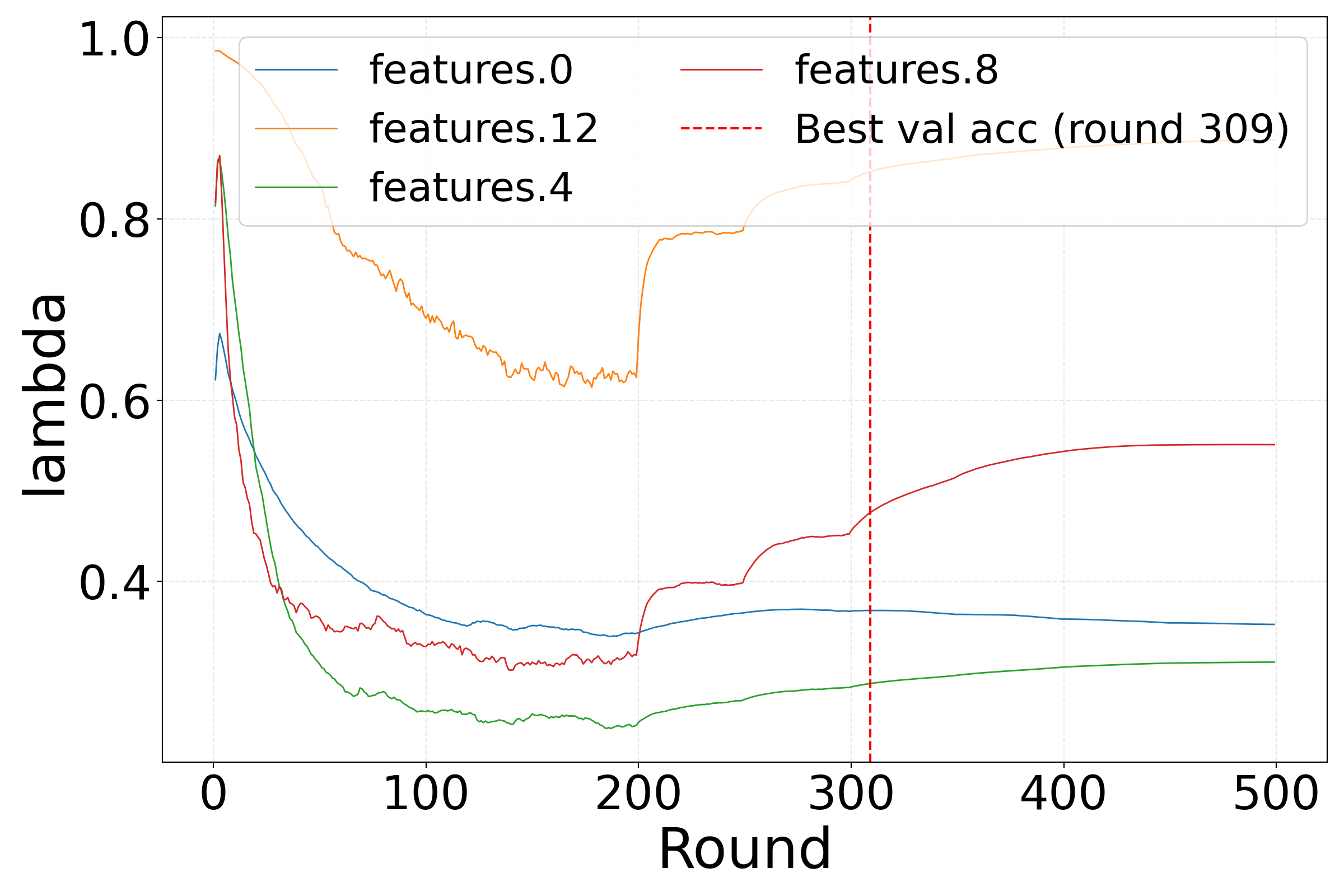}
\caption{IID: $\lambda$}
\end{subfigure}
\begin{subfigure}{0.32\textwidth}
\includegraphics[width=\linewidth]{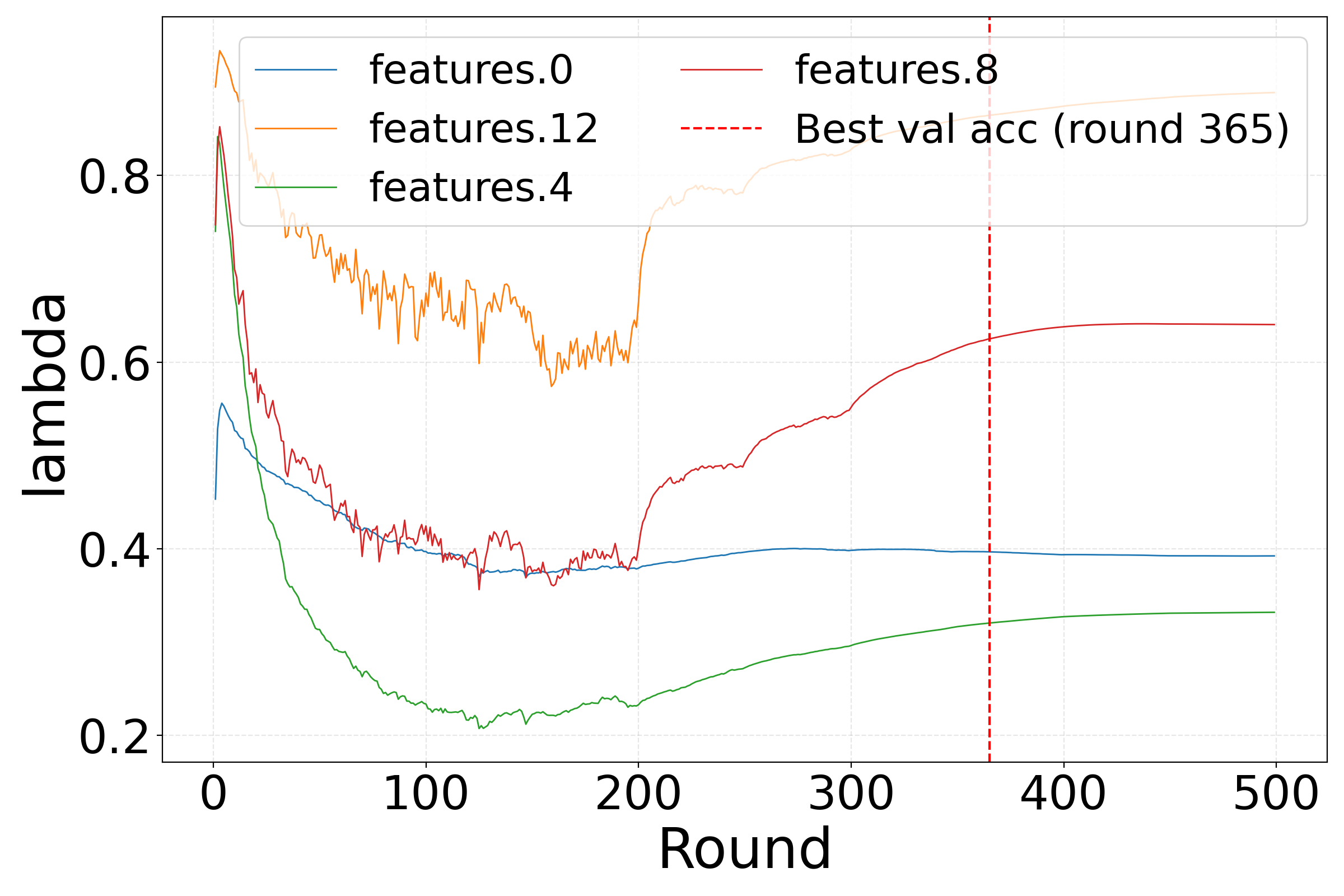}
\caption{Non-IID1: $\lambda$}
\end{subfigure}
\begin{subfigure}{0.32\textwidth}
\includegraphics[width=\linewidth]{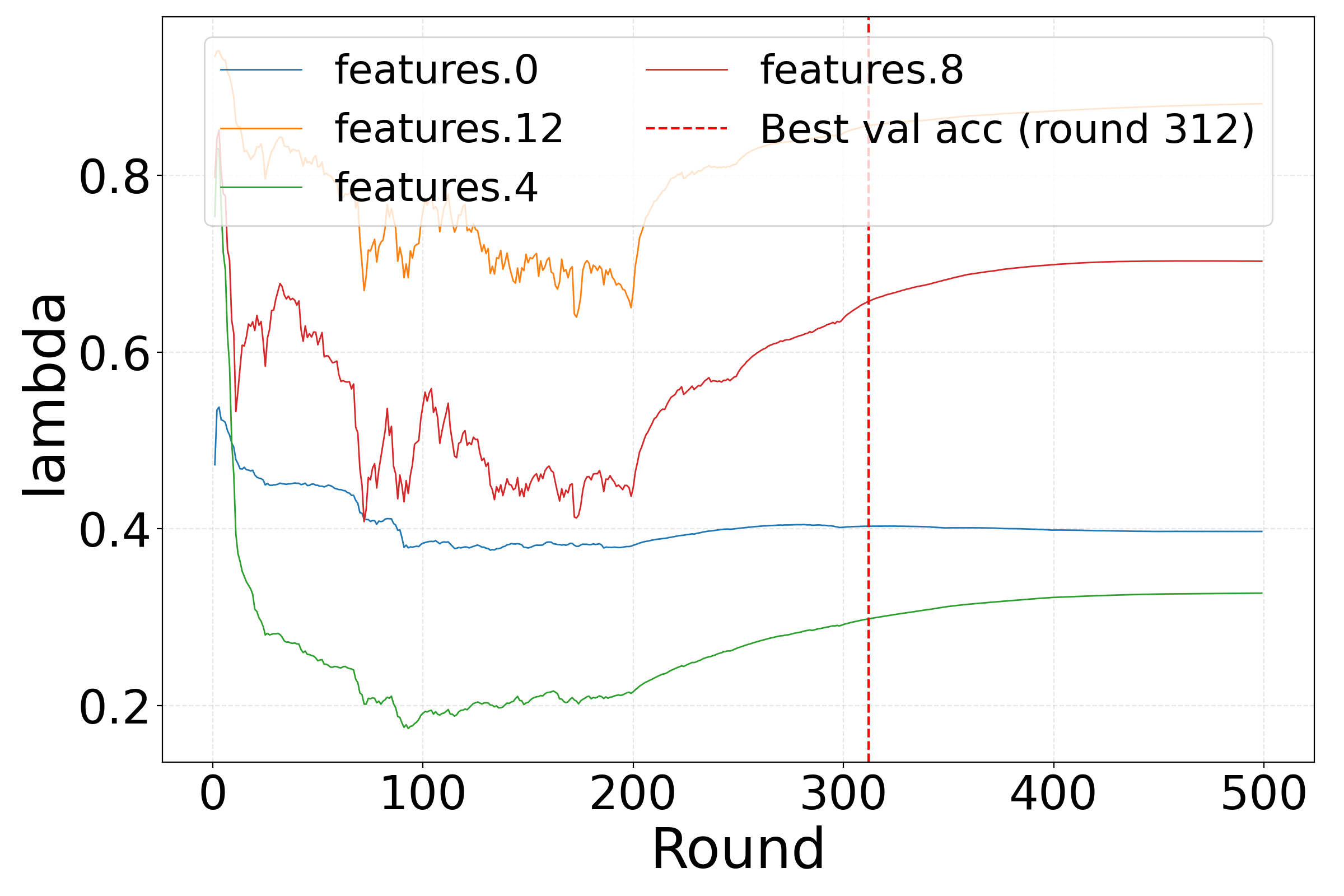}
\caption{Non-IID2: $\lambda$}
\end{subfigure}

\caption{Mixing parameter variation on SVHN under IID, Non-IID1, and Non-IID2.}
\label{alphabetalambdasvhn}
\end{figure*}

\begin{figure*}[htbp]
\centering

% Row 1: (1-alpha)
\begin{subfigure}{0.32\textwidth}
\includegraphics[width=\linewidth]{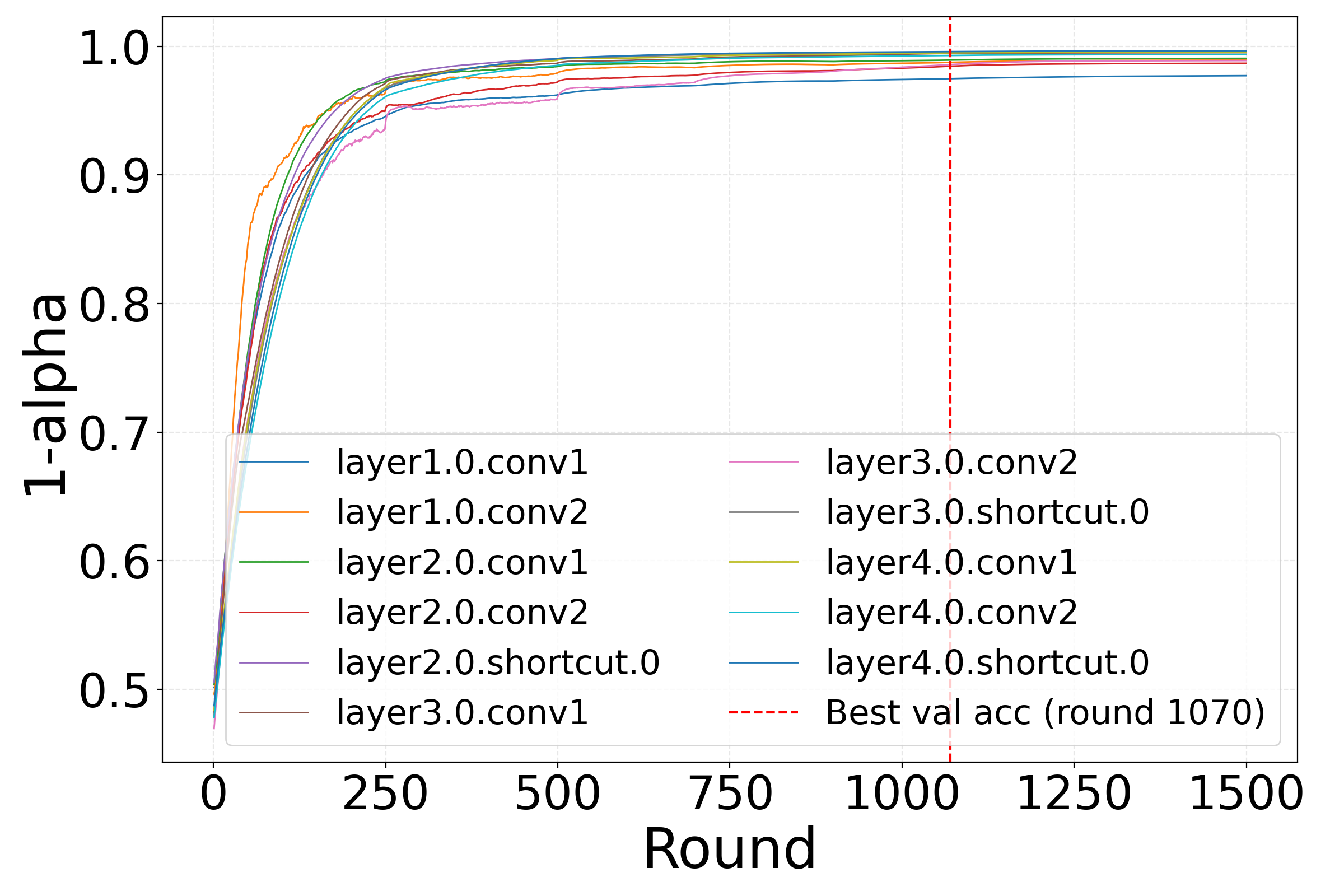}
\caption{IID: $(1-\alpha)$}
\end{subfigure}
\begin{subfigure}{0.32\textwidth}
\includegraphics[width=\linewidth]{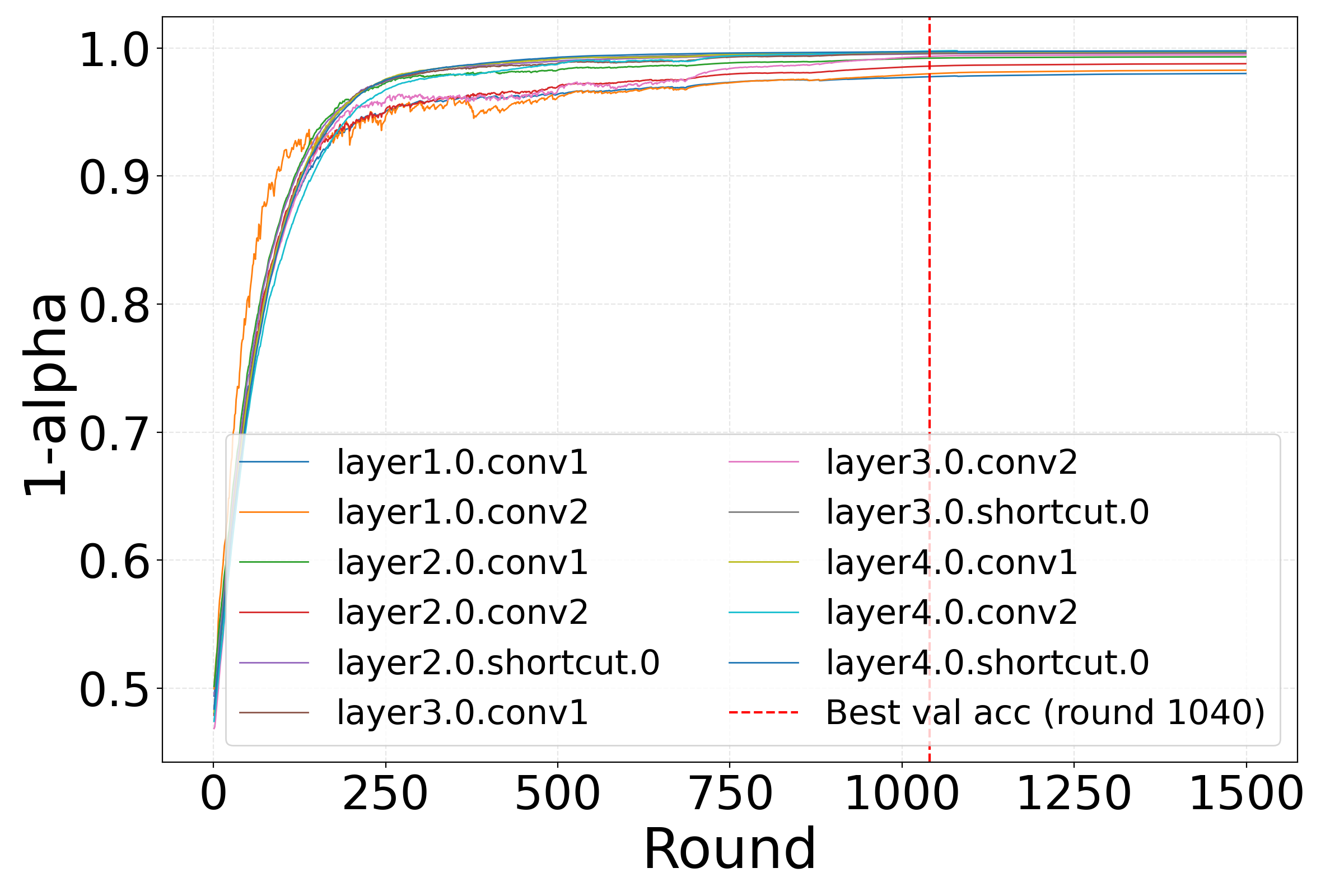}
\caption{Non-IID1: $(1-\alpha)$}
\end{subfigure}
\begin{subfigure}{0.32\textwidth}
\includegraphics[width=\linewidth]{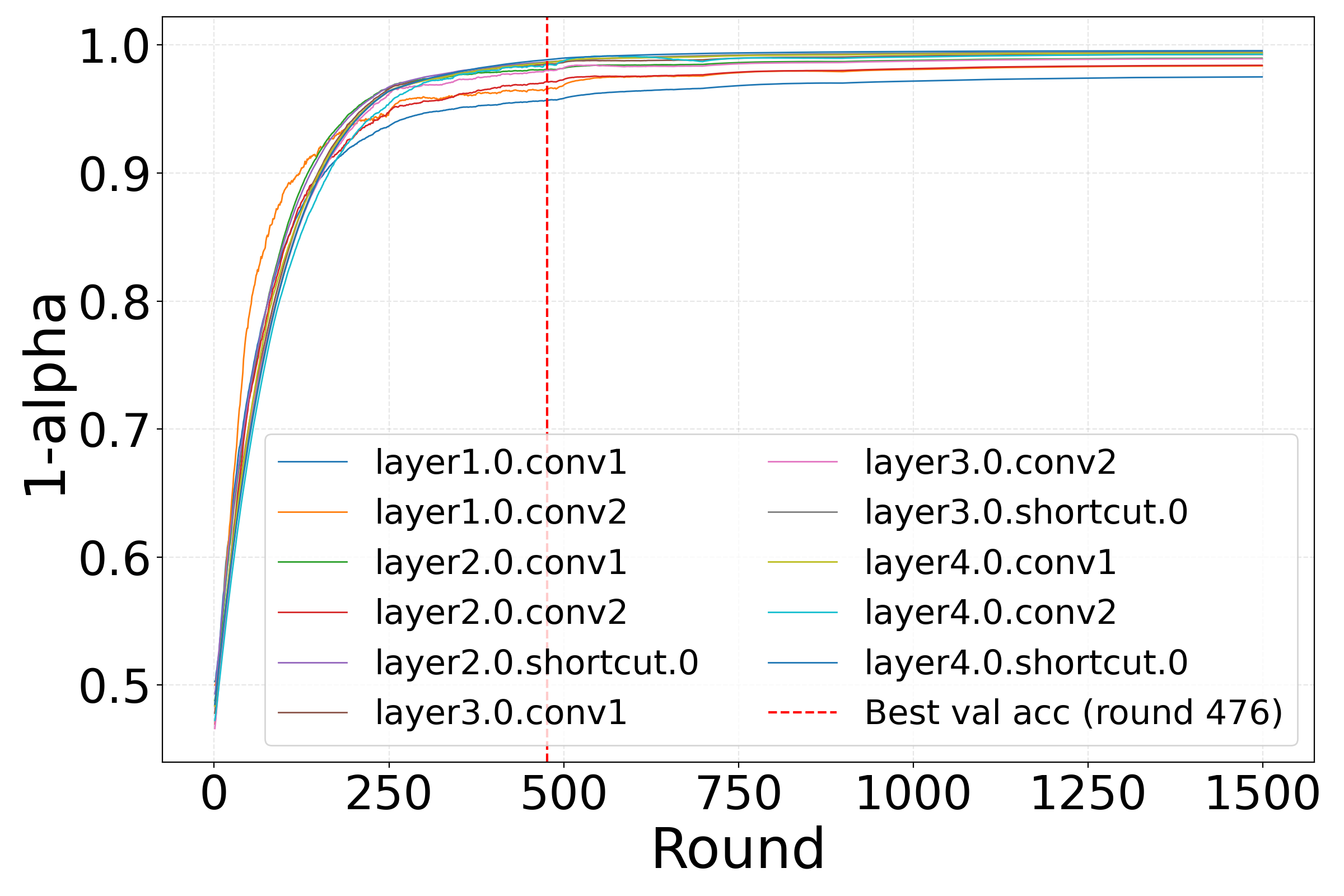}
\caption{Non-IID2: $(1-\alpha)$}
\end{subfigure}

\vspace{0.6em}

% Row 2: alpha beta
\begin{subfigure}{0.32\textwidth}
\includegraphics[width=\linewidth]{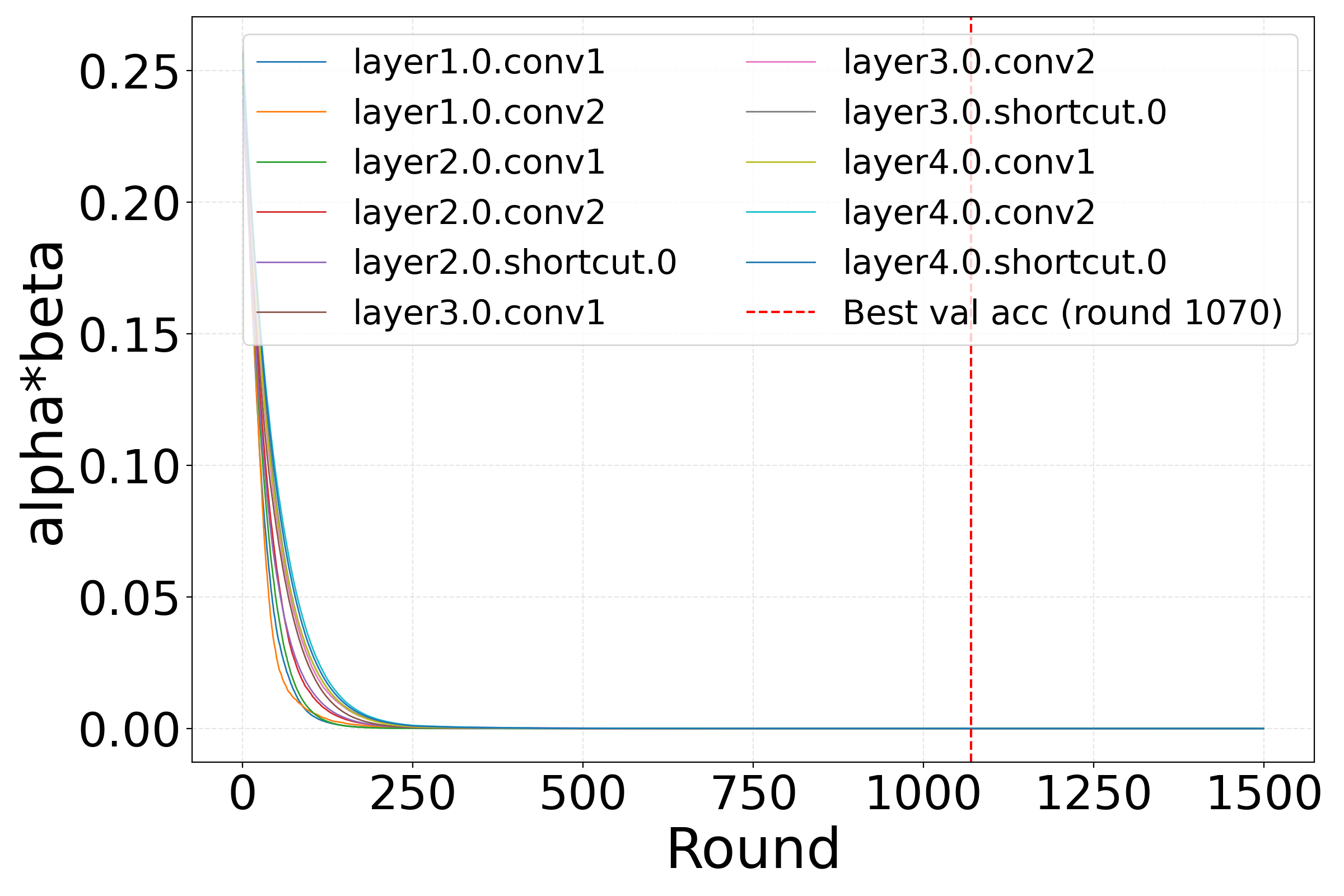}
\caption{IID: $\alpha\beta$}
\end{subfigure}
\begin{subfigure}{0.32\textwidth}
\includegraphics[width=\linewidth]{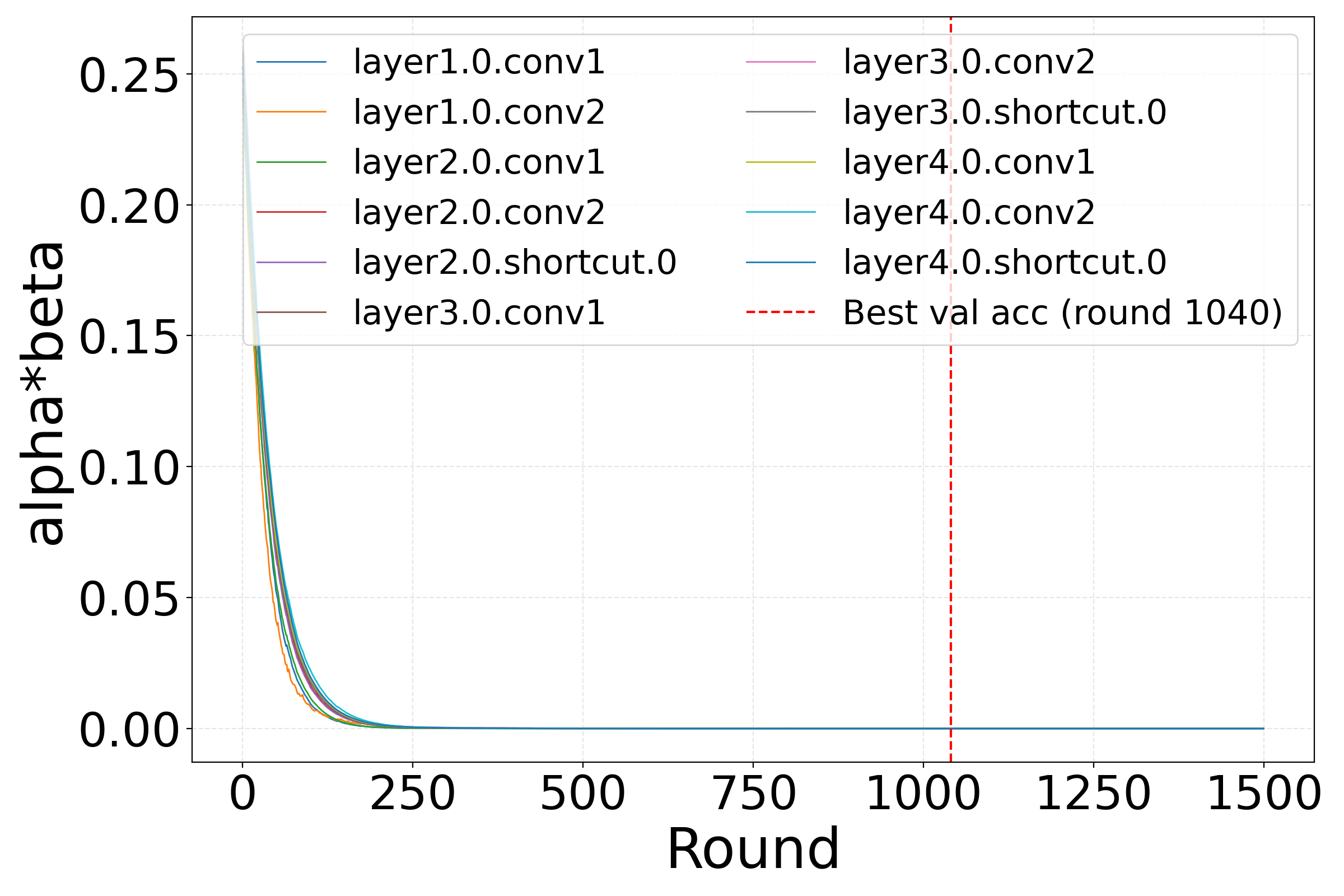}
\caption{Non-IID1: $\alpha\beta$}
\end{subfigure}
\begin{subfigure}{0.32\textwidth}
\includegraphics[width=\linewidth]{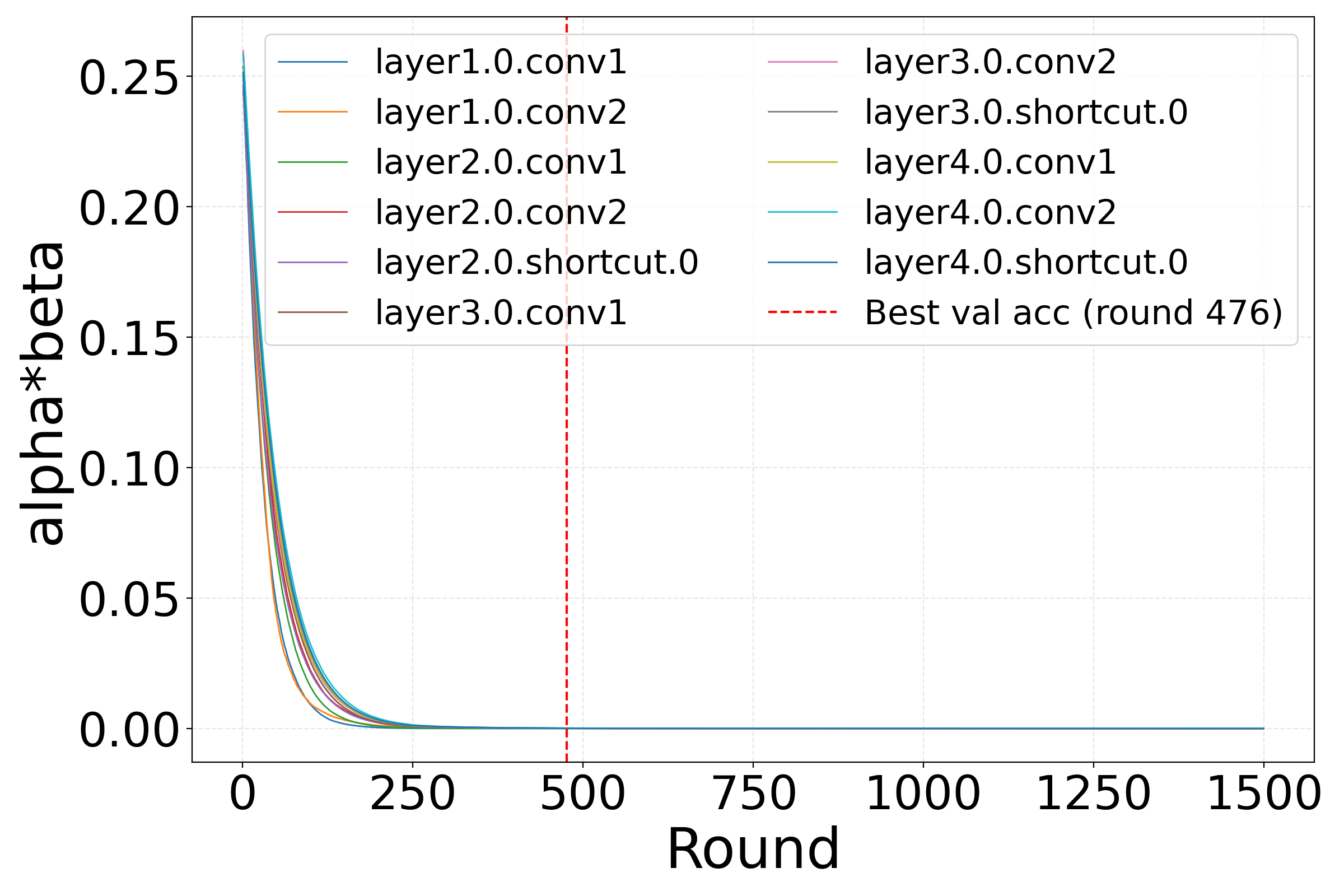}
\caption{Non-IID2: $\alpha\beta$}
\end{subfigure}

\vspace{0.6em}

% Row 3: alpha(1-beta)
\begin{subfigure}{0.32\textwidth}
\includegraphics[width=\linewidth]{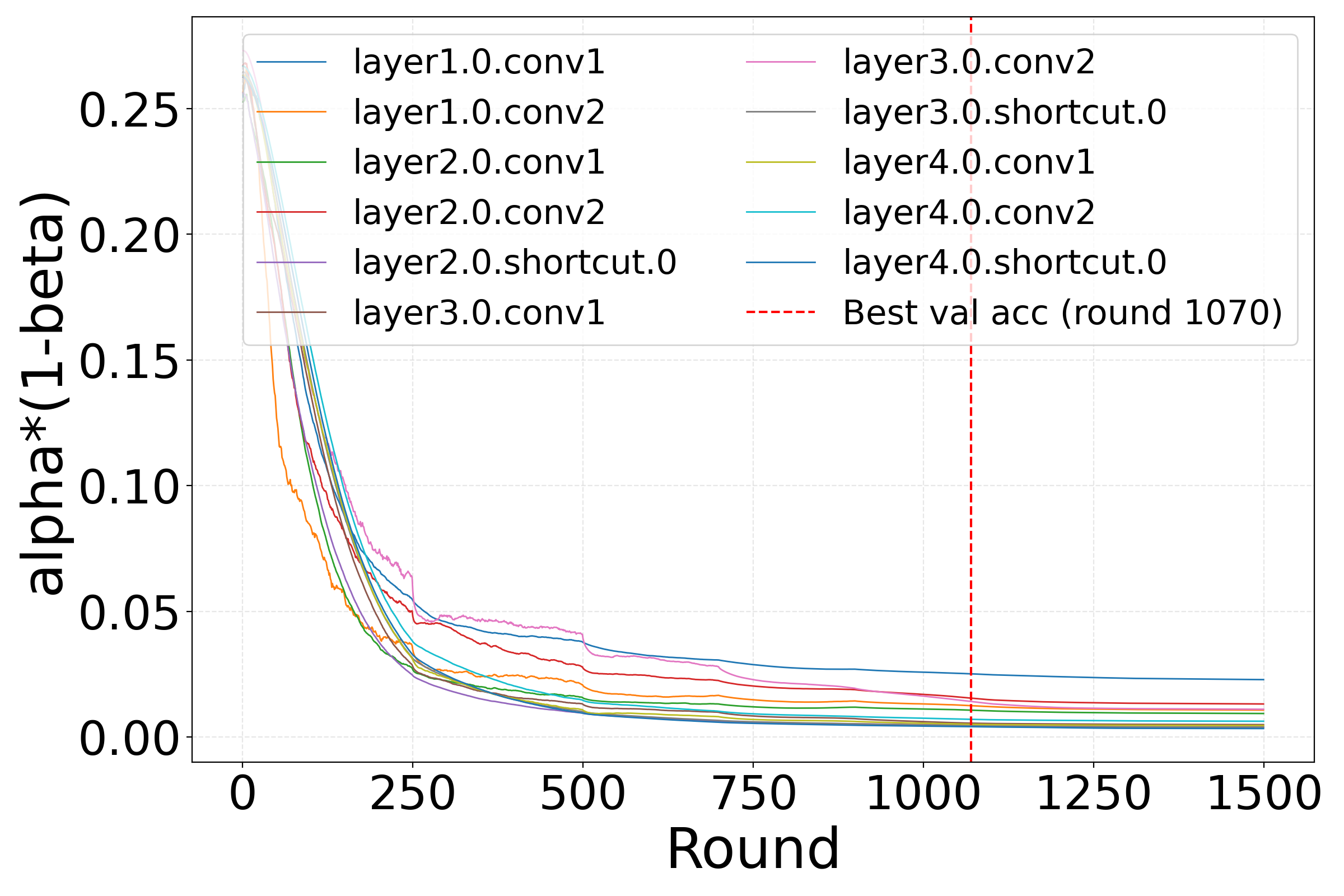}
\caption{IID: $\alpha(1-\beta)$}
\end{subfigure}
\begin{subfigure}{0.32\textwidth}
\includegraphics[width=\linewidth]{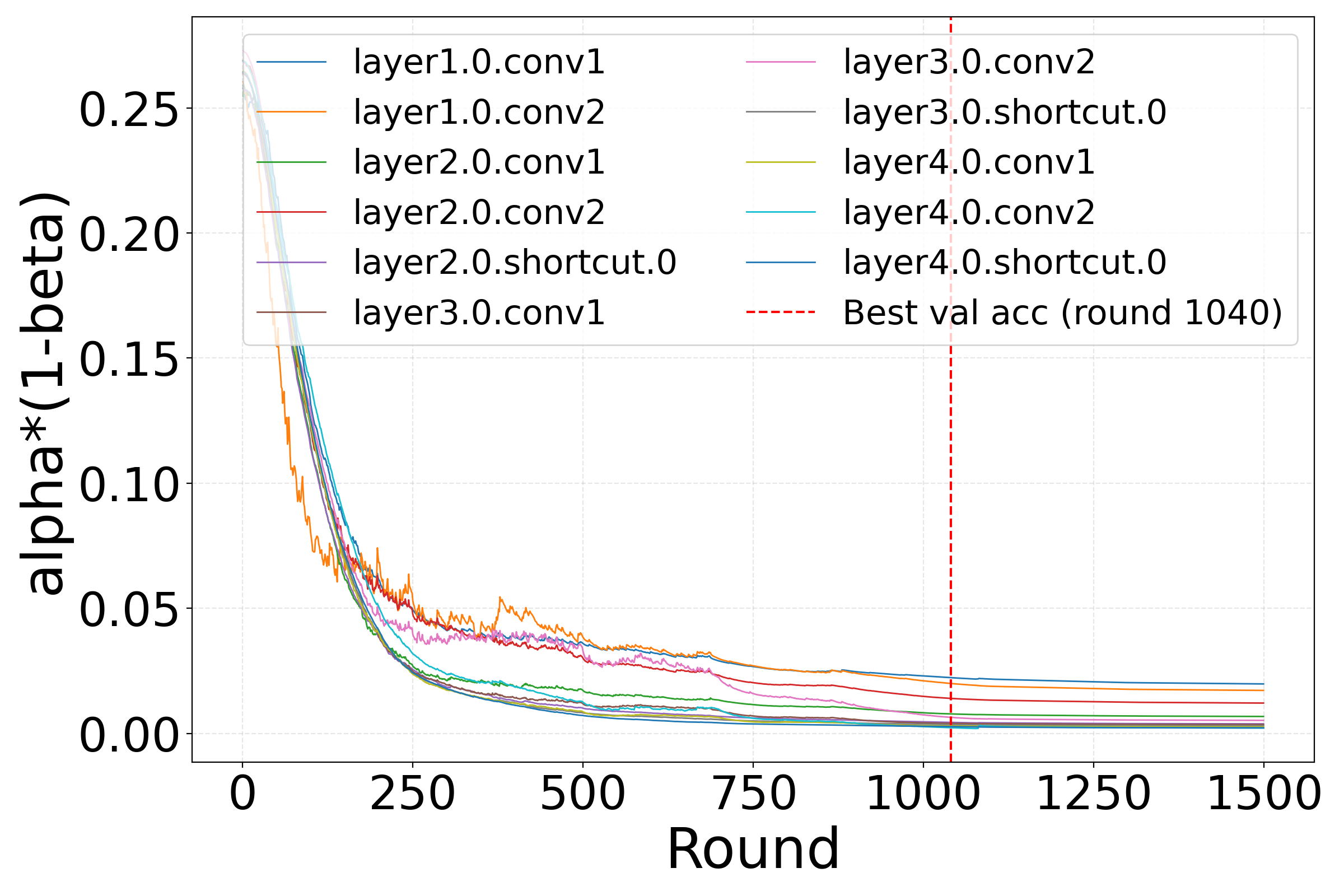}
\caption{Non-IID1: $\alpha(1-\beta)$}
\end{subfigure}
\begin{subfigure}{0.32\textwidth}
\includegraphics[width=\linewidth]{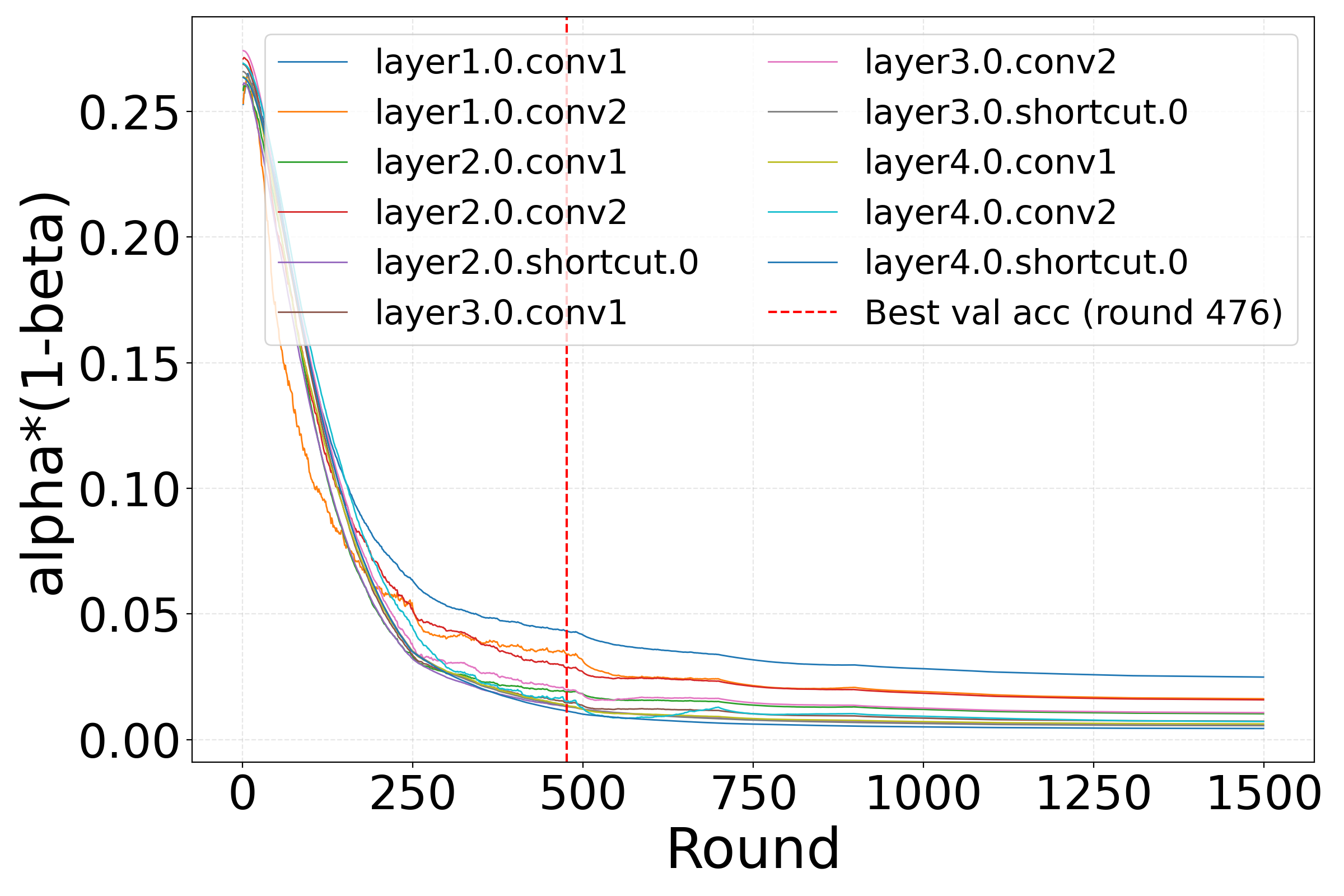}
\caption{Non-IID2: $\alpha(1-\beta)$}
\end{subfigure}

\vspace{0.6em}

% Row 4: lambda
\begin{subfigure}{0.32\textwidth}
\includegraphics[width=\linewidth]{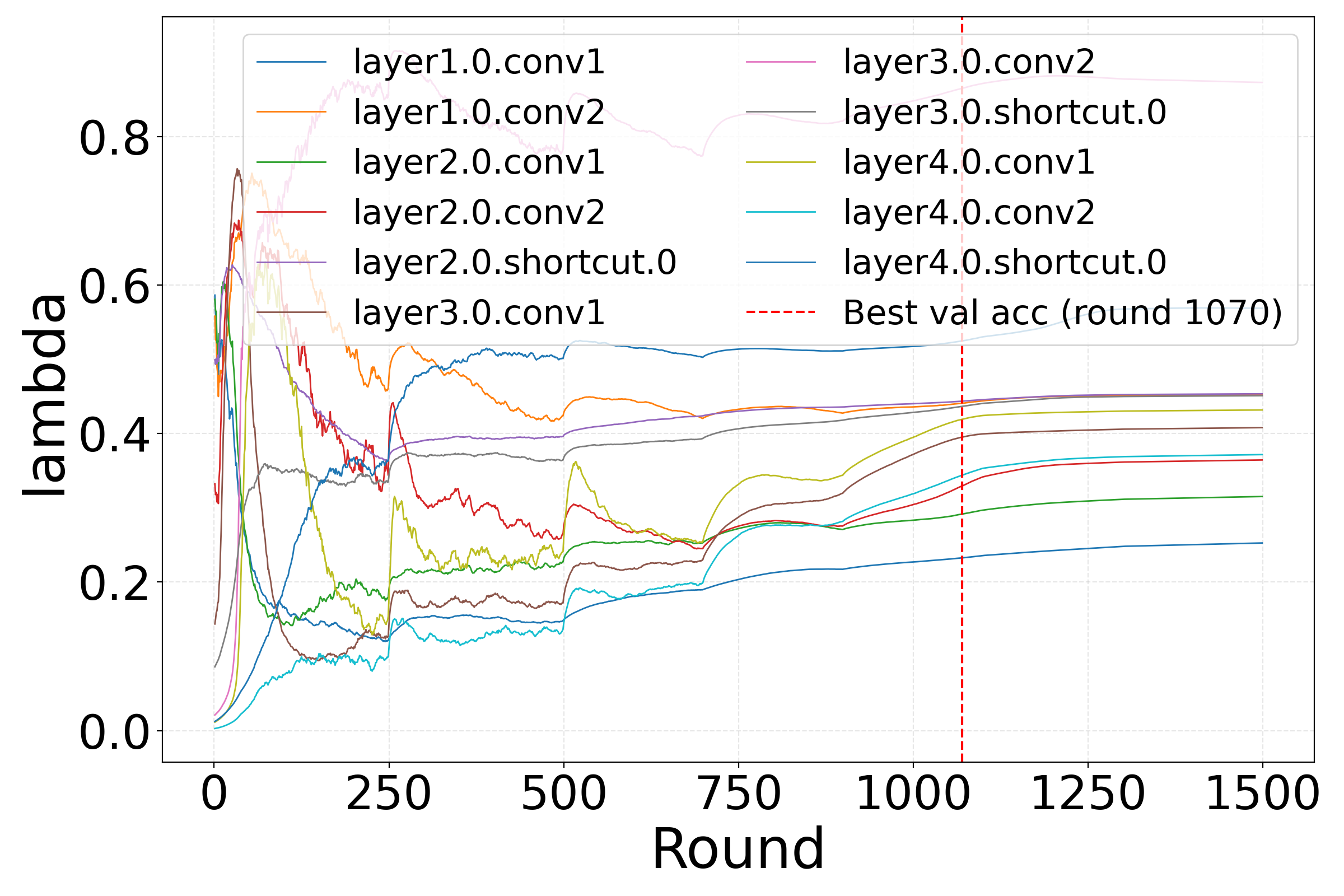}
\caption{IID: $\lambda$}
\end{subfigure}
\begin{subfigure}{0.32\textwidth}
\includegraphics[width=\linewidth]{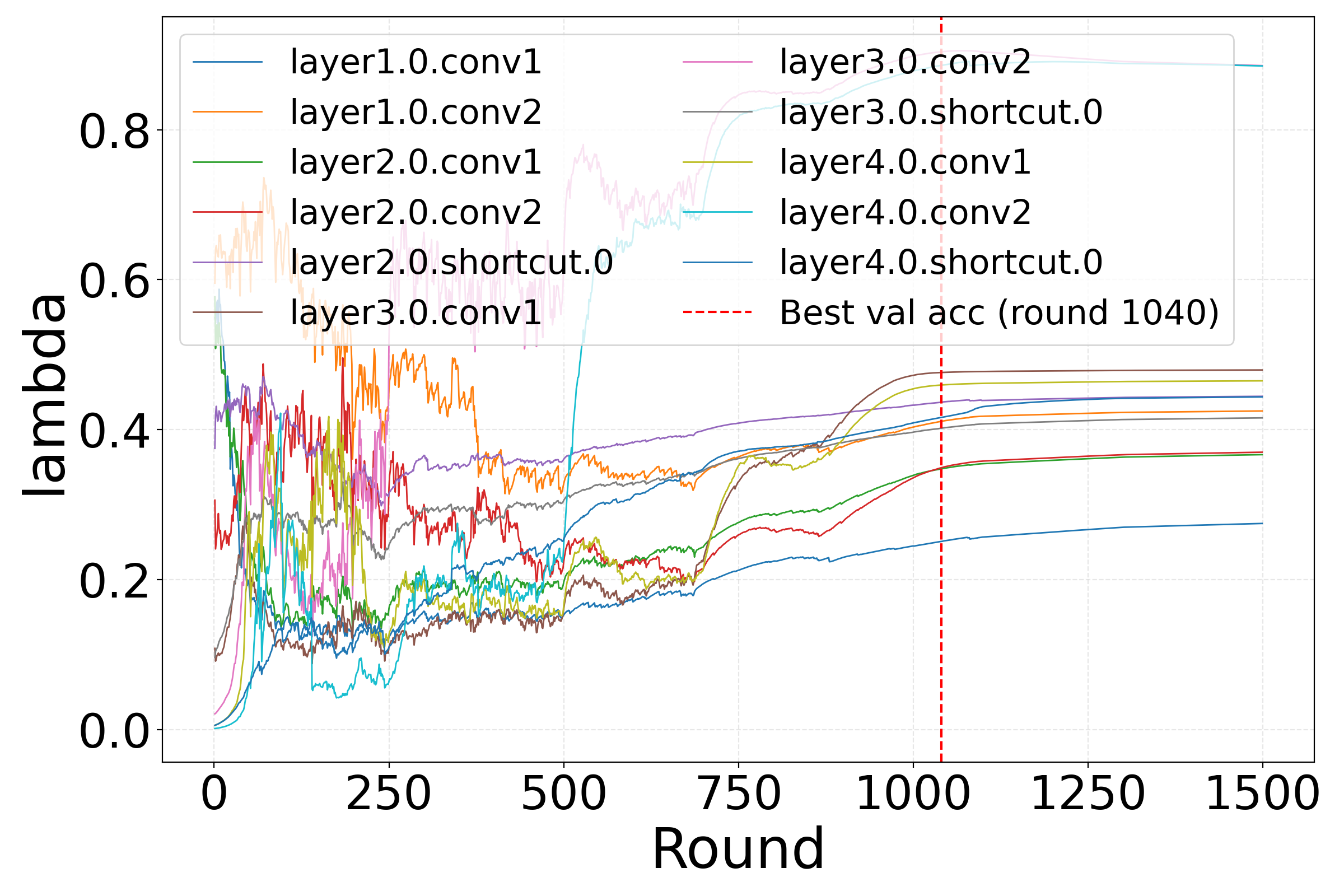}
\caption{Non-IID1: $\lambda$}
\end{subfigure}
\begin{subfigure}{0.32\textwidth}
\includegraphics[width=\linewidth]{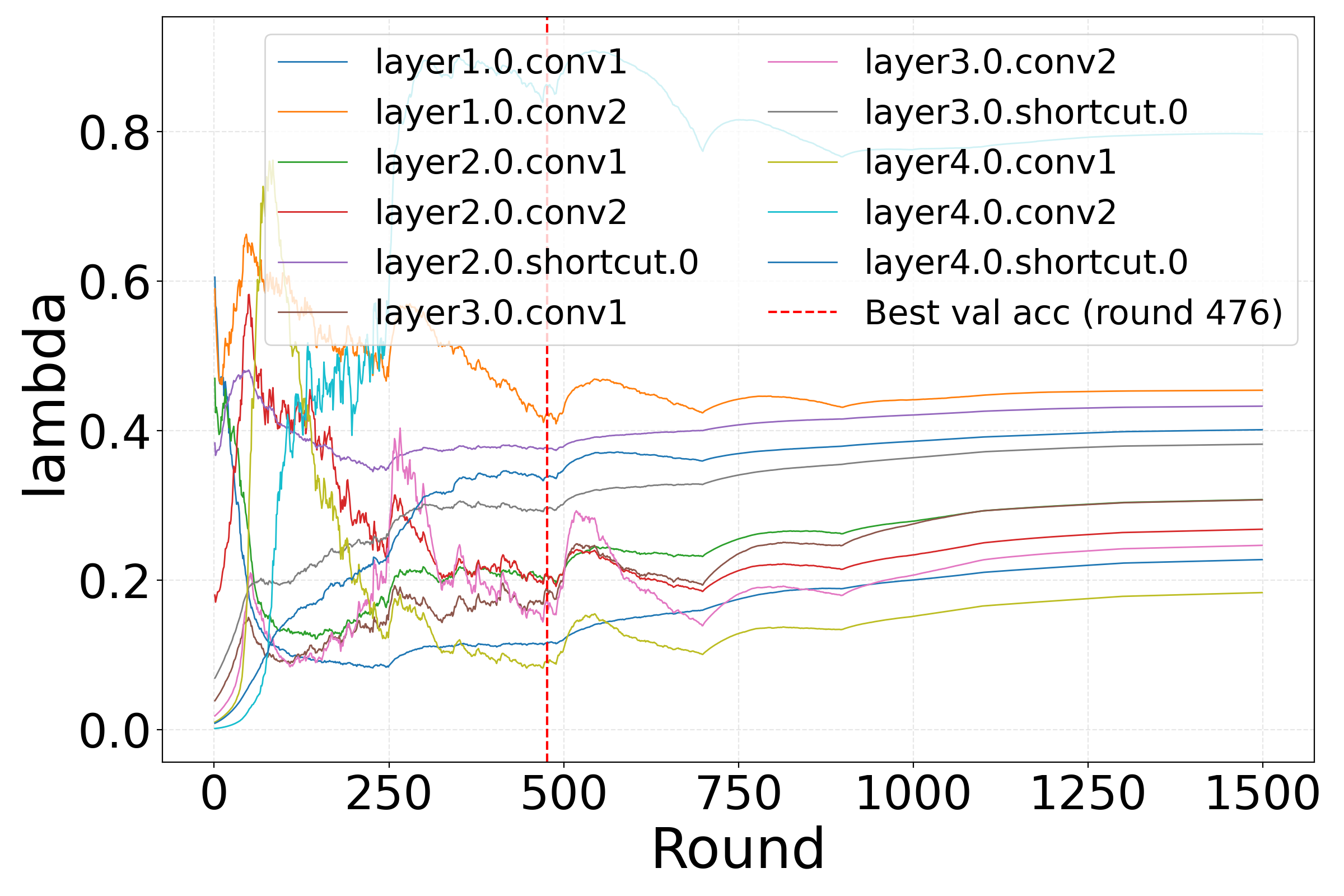}
\caption{Non-IID2: $\lambda$}
\end{subfigure}

\caption{Mixing parameter variation on CIFAR10 (ResNet10) under IID, Non-IID1, and Non-IID2 data heterogeneity settings.}
\label{alphabetalambdacifar10resnet10}
\end{figure*}

\begin{figure*}[t]
\centering

% -------- Row 1: (1-alpha) --------
\begin{subfigure}{0.32\textwidth}
\centering
\includegraphics[width=\linewidth]{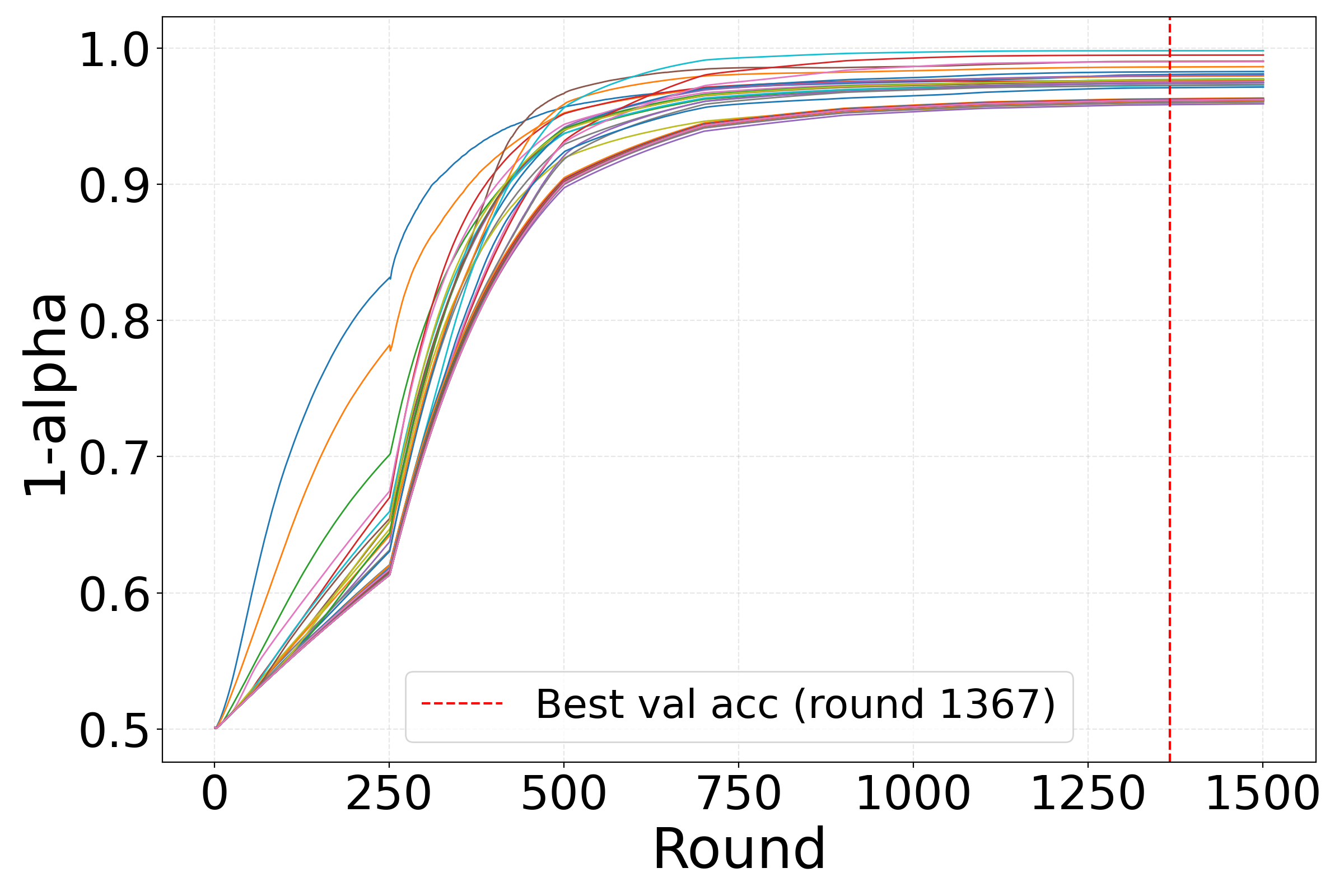}
\caption{IID: $(1-\alpha)$}
\end{subfigure}
\begin{subfigure}{0.32\textwidth}
\centering
\includegraphics[width=\linewidth]{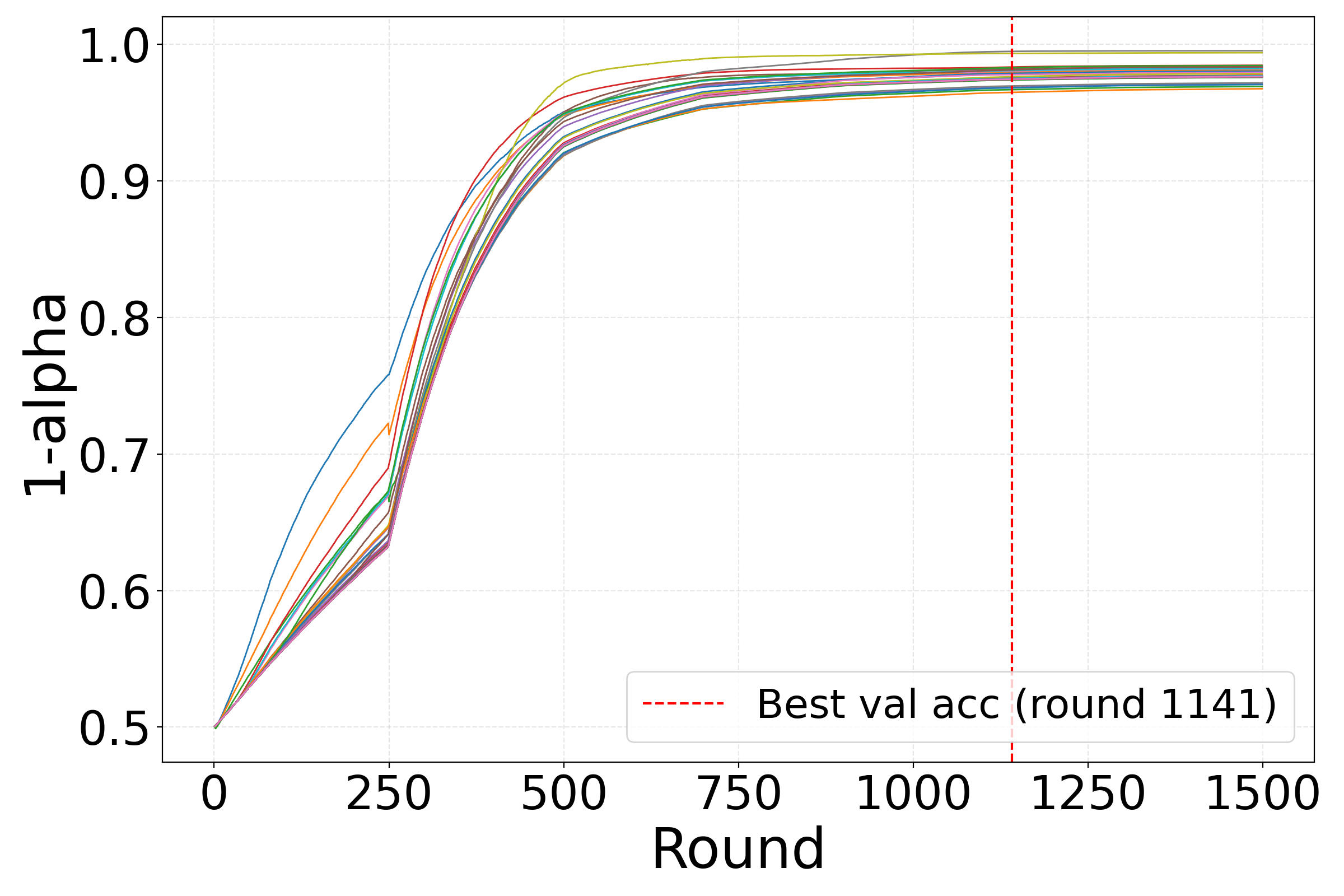}
\caption{Non-IID1: $(1-\alpha)$}
\end{subfigure}
\begin{subfigure}{0.32\textwidth}
\centering
\includegraphics[width=\linewidth]{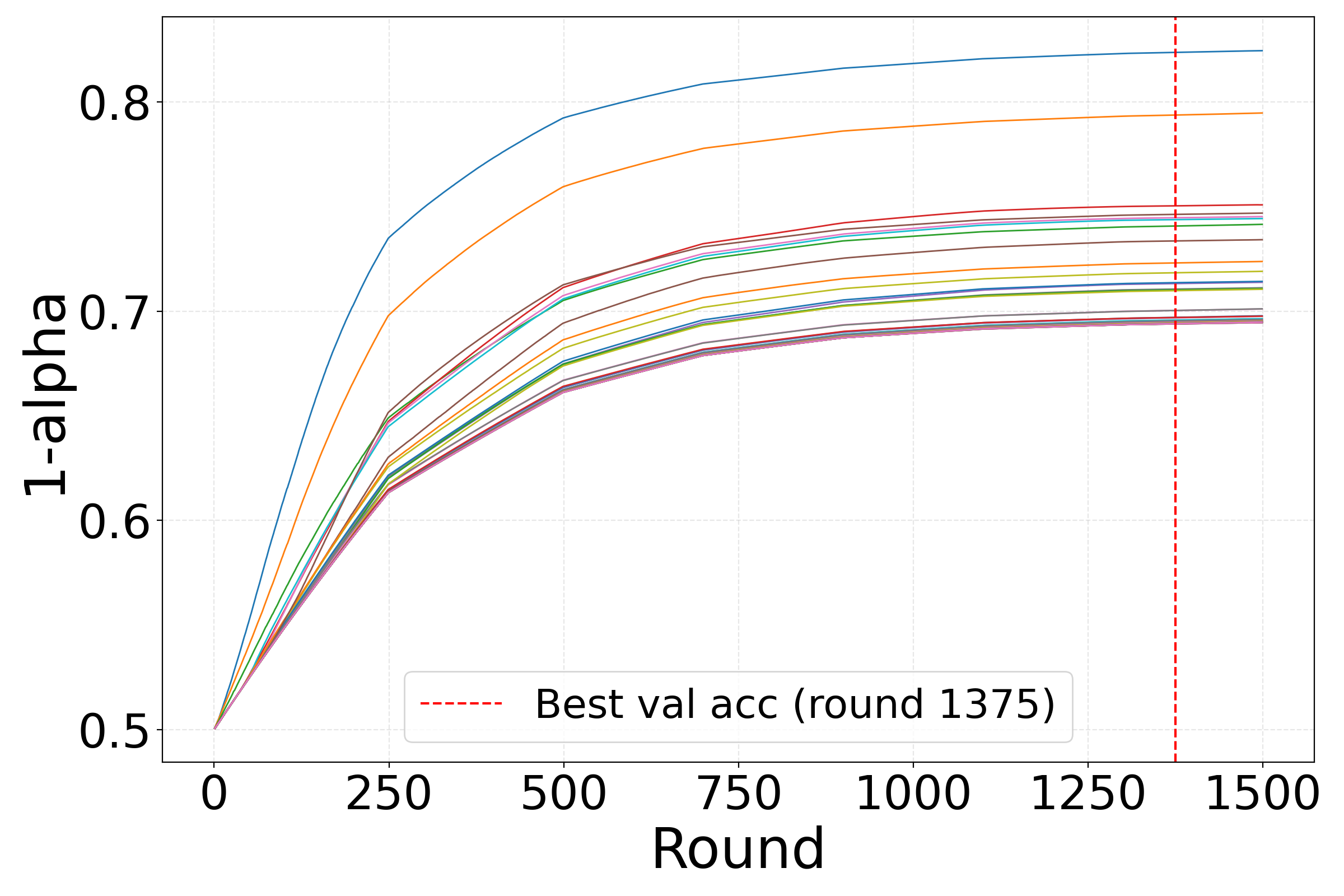}
\caption{Non-IID2: $(1-\alpha)$}
\end{subfigure}

\vspace{0.6em}

% -------- Row 2: alpha beta --------
\begin{subfigure}{0.32\textwidth}
\centering
\includegraphics[width=\linewidth]{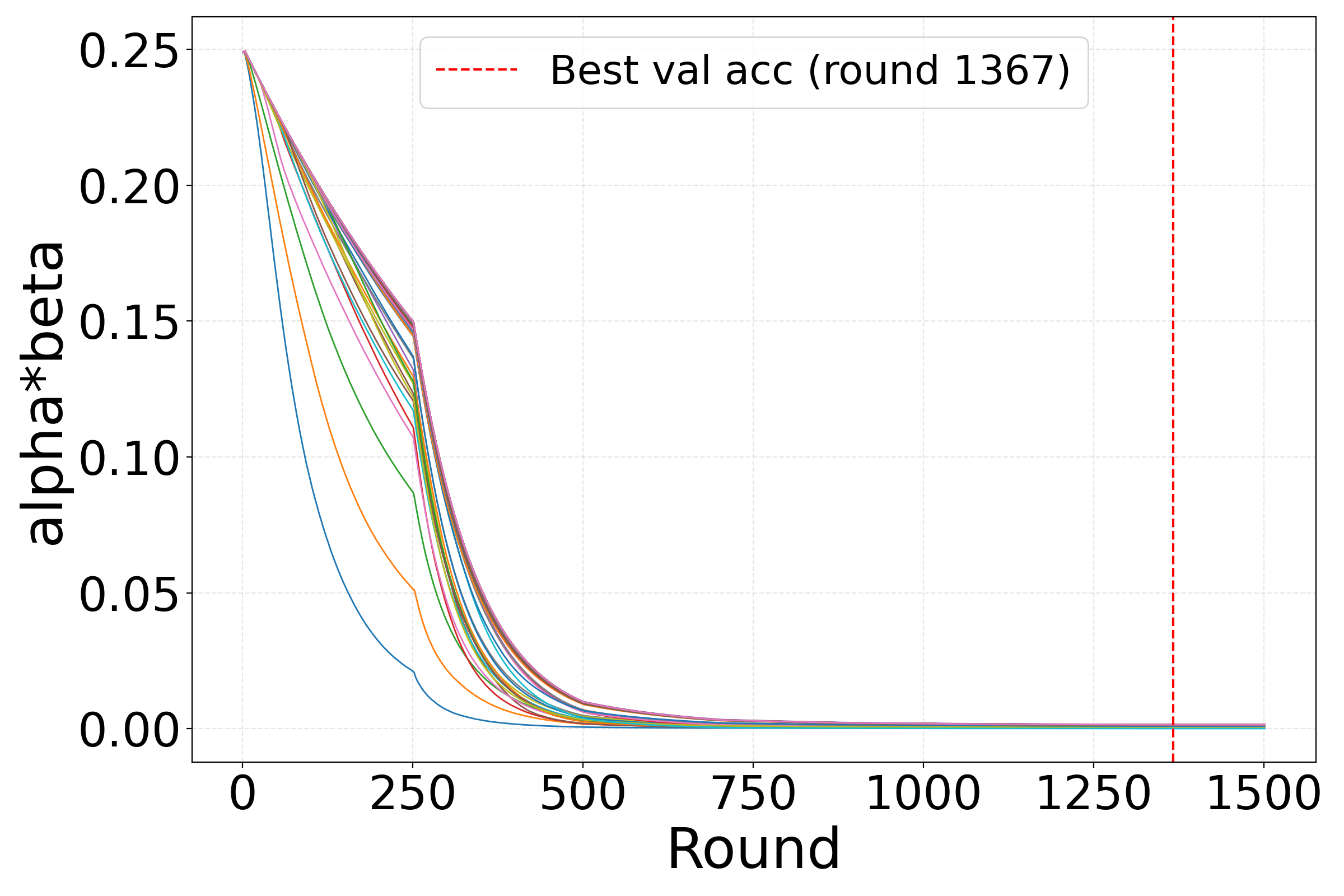}
\caption{IID: $\alpha\beta$}
\end{subfigure}
\begin{subfigure}{0.32\textwidth}
\centering
\includegraphics[width=\linewidth]{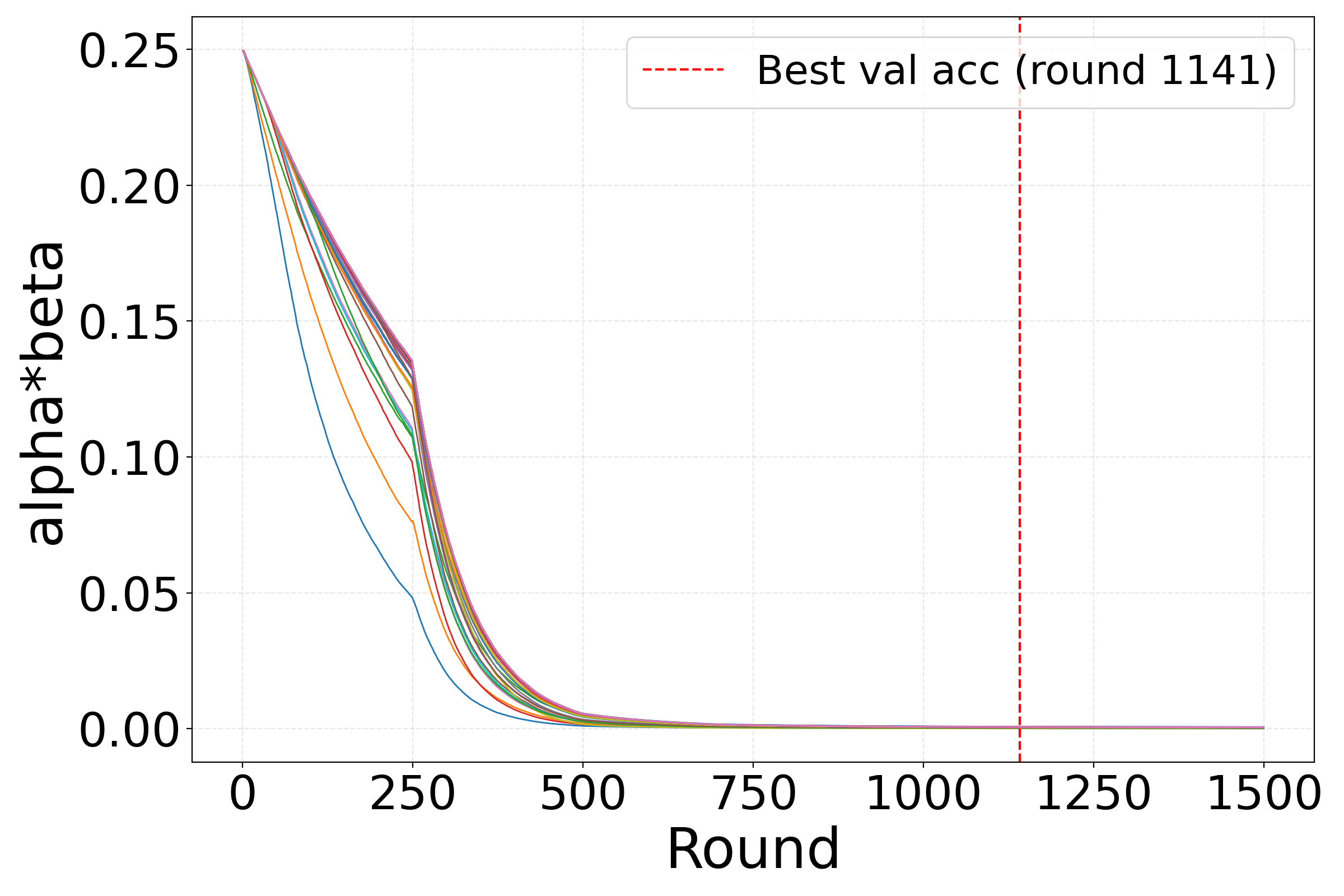}
\caption{Non-IID1: $\alpha\beta$}
\end{subfigure}
\begin{subfigure}{0.32\textwidth}
\centering
\includegraphics[width=\linewidth]{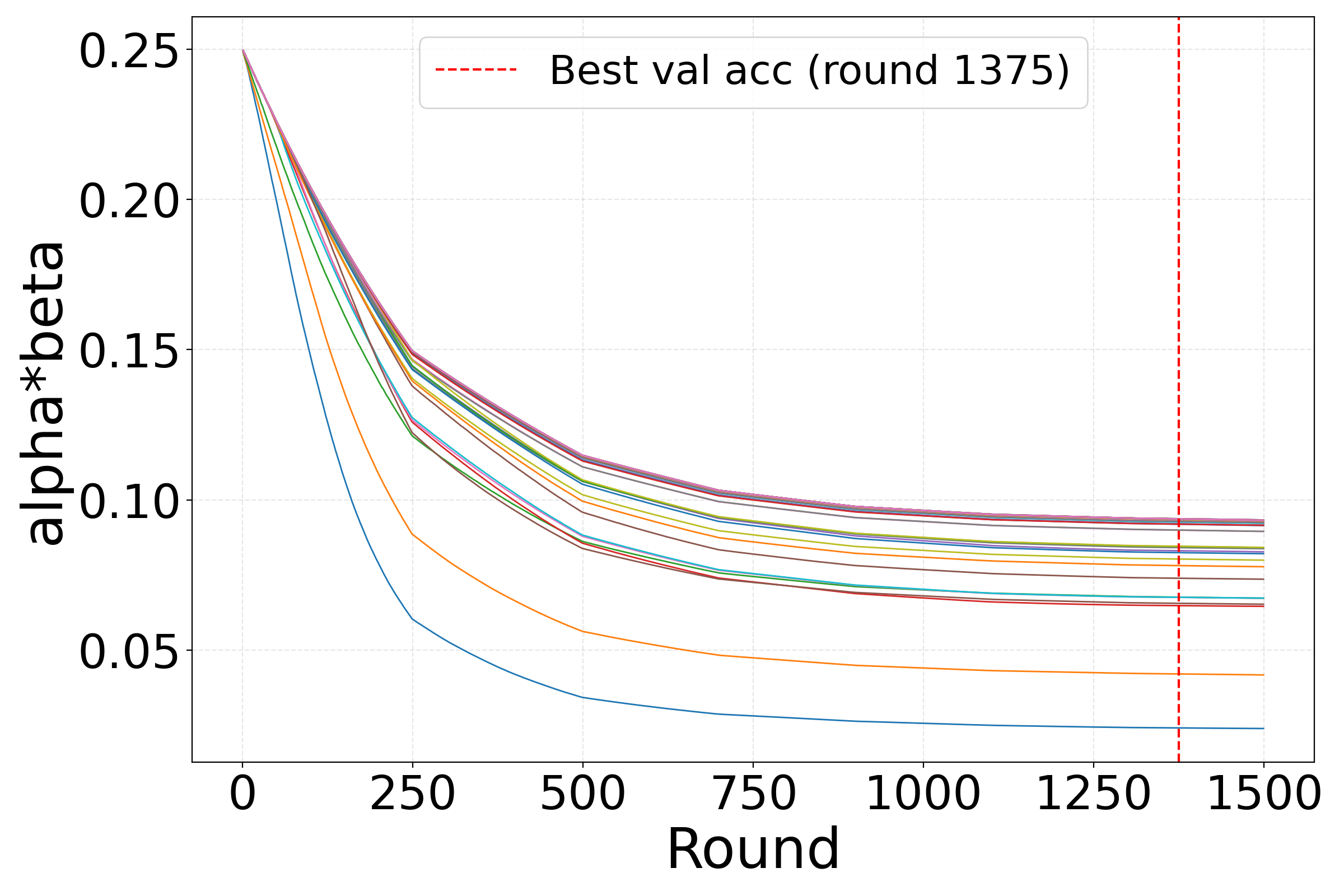}
\caption{Non-IID2: $\alpha\beta$}
\end{subfigure}

\vspace{0.6em}

% -------- Row 3: alpha(1-beta) --------
\begin{subfigure}{0.32\textwidth}
\centering
\includegraphics[width=\linewidth]{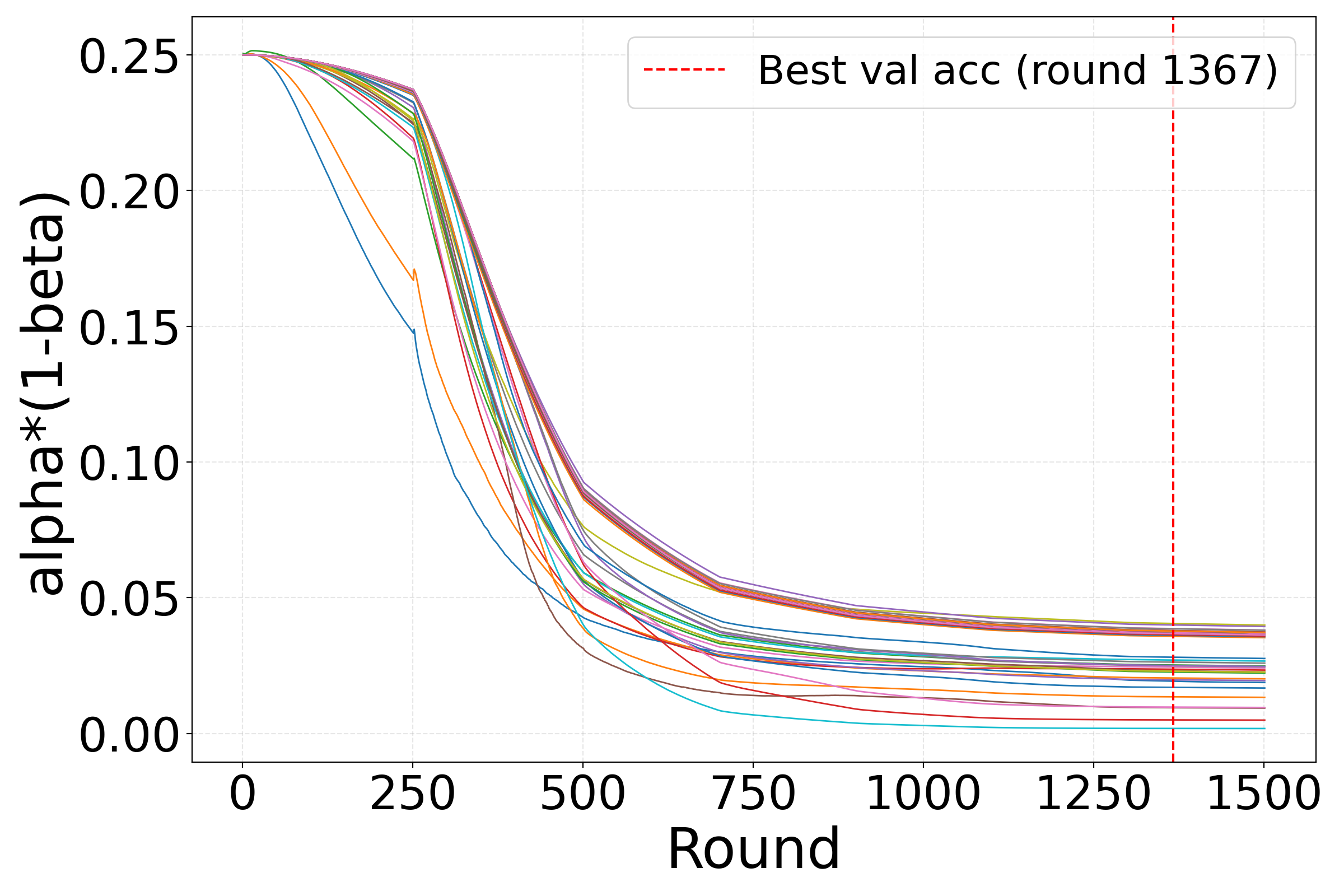}
\caption{IID: $\alpha(1-\beta)$}
\end{subfigure}
\begin{subfigure}{0.32\textwidth}
\centering
\includegraphics[width=\linewidth]{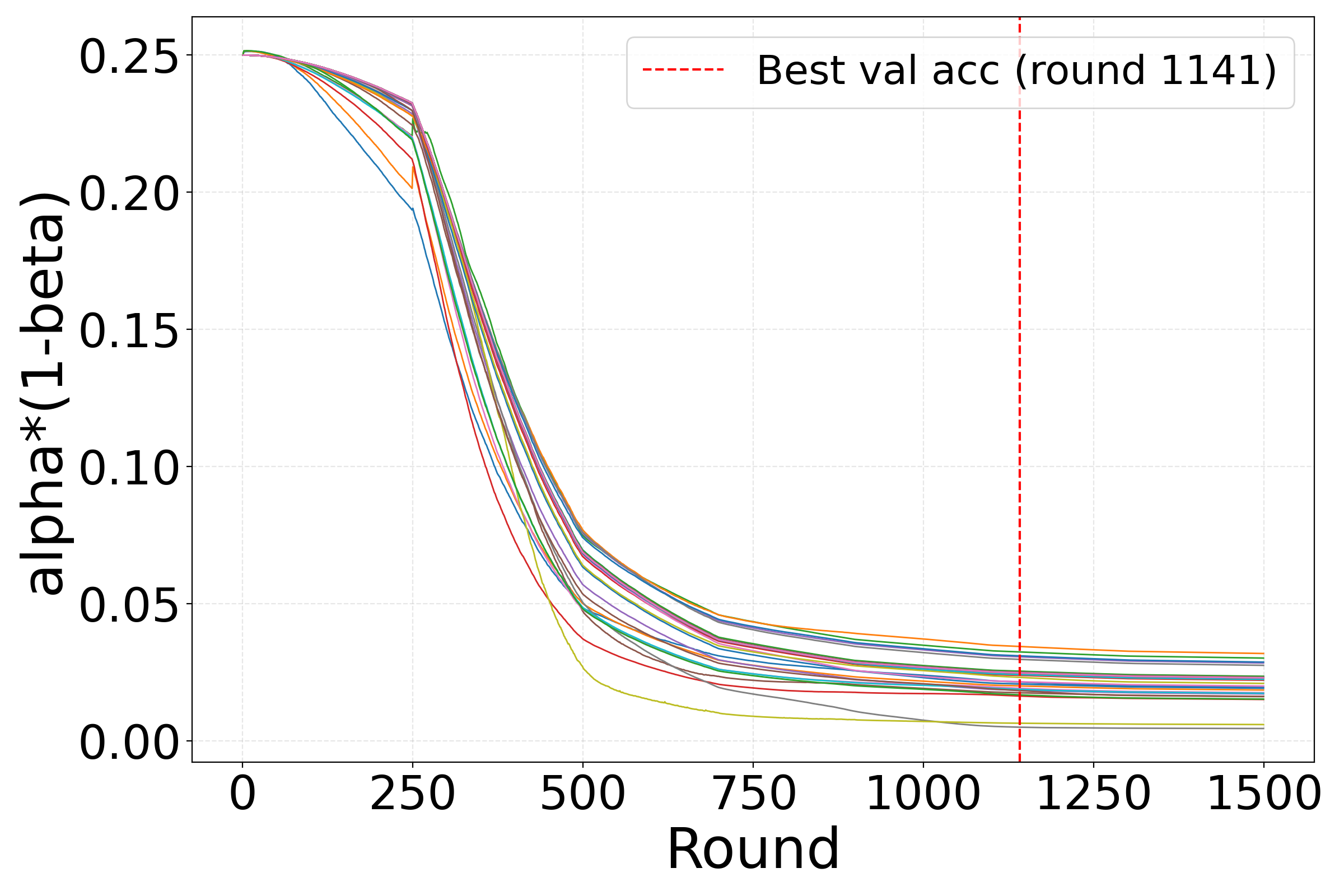}
\caption{Non-IID1: $\alpha(1-\beta)$}
\end{subfigure}
\begin{subfigure}{0.32\textwidth}
\centering
\includegraphics[width=\linewidth]{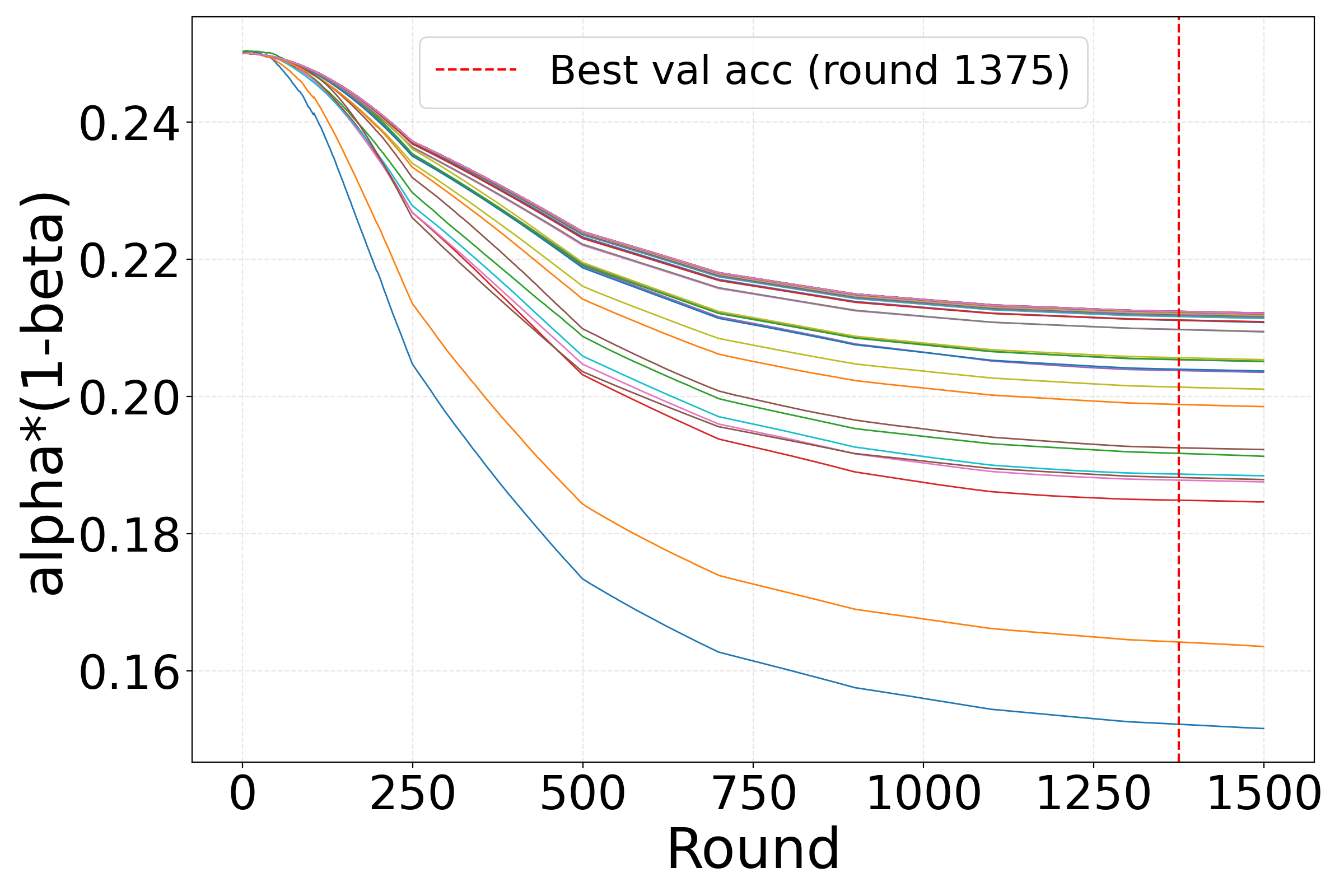}
\caption{Non-IID2: $\alpha(1-\beta)$}
\end{subfigure}

\vspace{0.6em}

% -------- Row 4: lambda --------
\begin{subfigure}{0.32\textwidth}
\centering
\includegraphics[width=\linewidth]{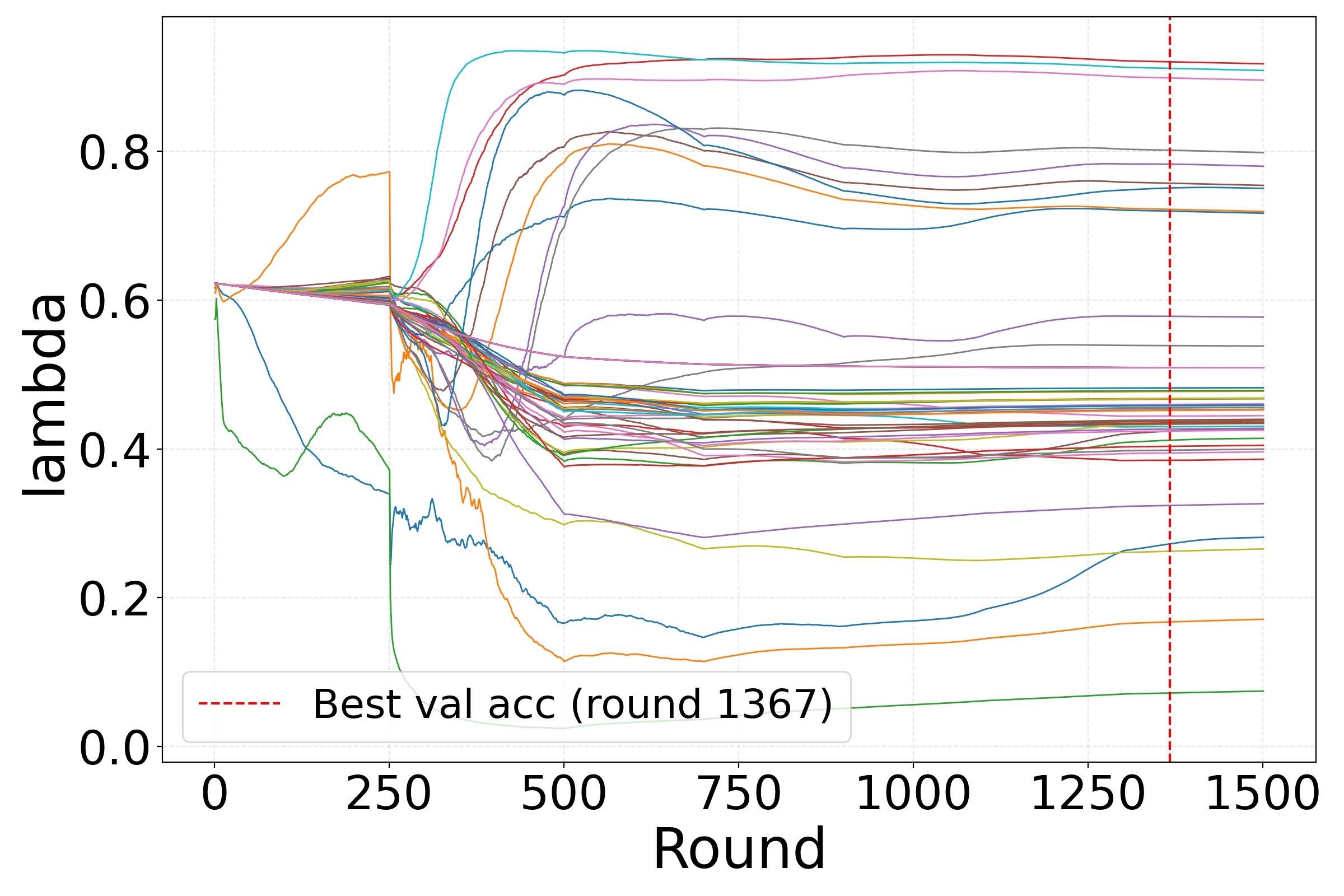}
\caption{IID: $\lambda$}
\end{subfigure}
\begin{subfigure}{0.32\textwidth}
\centering
\includegraphics[width=\linewidth]{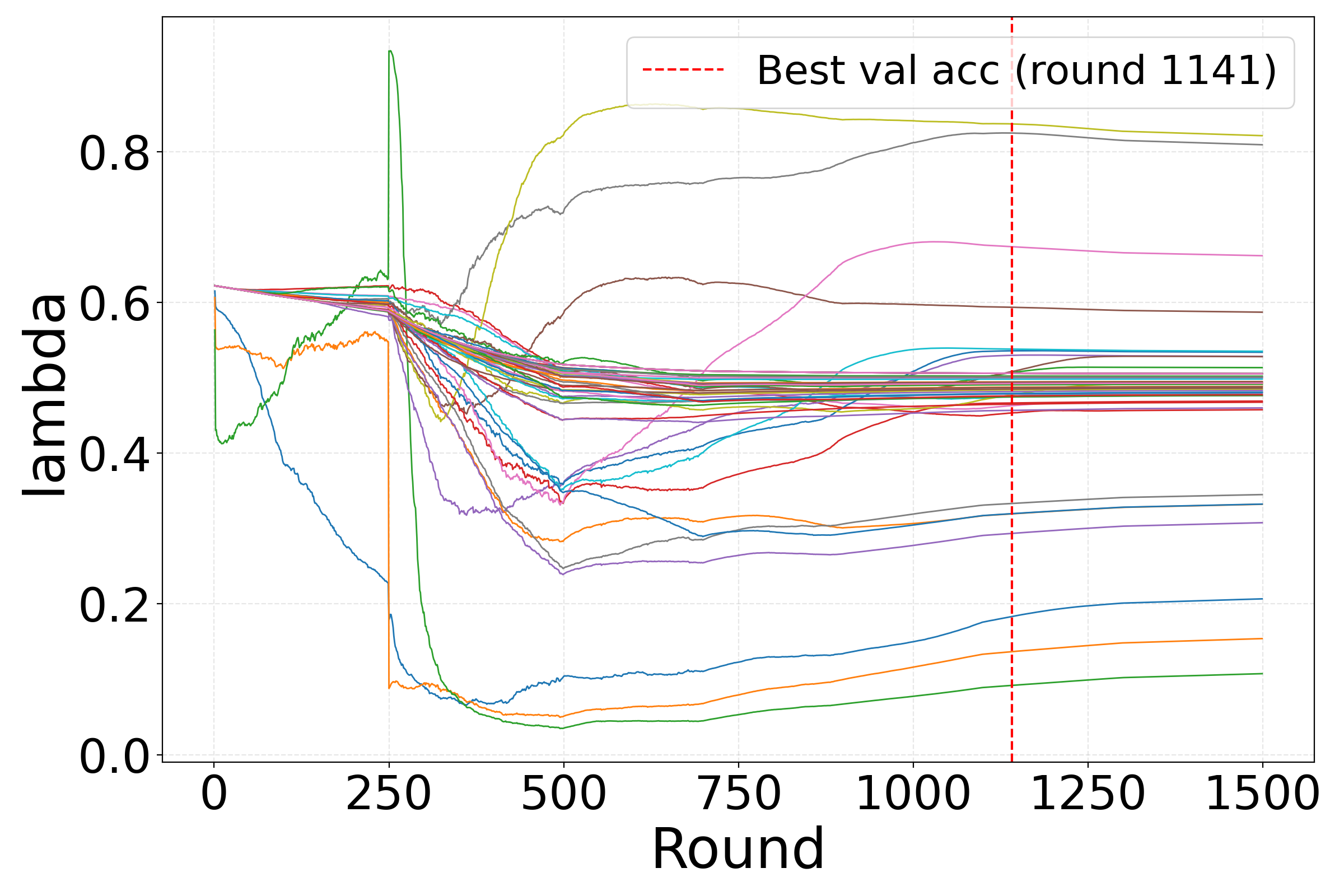}
\caption{Non-IID1: $\lambda$}
\end{subfigure}
\begin{subfigure}{0.32\textwidth}
\centering
\includegraphics[width=\linewidth]{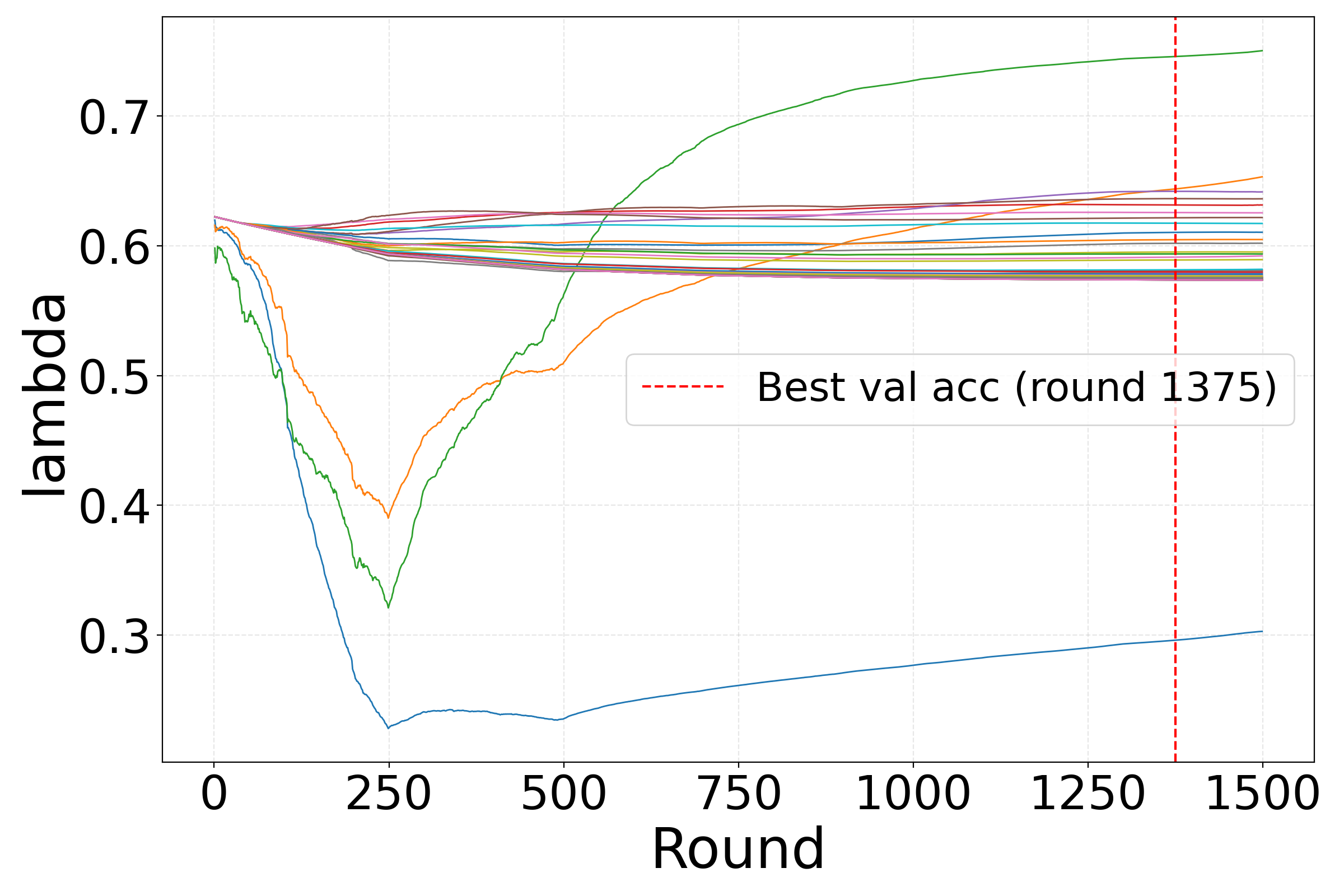}
\caption{Non-IID2: $\lambda$}
\end{subfigure}

\vspace{0.6em}
\begin{subfigure}{0.32\textwidth}
\centering
\includegraphics[width=\linewidth]{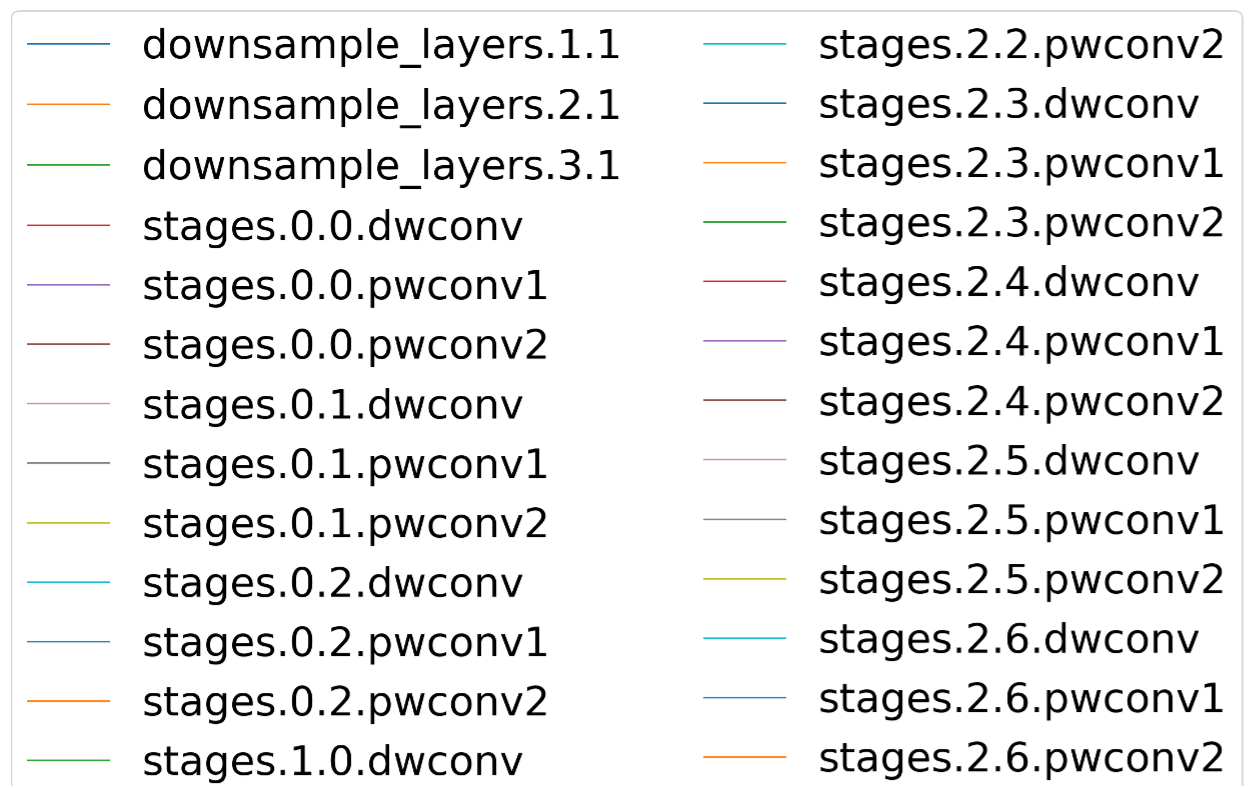}
\caption{Legend}
\end{subfigure}
\begin{subfigure}{0.32\textwidth}
\centering
\includegraphics[width=\linewidth]{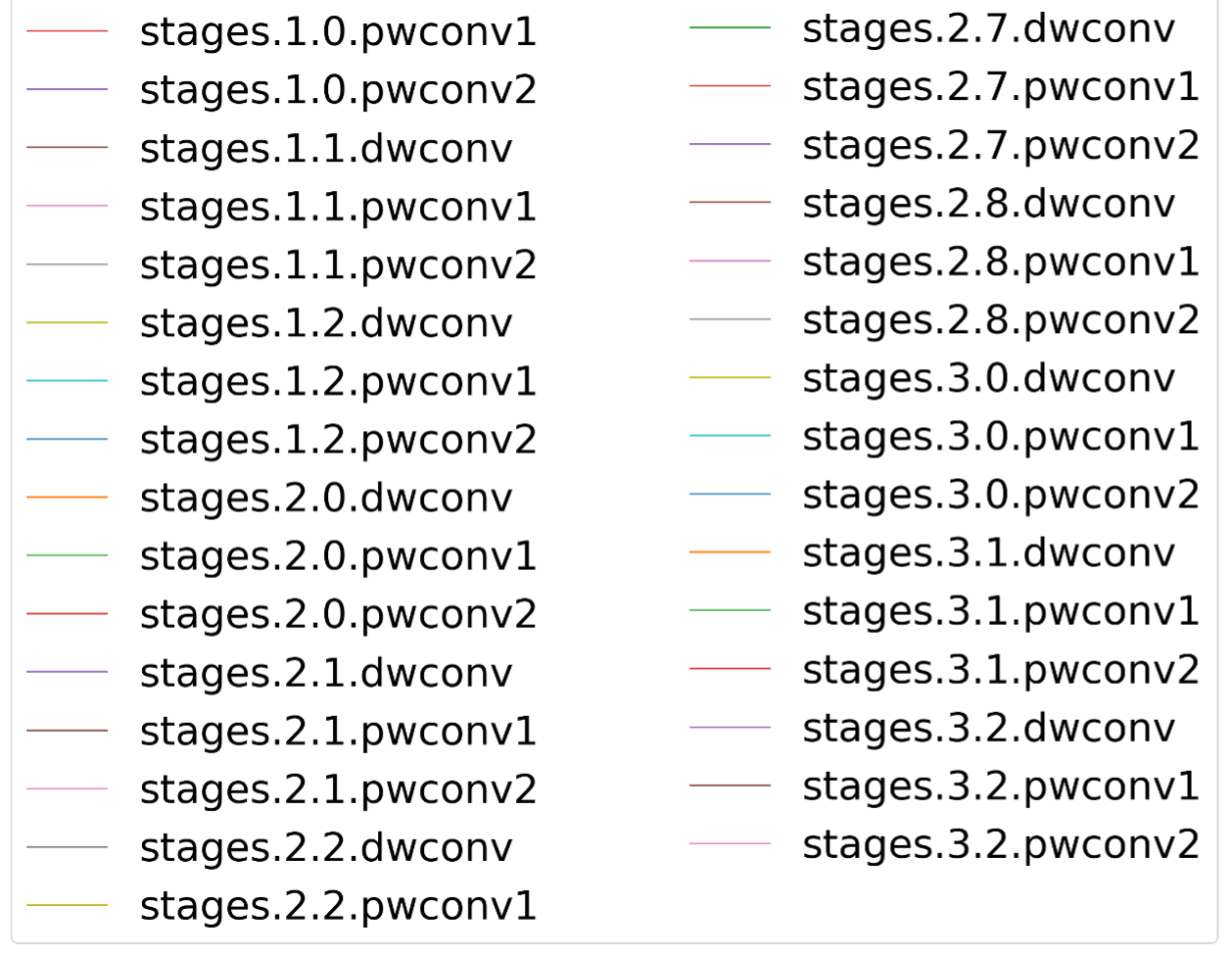}
\caption{Legend Continued}
\end{subfigure}
% -------- Row 5: legend --------

\caption{Variation analysis of the mixing parameters $(1-\alpha)$, $\alpha\beta$, $\alpha(1-\beta)$, and $\lambda$ on CIFAR10 (ConvNeXt-Tiny) under three data heterogeneity settings: IID, Non-IID1 (Dirichlet $\nu=0.3$), and Non-IID2.}
\label{alphabetalambdacifar10convnext}

\end{figure*}

\begin{figure*}[htbp]
\centering

% -------- Row 1: (1-alpha) --------
\begin{subfigure}{0.32\textwidth}
\includegraphics[width=\linewidth]{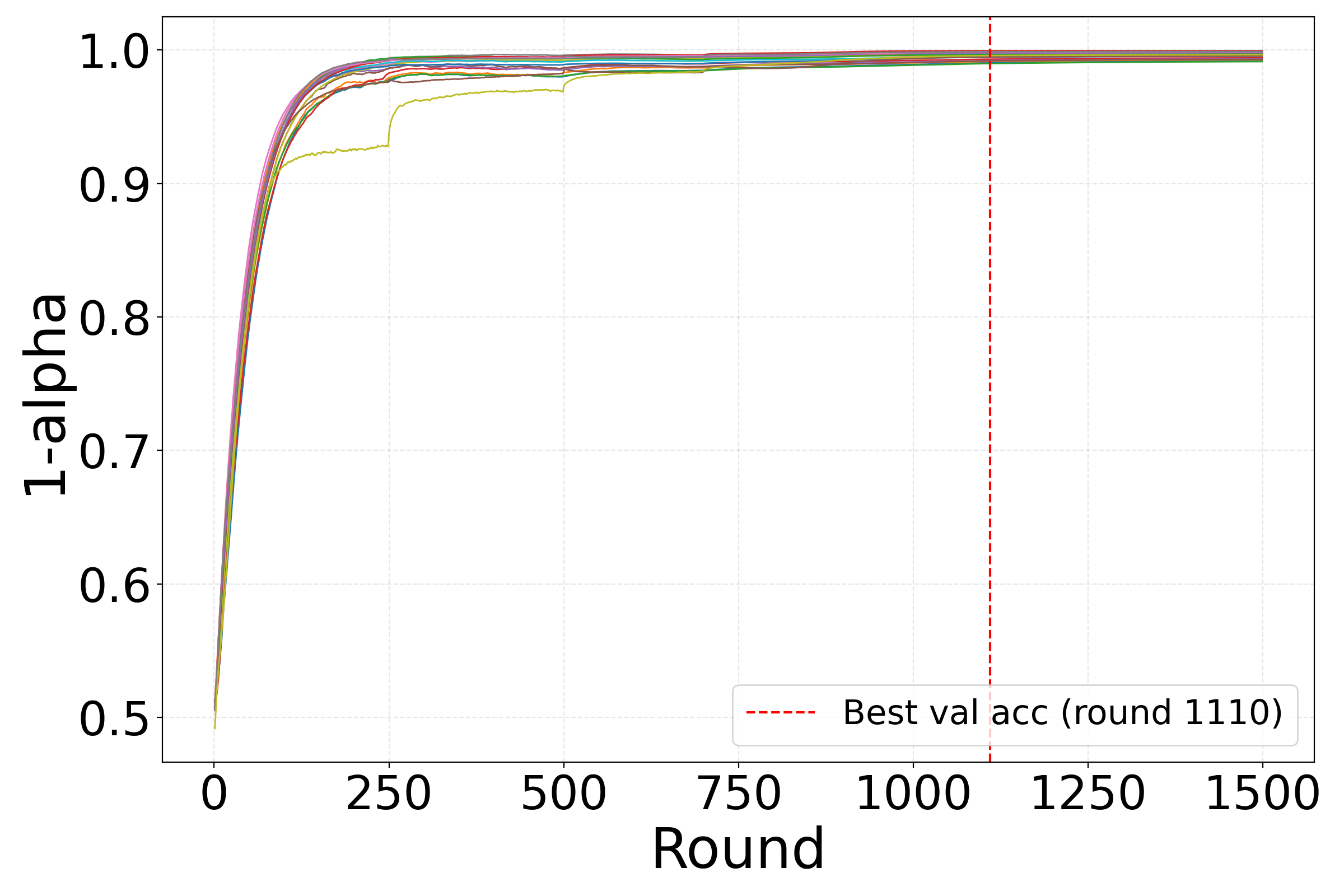}
\caption{IID: $(1-\alpha)$}
\end{subfigure}
\begin{subfigure}{0.32\textwidth}
\includegraphics[width=\linewidth]{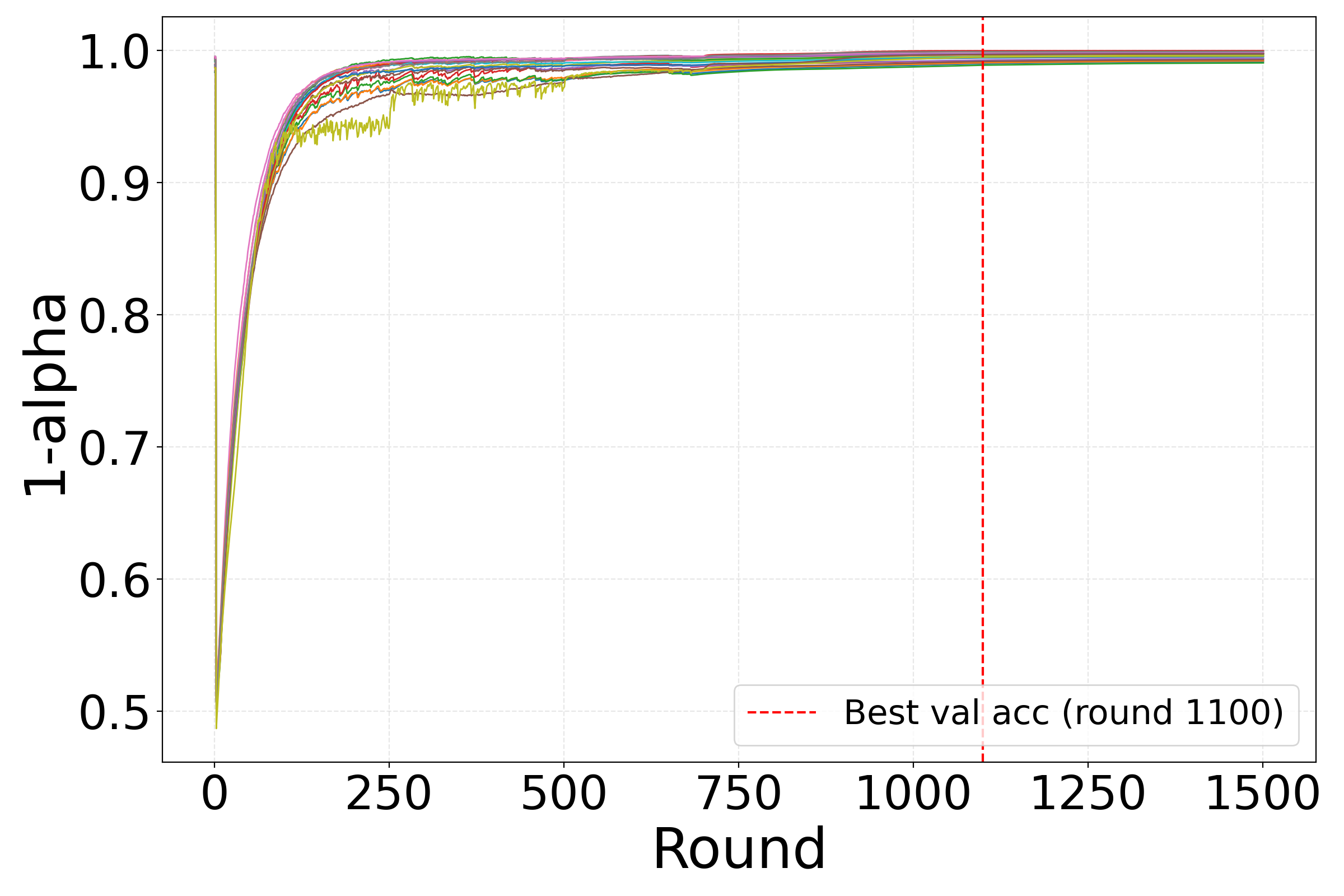}
\caption{Non-IID1: $(1-\alpha)$}
\end{subfigure}
\begin{subfigure}{0.32\textwidth}
\includegraphics[width=\linewidth]{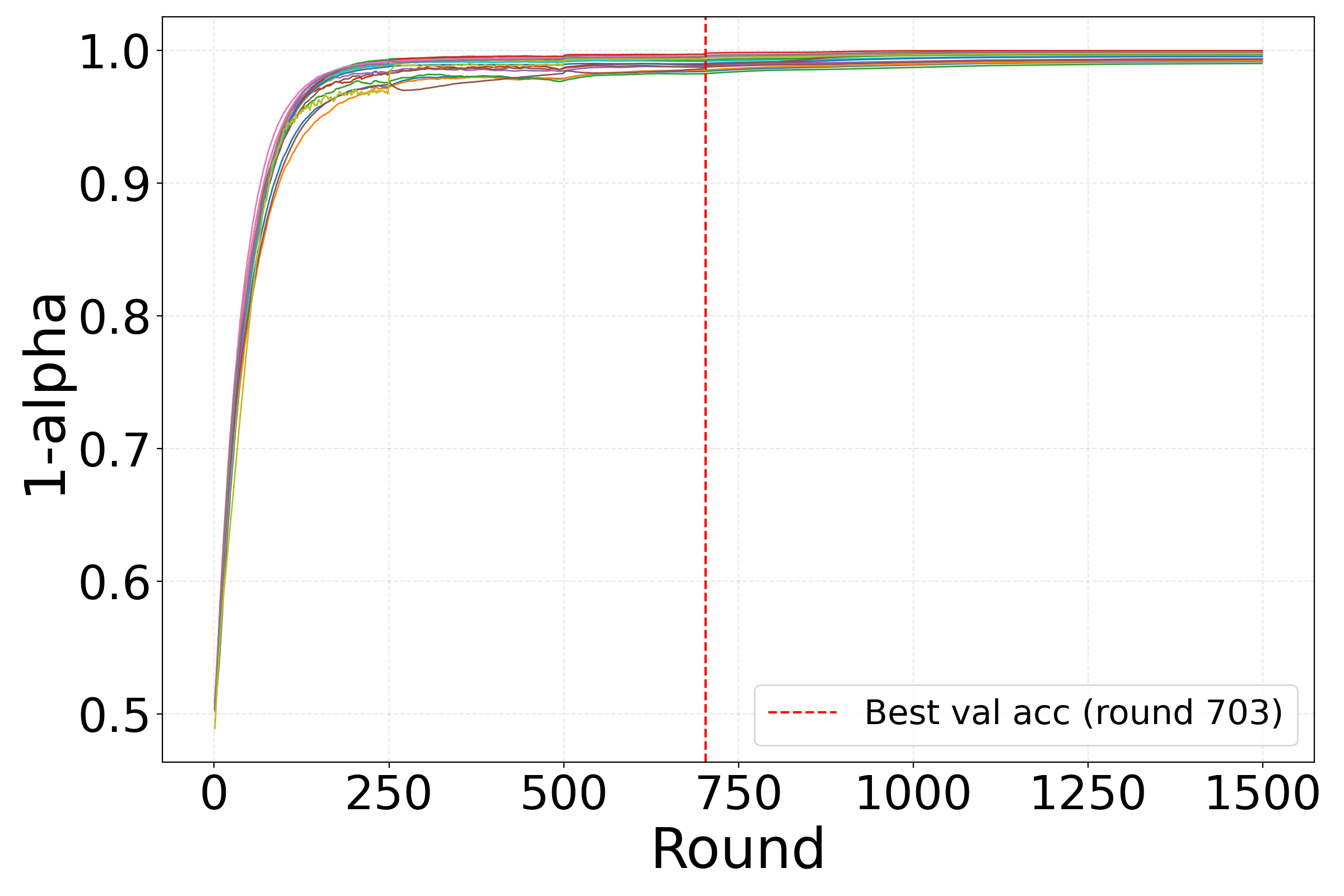}
\caption{Non-IID2: $(1-\alpha)$}
\end{subfigure}

\vspace{0.6em}

% -------- Row 2: alpha beta --------
\begin{subfigure}{0.32\textwidth}
\includegraphics[width=\linewidth]{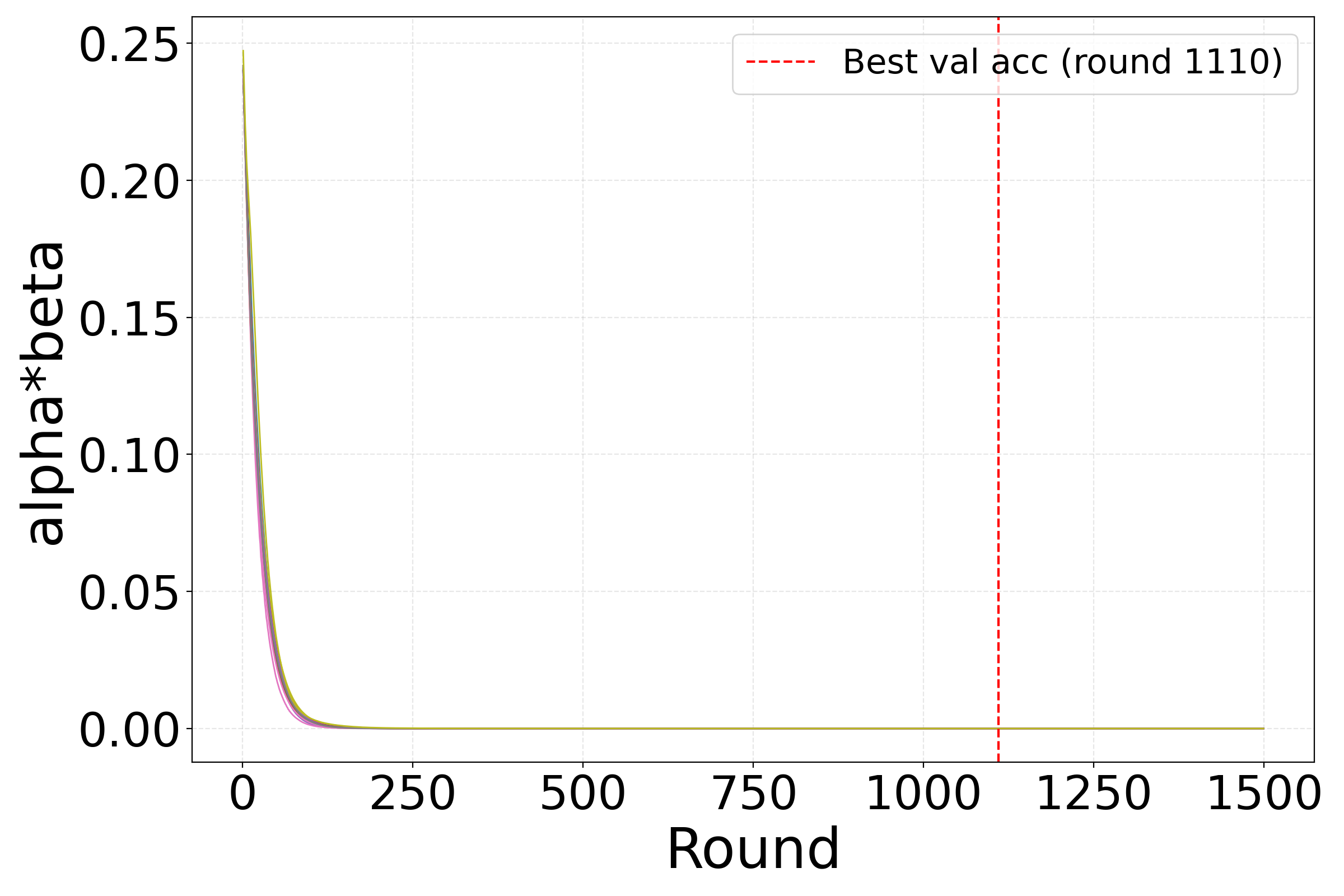}
\caption{IID: $\alpha\beta$}
\end{subfigure}
\begin{subfigure}{0.32\textwidth}
\includegraphics[width=\linewidth]{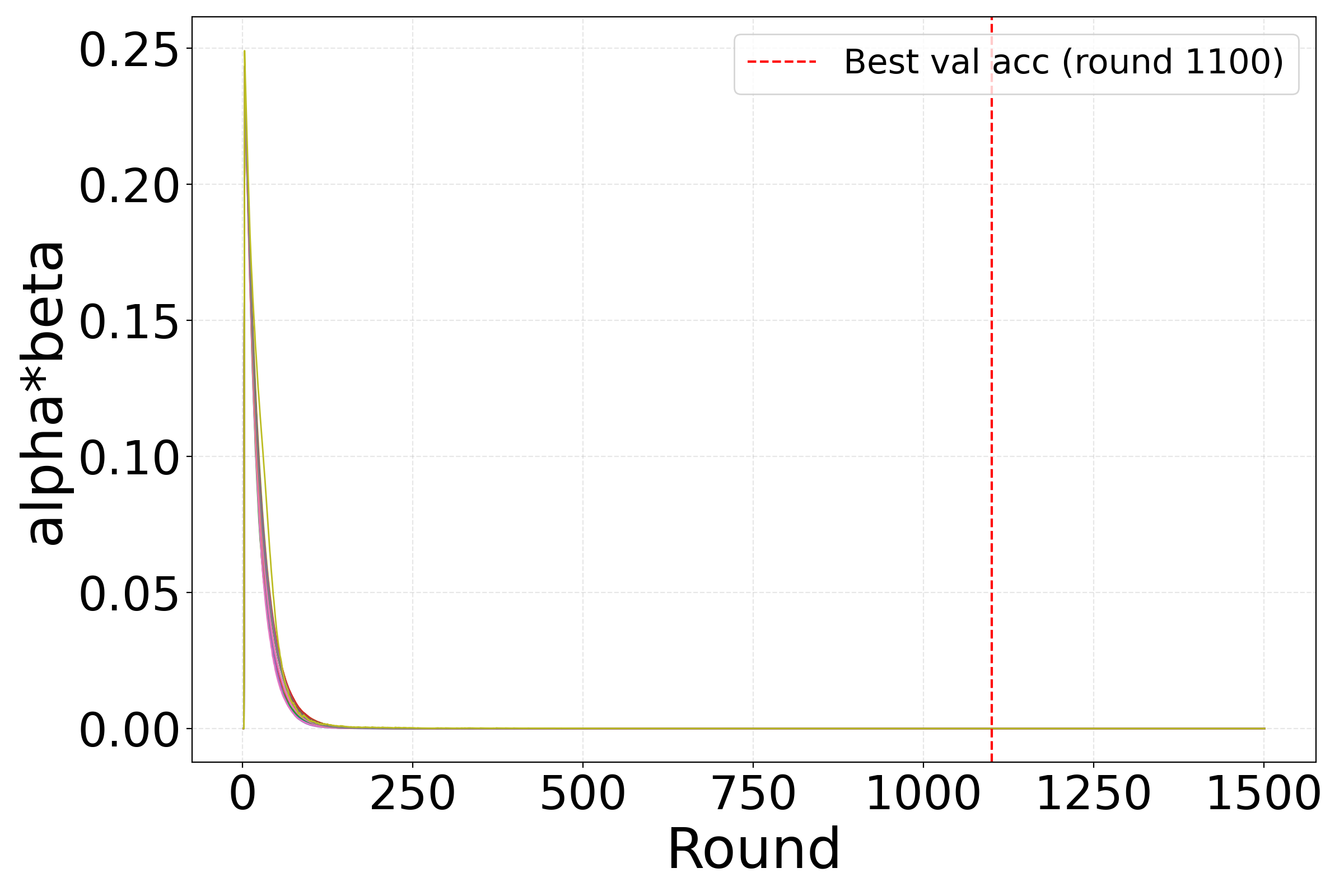}
\caption{Non-IID1: $\alpha\beta$}
\end{subfigure}
\begin{subfigure}{0.32\textwidth}
\includegraphics[width=\linewidth]{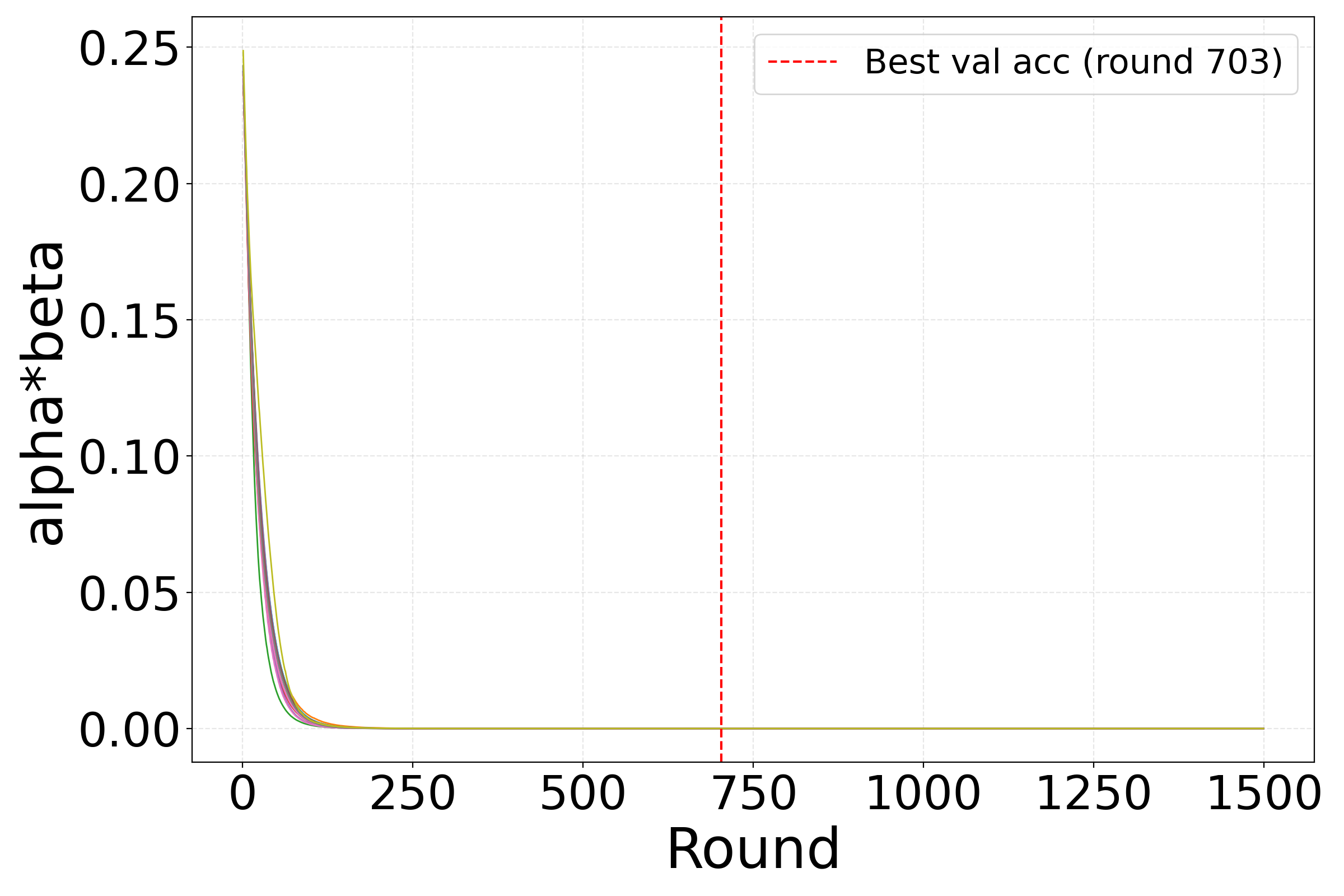}
\caption{Non-IID2: $\alpha\beta$}
\end{subfigure}

\vspace{0.6em}

% -------- Row 3: alpha(1-beta) --------
\begin{subfigure}{0.32\textwidth}
\includegraphics[width=\linewidth]{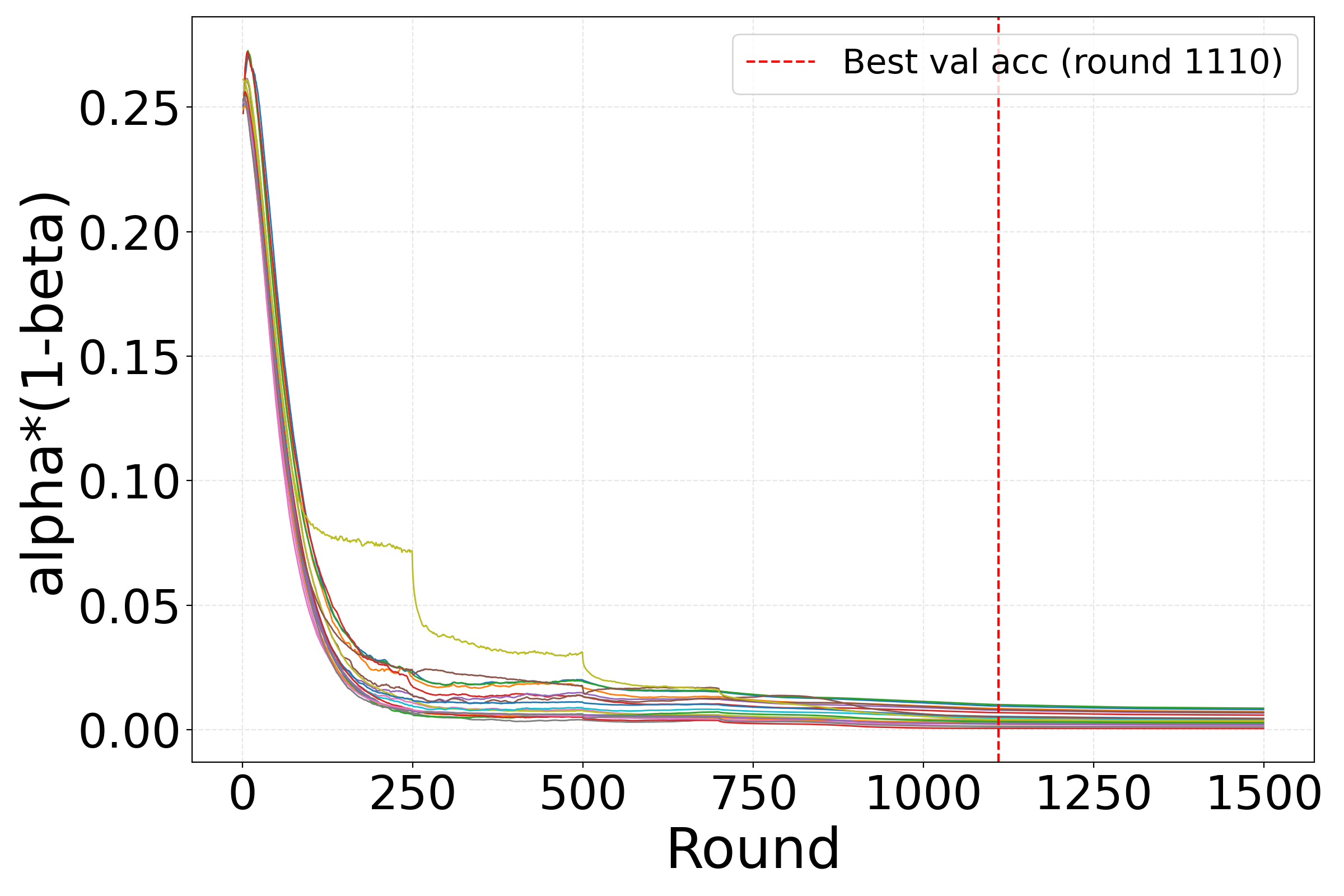}
\caption{IID: $\alpha(1-\beta)$}
\end{subfigure}
\begin{subfigure}{0.32\textwidth}
\includegraphics[width=\linewidth]{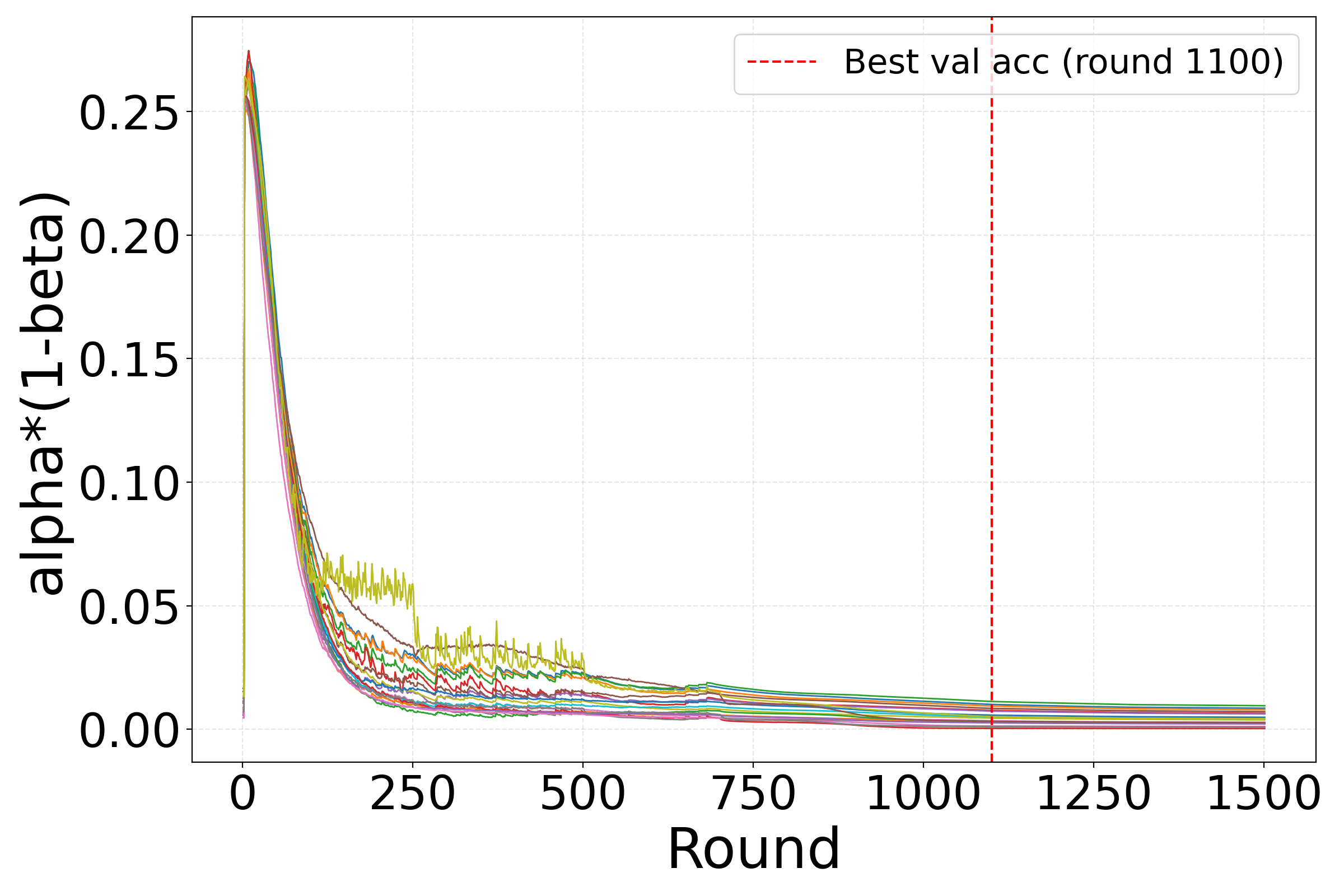}
\caption{Non-IID1: $\alpha(1-\beta)$}
\end{subfigure}
\begin{subfigure}{0.32\textwidth}
\includegraphics[width=\linewidth]{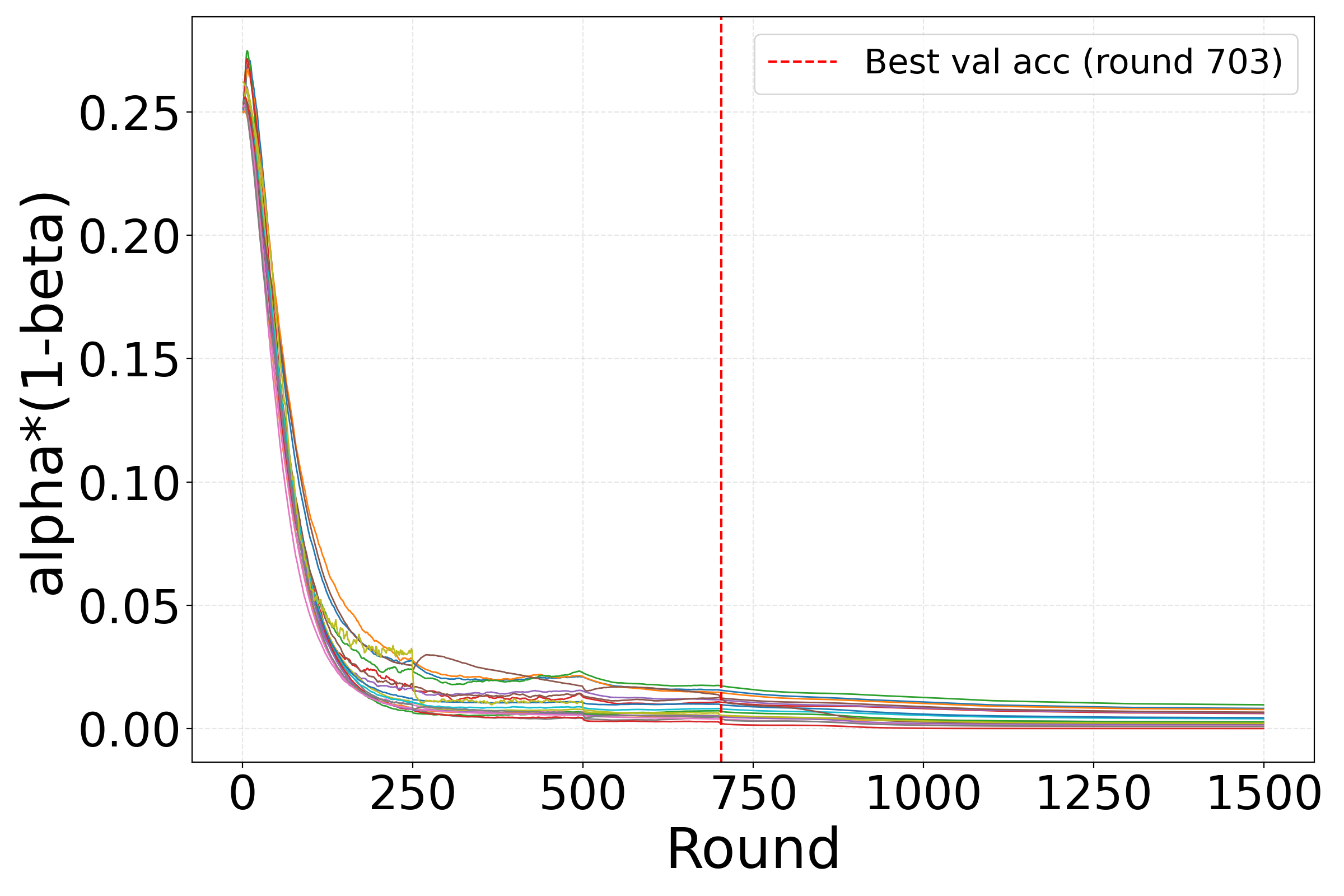}
\caption{Non-IID2: $\alpha(1-\beta)$}
\end{subfigure}

\vspace{0.6em}

% -------- Row 4: lambda --------
\begin{subfigure}{0.32\textwidth}
\includegraphics[width=\linewidth]{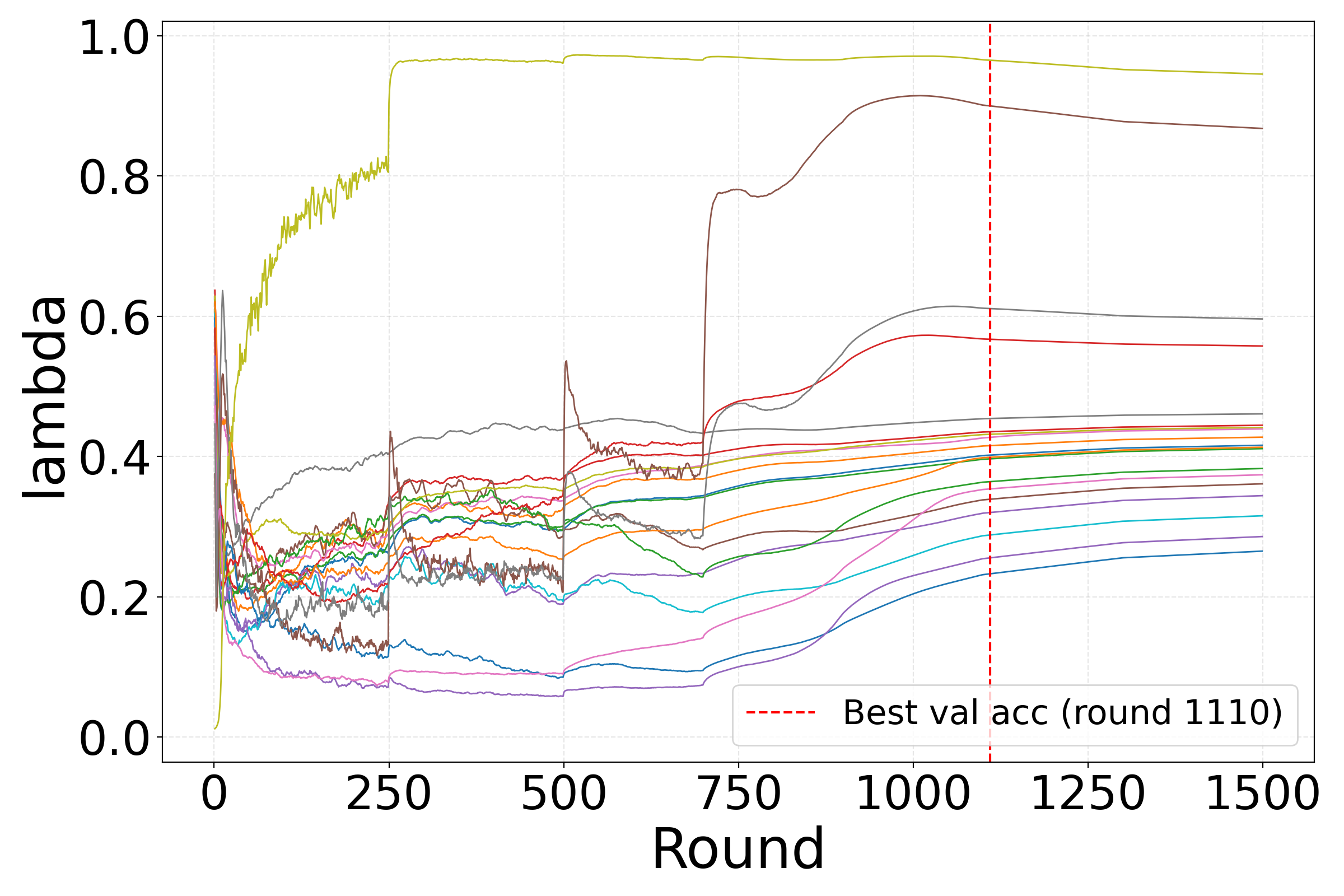}
\caption{IID: $\lambda$}
\end{subfigure}
\begin{subfigure}{0.32\textwidth}
\includegraphics[width=\linewidth]{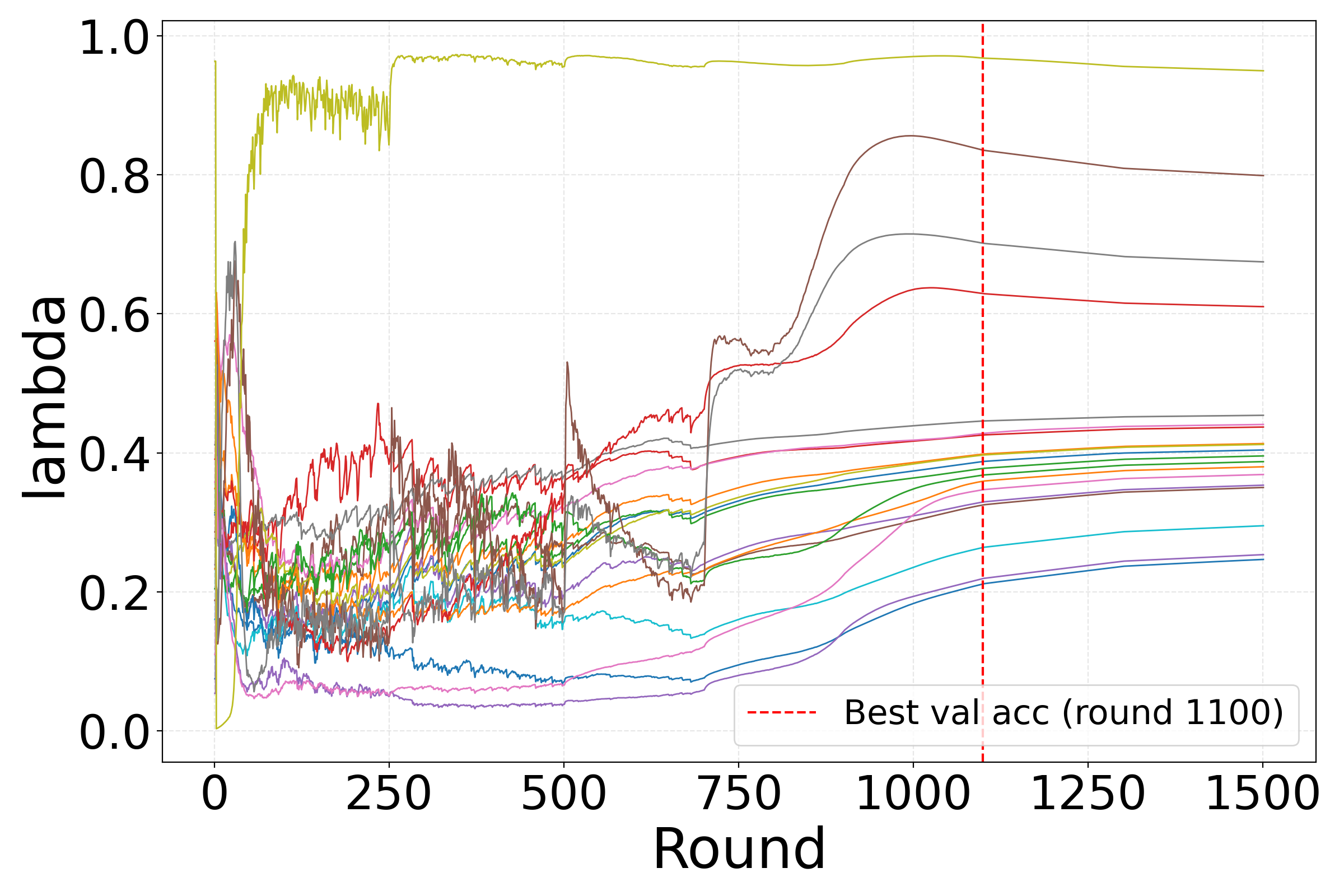}
\caption{Non-IID1: $\lambda$}
\end{subfigure}
\begin{subfigure}{0.32\textwidth}
\includegraphics[width=\linewidth]{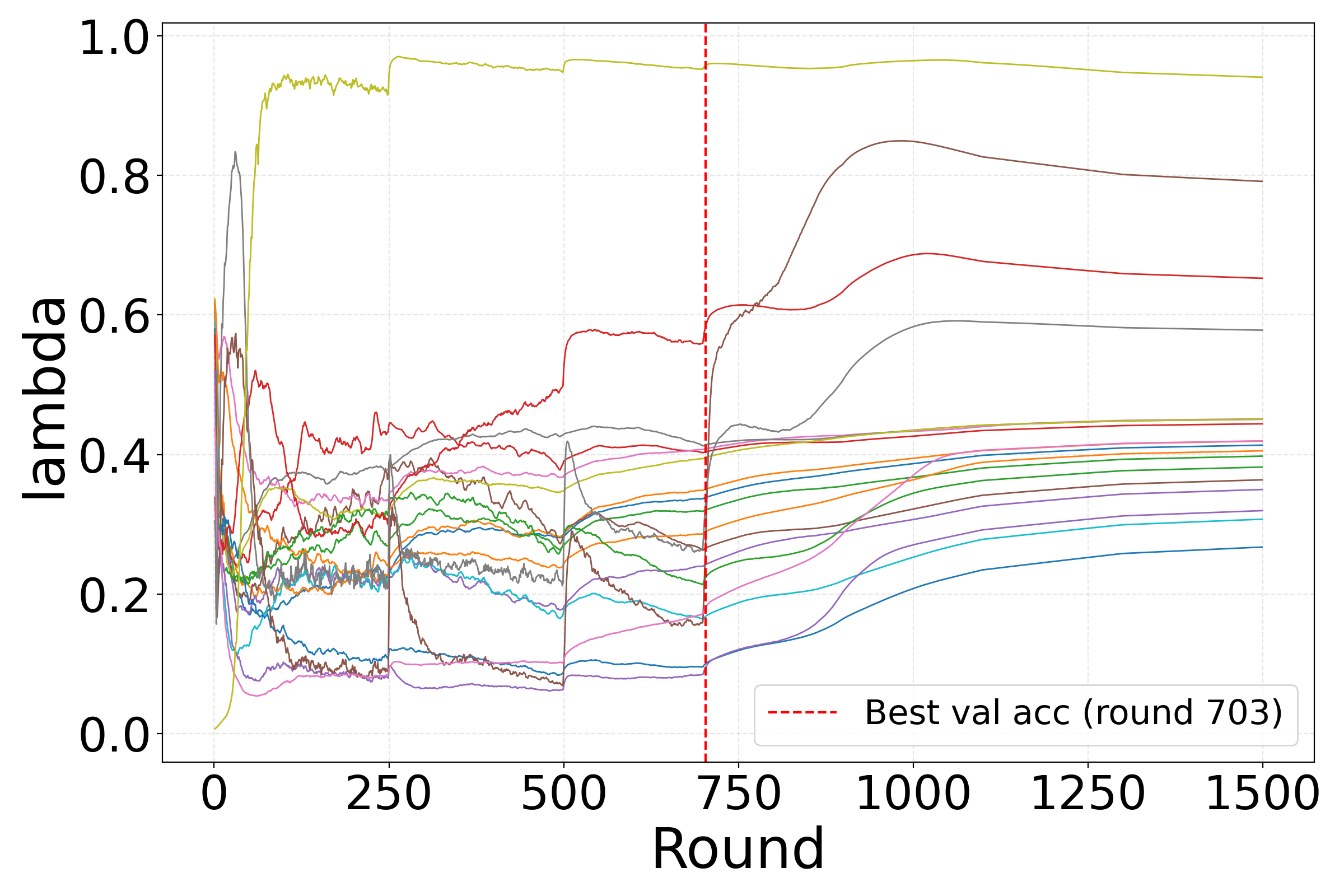}
\caption{Non-IID2: $\lambda$}
\end{subfigure}

\vspace{0.6em}

% -------- Row 4: legend --------
\begin{subfigure}{0.32\textwidth}
\includegraphics[width=\linewidth]{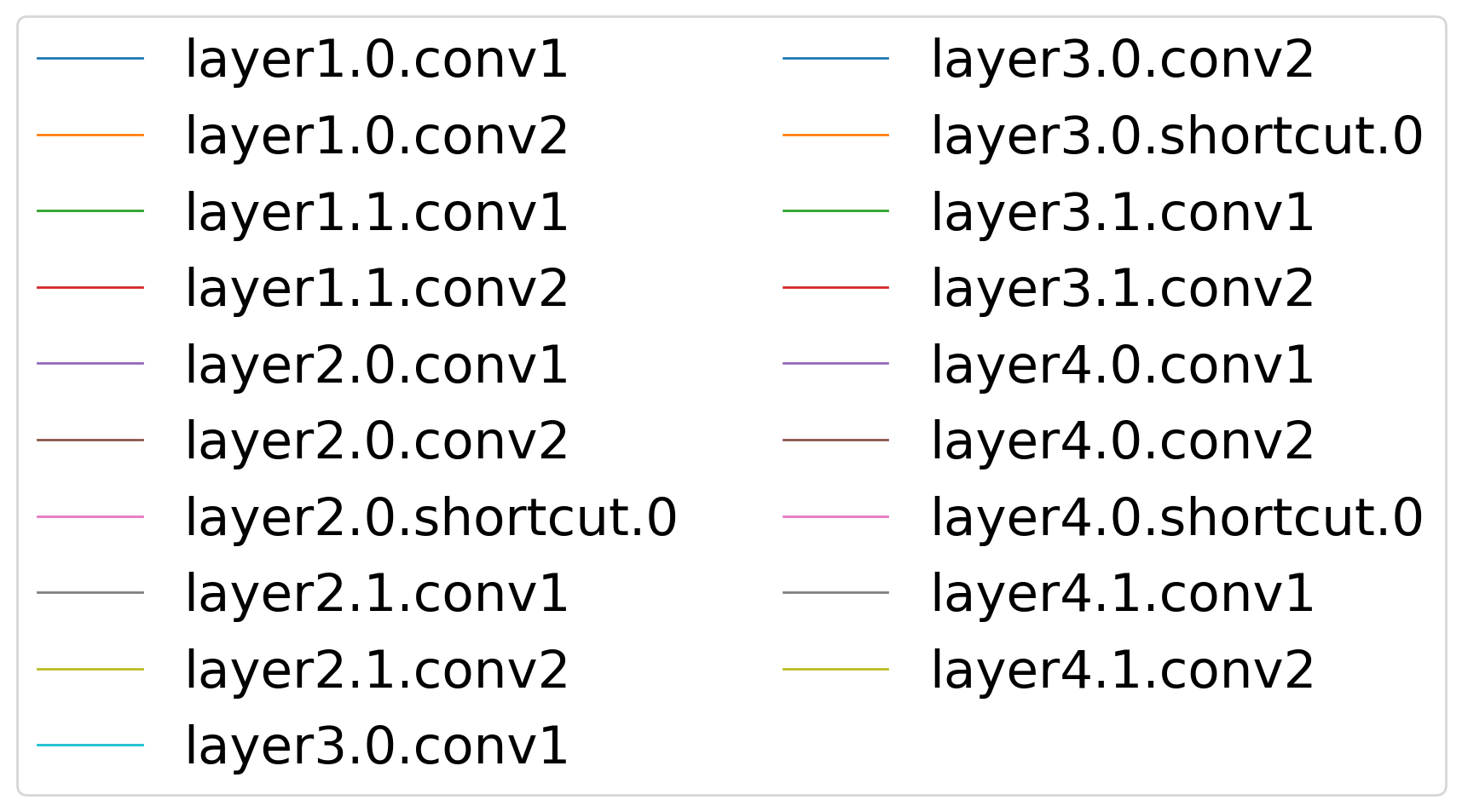}
\caption{Legend}
\end{subfigure}
\caption{Mixing parameter variation on TinyImageNet under IID, Non-IID1, and Non-IID2 settings.}
\label{alphabetalambdatinyimagenet}
\end{figure*}

\begin{figure*}[t]
\centering

% -------- Row 1: (1-alpha) --------
\begin{subfigure}{0.32\textwidth}
\includegraphics[width=\linewidth]{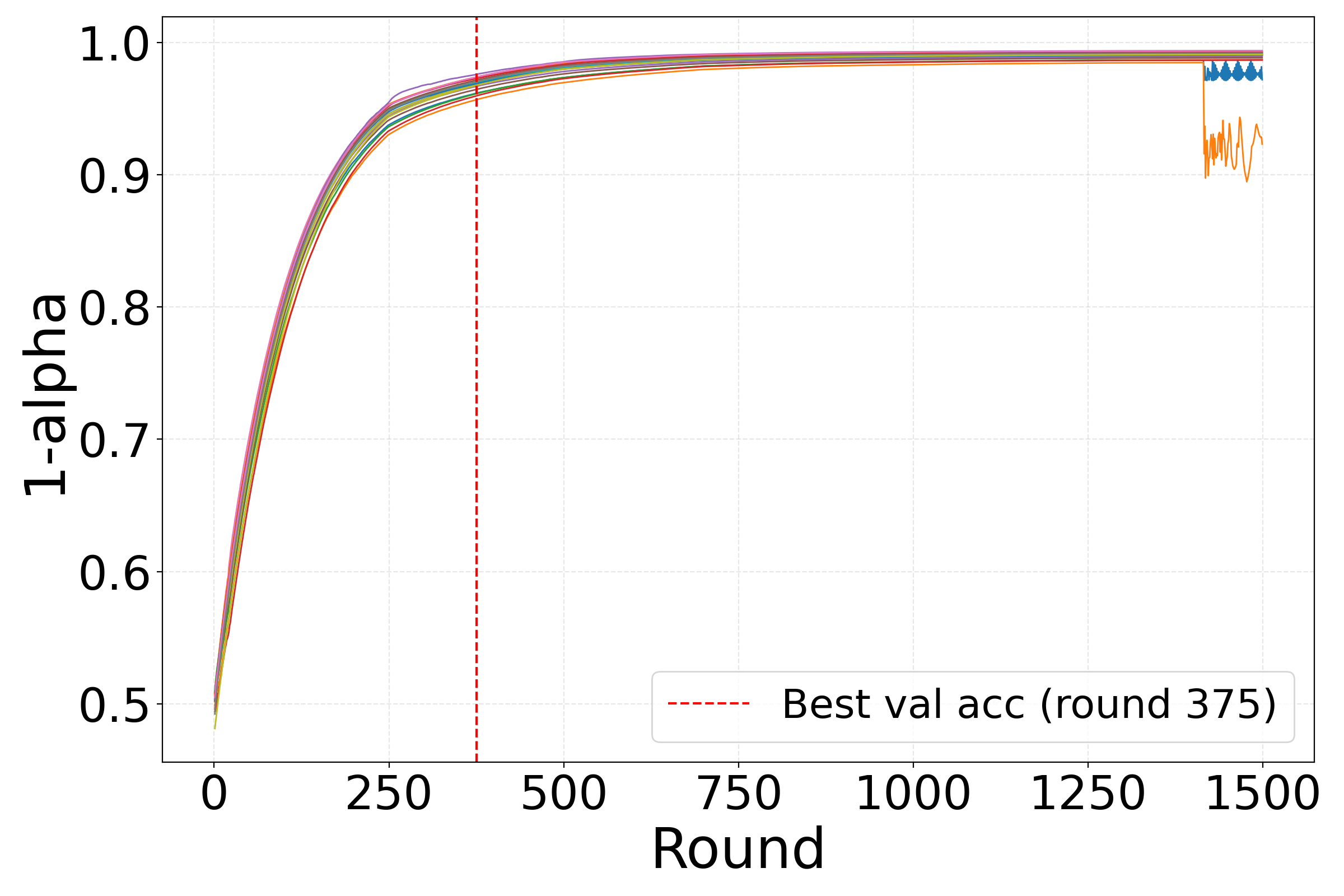}
\caption{IID: $(1-\alpha)$}
\end{subfigure}
\begin{subfigure}{0.32\textwidth}
\includegraphics[width=\linewidth]{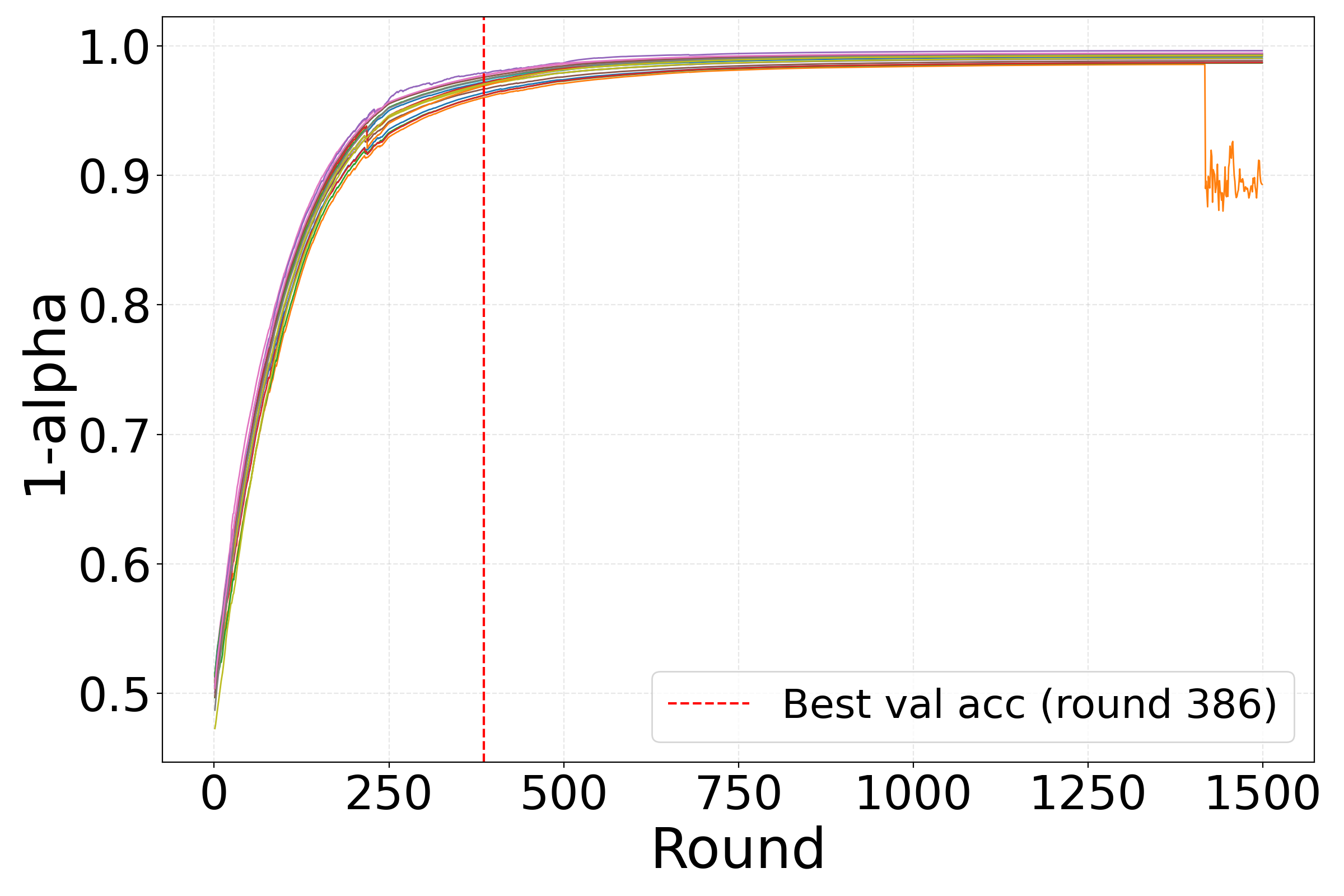}
\caption{Non-IID1: $(1-\alpha)$}
\end{subfigure}
\begin{subfigure}{0.32\textwidth}
\includegraphics[width=\linewidth]{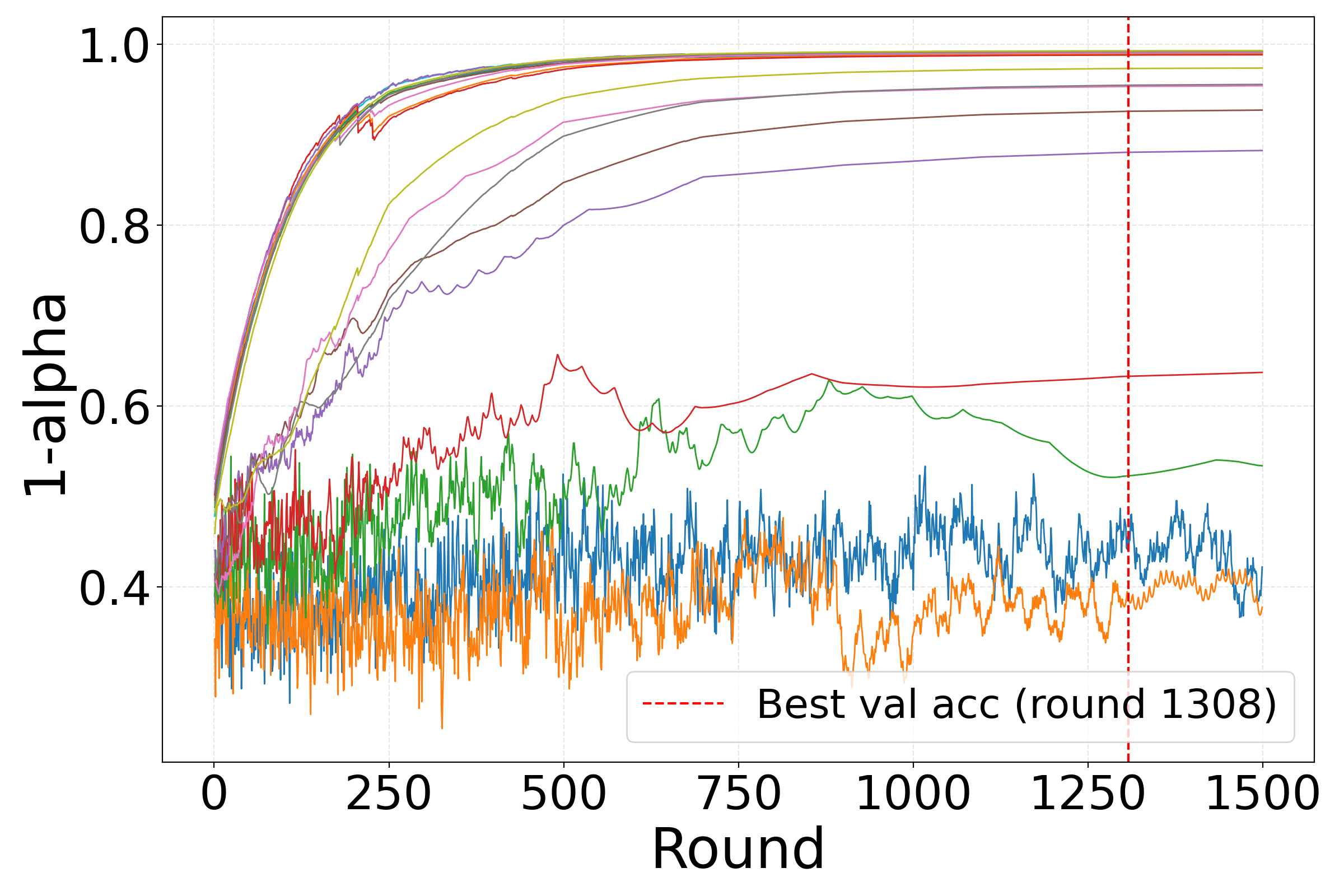}
\caption{Non-IID2: $(1-\alpha)$}
\end{subfigure}

\vspace{0.6em}

% -------- Row 2: alpha beta --------
\begin{subfigure}{0.32\textwidth}
\includegraphics[width=\linewidth]{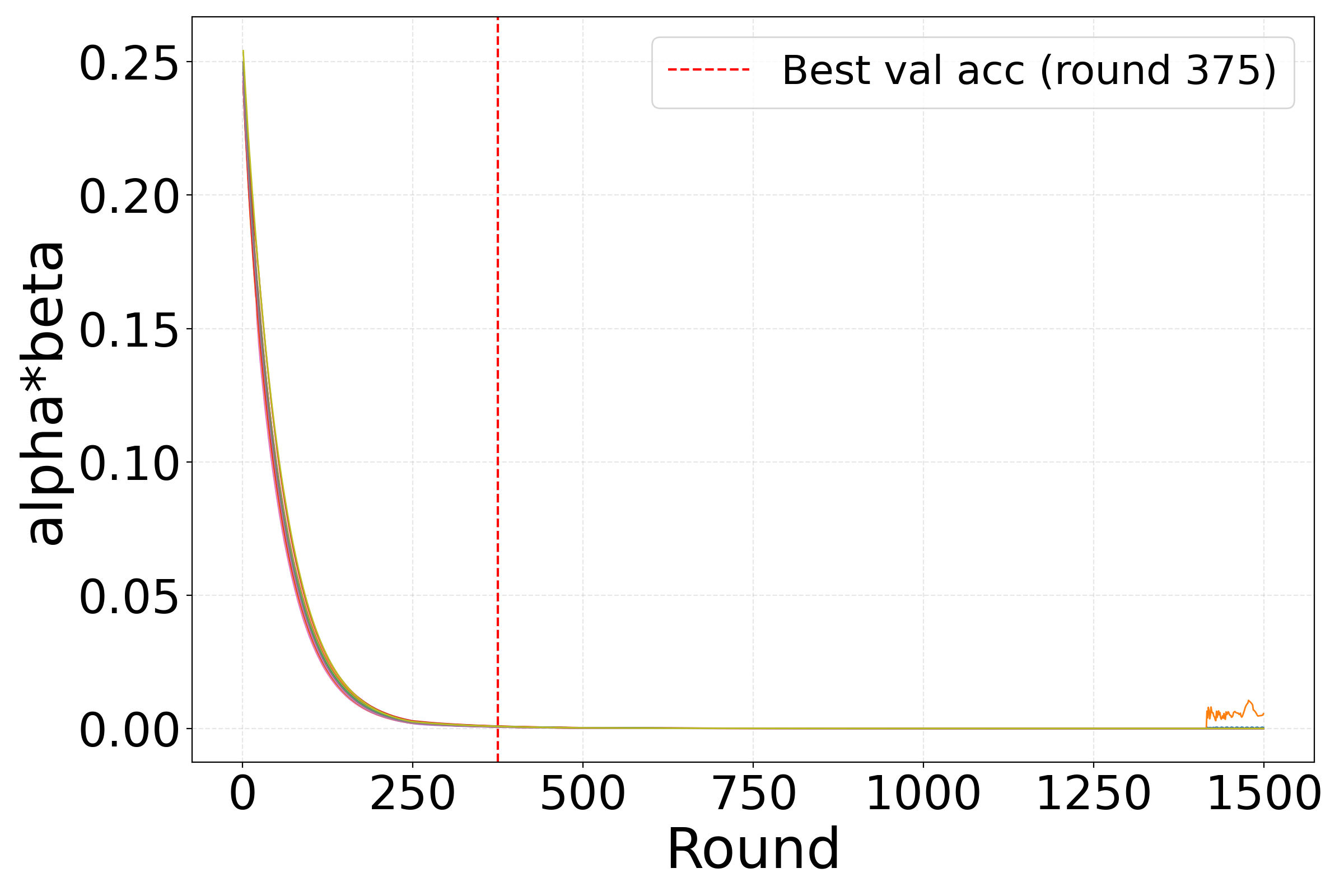}
\caption{IID: $\alpha\beta$}
\end{subfigure}
\begin{subfigure}{0.32\textwidth}
\includegraphics[width=\linewidth]{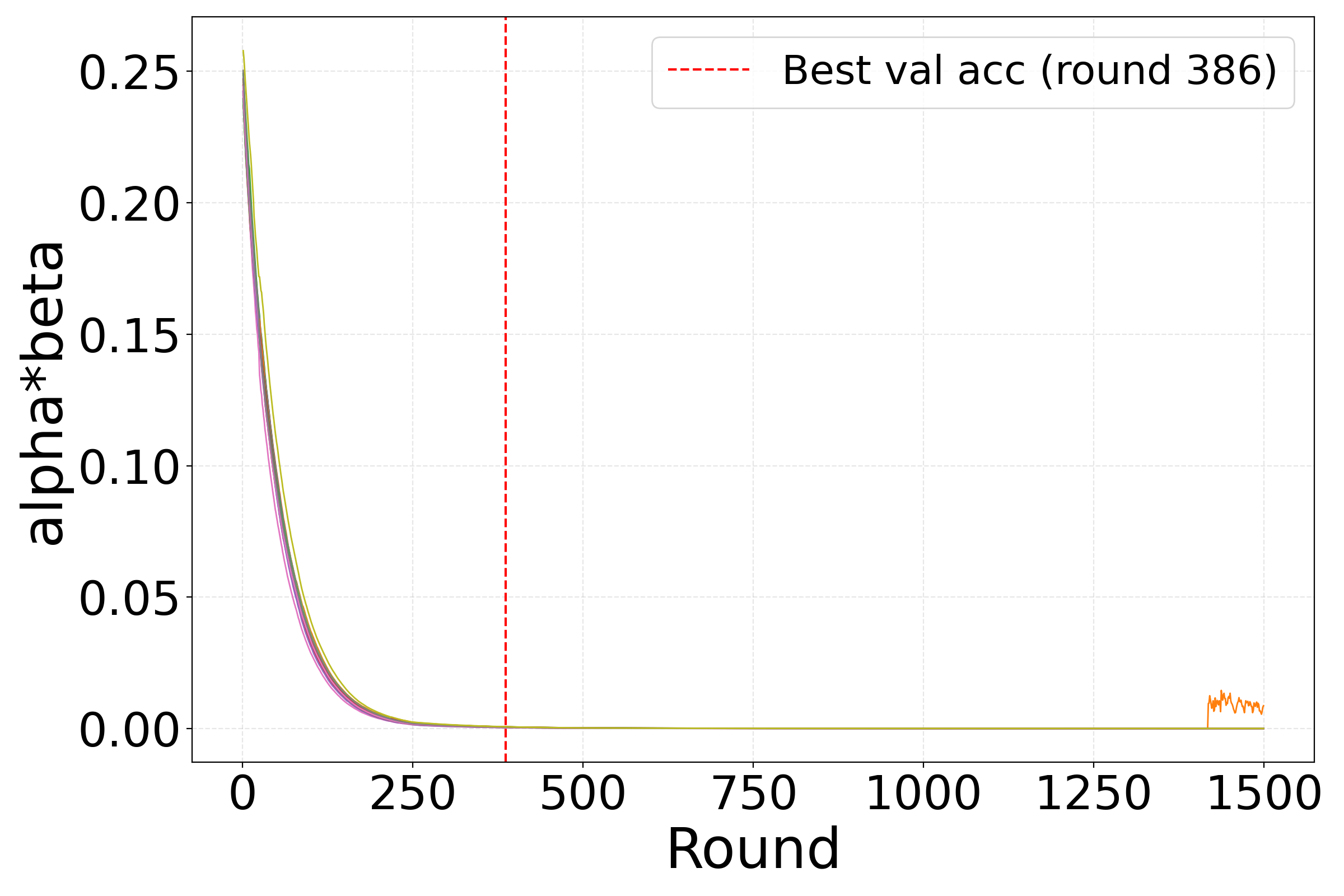}
\caption{Non-IID1: $\alpha\beta$}
\end{subfigure}
\begin{subfigure}{0.32\textwidth}
\includegraphics[width=\linewidth]{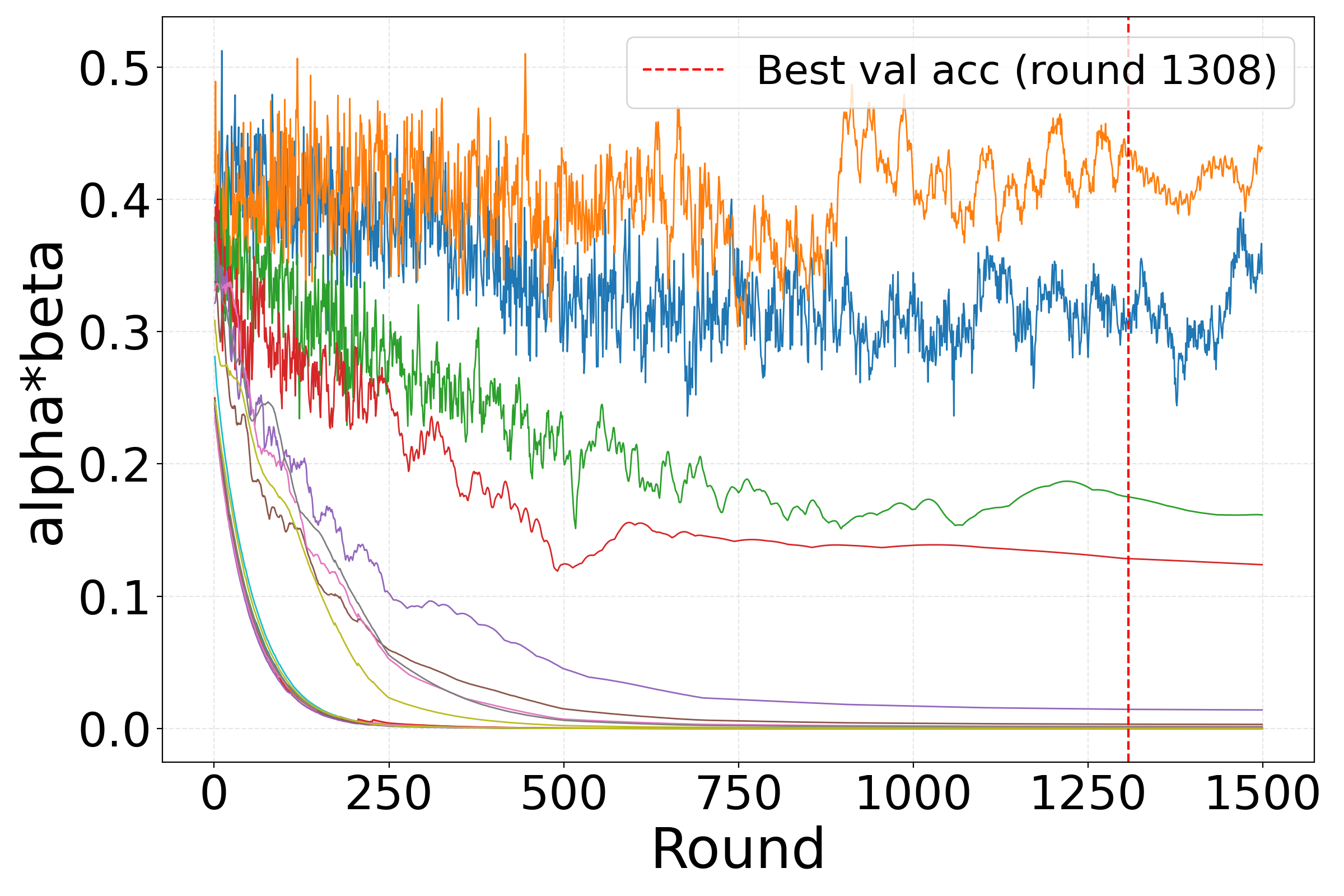}
\caption{Non-IID2: $\alpha\beta$}
\end{subfigure}

\vspace{0.6em}

% -------- Row 3: alpha(1-beta) --------
\begin{subfigure}{0.32\textwidth}
\includegraphics[width=\linewidth]{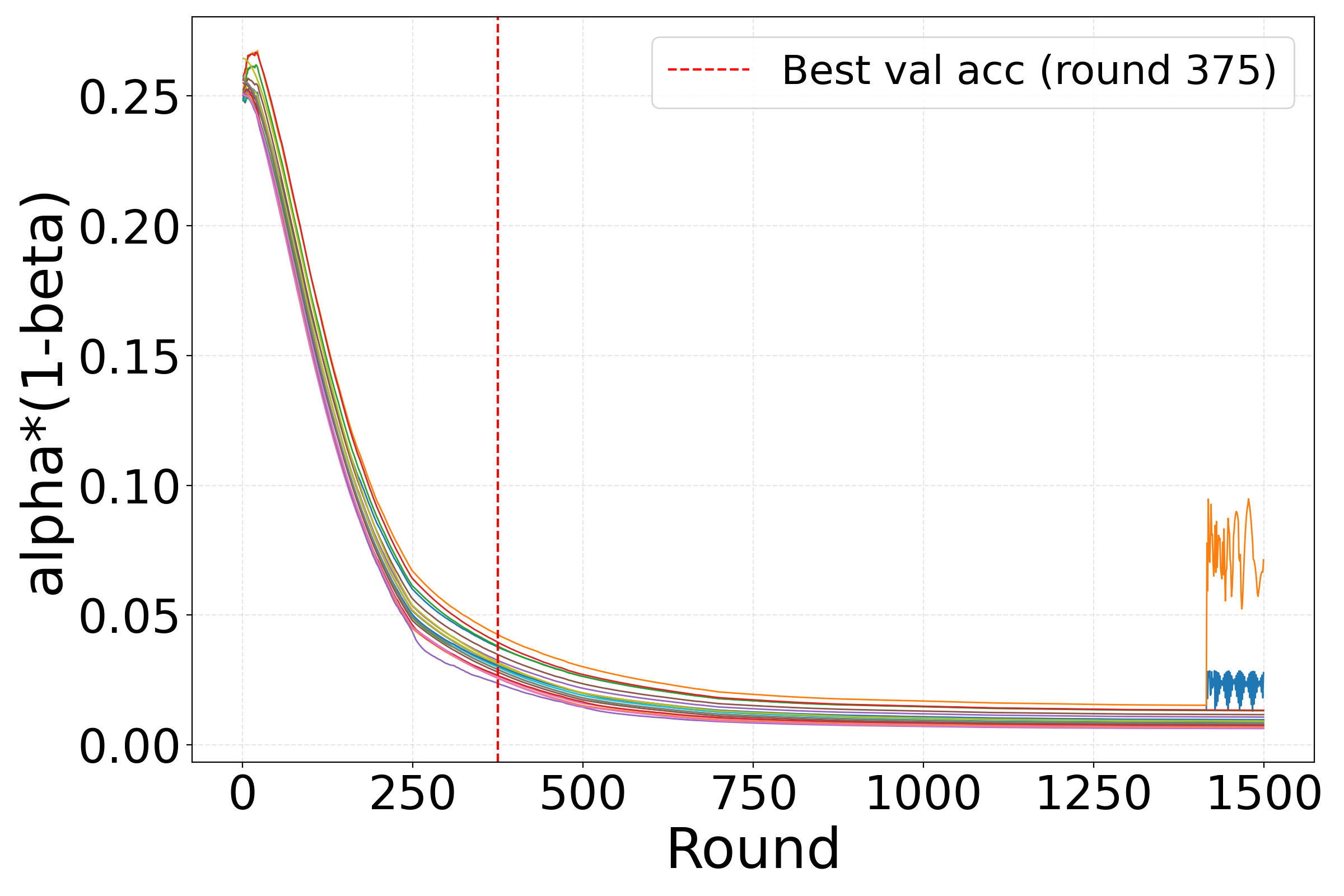}
\caption{IID: $\alpha(1-\beta)$}
\end{subfigure}
\begin{subfigure}{0.32\textwidth}
\includegraphics[width=\linewidth]{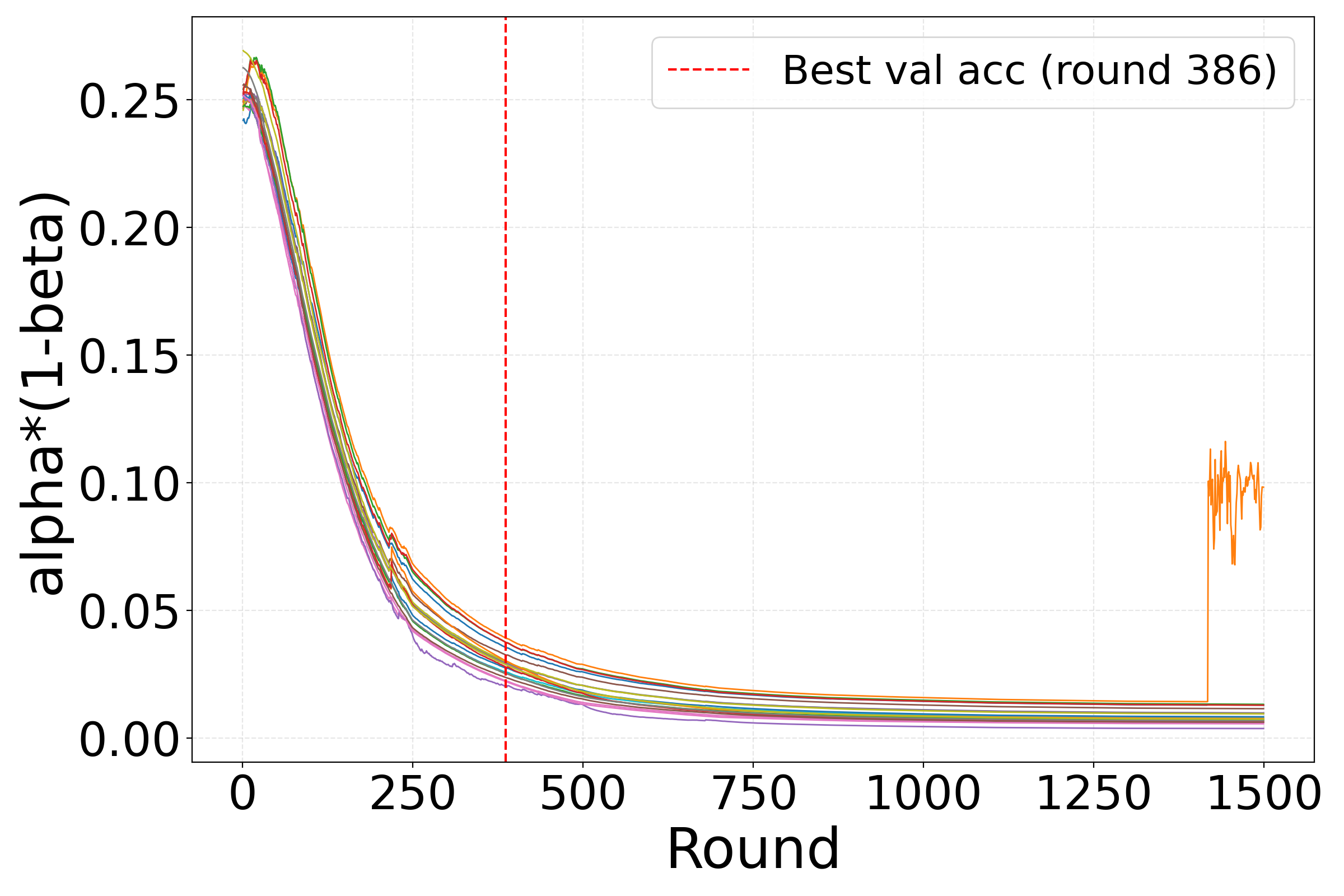}
\caption{Non-IID1: $\alpha(1-\beta)$}
\end{subfigure}
\begin{subfigure}{0.32\textwidth}
\includegraphics[width=\linewidth]{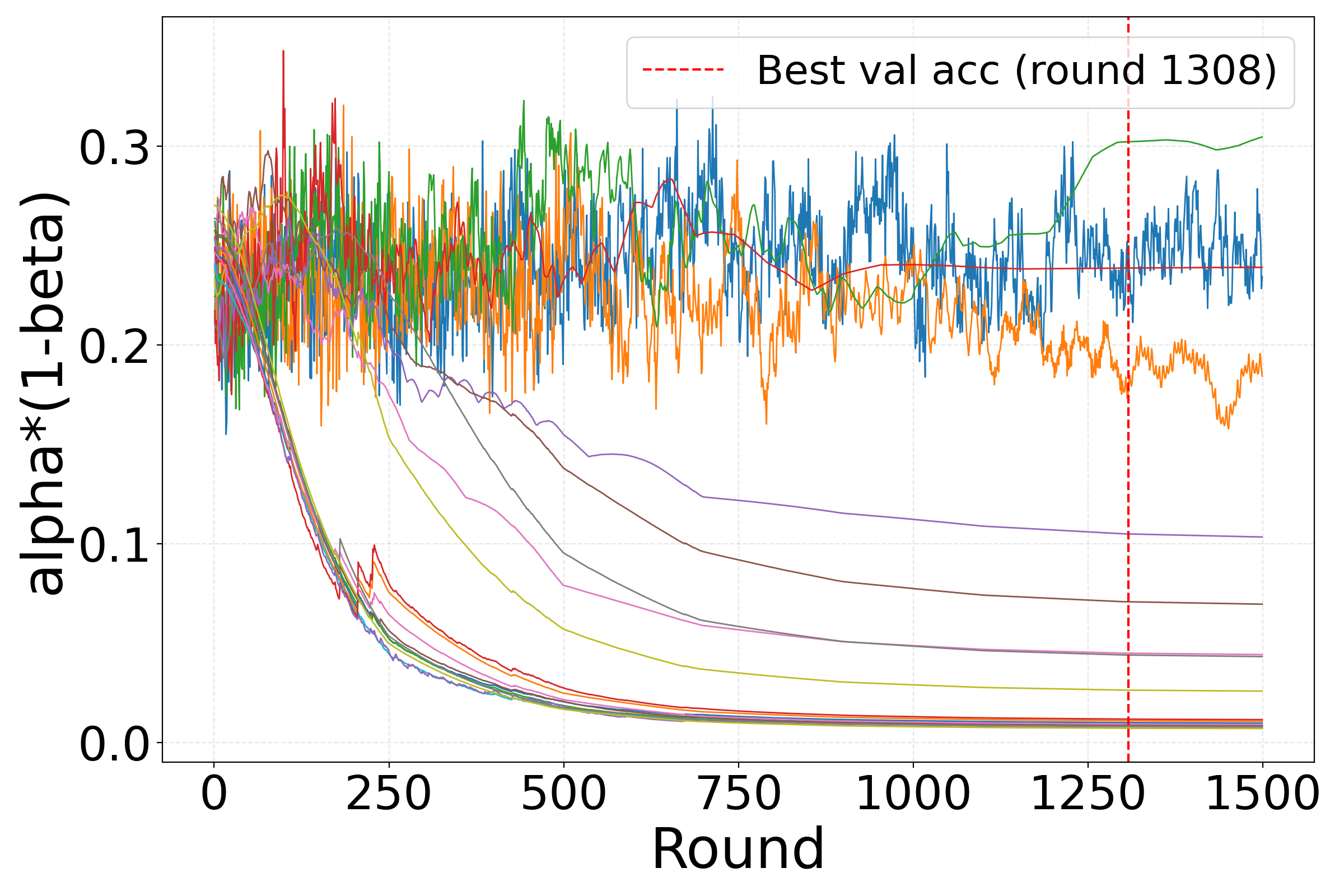}
\caption{Non-IID2: $\alpha(1-\beta)$}
\end{subfigure}

\vspace{0.6em}

% -------- Row 4: lambda --------
\begin{subfigure}{0.32\textwidth}
\includegraphics[width=\linewidth]{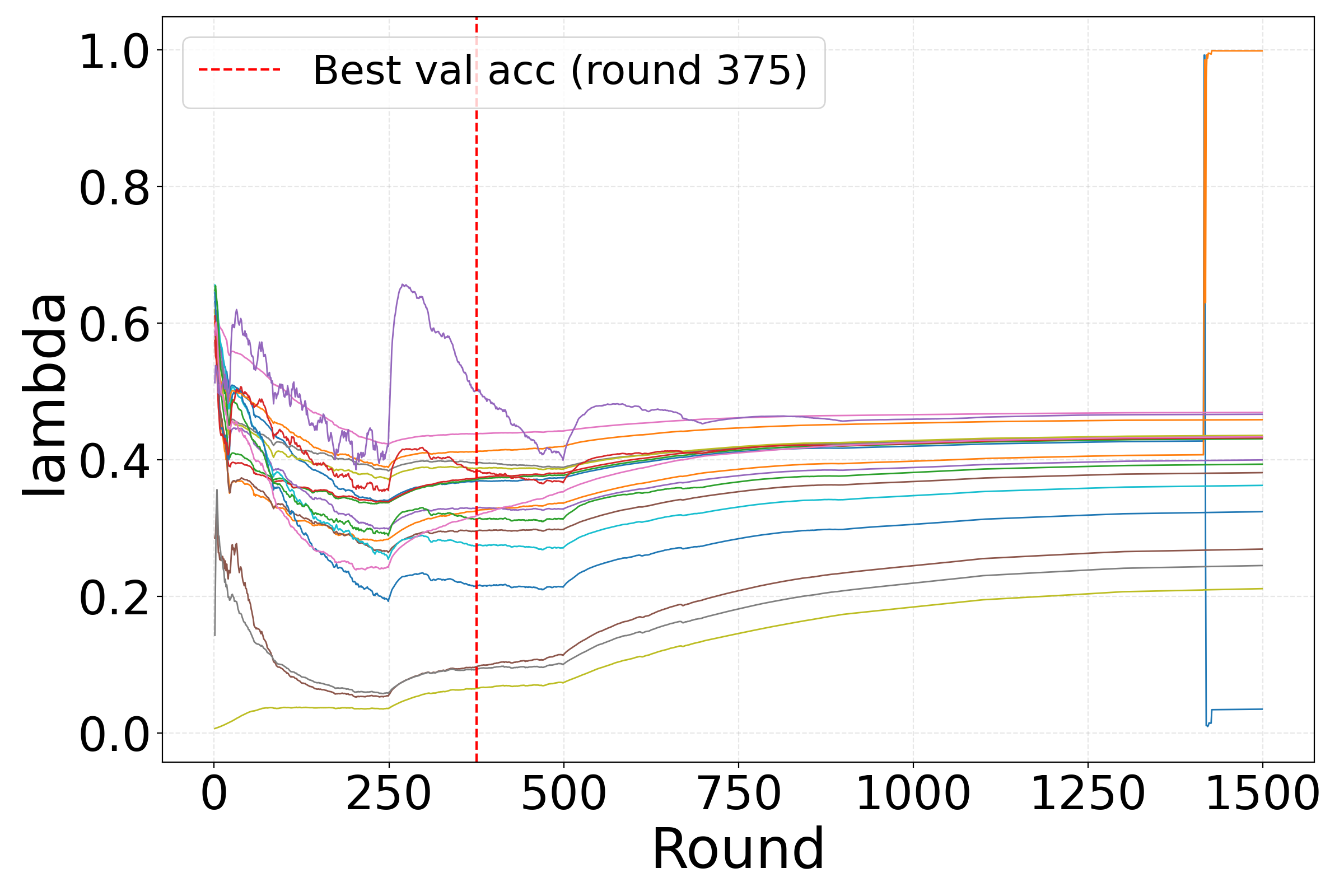}
\caption{IID: $\lambda$}
\end{subfigure}
\begin{subfigure}{0.32\textwidth}
\includegraphics[width=\linewidth]{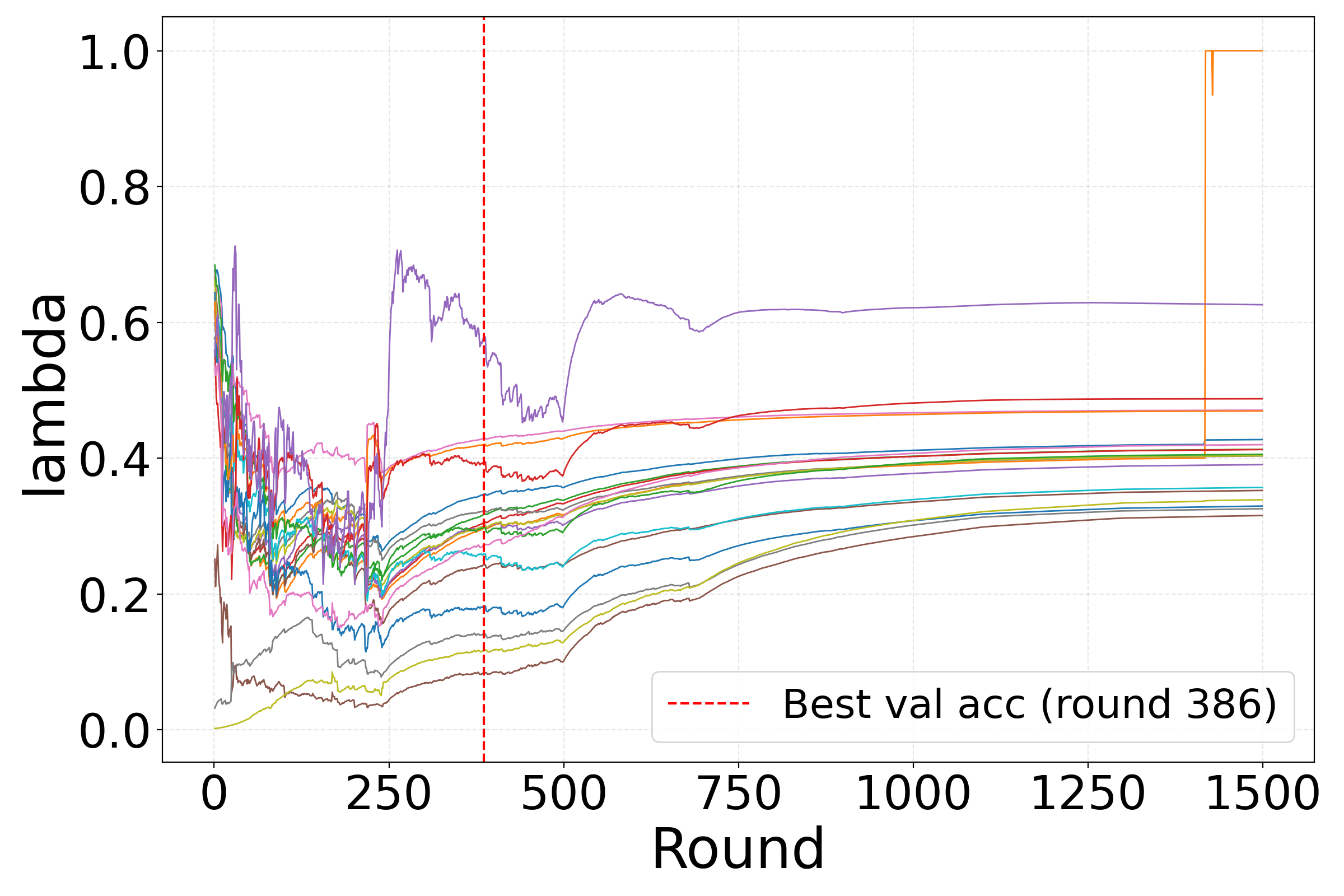}
\caption{Non-IID1: $\lambda$}
\end{subfigure}
\begin{subfigure}{0.32\textwidth}
\includegraphics[width=\linewidth]{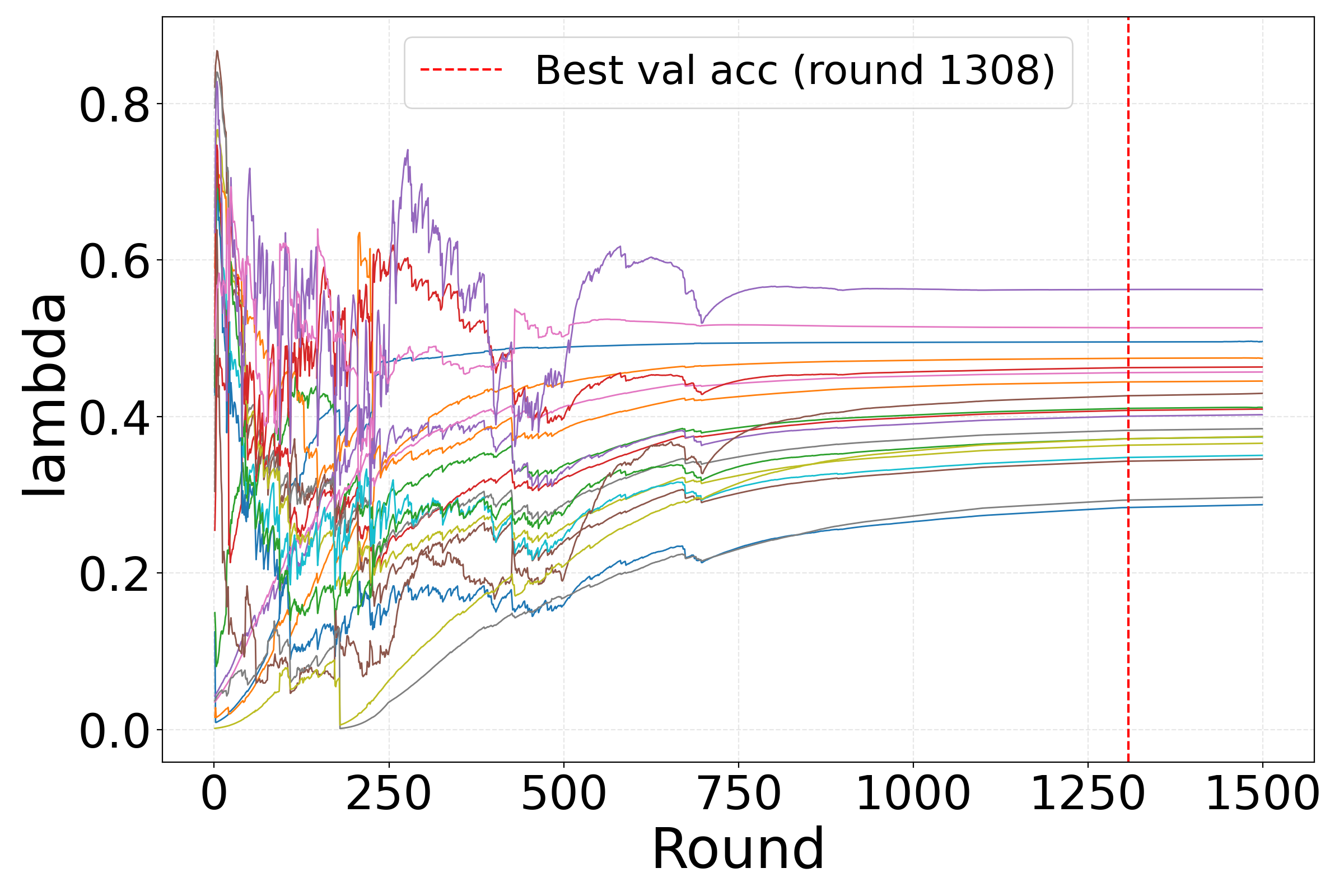}
\caption{Non-IID2: $\lambda$}
\end{subfigure}

% -------- Row 5: legend --------
\begin{subfigure}{0.32\textwidth}
\includegraphics[width=\linewidth]{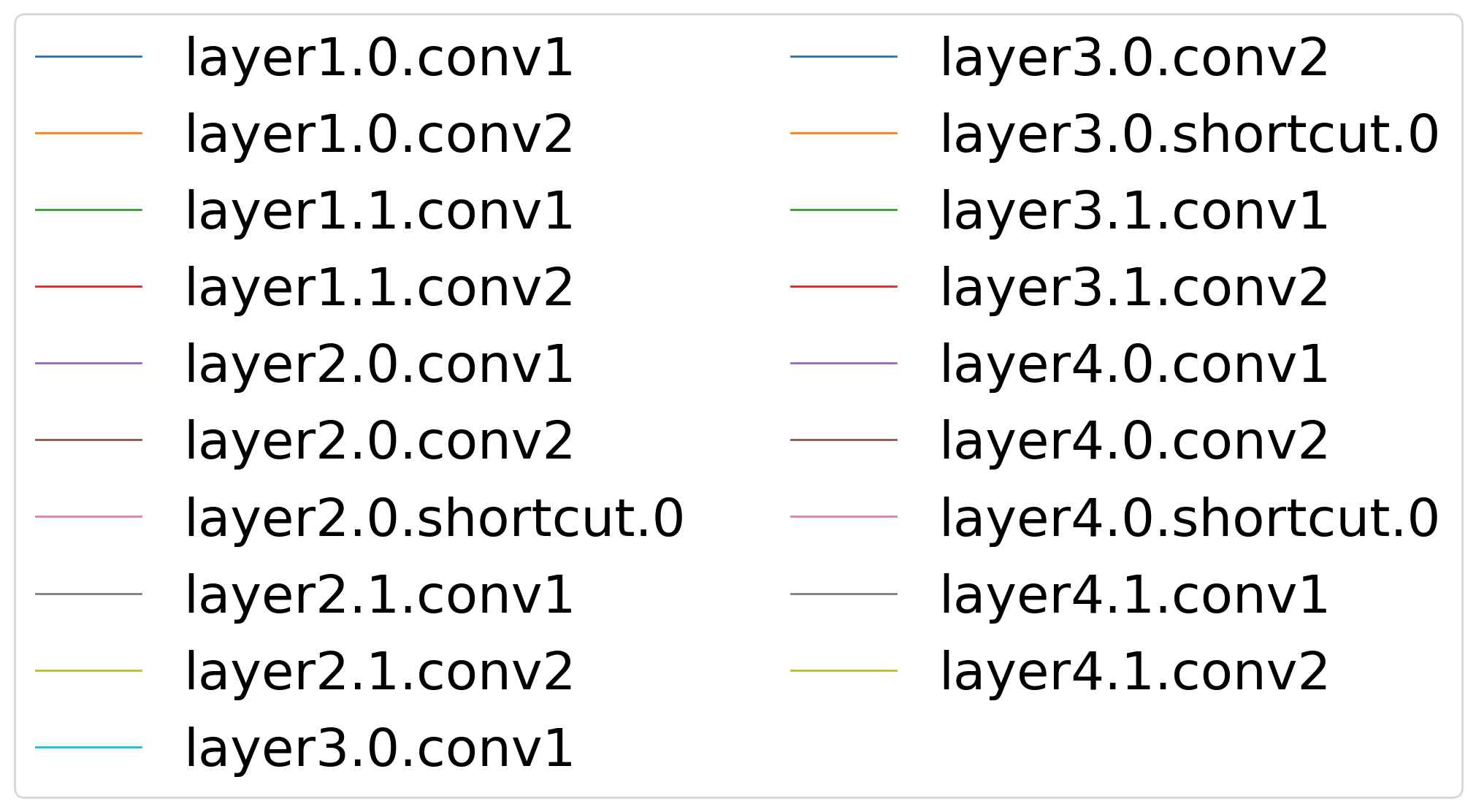}
\caption{Legend}
\end{subfigure}

\caption{Mixing parameter variation on FEMNIST under IID, Non-IID1, and Non-IID2 settings.}
\label{alphabetalambdafemnist}
\end{figure*}

{
We analyze the evolution of FedBNN's aggregation parameters: $\lambda$, $\alpha$, and $\beta$, over communication rounds and across layers, for multiple datasets (FMNIST, SVHN, CIFAR10, FEMNIST, TinyImageNet) and data-heterogeneity settings (IID, Non-IID-1 and Non-IID-2). The plots in Figure~\ref{alphabetalambdafmnist}, \ref{alphabetalambdasvhn}, \ref{alphabetalambdacifar10resnet10}, \ref{alphabetalambdacifar10convnext}, \ref{alphabetalambdatinyimagenet} and \ref{alphabetalambdafemnist} show, per round and per layer, four quantities: $\lambda$ (client weight contribution before rotation optimization of fused weight), $1-\alpha$ (client fused weight contribution), $\alpha\beta$ (rotated weight contribution), and $\alpha(1-\beta)$ (server contribution). A vertical line marks the round at which the model attains its best validation accuracy. These visualizations reveal how the balance between server and client contributions and the balance between rotated and non-rotated terms evolves during training; the round with the best validation performance often coincides with a favorable regime of these quantities, supporting the role of the rotation mechanism in FedBNN's convergence. The per-layer variation indicates that different layers adapt their client-versus-server and rotation weighting differently under federated training, which is consistent with the design of client-specific rotations that help align binary representations across non-IID clients. {Furthermore, the adaptive coefficients exhibit an annealing behavior: $1-\alpha \rightarrow 1$ and $\alpha\beta \rightarrow 0$ as optimization converges. This demonstrates that the proposed rotation correction automatically diminishes during late-stage training, suggesting that its primary role is to facilitate optimization rather than permanently modify the learned binary representation.} Overall, the $\alpha$-$\beta$-$\lambda$ analysis provides empirical evidence that FedBNN's parameterization is well-behaved across datasets and levels of heterogeneity and that the rotation-based aggregation is a key factor in its performance.
}

\newpage

\clearpage

% \appendix
% \section{Appendix}

\end{document}